\documentclass[cspaper,compsoc]{IEEEtran} % for computer society journals

\ifCLASSOPTIONcompsoc
  \usepackage[nocompress]{cite}
\else
  \usepackage{cite}
\fi
\ifCLASSINFOpdf
\else
\fi

\usepackage{ifpdf}

\usepackage[pdftex]{graphicx}
\usepackage{epsfig} 
\usepackage{amsthm}
\usepackage{amsmath}
\usepackage{amsfonts}
\usepackage{amssymb}
\usepackage{bm}
\usepackage{cases}
\usepackage{multirow}

\usepackage{multicol}
\usepackage{xcolor}
\usepackage{float,subfig}
\usepackage{caption}
\usepackage{cite}
\usepackage{url}
\usepackage{here}
\usepackage{stfloats} 
\usepackage{braket}
\usepackage[linesnumbered,ruled,vlined]{algorithm2e}
\usepackage{algorithmic}

\usepackage{multicol}
\usepackage[export]{adjustbox}
\usepackage{hhline}
\usepackage{colortbl}

\usepackage{hyperref}
\usepackage{xurl}
\usepackage[hang,flushmargin]{footmisc}
\hypersetup{
    colorlinks=true,
    linkcolor=blue, 
    urlcolor=magenta,
    }
\makeatletter
\let\saved@hyper@linkurl\hyper@linkurl
\let\saved@hyper@link@\hyper@link@
\AtBeginDocument{%
  \NoHyper % activate hyperlink for references
  
  \let\hyper@linkurl\saved@hyper@linkurl % needed by \url
  \let\hyper@link@\saved@hyper@link@ % needed by \href{<url>}
}
\makeatother

\usepackage{array}
\newcolumntype{C}[1]{>{\hfil}m{#1}<{\hfil}}

\usepackage{lscape} % figure/table rotation

\newcounter{sfig}     % no automatic reset
\newcounter{stab}
\newcommand{\sfiglabel}[1]{% #1 = label name
  \refstepcounter{sfig}% make this the current label value
  \label{#1}%
}
\newcommand{\stablabel}[1]{%
  \refstepcounter{stab}%
  \label{#1}%
}
\newcommand{\sfigref}[1]{\ref{#1}}
\newcommand{\stabref}[1]{\ref{#1}}

\begin{document}
\title{An Adaptive Resonance Theory-based Topological Clustering Algorithm \\ with a Self-Adjusting Vigilance Parameter}

\author{
	% Naoki~Masuyama,~\IEEEmembership{Member,~IEEE,}
	% Takanori Takebayashi,
	% Yusuke~Nojima,~\IEEEmembership{Member,~IEEE,} \\
	% Chu~Kiong~Loo,~\IEEEmembership{Senior Member,~IEEE,}
	% Hisao~Ishibuchi,~\IEEEmembership{Fellow,~IEEE,} \\
	% and~Stefan~Wermter,~\IEEEmembership{Member,~IEEE}
	Naoki~Masuyama,
	Yuichiro~Toda,
	Yusuke~Nojima,
	and~Hisao~Ishibuchi
	\thanks{N. Masuyama, and Y. Nojima are with the Graduate School of Informatics, Department of Core Informatics, Osaka Metropolitan University, 1-1 Gakuen-cho Naka-ku, Sakai-Shi, Osaka 599-8531, Japan, e-mails: masuyama@omu.ac.jp, nojima@omu.ac.jp.}% <-this % stops a space
	\thanks{Y. Toda is with the Faculty of Environmental, Life, Natural Science and Technology, Okayama University, Okayama-shi, Okayama 700-8530, Japan, e-mail: ytoda@okayama-u.ac.jp}% <-this % stops a space
	\thanks{H. Ishibuchi is with the Guangdong Provincial Key Laboratory of Brain-inspired Intelligent Computation, Department of Computer Science and Engineering, Southern University of Science and Technology, Shenzhen 518055, China, e-mail: hisao@sustech.edu.cn.}% <-this % stops a space
	\thanks{Corresponding author: Naoki Masuyama (e-mail: masuyama@omu.ac.jp).}
	\thanks{Manuscript received April 19, 2005; revised August 26, 2015.}
}

% \author{
% 	% Naoki~Masuyama,~\IEEEmembership{Member,~IEEE,}
% 	% Takanori Takebayashi,
% 	% Yusuke~Nojima,~\IEEEmembership{Member,~IEEE,} \\
% 	% Chu~Kiong~Loo,~\IEEEmembership{Senior Member,~IEEE,}
% 	% Hisao~Ishibuchi,~\IEEEmembership{Fellow,~IEEE,} \\
% 	% and~Stefan~Wermter,~\IEEEmembership{Member,~IEEE}
% 	Anonymous Authors
% }


\markboth{Journal of \LaTeX\ Class Files,~Vol.~14, No.~8, August~2015}%
{Shell \MakeLowercase{\textit{et al.}}: Bare Demo of IEEEtran.cls for Computer Society Journals}



\IEEEtitleabstractindextext{%
\begin{abstract}
  Clustering in stationary and nonstationary settings, where data distributions remain static or evolve over time, requires models that can adapt to distributional shifts while preserving previously learned cluster structures. This paper proposes an Adaptive Resonance Theory (ART)-based topological clustering algorithm that autonomously adjusts its recalculation interval and vigilance threshold through a diversity-driven adaptation mechanism. This mechanism enables hyperparameter-free learning that maintains cluster stability and continuity in dynamic environments. Experiments on 24 real-world datasets demonstrate that the proposed algorithm outperforms state-of-the-art methods in both clustering performance and continual learning capability. These results highlight the effectiveness of the proposed parameter adaptation in mitigating catastrophic forgetting and maintaining consistent clustering in evolving data streams. Source code is available at \url{https://github.com/Masuyama-lab/IDAT}
\end{abstract}

}

\maketitle

\IEEEdisplaynontitleabstractindextext

% \fi
\IEEEpeerreviewmaketitle

\IEEEraisesectionheading{\section{Introduction}\label{sec:introduction}}

\IEEEPARstart{C}{lustering} is a fundamental approach in data analysis that groups similar data to reveal underlying structures, simplify large datasets, and support subsequent tasks such as visualization and anomaly detection. Traditional clustering methods such as $k$-means \cite{lloyd82}, spectral clustering \cite{ng01}, and hierarchical clustering \cite{ward63} have demonstrated their effectiveness across a wide range of domains. In recent years, deep learning-based clustering approaches have also been proposed \cite{jabi19, li22, williams25}, and they are capable of effective clustering of high-dimensional data such as images and texts. Although various improvements have been proposed for conventional methods, they have difficulty adapting to situations where the data distribution changes over time, new clusters emerge, or new classes are introduced in a class-incremental manner. In contrast, in most real-world applications, data are continually generated over time, and this situation emphasizes the need for online clustering methods.

Representative online clustering methods are growing self-organizing neural approaches such as Growing Neural Gas (GNG) \cite{fritzke95}, Self-Organizing Incremental Neural Network (SOINN) \cite{furao06}, and Adaptive Resonance Theory (ART)-based clustering methods \cite{carpenter88, carpenter91b, grossberg23}. These methods adaptively update their structures to accommodate newly arriving data and changing data distributions. However, many existing approaches depend on data-dependent parameters and have limited applicability in nonstationary or open-world environments. Recently, deep learning-based clustering methods have been proposed as versatile methods. However, it is difficult for them to handle data arriving in a streaming manner since they require large amounts of data and are mostly designed for batch learning \cite{ren24}.

ART-based clustering has the unique advantage of effectively addressing the plasticity-stability dilemma \cite{carpenter88}. This property allows clustering models to retain previously acquired knowledge while adapting to new data and mitigates catastrophic forgetting. ART algorithms also support stream learning, in which each sample is processed only once without iterative retraining. The main limitation of ART-based clustering methods is that their performance depends on parameter values (e.g., vigilance threshold) and appropriate parameter specifications are not easy \cite{masuyama22b, yelugam23}. To address this limitation, this paper proposes a novel parameter-free clustering method called Inverse Distance-based ART with Topology (IDAT). Similar to GNG and SOINN, IDAT represents data by nodes corresponding to local regions, and clusters are defined as connected components of these nodes in the topology. 

IDAT builds on ART and adjusts its parameters based on the current data distribution, which removes manual parameter specification difficulty. It maintains a streaming buffer and updates both the recalculation interval and the vigilance using statistics from recently observed samples without referencing future observations. This mechanism enables continual adaptation to distributional changes while incrementally updating a topological network in a streaming manner. Note that continual learning is commonly categorized into domain-incremental, task-incremental, and class-incremental settings \cite{van19, wiewel19}. This paper focuses on the class-incremental setting, in which new classes are introduced one at a time without forgetting prior knowledge.

The main contributions of this paper are summarized as follows:
\begin{itemize}
	\vspace{-1mm}
	\item[(i)] IDAT is proposed as a parameter-free ART-based topological clustering algorithm that automatically adjusts its recalculation interval $\Lambda$ and vigilance threshold $V_{\text{threshold}}$, thereby eliminating the need for manual parameter setting.

	\item[(ii)] A diversity-driven adaptation mechanism is introduced to control $\Lambda$ and $V_{\text{threshold}}$ in both the incremental and decremental directions. This mechanism achieves a balance between stability and plasticity under evolving data distributions while maintaining consistent topological organization.

	\item[(iii)] Extensive experiments on 24 real-world datasets show that IDAT outperforms a wide range of state-of-the-art clustering algorithms under both stationary and nonstationary settings.
	\vspace{-1mm}
\end{itemize}

The paper is organized as follows. Section \ref{sec:literature} reviews batch, self-organizing, and deep learning-based clustering methods. Section \ref{sec:proposedAlgorithm} details the learning procedure of IDAT. Section \ref{sec:experiment} reports experimental results on real-world datasets. Section \ref{sec:conclusion} concludes the paper.

\section{Literature Review}
\label{sec:literature}

\subsection{Batch Clustering}
\label{sec:batchclustering}
Representative batch clustering algorithms include $k$-means \cite{lloyd82}, Gaussian mixture model \cite{mclachlan19}, spectral clustering \cite{ng01}, and hierarchical clustering \cite{ward63}. These methods typically require the number of clusters to be specified before training. Although this constraint limits flexibility in dynamic environments, its simplicity and high clustering performance have supported wide adoption.

Numerous variants and extensions have been developed to improve clustering performance and its functionality \cite{wang24, zhou24bsurvey, yu24, ding24, ran23}. Density-based clustering methods \cite{kumar24, hu25} identify clusters based on the spatial distribution of data points rather than relying on predefined centroids. A representative state-of-the-art method is Self-Regulating Possibilistic C-Means with High-Density Points (SR-PCM-HDP) \cite{hu25}, which identifies high-density representatives and regulates boundaries to enhance robustness to noise and outliers. Graph-based clustering methods \cite{sarfraz19, yang25} represent data as nodes with edges and extract clusters via graph partitioning or connectivity analysis. A leading state-of-the-art method is Torque Clustering (TC) \cite{yang25}, which defines a torque measure between nodes to capture structural relations and detects natural cluster boundaries without a predefined number of clusters.

\subsection{Growing Self-Organizing Clustering}
\label{sec:selforgnizing}
Growing self-organizing methods adaptively generate and update centroids (i.e., nodes) in response to sequentially arriving data. Unlike batch methods, they learn the data structure in a streaming manner and adaptively adjust both the number and positions of clusters in response to evolving data distributions.

Representative algorithms include GNG \cite{fritzke95} and SOINN-based variants \cite{furao06, shen08}. These algorithms incrementally construct topological networks (i.e., nodes and edges) that represent the underlying data distribution. As the network expands to learn new samples, previously acquired knowledge can be overwritten. This phenomenon is known as the plasticity-stability dilemma \cite{carpenter88}. The Grow When Required (GWR) algorithm \cite{marsland02} mitigates this dilemma by inserting new nodes when the existing network does not adequately represent newly arriving data. SOINN+ \cite{wiwatcharakoses20}, an improved SOINN-based method, can identify clusters of arbitrary shapes in noisy streams without predefined parameters. However, as the number of nodes increases, the cost of per-node threshold calculation grows and learning efficiency decreases.

ART-based clustering provides an effective mechanism for maintaining a balance between plasticity and stability \cite{carpenter88}. It preserves prior knowledge while adapting to new data and supports stream learning in which each sample is processed once without iterative retraining \cite{carpenter91b, vigdor07, wang19, masuyama18, masuyama19a, masuyama19b, da19, da20, masuyamaFTCA, masuyama22b}. Recent studies have incorporated topological representations (i.e., nodes and edges) into ART \cite{tscherepanow10, masuyama22a, yelugam23, masuyama24a} to capture complex data geometries while maintaining the adaptive clustering capabilities of ART. In addition, a notable recent development is the use of the Correntropy-Induced Metric (CIM) \cite{liu07}, which provides a kernel-based similarity measure. CIM mitigates the problem of distance concentration in high-dimensional spaces and preserves local manifold structures. CIM-based ART with Edge and Age (CAEA) \cite{masuyama22a} and its enhanced variant called CIM-based ART with Edge (CAE) \cite{masuyama23} demonstrate the superior self-organizing performance compared with conventional ART-based clustering algorithms. A representative state-of-the-art method of ART-based clustering is Topological Biclustering ARTMAP with a Pearson correlation (TBM-P) \cite{yelugam23}, which extends Biclustering ARTMAP to learn topology via TopoART \cite{tscherepanow10}.

Despite these progresses, the performance of ART-based clustering still depends heavily on data-dependent parameters (e.g., vigilance parameter). Several studies have attempted to alleviate this dependency through adaptive scaling \cite{meng15}, indirect specification \cite{da22a, da22b, masuyama22a}, and direct specification \cite{meng13, majeed18}. Although such approaches reduce parameter sensitivity to some extent, their reliance on pre-specified settings continues to limit their practical usefulness in continual learning. This limitation motivates our parameter-free algorithm proposed in this paper, which aims to enhance adaptability and robustness in dynamic environments.

From an application perspective, growing self-organizing clustering has been widely explored in continual learning \cite{parisi18, george19, hafez23}. It has also been applied to simultaneous localization and mapping \cite{chin18, toda24} and to knowledge acquisition for robots \cite{hafez23b, dawood25}, demonstrating versatility across diverse domains. Nevertheless, many existing systems require careful parameter specifications to achieve high performance, which further underscores the need for our parameter-free algorithm proposed in this paper.

\subsection{Deep Learning-based Clustering}
\label{sec:deepclustering}
The rapid progress of deep learning has inspired many clustering approaches that use deep neural networks. These methods have shown high performance, particularly on high-dimensional data such as images and texts \cite{jabi19, li22, williams25, ren24}. However, they do not consistently outperform traditional methods on tabular data. Classical machine learning techniques often achieve competitive results in this domain \cite{shwartz22}, indicating that the advantages of deep learning are not universal across all data modalities. Recent studies have proposed deep learning-based techniques specifically for tabular data. Table Deep Clustering (TableDC) \cite{rauf25} extends deep clustering to tabular domains using an autoencoder with Mahalanobis distance for feature correlation and a Cauchy distribution for robust similarity estimation. Gaussian Cluster Embedding in Autoencoder Latent Space (G-CEALS) \cite{rabbani25} replaces $t$-distributions with multivariate Gaussian mixtures and learns tabular embeddings through independent target and latent distributions for flexible cluster-aware representations. These algorithms are designed for batch learning and assume that all input samples are available in advance. They rely on iterative retraining to integrate new information rather than dynamically updating representations or cluster structures as new classes are introduced. As a result, they require large datasets and are unsuitable for class-incremental learning.

Although continual learning has been extensively studied in deep learning \cite{zhou24a}, effective methods for the class-incremental setting remain limited. In this setting, new classes are introduced one at a time, and models must learn them sequentially while retaining previously acquired knowledge. In addition to supervised learning, continual learning has also been explored in unsupervised contexts \cite{sadeghi24, solomon24, aghasanli25, zhu25}, where the absence of labels makes stable adaptation more difficult. Most existing approaches still depend on iterative retraining and repeated access to previously seen data to maintain clustering performance over time. To the best of our knowledge, there are currently no deep learning approaches that learn stably and efficiently in a streaming manner without retraining or data replay under the class-incremental setting \cite{zhou24survey}.

\section{Proposed Algorithm}
\label{sec:proposedAlgorithm}

\subsection{Overview}
\label{sec:overview}
Table \ref{tab:idat_notations} summarizes the main notations used in this paper, and Algorithm \ref{alg:fit_idat} presents the overall learning procedure of the proposed IDAT algorithm.

The proposed IDAT algorithm extends ART-based topological clustering by introducing automatic parameter adaptation for continual learning. IDAT retains data from local regions as nodes, and clusters are defined by connected components formed by these nodes and their edges in the topology. To realize continual adaptation, the recalculation interval $\Lambda$, which determines the diversity of nodes used to estimate the vigilance parameter $V_{\text{threshold}}$, and $V_{\text{threshold}}$ itself is jointly adjusted during learning.

The following subsections describe the key learning processes of IDAT.

\begin{table}[htbp]
\centering
\caption{Summary of main notations used in IDAT}
\label{tab:idat_notations}
\renewcommand{\arraystretch}{1.2}
\begin{tabular}{ll}
\hline\hline
\textbf{Notation}                                  & \textbf{Description} \\ \hline
$\mathbf{x}_t$                                    & input sample observed at time $t$ \\
$\mathcal{X}=\{\mathbf{x}_t\}_{t=1}^{N}\subset\mathbb{R}^d$  & sequence of $d$-dimensional samples \\
% $N$                                               & total number of processed samples \\
$\mathcal{Y}=\{\mathbf{y}_k\}_{k=1}^{K}$         & set of current nodes \\
% $K=|\mathcal{Y}|$                                 & number of nodes \\
$E\in\{0,1\}^{K\times K}$                         & adjacency matrix of the topological graph \\
$E^{\mathrm{cand}}\in\mathbb{R}^{K\times K}$      & matrix of edge-candidate counts \\
$\mathcal{G}=(\mathcal{Y},E)$                     & topological graph \\
$C$                                               & number of connected components in $\mathcal{G}$ \\
$M_k$                                             & update count of node $k$ \\
$\mathcal{M}=\{M_k\}_{k=1}^{K}$                  & set of update counts \\
$a_k\in\{0,1\}$                                   & activity flag of node $k$ (1: active, 0: inactive) \\
$\mathcal{A}=\{a_k\}_{k=1}^{K}$                  & set of node activity states \\
$U_k$                                             & inactivity counter of node $k$ \\
$\sigma_k$                                       & local scaling factor associated with node $k$ \\
$\mathcal{S}=\{\sigma_k\}_{k=1}^{K}$             & set of scaling factors \\
% $\alpha_k=1/\sigma_k$                            & inverse scaling factor of node $k$ \\
% $\alpha^\star=1/\max_k\sigma_k$                  & global inverse scaling factor used for similarity computation \\
% $d_k=\|\mathbf{x}-\mathbf{y}_k\|_2$              & Euclidean distance between sample $\mathbf{x}$ and node $\mathbf{y}_k$ \\
% $V_k=\dfrac{1}{1+\alpha_k d_k}$                  & inverse-distance similarity for vigilance evaluation \\
$s_1, s_2$                                       & indices of the 1st and 2nd nearest nodes \\
$V_{\text{threshold}}$                           & vigilance threshold \\
% $V_{\text{threshold}}^{\text{new}}$              & updated vigilance threshold after recalculation \\
$\Lambda$                                        & recalculation interval \\
% $\Lambda^{\text{new}}$                           & updated recalculation interval \\
% $W_t=\{\mathbf{x}_{t-L+1},\dots,\mathbf{x}_t\}$  & most recent buffer of $L$ samples \\
$L$                                              & window length for diversity evaluation \\
% $L^\star$                                        & largest window length satisfying the diversity condition \\
% $L_{\max}=2\Lambda$                              & maximum window length in incremental adjustment \\
$S_L\in\mathbb{R}^{L\times L}$                   & similarity matrix within the window $L$ \\
$\widehat{\det}(S_L)$                            & determinant of similarity matrix $S_L$ \\
% $\varepsilon_{\det}=1.0\times10^{-6}$            & tolerance for numerical stability (Eq.~\eqref{eq:diversity_condition}) \\
% $\textit{is\_stable}$                            & flag indicating whether the diversity condition is satisfied \\
% $\widetilde{S}_L=S_L-\operatorname{diag}(S_L)$   & similarity matrix with diagonal self-similarities removed (Eq.~\eqref{eq:remove_diag}) \\
$u^{(L)}$                               & row-wise maxima vector of $S_L$ \\
$q$                                              & quantile level for the threshold $V_{\text{threshold}}$ \\
% $q^{\text{prev}}$                                & quantile level from the previous recalculation cycle \\
% $q^{\text{new}}$                                 & updated quantile level computed from graph connectivity (Eq.~\eqref{eq:quantile_level}) \\
$r$                                              & recomputation counter for $V_{\text{threshold}}$ \\
$T_{\text{active}}$                              & threshold for node activation \\
$T_{\text{edge}}$                                & threshold for edge formation \\
$T_{\text{demote}}$                              & threshold for node demotion \\
$P_k$                                            & activity potential for node $k$ \\
% $\ell_{ij}=\|\mathbf{y}_i-\mathbf{y}_j\|_2$      & edge length between nodes $i$ and $j$ (Eq.~\eqref{eq:edge_length}) \\
$Q_1, Q_3$                                       & 1st and 3rd quartiles \\
$\mathrm{IQR}=Q_3-Q_1$                           & interquartile range \\
$\widehat{c}$                                  & predicted label for sample $\mathbf{x}$ \\
\hline\hline
\end{tabular}
\end{table}

\subsection{Initialization}
\label{sec:initialization}
IDAT initializes its internal states independently of the input data, setting most of them to zero or empty values, while fixing $\Lambda_{\text{init}}$ to 2 to provide a minimal buffer length for the initial evaluation of local diversity. IDAT then processes incoming samples sequentially in a streaming manner while incrementally updating its nodes, edges, and internal statistics. This design enables the algorithm to autonomously adapt to changes in the data distribution while maintaining continual learning in the nonstationary setting.

% Main Procedure (fit)
\begin{algorithm*}[htbp]
\footnotesize
\DontPrintSemicolon
\setlength{\algomargin}{1.2em} % reduce left margin
\KwIn{ \\
  \quad samples $\mathcal{X}=\{\mathbf{x}_t\}_{t=1}^{N}\subset\mathbb{R}^d$;\;
  \quad carried-over internal state consisting of:\;
  \quad \quad\quad \textit{Topology:} $\mathcal{Y}=\{\mathbf{y}_k\}$ (nodes), $E$ (adjacency matrix), $E^{\mathrm{cand}}$ (edge-candidate matrix);\;
  \quad \quad\quad \textit{Parameters:} $V_{\textup{threshold}}$ (vigilance threshold), $\mathcal{S}=\{\sigma_k\}$ (scaling factors), $\mathcal{M}=\{M_k\}$ (update counts), $q$ (quantile level);\;
  \quad \quad\quad \textit{Activity:} $\mathcal{A}=\{a_k\}$ (node activity states, $a_k\in\{0,1\}$), $\{U_k\}$ (inactivity counters);\;
  \quad \quad\quad \textit{Snapshots:} $M^{\text{prev}}$, $E^{\mathrm{cand,prev}}$;\;
  \quad \quad\quad \textit{Configuration:} recalculation interval $\Lambda$
}
\KwOut{ \\
  \quad updated internal state: $\mathcal{Y}, E, E^{\mathrm{cand}}, \Lambda, V_{\textup{threshold}}, \mathcal{S}, \mathcal{M}, q, r$
}

\tcp{Initialization on the very first training call; otherwise keep the carried-over state}
\uIf{\normalfont{this is the first training call}}{
  Initialize $\mathcal{Y}\leftarrow\emptyset$, $E\leftarrow\emptyset$, $E^{\mathrm{cand}}\leftarrow\emptyset$, $V_{\textup{threshold}}\leftarrow 0$, $\mathcal{S}\leftarrow\emptyset$, $\mathcal{M}\leftarrow\emptyset$, $q\leftarrow 0$, $r\leftarrow 0$;\;
  $\Lambda_{\text{init}} \leftarrow 2$; \tcp{minimum buffer size}
}
\Else{
  Retain the carried-over internal state from the previous session;\;
}

\SetKw{And}{and}
\While{\normalfont{existing samples to be trained}}{
  Input a sample $\mathbf{x}$ ($\mathbf{x}\in\mathcal{X}$);\;
  Append $\mathbf{x}$ to the buffer of length $2\Lambda$;\;

  \tcp{Perform clustering for the current sample}
  $(\mathcal{Y}, E, E^{\mathrm{cand}}, \mathcal{S}, \mathcal{M}, \mathcal{A}, \{U_k\}, M^{\text{prev}}, E^{\mathrm{cand,prev}}) \leftarrow 
  \textsc{ClusteringStep}(\mathbf{x}, \mathcal{Y}, E, E^{\mathrm{cand}}, V_{\textup{threshold}}, \mathcal{S}, \mathcal{M}, \mathcal{A}, \{U_k\}, M^{\text{prev}}, E^{\mathrm{cand,prev}}, \Lambda)$; 
  \tcp*{Algorithm \ref{alg:clusteringstep_idat}}

  Increment $r \leftarrow r+1$;\;
  Set $K \leftarrow |\mathcal{Y}|$;\;

  \If{$r \ge \Lambda$ \And $K>2$}{
    \tcp{Graph stats for quantile update and active window extraction}
    Count connected components in $\mathcal{G}=(\mathcal{Y},E)$ and set $C$;\;
    Extract the most recent $m=\min(2\Lambda,\,\text{buffer length})$ samples as $W_t=\{\mathbf{x}_{t-m+1},\dots,\mathbf{x}_t\}$;\;

    \tcp{Adjustment of $\Lambda$ and recalculation of $V_{\textup{threshold}}$}
    $(\Lambda^{\text{new}},\, V_{\textup{threshold}}^{\text{new}},\, q^{\text{new}}) \leftarrow 
    \textsc{UpdateLambdaAndVigilance}(W_t, \Lambda, \mathcal{S}, q, C, K)$; \tcp*{Algorithm \ref{alg:lambda_unified_with_vigilance}}

    \tcp{Commit updates and reset recomputation counter}
    Update $\Lambda \leftarrow \Lambda^{\text{new}}$;\;
    Update $V_{\textup{threshold}} \leftarrow V_{\textup{threshold}}^{\text{new}}$;\;
    Update $q \leftarrow q^{\text{new}}$;\;
    Reset counter $r \leftarrow 0$;\;
  }
}

\Return{$\mathcal{Y}, E, E^{\mathrm{cand}}, \Lambda, V_{\textup{threshold}}, \mathcal{S}, \mathcal{M}, q, r$};

\caption{Overall Learning Procedure of IDAT}
\label{alg:fit_idat}
\end{algorithm*}

% Procedure of Clustering Step
\begin{algorithm*}[htbp]
\footnotesize
\DontPrintSemicolon
\KwIn{ \\
  \quad sample $\mathbf{x}$;\\
  \quad carried-over internal state consisting of:\\
  \quad \quad \textit{Topology:} $\mathcal{Y}=\{\mathbf{y}_k\}$ (nodes), $E$ (adjacency matrix), $E^{\mathrm{cand}}$ (edge-candidate matrix);\\
  \quad \quad \textit{Parameters:} $V_{\textup{threshold}}$ (vigilance threshold), $\mathcal{S}=\{\sigma_k\}$ (scaling factors), $\mathcal{M}=\{M_k\}$ (update counts);\\
  \quad \quad \textit{Activity:} $\mathcal{A}=\{a_k\}$ (node activity states, $a_k\in\{0,1\}$), $\{U_k\}$ (inactivity counters);\\
  \quad \quad \textit{Snapshots:} $M^{\text{prev}}$, $E^{\mathrm{cand,prev}}$;\\
  \quad \quad \textit{Configuration:} recalculation interval $\Lambda$.
}
\KwOut{ \\
  \quad updated state: $\mathcal{Y}, E, E^{\mathrm{cand}}, \mathcal{S}, \mathcal{M}, \mathcal{A}, \{U_k\}, M^{\text{prev}}, E^{\mathrm{cand,prev}}$
}

\vspace{2mm}
\tcp{(1) Initial Node Creation}
Set $K \leftarrow |\mathcal{Y}|$;\;
\uIf{$K < 3$}{
  Initialize a new node $\mathbf{y}_{K+1}$ from $\mathbf{x}$ as in (\ref{eq:initnode_y_v2})--(\ref{eq:initnode_s_v2});\;
  Append $\mathbf{y}_{K+1}$ to $\mathcal{Y}$, and its corresponding $M_{K+1}$ and $\sigma_{K+1}$ to $\mathcal{M}$ and $\mathcal{S}$;\;
  Set $a_{K+1}\leftarrow 0$ and $U_{K+1}\leftarrow 0$;\;
  \If{$K+1=2$}{Copy $\sigma_2$ to $\sigma_1$ to stabilize subsequent computations;\;}
  \textbf{return} updated state $(\mathcal{Y}, E, E^{\mathrm{cand}}, \mathcal{S}, \mathcal{M}, \mathcal{A}, \{U_k\}, M^{\text{prev}}, E^{\mathrm{cand,prev}})$;\;
}

% \vspace{1mm}
\tcp{(2) Similarity Computation for Nearest Node Selection}
Compute pairwise distances and similarities as in (\ref{eq:dist}) and (\ref{eq:sim});\;
Sort nodes by ascending $d_k$ and assign the nearest and second-nearest indices $s_1$ and $s_2$;\;

% \vspace{1mm}
\tcp{(3) Vigilance Test and Node Learning}
\uIf{$V_{s_1}<V_{\textup{threshold}}$}{
  \tcp{Case I: Create a new node.}
  Initialize a new node $\mathbf{y}_{K+1}$ from $\mathbf{x}$ as in (\ref{eq:initnode_y_v2})--(\ref{eq:initnode_s_v2});\;
  Set $a_{K+1}\leftarrow 0$ and $U_{K+1}\leftarrow 0$;\;
}
\Else{
  \tcp{Case II: Update existing nodes.}
  Update $M_{s_1}$, $\mathbf{y}_{s_1}$, and $\sigma_{s_1}$ as in (\ref{eq:case2_m})--(\ref{eq:case2_sigma});\;

  \If{$V_{s_2}>V_{\textup{threshold}}$}{
    \tcp{Case III: Form edge candidates.}
    Update $M_{s_2}$ and $\mathbf{y}_{s_2}$ as in (\ref{eq:case3_m}) and (\ref{eq:case3_y});\;
    Compute $T_{\text{active}}$ as in (\ref{eq:case3_Tact});\;
    \If{$M_{s_1}>T_{\text{active}}$ \textbf{and} $M_{s_2}>T_{\text{active}}$}{
      Set $a_{s_1}\leftarrow 1$ and $a_{s_2}\leftarrow 1$;\;
    }
    Update $E^{\mathrm{cand}}_{s_1 s_2}$ as in (\ref{eq:case3_ecand});\;
    Compute $T_{\text{edge}}$ as in (\ref{eq:case3_Tedge});\;
    \If{$E^{\mathrm{cand}}_{s_1 s_2}>T_{\text{edge}}$}{
      Set $E_{s_1 s_2}=E_{s_2 s_1}=1$ as in (\ref{eq:case3_edge});\;
    }
  }
}

% \vspace{1mm}
\tcp{(4) Periodic Topology Maintenance}
\If{the number of processed samples is a multiple of $\Lambda$}{
  \tcp{(i) Edge Pruning}
  Compute edge lengths $\ell_{ij}$ for all connected node pairs;\;
  \If{$\ell_{ij} > Q_3 + 1.5\,\mathrm{IQR}$}{
    Remove the edge from $E$ and reset the corresponding $E^{\mathrm{cand}}_{ij}$;\;
  }

  \tcp{(ii) Node Deletion}
  \If{the number of nodes with $M_k=1$ exceeds $\Lambda$}{
    Retain the latest $\Lambda$ nodes with $M_k=1$ and remove the others together with related data;\;
  }

  \tcp{(iii) Node Demotion}
  Compute the activity potential $P_k$ for each node as in (\ref{eq:activity_potential});\;
  Compute the lower bound $\mathrm{LB} \leftarrow (Q_1 - 1.5\,\mathrm{IQR})$ from positive $\{P_k\}$ values;\;

  \For{$k \leftarrow 1$ \KwTo $K$}{
    \If{$a_k = 1$}{ % active nodes only
      Let $\operatorname{deg}(\mathbf{y}_k)$ be the degree of node $\mathbf{y}_k$ in $E$ after pruning;\;
      \uIf{$\operatorname{deg}(\mathbf{y}_k)=0$ \textbf{and} $P_k < \mathrm{LB}$}{ % isolated and low-activity in this interval
        Increment $U_k \leftarrow U_k + 1$;\;
      }
      \Else{
        Reset $U_k \leftarrow 0$;\;
      }
      \If{$U_k \ge T_{\text{demote}}$}{
        Set $a_k \leftarrow 0$;\; % demote
      }
    }
  }

  Update $M^{\text{prev}}$ and $E^{\mathrm{cand,prev}}$ for the next maintenance interval;\;
}

% \vspace{1mm}
\Return{updated state $\mathcal{Y}, E, E^{\mathrm{cand}}, \mathcal{S}, \mathcal{M}, \mathcal{A}, \{U_k\}, M^{\text{prev}}, E^{\mathrm{cand,prev}}$};

\caption{Procedure of Clustering Step}
\label{alg:clusteringstep_idat}
\end{algorithm*}

\subsection{Clustering Step}
\label{sec:clustering}
This subsection describes the clustering step for a single incoming sample $\mathbf{x}\in\mathbb{R}^d$. Algorithm \ref{alg:clusteringstep_idat} presents the procedure of the clustering step.

\subsubsection*{(1) Initial Node Creation}
If the number of nodes is less than three (i.e., $K=|\mathcal{Y}|<3$), a new node is created from the input. The new node is initialized as
\begin{equation}
\mathbf{y}_{K+1}\leftarrow\mathbf{x},
\label{eq:initnode_y_v2}
\end{equation}
\begin{equation}
M_{K+1}\leftarrow 1,
\label{eq:initnode_m_v2}
\end{equation}
\begin{equation}
\sigma_{K+1}\leftarrow \text{WelfordUpdate}(\mathbf{x}),
\label{eq:initnode_s_v2}
\end{equation}
where Welford’s online variance update provides a scaling factor used for computing similarity between an input and a node. If the number of nodes is two after creating a new node (i.e., $K+1=2$), the scaling factor of the first node (i.e., $\sigma_{1}=0$) is copied from the second, i.e., $\sigma_{1}\leftarrow\sigma_{2}$, to stabilize subsequent computations.

\subsubsection*{(2) Similarity Computation for Nearest Node Selection}
When $K\!\ge\!3$, the algorithm first evaluates the Euclidean distance between the input sample $\mathbf{x}$ and each node $\mathbf{y}_k$ as  
\begin{equation}
d_k = \|\mathbf{x}-\mathbf{y}_k\|_2,
\label{eq:dist}
\end{equation}
and then converts it into an inverse-distance similarity to emphasize the relative closeness between samples and nodes:  
\begin{equation}
V_k = \frac{1}{1+\alpha_k d_k},
\label{eq:sim}
\end{equation}
where $\alpha_k = 1/\sigma_k$ represents the scaling factor of node $\mathbf{y}_k$.

The two nearest nodes are selected according to their Euclidean distances as  
\begin{equation}
s_1 = \arg\min_{k \in \mathcal{K}} d_k,
\label{eq:s1}
\end{equation}
\begin{equation}
s_2 = \arg\min_{k \in \mathcal{K}\setminus\{s_1\}} d_k,
\label{eq:s2}
\end{equation}
where $\mathcal{K}={1,\dots,K}$ denotes the set of all node indices.

Note that the ordering $(s_1,s_2)$ is defined by the Euclidean distances $d_k$, whereas the vigilance test in the next step is based on the inverse-distance similarities $V_k$. The formulation in (\ref{eq:sim}) guarantees that the similarity to each node decreases monotonically with the distance and remains bounded within $(0,1]$ since $\sigma_k$ is positive. This property provides a stable foundation for the vigilance test.

\subsubsection*{(3) Vigilance Test and Node Learning}

The vigilance test and node learning determine the update mechanism of the network based on the current input sample $\mathbf{x}$ and the similarity values $V_{s_1}$ and $V_{s_2}$. The algorithm classifies each input into one of three cases depending on whether the vigilance condition is satisfied for the nearest node $\mathbf{y}_{s_1}$ and the second-nearest node $\mathbf{y}_{s_2}$, where the condition holds when the similarity exceeds the vigilance threshold $V_{\text{threshold}}$.

\begin{itemize}
[
\setlength{\IEEElabelindent}{\dimexpr-\labelwidth-\labelsep}% Wrap text after label
\setlength{\itemindent}{\dimexpr\labelwidth+\labelsep}% Indent subsequent lines
\setlength{\listparindent}{\parindent}% Normal paragraph indentation
]

\item \textbf{Case I} \\
\indent
The vigilance condition is not satisfied by the nearest node $\mathbf{y}_{s_1}$, i.e.,
\begin{equation}
V_{s_1} < V_{\text{threshold}}.
\label{eq:case1_cond_revised}
\end{equation}

Although a scaling factor $\alpha_k$ in (\ref{eq:sim}) may occasionally produce $V_{s_2} \ge V_{\text{threshold}}$ even when $V_{s_1} < V_{\text{threshold}}$, IDAT adopts a nearest-neighbor-first policy in which any input sample whose nearest node fails the vigilance test is treated as novel information. This policy prevents the assignment of the input to a more distant node that appears similar only because $\sigma_{s_2}$ is small. It also reflects the ART principle that node selection (i.e., category choice) and the vigilance test are treated as distinct processes. Consequently, the algorithm creates a new node at the position of the input sample $\mathbf{x}$ in the same manner as in (\ref{eq:initnode_y_v2})-(\ref{eq:initnode_s_v2}), and the internal representation expands as new samples are observed. Case II and Case III apply only when the nearest node $\mathbf{y}_{s_1}$ satisfies the vigilance condition (i.e., $V_{s_1} \ge V_{\text{threshold}}$).

\vspace{2mm}

\item \textbf{Case II} \\ 
\indent 
The vigilance condition is satisfied by the nearest node $\mathbf{y}_{s_1}$ but not by the second-nearest node $\mathbf{y}_{s_2}$, i.e.,
\begin{equation}
V_{s_1} \ge V_{\text{threshold}} > V_{s_2}.
\label{eq:case2_cond}
\end{equation}

In this case, the internal representation of the node $\mathbf{y}_{s_1}$ is updated to incorporate the current input sample $\mathbf{x}$ as  
\begin{equation}
M_{s_1}\leftarrow M_{s_1}+1,
\label{eq:case2_m}
\end{equation}
\begin{equation}
\mathbf{y}_{s_1}\leftarrow \mathbf{y}_{s_1}+\frac{1}{M_{s_1}}(\mathbf{x}-\mathbf{y}_{s_1}),
\label{eq:case2_y}
\end{equation}
\begin{equation}
\sigma_{s_1}\leftarrow \text{WelfordUpdate}(\mathbf{x}).
\label{eq:case2_sigma}
\end{equation}

This operation refines the position and scale of the winning node according to the new input. It allows the model to adjust locally while maintaining a compact set of nodes representing the current data distribution.

\vspace{2mm}

\item \textbf{Case III} \\
\indent 
The vigilance condition is satisfied by both nodes $\mathbf{y}_{s_1}$ and $\mathbf{y}_{s_2}$, i.e.,
\begin{equation}
V_{s_1} > V_{\text{threshold}} \;\text{and}\; V_{s_2} > V_{\text{threshold}}.
\label{eq:case3_cond}
\end{equation}

In this case, the node $\mathbf{y}_{s_1}$ is updated in the same manner as in (\ref{eq:case2_m})--(\ref{eq:case2_sigma}), and the node $\mathbf{y}_{s_2}$ is updated as
\begin{equation}
M_{s_2}\leftarrow M_{s_2}+1,
\label{eq:case3_m}
\end{equation}
\begin{equation}
\mathbf{y}_{s_2}\leftarrow \mathbf{y}_{s_2}+\frac{1}{M_{s_1}+M_{s_2}}(\mathbf{x}-\mathbf{y}_{s_2}).
\label{eq:case3_y}
\end{equation}

In IDAT, a newly created node is initially treated as inactive. Here, both nodes $\mathbf{y}_{s_1}$ and $\mathbf{y}_{s_2}$ are promoted from inactive to active status only when they jointly satisfy
\begin{equation}
M_{s_1} > T_{\text{active}} \;\text{and}\; M_{s_2} > T_{\text{active}},
\label{eq:case3_Tcond}
\end{equation}
where
\begin{equation}
T_{\text{active}} = \mathrm{mean}\{\,M_k \mid M_k > 1\,\}.
\label{eq:case3_Tact}
\end{equation}

When this condition holds, the activation states of $\mathbf{y}_{s_1}$ and $\mathbf{y}_{s_2}$ are updated as
\begin{equation}
a_{s_1} \leftarrow 1 \;\text{and}\; a_{s_2} \leftarrow 1,
\label{eq:activation_update}
\end{equation}
where both nodes become active and are incorporated into the stable topology of the network.

Through continuous presentation of new input samples, these nodes are reinforced and become stable components of the topology. Active nodes constitute the core structure of the topological graph and function as cluster centers that summarize locally consistent data regions. In contrast, inactive nodes ($a_k = 0$) remain candidate elements of the network until they accumulate sufficient evidence to meet the activation condition in~(\ref{eq:case3_Tcond}).

The relational strength between $\mathbf{y}_{s_1}$ and $\mathbf{y}_{s_2}$ is then accumulated in the edge-candidate matrix as
\begin{equation}
E^{\mathrm{cand}}_{s_1 s_2}\leftarrow E^{\mathrm{cand}}_{s_1 s_2} + 1.
\label{eq:case3_ecand}
\end{equation}

When $E^{\mathrm{cand}}_{s_1 s_2}$ exceeds the adaptive edge threshold, i.e.,
\begin{equation}
E^{\mathrm{cand}}_{s_1 s_2} > T_{\text{edge}},
\label{eq:case3_Tcond_edge}
\end{equation}
where
\begin{equation}
T_{\text{edge}} = \mathrm{mean}\{\,E^{\mathrm{cand}}_{ij} \mid E^{\mathrm{cand}}_{ij}>0,\, i,j \in \{1,\dots,K\}\,\}.
\label{eq:case3_Tedge}
\end{equation}

If the condition in (\ref{eq:case3_Tcond_edge}) holds, a bidirectional connection is formed by
\begin{equation}
E_{s_1 s_2}=E_{s_2 s_1}=1.
\label{eq:case3_edge}
\end{equation}

This mechanism reinforces connections between node pairs that are frequently selected together and suppresses spurious links caused by temporary similarities.

All of these process are performed only when both nodes satisfy the vigilance condition (i.e., in Case III).

\end{itemize}

\subsubsection*{(4) Periodic Topology Maintenance}

A periodic maintenance process is performed to reorganize the topology by removing redundant edges and inactive nodes. This process is triggered every $\Lambda$ samples and consists of three stages, namely, edge pruning, node deletion, and node demotion, as summarized below.

\begin{itemize}
[
\setlength{\IEEElabelindent}{\dimexpr-\labelwidth-\labelsep}
\setlength{\itemindent}{\dimexpr\labelwidth+\labelsep+4mm}
\setlength{\listparindent}{\parindent}
]

\item[(i)] \textbf{Edge Pruning} \\ 
\indent 
All existing edges in the adjacency matrix $E$ are examined by their Euclidean lengths:
\begin{equation}
\ell_{ij} = \|\mathbf{y}_i - \mathbf{y}_j\|_2, \quad \text{for all } E_{ij}>0.
\label{eq:edge_length}
\end{equation}

An Interquartile Range (IQR)-based outlier detection is applied to identify abnormally long connections. The IQR, defined as the difference between the first and third quartiles, provides a robust measure of dispersion against extreme values. Let $Q_1$ and $Q_3$ be the first and third quartiles of $\{\ell_{ij}\}$, and define $\mathrm{IQR}=Q_3-Q_1$. An edge is regarded as an outlier if its length exceeds the upper bound defined by the IQR criterion. The upper bound is given by
\begin{equation}
\ell_{ij} > Q_3 + 1.5\,\mathrm{IQR}.
\label{eq:edge_prune}
\end{equation}

The edges satisfying this condition are removed from $E$, and their corresponding counters in $E^{\mathrm{cand}}$ are reset to zero. 
This process prevents structural distortion caused by excessively long links.

\vspace{2mm}

\item[(ii)] \textbf{Node Deletion} \\ 
\indent 
Nodes that have been updated only once ($M_k=1$) are regarded as weakly representative of the current data distribution. When the number of such nodes exceeds the size of $\Lambda$, only the most recent $\Lambda$ nodes are retained, and the remaining ones are removed along with their associated information.

This deletion policy removes unstable or immature nodes and keeps the network compact and stable.

\vspace{2mm}

\item[(iii)] \textbf{Node Demotion} \\ 
\indent 
IDAT preserves an active and meaningful topology by gradually demoting nodes that exhibit both isolation and low activity. The procedure consists of two stages. First, for each node $k$, the activity potential $P_k$ is computed as
\begin{equation}
P_k = (M_k - M^{\text{prev}}_k) + \Bigl(\sum_{j} E^{\mathrm{cand}}_{kj} - \sum_{j} E^{\mathrm{cand,prev}}_{kj}\Bigr),
\label{eq:activity_potential}
\end{equation}
where $M^{\text{prev}}_k$ and $E^{\mathrm{cand,prev}}_{kj}$ denote the snapshots of node usage and candidate edge interactions taken at the previous maintenance interval. This value represents the degree to which node $\mathbf{y}_{s_1}$ has been updated or co-activated during the current interval.

Next, the algorithm identifies nodes that satisfy three conditions: the node remains active ($a_k = 1$), the node has no incident edges, and the activity potential satisfies $P_k \le Q_1 - 1.5\,\mathrm{IQR}$, where the bound is computed from the positive $\{P_k\}$ values. Nodes that satisfy these conditions are regarded as inactive. Each active node holds a tolerance counter $U_k$. The counter increases by one when the node is inactive and is reset to zero when the node shows sufficient activity. A node is demoted (i.e., $a_k$ is updated from $1$ to $0$) when its counter reaches or exceeds the demotion threshold:
\begin{equation}
U_k \ge T_{\text{demote}}.
\label{eq:demotion_cond}
\end{equation}

In this study, $T_{\text{demote}}$ is set to $2$ because the buffer retains samples for a duration of $2\Lambda$, while the maintenance operation is executed once every $\Lambda$ samples. Consequently, a node that stays inactive for two consecutive maintenance intervals is regarded as unresponsive during one full buffer cycle and is therefore demoted. 

This process combines statistical detection through the activity potential with temporal persistence tracking through the tolerance counter. The combination prevents premature demotion caused by short-term inactivity and suppresses the continued presence of nodes that remain dormant across multiple maintenance cycles.

\end{itemize}

Overall, the periodic maintenance processes maintain a compact and adaptive topology that reflects the evolving data stream.

\subsection{Diversity-Driven Adaptation of $\Lambda$ and $V_{\text{threshold}}$}
\label{sec:adjustment}

This subsection describes the diversity-driven adaptation of the recalculation interval $\Lambda$ and the vigilance threshold $V_{\text{threshold}}$. A buffer continually stores up to $2\Lambda$ recent samples, and the algorithm evaluates their geometric consistency through a window of length $L$ $(2 \le L \le 2\Lambda)$. In this work, diversity refers to the numerical stability of the inverse-distance similarity matrix constructed from the window, rather than to the global variability of the data distribution. The stability of this matrix serves as an indicator of the degree to which the recent samples preserve their pairwise distance relationships.

When the environment is nonstationary, samples collected within the same window may originate from different distributions, and the resulting geometric relationships can become distorted. As a consequence, the inverse-distance similarity matrix often becomes nearly singular or poorly conditioned even for moderate window sizes. This numerical instability indicates insufficient diversity in the sense of geometric consistency, and the algorithm decreases the window length $L$ so that diversity is evaluated only from the most recent samples. This adjustment improves responsiveness to distributional changes.

When the environment is stationary, the relative distances among samples remain coherent over time. The similarity matrix therefore remains numerically stable under progressively larger window sizes and passes the associated Cholesky-based stability test. This behavior implies stable geometric relationships, and the algorithm increases $L$ toward $2\Lambda$ to incorporate a broader temporal context and to strengthen the stability of the recalculation interval.

The following subsections first describe the stability-based assessment of diversity, then present the adaptive adjustment of $\Lambda$, and finally explain the computation of $V_{\text{threshold}}$ from the resulting interval.

\subsubsection{Diversity Assessment via Matrix Stability}
\label{sec:stability_condition}

To assess the diversity of buffered samples, the algorithm analyzes their pairwise similarity relationships within the current window of length $L$ $(2 \le L \le 2\Lambda)$. In this subsection, diversity is interpreted as the numerical stability of the similarity matrix, which reflects the consistency of the geometric relationships preserved by the samples.

Within a window $W_t=\{\mathbf{x}_{t-L+1},\ldots,\mathbf{x}_t\}$, the pairwise inverse-distance similarity matrix $S_L \in \mathbb{R}^{L \times L}$ is defined as
\begin{equation}
[S_L]_{ij} = \frac{1}{1+\alpha^\star\lVert\mathbf{x}_{t-L+i}-\mathbf{x}_{t-L+j}\rVert_2},
\label{eq:similarity_matrix}
\end{equation}
where $\alpha^\star = 1/\max_k\sigma_k$ denotes a global scaling factor derived from the largest local scale $\sigma_k$. The function in \eqref{eq:similarity_matrix} is a positive definite kernel \cite{gneiting13}, and evaluating it on a finite set of samples produces a Gram matrix. As a result, $S_L$ is positive semidefinite in exact arithmetic. In practice, however, finite-precision computation and strong sample dependencies may cause $S_L$ to become numerically ill-conditioned.

The Gram matrix acquires a low effective rank when the samples in the window are strongly dependent. This reduction yields eigenvalues that approach zero, and this spectral collapse is a mathematical consequence of limited geometric variation in the window. Eigenvalues of this scale are sensitive to finite-precision arithmetic, so a matrix that is positive semidefinite in exact arithmetic can appear indefinite in numerical computation. This sensitivity motivates the use of a numerical test for the stability of $S_L$. The numerical stability of $S_L$ is evaluated through the Cholesky decomposition as
\begin{equation}
S_L = R R^\top,
\label{eq:chol_decomp}
\end{equation}
where $R$ is an upper triangular factor.

When the decomposition does not succeed, $S_L$ is treated as numerically unstable. When the decomposition succeeds, the matrix may still be close to singular, and the determinant is therefore approximated by
\begin{equation}
\widehat{\det}(S_L) = \left(\prod_{i=1}^{L} R_{ii}\right)^2,
\label{eq:stability_criterion}
\end{equation}
which provides a continuous measure of conditioning.

A window is regarded as diverse when $S_L$ passes the stability test consisting of a successful Cholesky decomposition and the determinant satisfying
\begin{equation}
\widehat{\det}(S_L) \ge \varepsilon_{\det},
\label{eq:diversity_condition}
\end{equation}
where $\varepsilon_{\det}=1.0\times10^{-6}$ separates well-conditioned matrices from nearly singular ones.

\subsubsection{Adaptive Adjustment of $\Lambda$}
\label{sec:lambda_adjustment}
The recalculation interval $\Lambda$ defines the update frequency of the vigilance threshold $V_{\text{threshold}}$. To adapt $\Lambda$ to changing data conditions, the algorithm evaluates diversity within a window of length $L$ $(2 \le L \le 2\Lambda)$ by applying the stability condition in~(\ref{eq:diversity_condition}). In this work, diversity refers to the geometric consistency of recent samples, which is assessed through the stability of the inverse-distance similarity matrix. This notion differs from global variability in the data distribution and instead focuses on the preservation of pairwise distance relationships.

When samples within the window originate from inconsistent distributions, the similarity matrix becomes numerically unstable. This instability indicates insufficient diversity in the geometric sense and characterizes a nonstationary environment. The algorithm therefore reduces the window length $L$ and updates the recalculation interval to improve responsiveness to recent changes (i.e., decremental direction). When the similarity matrix remains stable across progressively larger windows, the geometric structure is interpreted as consistent over time. This situation reflects a stationary environment, and the algorithm expands $L$ to incorporate longer temporal context and to stabilize the recalculation interval (i.e., incremental direction).

\vspace{1mm}
\noindent\textbf{Decremental Direction:}
The algorithm first searches for the largest window length $L$ within $\{2,\ldots,\Lambda\}$ that satisfies the stability condition in~(\ref{eq:diversity_condition}). The optimal window length is defined as
\begin{equation}
L^\star = \max\!\left\{\,L \in \{2,\ldots,\Lambda\} \,\big|\, \widehat{\det}(S_L) \ge \varepsilon_{\det} \,\right\}.
\label{eq:L_star}
\end{equation}

The recalculation interval is updated as
\begin{equation}
\Lambda^{\text{new}} = L^\star.
\label{eq:decremental_rule}
\end{equation}

If all $L$ within $\{2,\Lambda\}$ satisfy the condition, the algorithm proceeds to the incremental direction without reducing $\Lambda$. 

Note that the stability check evaluates prefix windows $L \in [2,\Lambda]$ sequentially at each update step. The method does not assume that all shorter windows are stable, and the decremental step remains valid even when the current $\Lambda$ is unstable.

\vspace{1mm}
\noindent\textbf{Incremental Direction:}
After identifying the largest stable prefix window in the decremental direction, the algorithm evaluates longer windows to examine whether geometric consistency can be maintained over longer spans. The candidate window length $L$ starts from $\Lambda^{\text{new}}+1$ and increases in an exponential manner up to $L_{\max} = \min(2\Lambda,\, |W_t|)$, where $|W_t|$ denotes the number of buffered samples. For each candidate length $L$, the stability condition is assessed on the last $L$ samples in the buffer. Let $L^\star$ denote the largest tested length for which the similarity matrix remains stable. If the stability condition first fails at some length $\bar{L}$, then $L^\star < \bar{L}$ and the recalculation interval is updated as
\begin{equation}
\Lambda^{\text{new}} = L^\star.
\label{eq:incremental_rule}
\end{equation}

If the condition remains valid for all lengths up to $L_{\max}$, then $L^\star = L_{\max}$ and the maximum stable window is adopted:
\begin{equation}
\Lambda^{\text{new}} = L_{\max}.
\label{eq:incremental_rule_max}
\end{equation}

In summary, the recalculation interval $\Lambda$ expands when the similarity matrix remains stable for longer windows and contracts when the matrix becomes unstable for shorter windows.
This adaptive mechanism aligns $\Lambda$ with the time scale over which geometric stability persists. Smaller intervals tend to arise under nonstationary settings, whereas larger intervals emerge when the environment is approximately stationary.

% Diversity assessment by similarity-matrix stability
\begin{algorithm}[htbp]
\footnotesize
\DontPrintSemicolon
\KwIn{ \\
  \quad $W_t=\{\mathbf{x}_{t-L+1},\dots,\mathbf{x}_t\}$ (window of $L$ buffered samples);\;
  \quad $\mathcal{S}=\{\sigma_k\}$ (set of scaling factors for computing $\alpha^\star$)
}
\KwOut{ \\
  \quad $\textit{is\_stable}$ (flag; true if the condition in \eqref{eq:diversity_condition} is satisfied)
}

\vspace{2mm}

Construct $S_L$ as in \eqref{eq:similarity_matrix};\;
Apply the Cholesky decomposition as in \eqref{eq:chol_decomp};\;
\uIf{factorization fails}{
  $\textit{is\_stable} \leftarrow \textbf{false}$;\;
  \textbf{return} $\textit{is\_stable}$;\;
}
Compute $\widehat{\det}(S_L)$ as in \eqref{eq:stability_criterion};\;
Check the diversity condition as in \eqref{eq:diversity_condition};\;
\If{$\widehat{\det}(S_L) \ge \varepsilon_{\det} = 1.0\times10^{-6}$}{
  $\textit{is\_stable} \leftarrow \textbf{true}$;\;
}
\Else{
  $\textit{is\_stable} \leftarrow \textbf{false}$;\;
}

\Return{$\textit{is\_stable}$};\;

\caption{Procedure for Assessing Diversity via Similarity-Matrix Stability}
\label{alg:assess_diversity_matrix_stability}
\end{algorithm}

% Unified Procedure for Adjustment of Lambda and Recalculation of V_threshold
\begin{algorithm}[htbp]
\footnotesize
\DontPrintSemicolon
\KwIn{ \\
  \quad $W_t=\{\mathbf{x}_{t-m+1},\dots,\mathbf{x}_t\}$ (buffered samples; $m$ up to $2\Lambda$);\;
  \quad $\Lambda$ (current recalculation interval);\;
  \quad $\mathcal{S}=\{\sigma_k\}$ (set of scaling factors for computing $\alpha^\star$);\;
  \quad $q^{\text{prev}}$ (previous quantile level);\;
  \quad $C$ (number of connected components);\;
  \quad $K$ (number of nodes)
}
\KwOut{ \\
  \quad $\Lambda^{\text{new}}$ (updated interval), $V_{\textup{threshold}}^{\text{new}}$ (vigilance threshold), $q^{\text{new}}$ (updated quantile level)
}

\vspace{2mm}

\tcp{Decremental Adjustment}
Set $L^\star \leftarrow 0$;\;
Set $\textit{need\_incremental} \leftarrow \textbf{true}$;\;

\For{$L \leftarrow \Lambda$ \KwTo $2$ \KwRet{$-1$}}{
  Extract the last $L$ samples from $W_t$ to form a length-$L$ window, denoted by $W^{(L)}$;\;
  $\textit{is\_stable} \leftarrow \textsc{AssessDiversityByMatrixStability}(W^{(L)}, \mathcal{S})$; \tcp*{Algorithm \ref{alg:assess_diversity_matrix_stability}}
  \If{$\textit{is\_stable}=\textbf{true}$ \textbf{and} $L^\star=0$}{
    Update $L^\star \leftarrow L$;\;
  }
  \If{$\textit{is\_stable}=\textbf{false}$}{
    Update $\textit{need\_incremental} \leftarrow \textbf{false}$;\;
  }
}

\uIf{$L^\star>0$}{
  Update $\Lambda^{\text{new}} \leftarrow L^\star$ as in \eqref{eq:decremental_rule};\;
}
\Else{
  Update $\Lambda^{\text{new}} \leftarrow 2$;\;
}

\tcp{Incremental adjustment}
\If{$\textit{need\_incremental}=\textbf{true}$}{
  Set $L_{\max} \leftarrow \min(2\Lambda,\, |W_t|)$;\;
  Set $\textit{renew} \leftarrow \textbf{false}$;\;
  Set $s \leftarrow 1$;\;
  Set $L_{\text{low}} \leftarrow \Lambda^{\text{new}}$;\;
  Set $L_{\text{high}} \leftarrow \min(L_{\text{low}} + s,\, L_{\max})$;\;

  \While{$L_{\text{high}} \le L_{\max}$ \textbf{and} $\textit{renew}=\textbf{false}$}{
    Extract the last $L_{\text{high}}$ samples from $W_t$ to form a length-$L_{\text{high}}$ window, denoted by $W^{(L_{\text{high}})}$;\;
    $\textit{is\_stable} \leftarrow \textsc{AssessDiversityByMatrixStability}(W^{(L_{\text{high}})}, \mathcal{S})$; \tcp*{Algorithm \ref{alg:assess_diversity_matrix_stability}}
    \If{$\textit{is\_stable}=\textbf{false}$}{
      Update $\Lambda^{\text{new}} \leftarrow L_{\text{low}}$, and $\textit{renew} \leftarrow \textbf{true}$;\;
      \textbf{break}\;
    }
    Update $L_{\text{low}} \leftarrow L_{\text{high}}$;\;
    Update $s \leftarrow \min(2s,\, L_{\max}-\Lambda^{\text{new}})$;\;
    Update $L_{\text{high}} \leftarrow \min(\Lambda^{\text{new}} + s,\, L_{\max})$;\;
  }

  \If{$\textit{renew}=\textbf{false}$}{
    Update $\Lambda^{\text{new}} \leftarrow L_{\max}$;\;
  }
}

\tcp{Recompute Vigilance Parameter by Updated $\Lambda$}
Compute $q^{\text{new}}$ as in \eqref{eq:quantile_level} using $(q^{\text{prev}},\, \Lambda^{\text{new}},\, C,\, K)$;\;
% Construct $S_L$ on the last $\Lambda^{\text{new}}$ samples as in \eqref{eq:similarity_matrix};\;
% Apply the diagonal removal as in \eqref{eq:remove_diag} to obtain $\widetilde{S}_L$;\;
Compute $u^{(L)}$ as in \eqref{eq:rowwise_maxima};\;
Compute $V_{\textup{threshold}}^{\text{new}}$ as in \eqref{eq:quantile_threshold} using $(u^{(L)},\, q^{\text{new}})$;\;

\Return{$\Lambda^{\text{new}},\, V_{\textup{threshold}}^{\text{new}},\, q^{\text{new}}$};

\caption{Procedure for Adjustment of $\Lambda$ and Recalculation of $V_{\textup{threshold}}$}
\label{alg:lambda_unified_with_vigilance}
\end{algorithm}

\subsubsection{Estimating $V_{\text{threshold}}$ from Similarity Matrix Stability}
\label{sec:vigilance_threshold_estimation}

The vigilance threshold $V_{\text{threshold}}$ is estimated from the similarity structure within the current $L$-sample window $W_t=\{\mathbf{x}_{t-L+1},\ldots,\mathbf{x}_t\}$. The similarity matrix $S_L$ is defined as in~(\ref{eq:similarity_matrix}). When the stability condition in~(\ref{eq:diversity_condition}) is not satisfied, the matrix is regarded as nearly singular, which indicates insufficient geometric diversity within the window. When the condition is satisfied, the diagonal elements of $S_L$ are removed and the maximum off-diagonal similarities are extracted as
\begin{equation}
\widetilde{S}_L = S_L - \operatorname{diag}(S_L),
\label{eq:remove_diag}
\end{equation}
\begin{equation}
u^{(L)}_i = \max_{j\neq i} [\widetilde{S}_L]_{ij}, \quad (i=1,\ldots,L).
\label{eq:rowwise_maxima}
\end{equation}
where $u^{(L)}_i$ represents the maximum similarity between sample $i$ and all other samples.

The vigilance threshold is determined by
\begin{equation}
V_{\text{threshold}}^{\text{new}} = \operatorname{Quantile}(u^{(L)},\,q),
\label{eq:quantile_threshold}
\end{equation}
where $q$ denotes the quantile level.

The quantile level $q$ is computed from the connectivity of the current topological graph. Let $C$ and $K$ denote the number of connected components and nodes, respectively. The ratio $C/K$ is incorporated through exponential smoothing as
\begin{equation}
q = 1 -\left[\left(1-\frac{1}{\Lambda}\right)q^{\text{prev}} + \frac{1}{\Lambda}\cdot\frac{C}{K}\right],
\label{eq:quantile_level}
\end{equation}
where $q^{\text{prev}}$ is the previous quantile level. A smaller interval $\Lambda$ yields faster smoothing and therefore adapts the quantile to recent structural changes. A larger interval slows the update and incorporates a broader temporal context.

This adaptive update connects the topological connectivity of the graph to the vigilance threshold. Higher connectivity (i.e., when $C/K$ is small) raises the threshold and yields finer structural resolution. Lower connectivity (i.e., when $C/K$ is large) decreases the threshold and stabilizes the learning process under stationary settings. It should be noted that this connectivity-based adjustment is complementary to the diversity-driven update of $\Lambda$, where diversity reflects the stability of the similarity matrix rather than the variability of the data distribution. Through this mechanism, the vigilance threshold adjusts to the evolving topology and maintains coherent cluster formation across changing environments.

\subsection{Prediction Procedure}
\label{sec:prediction}
The prediction procedure assigns a cluster label to an input sample based on the learned topological graph. Given a test sample $\mathbf{x}\in\mathbb{R}^d$, the algorithm identifies the most similar node within the learned network and assigns the corresponding cluster index as the predicted label, as summarized in Algorithm~\ref{alg:predict_idat}.

% Predict Procedure (single-sample version)
\begin{algorithm}[htbp]
\footnotesize
\DontPrintSemicolon
\KwIn{ \\
  \quad sample $\mathbf{x}\in\mathbb{R}^d$;\\
  \quad carried-over internal state consisting of:\\
  \quad\quad \textit{Topology:} $\mathcal{Y}=\{\mathbf{y}_k\}_{k=1}^{K}$ (nodes), $E$ (adjacency matrix);\\
  \quad\quad \textit{Parameters:} $\mathcal{S}=\{\sigma_k\}_{k=1}^{K}$ (scaling factors, $\alpha_k=1/\sigma_k$);\\
  \quad\quad \textit{Activity:} $\mathcal{A}=\{a_k\}_{k=1}^{K}$ (node activity states, $a_k\in\{0,1\}$)
}
\KwOut{\\
  \quad predicted label $\widehat{c}\in\mathbb{N}$
}

\vspace{2mm}
\tcp{(1) Graph-to-cluster mapping}
Construct the topological graph $\mathcal{G}=(\mathcal{Y},E)$;\;
Identify connected components in $\mathcal{G}$ and assign cluster indices $\mathbf{z}_k$ to nodes $\mathbf{y}_k$;\;

\vspace{1mm}
\tcp{(2) Active node selection}
\uIf{all nodes are inactive ($\sum_{k=1}^{K}a_k=0$)}{
  Set $\mathcal{I}\leftarrow\{1,\dots,K\}$;\;
}
\Else{
  Set $\mathcal{I}\leftarrow\{\,k\mid a_k=1\,\}$;\;
}

\vspace{1mm}
\tcp{(3) Similarity-based matching and label assignment}
Compute inverse-distance similarities $S_k$ between $\mathbf{x}$ and $\mathbf{y}_k$ for all $k\in\mathcal{I}$ as defined in~(\ref{eq:sim});\;
Find the nearest (best-matching) node $w \leftarrow \arg\max_{k\in\mathcal{I}} S_k$;\;
Assign the predicted label $\widehat{c} \leftarrow \mathbf{z}_w$;\;

\Return{$\widehat{c}$}

\caption{Label Assignment Procedure of IDAT}
\label{alg:predict_idat}
\end{algorithm}

The topological graph $\mathcal{G}=(\mathcal{Y},E)$ is reconstructed from the current node set $\mathcal{Y}$ and adjacency matrix $E$. Each node $\mathbf{y}_k$ is assigned to a cluster index $\mathbf{z}_k$ according to the connected component it belongs to. The algorithm then selects the active node indices $\mathcal{I}=\{k\mid a_k=1\}$ from the activity states $\mathcal{A}=\{a_k\}_{k=1}^{K}$. If all nodes are inactive, all existing nodes are used as a fallback to ensure label assignment. Finally, the input sample $\mathbf{x}$ is compared with all nodes $\{\mathbf{y}_k\}_{k\in\mathcal{I}}$ using the similarity measure defined in~(\ref{eq:sim}), and the label $\widehat{c}$ corresponding to the most similar (best-matching) node is returned as the prediction.

\subsection{Computational Complexity}
\label{sec:complexity}

The computational complexity of IDAT is analyzed with respect to three major procedures in Algorithm~\ref{alg:fit_idat}, which correspond to the clustering step, topological maintenance, and adaptive adjustment of $\Lambda$ and $V_{\text{threshold}}$. Let $N$ denote the total number of samples, and let $d$, $K$, $\Lambda$, and $|E|$ denote the sample dimension, the number of nodes in the network, the recalculation interval, and the number of edges in the topological graph, respectively.

In the clustering step, each sample $\mathbf{x}$ is compared with all nodes $\mathcal{Y}$ to compute Euclidean distances and similarities. This process requires $\mathcal{O}(K d)$ per sample. The node position and scale updates are performed in constant time.

Topology maintenance is executed every $\Lambda$ samples to update the topological structure. Edge pruning requires $\mathcal{O}(|E| d)$, and node deletion and demotion scale linearly with $K$. As a result, the average computational cost per sample for this process is $\mathcal{O}(|E| d/\Lambda + K/\Lambda)$. 

The adaptive adjustment procedure recalculates $\Lambda$ and $V_{\text{threshold}}$ after every $\Lambda$ samples. The construction of the similarity matrix and the Cholesky decomposition require $\mathcal{O}(\Lambda^2 d + \Lambda^3)$ per update. Accordingly, the average computational cost per sample is $\mathcal{O}(\Lambda d + \Lambda^2)$.

By combining these components, the computational cost per sample can be written as
\begin{equation}
\mathcal{O}\!\left(\max\{\,K d,\; |E| d/\Lambda,\; \Lambda d,\; \Lambda^2\,\}\right).
\end{equation}

Thus, the total computational cost for $N$ samples is
\begin{equation}
\mathcal{O}\!\left(N \cdot \max\{\,K d,\; |E| d/\Lambda,\; \Lambda d,\; \Lambda^2\,\}\right).
\end{equation}

In sparse-graph settings where the number of edges grows proportionally to the number of nodes, the dominant term is $\mathcal{O}(K d)$. This result indicates that the algorithm scales linearly with both $K$ and $d$. Therefore, IDAT maintains efficient learning performance under continual learning conditions.

\section{Simulation Experiments}
\label{sec:experiment}

This section presents the detailed experimental design, covering the datasets, hyperparameter settings, and evaluation metrics. The proposed IDAT is evaluated in both stationary and nonstationary settings in comparison with state-of-the-art algorithms across various clustering paradigms.

\subsection{Dataset}
\label{sec:datasetReal}

The experiments use 24 real-world datasets available from public repositories \cite{derrac15, dua19}, including a GitHub repository. Table~\ref{tab:datasets} summarizes the statistics of the datasets used in the experiments. Each dataset differs in dimensionality, sample size, and the number of classes, covering both low- and high-dimensional clustering scenarios.

Among the image-based datasets, MNIST10K and STL10 are unique in using high-level and high-dimensional representations from pre-trained convolutional networks instead of raw features. MNIST10K contains 10,000 handwritten-digit images from 10 categories, and 4,096 features are extracted from a pre-trained CNN \cite{Jaderberg14}. STL10 consists of 13,000 natural images from 10 object categories, and each image is represented by 2,048 features obtained from the average pooling layer of a ResNet-50 model pre-trained on ImageNet \cite{sarfraz19}. Further details can be found in \cite{sarfraz19, jiang17}.

\subsection{Compared Algorithms}
\label{sec:comparedAlgorithms}
The experiments involve 11 clustering algorithms categorized according to their methodological characteristics. 

GNG~\cite{fritzke95} and SOINN+~\cite{wiwatcharakoses20} represent self-organizing neural approaches that incrementally expand their networks in response to newly observed data. These models adapt their structures to capture evolving data manifolds, and SOINN+ further introduces adaptive thresholds and noise suppression for stable incremental learning. 

Distributed Dual Vigilance Fuzzy ART (DDVFA)~\cite{da20}, CAEA~\cite{masuyama22a}, TBM-P~\cite{yelugam23}, and CAE~\cite{masuyama23} are ART-based algorithms that enable continual learning without catastrophic forgetting by employing the resonance learning mechanism controlled by a vigilance parameter. DDVFA employs dual vigilance parameters to independently regulate data quantization and inter-cluster similarity. CAEA and CAE employ the CIM as a robust similarity measure for high-dimensional data, and TBM-P introduces a biclustering mechanism to represent more complex topological relationships within the ART framework. Note that DDVFA and TBM-P normalize all attributes to the range $[0,1]$ due to the requirement of their base model (i.e., Fuzzy ART~\cite{carpenter91b}).

\begin{table}[htbp]
  \centering
  \renewcommand{\arraystretch}{1.2}
  \caption{Statistics of real-world datasets}
  \label{tab:datasets}
  \begin{tabular}{l|c|c|c}
    \hline\hline
    Dataset           & \# Samples      & \# Attributes     & \#Classes  \\
    \hline
    Iris           & 150      & 4     & 3  \\
    Seeds          & 210      & 7     & 3  \\
    Dermatology    & 366      & 34    & 6  \\
    Pima           & 768      & 8     & 2  \\
    Mice Protein   & 1,077    & 77    & 8  \\
    Binalpha       & 1,404    & 320   & 36 \\
    Yeast          & 1,484    & 8     & 10 \\
    Semeion        & 1,593    & 256   & 10 \\
    MSRA25         & 1,799    & 64    & 12 \\
    Image Segmentation     & 2,310    & 19    & 7  \\
    Rice           & 3,810    & 7     & 2  \\
    TUANDROMD      & 4,464    & 241   & 2  \\
    Phoneme        & 5,404    & 5     & 2  \\
    Texture        & 5,500    & 40    & 11 \\
    OptDigits      & 5,620    & 64    & 10 \\
    Statlog        & 6,435    & 36    & 6  \\
    Anuran Calls   & 7,195    & 22    & 4  \\
    Isolet         & 7,797    & 617   & 26 \\
    MNIST10K       & 10,000   & 4,096 & 10 \\
    PenBased       & 10,992   & 16    & 10 \\
    STL10          & 13,000   & 2,048 & 10 \\
    Letter         & 20,000   & 16    & 26 \\
    Shuttle        & 58,000   & 9     & 7  \\
    Skin           & 245,057  & 3     & 2  \\
    \hline\hline
  \end{tabular}
  \footnotesize
  \begin{minipage}{\linewidth}
    \vspace{1mm}
    \setlength{\leftskip}{4mm}
      Anuran Calls dataset uses only the {\it Family} label. \\
      % Image Seg. stands for Image Segmentation. \\
  \end{minipage}
  \vspace{-5mm}
\end{table}

SR-PCM-HDP~\cite{hu25} and TC~\cite{yang25} belong to density- and graph-based clustering paradigms, respectively. SR-PCM-HDP identifies cluster centers from local density peaks and regulates membership boundaries to improve robustness to outliers. TC detects natural cluster boundaries in similarity graphs by defining a torque measure between connected nodes without specifying the number of clusters in advance. 

TableDC~\cite{rauf25} and G-CEALS~\cite{rabbani25} are deep clustering methods designed for tabular data. Both methods employ autoencoders to learn latent feature representations that improve clustering performance. TableDC uses Mahalanobis distance and a Cauchy distribution for noise-robust feature correlation. G-CEALS employs Gaussian mixture regularization in the latent space to enhance cluster separability. Both methods adopt a fully-connected autoencoder architecture of $d$-500-500-2000-10-2000-500-500-$d$ with ReLU activations, where $d$ is the input dimension. Further implementation details are provided in Section 2 of the supplementary materials.

Among these algorithms, SR-PCM-HDP, TC, TBM-P, TableDC, and G-CEALS are evaluated only under the stationary setting because they are designed for batch or static data processing, whereas GNG, SOINN+, DDVFA, CAEA, and CAE are evaluated in both stationary and nonstationary settings. The proposed IDAT is applicable to both stationary and nonstationary settings.

The source codes of SOINN+\footnote{\url{https://osf.io/6dqu9/}}, 
CAEA\footnote{\url{https://github.com/Masuyama-lab/HCAEA}}, 
CAE\footnote{\url{https://github.com/Masuyama-lab/CAE}}, 
SR-PCM-HDP\footnote{\url{https://github.com/windflowerHu/SRPCMHDP}}, 
TC\footnote{\url{https://github.com/JieYangBruce/TorqueClustering}}, 
TableDC\footnote{\url{https://github.com/hafizrauf/TableDC}}, 
and G-CEALS\footnote{\url{https://github.com/mdsamad001/G-CEALS---Deep-Clustering-for-Tabular-Data}} 
are provided by the authors of the original papers. 
The source code of the proposed IDAT is available on GitHub\footnote{\url{https://github.com/Masuyama-lab/IDAT}}.

\subsection{Hyperparameter Settings}
\label{sec:params}
Among the compared algorithms, TC, SOINN+, CAE, and the proposed IDAT do not require any explicit hyperparameter tuning, as they automatically adjust their internal parameters during learning.

In contrast, SR-PCM-HDP, TBM-P, GNG, DDVFA, and CAEA include several data-dependent hyperparameters that significantly affect their clustering performance. For these algorithms, the hyperparameters are optimized using the \texttt{bayesopt} function provided in the Statistics and Machine Learning Toolbox of MATLAB\footnote{\url{https://www.mathworks.com/help/stats/bayesopt.html}}. In both stationary and nonstationary settings, Bayesian optimization is performed for 25 iterations to maximize the average Adjusted Rand Index (ARI)~\cite{hubert85} obtained over 10 independent runs with different random seeds. In each run, the model is trained and evaluated using the parameter set proposed by the optimizer, and the mean ARI over 10 runs serves as the objective value. Other aspects of the experimental procedure, including the optimization protocol and evaluation criteria, are kept identical across the two settings.

Since TC, SOINN+, CAE, and IDAT learn without prior hyperparameter specification, they require no external optimization and run more efficiently than algorithms that rely on parameter tuning. If Bayesian optimization for parameter tuning does not complete 25 iterations within 12 hours, the corresponding model is regarded as infeasible for that dataset, and its result is reported as N/A. In addition, GNG, DDVFA, and CAE tend to generate an excessive number of nodes depending on the hyperparameter settings, and thus their maximum number of nodes is limited to 2,000.

% list the hyperparameters
\stablabel{st:srpcmdp-param-def}
\stablabel{st:tbm-param-def}
\stablabel{st:gng-param-def}
\stablabel{st:ddvfa-param-def}
\stablabel{st:caea-param-def}
% optimized hyperparameter values
% stationary
\stablabel{st:srpcmdp-stationary}
\stablabel{st:tbm-stationary}
\stablabel{st:gng-stationary}
\stablabel{st:ddvfa-stationary}
\stablabel{st:caea-stationary}
% nonstationary
\stablabel{st:gng-nonstationary}
\stablabel{st:ddvfa-nonstationary}
\stablabel{st:caea-nonstationary}

Due to page limitations, detailed parameter lists and their optimized values are shown in Tables \stabref{st:srpcmdp-param-def}-\ref{st:caea-nonstationary} of the supplementary materials. Tables \stabref{st:srpcmdp-param-def}-\stabref{st:caea-param-def} list the hyperparameters optimized for SR-PCM-HDP, TBM-P, GNG, DDVFA, and CAEA. Tables~\stabref{st:srpcmdp-stationary}-\ref{st:caea-stationary} present the optimized hyperparameters of these algorithms under the stationary setting, and Tables~\ref{st:gng-nonstationary}-\ref{st:caea-nonstationary} show those obtained under the nonstationary setting. The optimized parameter values vary across datasets under both stationary and nonstationary settings. These results suggest that the clustering performance of these algorithms strongly depends on their parameter specifications. Therefore, algorithms that work well without careful parameter tuning have a clear advantage because they can maintain high clustering performance without additional optimization.

\subsection{Evaluation Metrics}
\label{sec:metrics}

% Short version =======================
To evaluate clustering performance, we employ the ARI~\cite{hubert85} and the Adjusted Mutual Information (AMI)~\cite{vinh10}. Both metrics quantify the degree of agreement between the obtained clusters and the ground-truth classes while correcting for the similarity that may occur by chance under random labeling. This chance adjustment allows fair comparison even when the number of clusters differs across methods. The ARI takes values between $-0.5$ and $1$, and the AMI ranges from $0$ to $1$. In both metrics, a value of $1$ indicates perfect agreement between partitions, whereas their lowest values mean random labeling. Both metrics represent complementary aspects of clustering agreement. The ARI emphasizes pairwise consistency between samples and is sensitive to over-segmentation or merging. The AMI quantifies the information shared between partitions and is robust to differences in cluster size and number.

In general, the ARI and AMI evaluate the final clustering performance. In continual learning scenarios, however, it is also necessary to maintain previously acquired knowledge and prevent forgetting in addition to achieving high final clustering performance.

Average Incremental (AI) performance evaluates the mean performance of an algorithm over all incremental learning stages. It reflects both the acquisition of new classes and the retention of previously learned ones. For a class-incremental setting with $C$ classes, let $M_{c}$ denote the performance after training on the $c$-th class and evaluating on all classes learned up to that point. The AI is defined as
\begin{equation}
\text{AI} = \frac{1}{C}\sum_{c=1}^{C} M_{c}.
\end{equation}

Backward Transfer (BWT) measures how performance on previously learned classes changes after training on subsequent classes. It quantifies the extent of forgetting or improvement caused by later updates. Let $R_{i,j}$ denote the performance on class $i$ after training up to class $j$ $(j \geq i)$. Then, BWT is defined as
\begin{equation}
\text{BWT} = \frac{1}{C-1}\sum_{i=1}^{C-1}\big(R_{i,C} - R_{i,i}\big)
\end{equation}
where, $R_{i,i}$ represents the performance immediately after learning class $i$, and $R_{i,C}$ represents the final performance on the same class after all training.  
A negative BWT indicates forgetting, whereas a positive value implies that subsequent learning has improved earlier knowledge.

In this study, ARI and AMI are used as the base performance measures for computing AI and BWT. Accordingly, the results are reported as AI-ARI, AI-AMI, BWT-ARI, and BWT-AMI. The average values of AI and BWT are calculated over multiple runs with a random order of classes. 

In addition to evaluating clustering performance, it is also important to assess the number of clusters generated by each algorithm. This assessment requires a measure of the difference from the true number of classes. For this purpose, the cluster error is used, and it is defined as
\begin{equation}
\label{eq:clusterError}
\mathrm{Cluster\ Error} = \frac{\lvert \text{\#Clusters} - \text{\#Classes} \rvert}{\text{\#Classes}}.
\end{equation}

This metric represents the proportional difference between the number of clusters and the true number of classes. Smaller values correspond to a smaller discrepancy between these quantities. Note that an error value of zero does not guarantee a one-to-one correspondence between clusters and classes, because the metric evaluates numerical deviation rather than semantic alignment. A complete assessment of clustering performance therefore requires additional measures such as ARI and AMI, which quantify the agreement between cluster assignments and true class labels.

\subsection{Comparisons of Clustering Performance}
\label{sec:comparisons}

The proposed IDAT is compared with state-of-the-art approaches across diverse clustering paradigms, including density-based, graph-based, ART-based, and deep learning-based methods.
The clustering performance is evaluated in both stationary and nonstationary settings. Since the clustering process is unsupervised, the same dataset is used for both training and testing, and class labels are used only for evaluating clustering performance. The results are averaged over 30 independent runs with different random seeds to ensure fair and reliable comparisons. All experiments are conducted using MATLAB 2025b on an Apple M2 Ultra processor with 128GB of RAM running macOS.

\subsubsection{Stationary Setting}
\label{sec:stationary}

This section compares the proposed IDAT with the 11 algorithms listed in Section~\ref{sec:comparedAlgorithms}. In the stationary setting, all data are assumed to originate from a fixed distribution, and the entire dataset is available for training. The ARI and AMI are adopted as evaluation metrics to evaluate the fundamental clustering performance of each algorithm.

\stablabel{st:results_stationary}
Due to page limitations, the results of ARI and AMI are provided in Table \stabref{st:results_stationary} of the supplementary materials. This section presents the results of statistical significance tests based on the average ARI and AMI reported in Table \stabref{st:results_stationary}. For statistical comparisons, the Friedman test and the Nemenyi post-hoc analysis \cite{demvsar06} are used. The Friedman test examines the null hypothesis that all algorithms perform equally. If the null hypothesis is rejected, the Nemenyi post-hoc test is conducted for all pairwise algorithm comparisons based on the ranks of the performance metrics across all datasets. A difference in performance between two algorithms is considered statistically significant when the $p$-value from the Nemenyi post-hoc test is below the significance level. In this study, the significance level is set to $0.05$ for both the Friedman test and the Nemenyi post-hoc analysis.

\begin{figure}[htbp]
  \centering
  \subfloat[ARI]{%
    \includegraphics[width=2.8in]{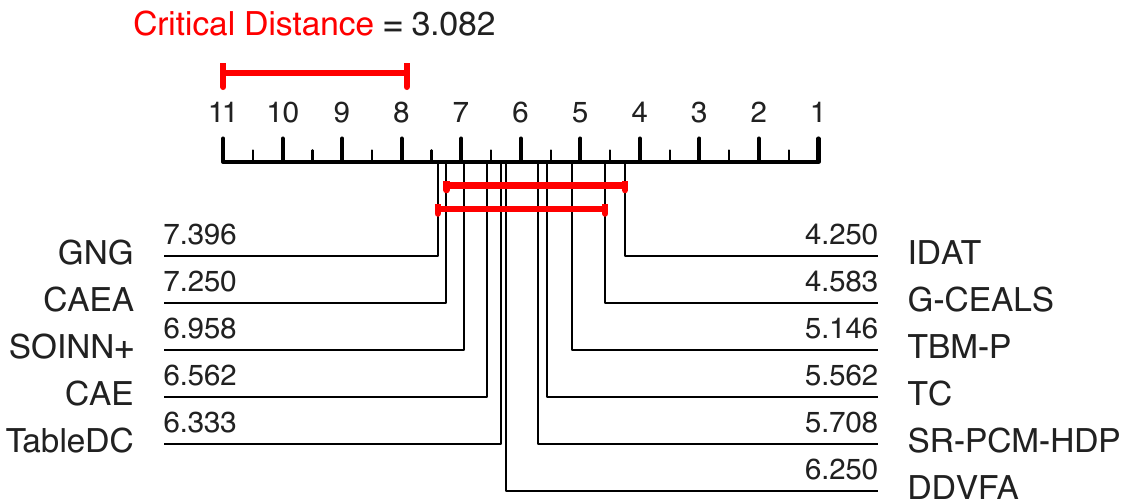}
    \label{fig:CD_ARI_stationary_p008}
  } \\
  \subfloat[AMI]{%
    \includegraphics[width=2.8in]{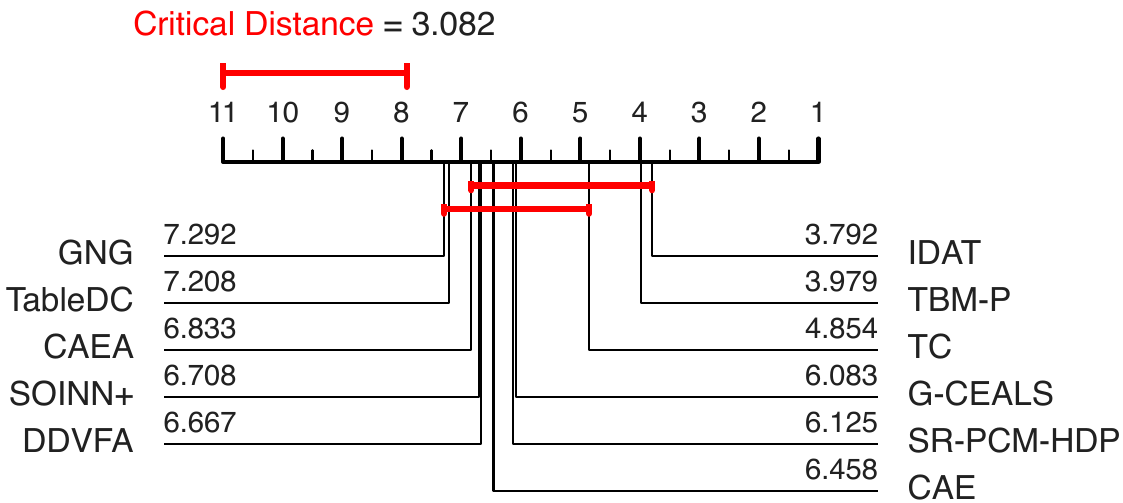}
    \label{fig:CD_AMI_stationary_p000}
  }
  % \vspace{2mm}
  \caption{Critical difference diagrams based on the average ARI and AMI in the stationary setting.}
  \label{fig:cd_stationary_all}
  \vspace{-3mm}
\end{figure}

Fig.~\ref{fig:cd_stationary_all} shows critical difference diagrams based on the average ARI and AMI across all datasets in the stationary setting. As shown in Fig.~\ref{fig:cd_stationary_all}, IDAT achieves the lowest (i.e., best) average ranks for both metrics. Although several algorithms (e.g., G-CEALS and TBM-P) attain comparable ranks within the critical distance, their clustering performance depends heavily on data-dependent hyperparameter optimization or pretraining. In contrast, IDAT provides high-quality clustering results (i.e., precise and stable partitioning) across datasets without any optimization procedure. These observations indicate that IDAT is capable of robust and practical clustering, demonstrating both effectiveness and efficiency under the stationary setting.

In clustering, the ability to estimate the number of clusters is also an important aspect of performance. Among the algorithms compared in this section, TBM-P, TC, GNG, SOINN+, DDVFA, CAEA, CAE, and IDAT can automatically determine the number of clusters without specifying the true number as a hyperparameter. Here, the number of clusters generated by these algorithms after training is compared to evaluate their cluster estimation capability. Note that TBM-P, GNG, SOINN+, CAEA, CAE, and IDAT represent clusters as connected components of nodes in the topology, where each node corresponds to a local region of the data. In contrast, TC and DDVFA handle clusters directly without explicitly maintaining node representations.

Table~\ref{tab:NodesClustersStationary} summarizes the average number of nodes and clusters obtained by each algorithm in the stationary setting. In addition, the cluster error as in (\ref{eq:clusterError}) is also provided and reported as an average over all 24 real-world datasets.

Several algorithms produce far more clusters than the true number of classes, and their average cluster error becomes large. IDAT shows a different tendency. The number of clusters in IDAT remains close to the true number of classes, and the number of nodes stays compact relative to the training sample size. This tendency indicates that IDAT avoids over-segmentation of the data space and represents the data distribution through a topological structure (i.e., nodes and edges).

GNG and SOINN+ often produce a moderate number of clusters because of their self-organizing mechanisms. However, these algorithms generate a large number of nodes in several datasets, which reduces computational efficiency and complicates the resulting topological structure. TBM-P, DDVFA, CAEA, and CAE often produce substantially more clusters than the true number of classes, which results in over-segmentation. TC succeeds in a subset of datasets by detecting structural boundaries in similarity graphs, but its performance decreases when the assumptions behind the torque measure do not hold.

The overall results indicate that IDAT achieves superior clustering performance to that of state-of-the-art algorithms in the stationary setting. IDAT attains high ARI and AMI scores while generating a number of clusters that remains close to the true number of classes across a variety of datasets.

% Average number of nodes and clusters in the stationary setting
\begin{table*}[htbp]
\centering
\caption{Average number of nodes and clusters in the stationary setting}
\label{tab:NodesClustersStationary}
\footnotesize
\renewcommand{\arraystretch}{1.1}
\scalebox{0.8}{
\begin{tabular}{ll|*{7}{C{17mm}|}C{17mm}}  \hline\hline
Dataset & Metric & TBM\textnormal{-}P & TC & GNG & SOINN+ & DDVFA & CAEA & CAE & IDAT \\ \hline

Iris & \#Nodes
& \textnormal{5.0 (0.2)} & \textnormal{--} & \textnormal{50.8 (0.9)} & \textnormal{24.9 (7.0)} 
& \textnormal{--} & \textnormal{40.4 (2.7)} & \textnormal{62.7 (28.3)} & \textnormal{8.8 (1.5)} \\
(\#Classes: 3) & \#Clusters
& \textnormal{5.0 (0.2)} & \textnormal{2.0 (0.0)} & \textnormal{4.6 (2.2)} & \textnormal{9.3 (3.8)} 
& \textnormal{2.2 (0.5)} & \textnormal{25.7 (3.3)} & \textnormal{48.1 (28.7)} & \textnormal{5.6 (1.4)} \\
\hline

Seeds & \#Nodes
& \textnormal{4.6 (1.0)} & \textnormal{--} & \textnormal{32.0 (0.0)} & \textnormal{29.2 (9.1)} 
& \textnormal{--} & \textnormal{54.5 (5.1)} & \textnormal{187.4 (14.7)} & \textnormal{12.1 (2.4)} \\
(\#Classes: 3) & \#Clusters
& \textnormal{3.7 (0.9)} & \textnormal{7.0 (0.0)} & \textnormal{1.0 (0.2)} & \textnormal{7.9 (5.3)} 
& \textnormal{13.7 (1.4)} & \textnormal{38.8 (2.1)} & \textnormal{179.2 (19.8)} & \textnormal{6.8 (1.9)} \\
\hline

Dermatology & \#Nodes
& \textnormal{8.8 (0.4)} & \textnormal{--} & \textnormal{54.0 (0.0)} & \textnormal{46.8 (14.3)} 
& \textnormal{--} & \textnormal{4.7 (4.7)} & \textnormal{285.7 (79.9)} & \textnormal{12.7 (2.9)} \\
(\#Classes: 6) & \#Clusters
& \textnormal{8.8 (0.4)} & \textnormal{10.0 (0.0)} & \textnormal{1.0 (0.2)} & \textnormal{16.9 (7.1)} 
& \textnormal{13.7 (1.8)} & \textnormal{1.2 (0.6)} & \textnormal{253.7 (90.5)} & \textnormal{6.1 (2.0)} \\
\hline

Pima & \#Nodes
& \textnormal{7.5 (0.6)} & \textnormal{--} & \textnormal{110.8 (0.4)} & \textnormal{53.9 (11.7)} 
& \textnormal{--} & \textnormal{312.8 (9.0)} & \textnormal{300.6 (160.0)} & \textnormal{13.1 (3.4)} \\
(\#Classes: 2) & \#Clusters
& \textnormal{7.5 (0.6)} & \textnormal{5.0 (0.0)} & \textnormal{1.2 (0.4)} & \textnormal{11.6 (4.1)} 
& \textnormal{112.8 (5.6)} & \textnormal{234.9 (9.1)} & \textnormal{218.1 (161.8)} & \textnormal{6.1 (1.7)} \\
\hline

Mice Protein & \#Nodes
& \textnormal{113.3 (3.6)} & \textnormal{--} & \textnormal{35.6 (0.6)} & \textnormal{76.7 (18.6)} 
& \textnormal{--} & \textnormal{66.0 (0.0)} & \textnormal{876.4 (282.1)} & \textnormal{21.4 (3.9)} \\
(\#Classes: 8) & \#Clusters
& \textnormal{105.0 (4.6)} & \textnormal{15.0 (0.0)} & \textnormal{1.6 (0.8)} & \textnormal{15.5 (7.5)} 
& \textnormal{374.8 (4.9)} & \textnormal{66.0 (0.0)} & \textnormal{843.8 (313.9)} & \textnormal{9.3 (2.1)} \\
\hline

Binalpha & \#Nodes
& \textnormal{165.9 (3.9)} & \textnormal{--} & \textnormal{235.9 (0.3)} & \textnormal{77.6 (24.8)} 
& \textnormal{--} & \textnormal{63.2 (8.3)} & \textnormal{312.6 (159.9)} & \textnormal{27.5 (14.9)} \\
(\#Classes: 36) & \#Clusters
& \textnormal{132.9 (6.2)} & \textnormal{327.0 (0.0)} & \textnormal{1.0 (0.0)} & \textnormal{39.9 (13.4)} 
& \textnormal{224.3 (11.5)} & \textnormal{16.7 (4.4)} & \textnormal{213.0 (135.2)} & \textnormal{9.6 (5.9)} \\
\hline

Yeast & \#Nodes
& \textnormal{47.7 (2.7)} & \textnormal{--} & \textnormal{484.0 (4.0)} & \textnormal{61.4 (16.0)} 
& \textnormal{--} & \textnormal{18.8 (3.5)} & \textnormal{230.9 (83.4)} & \textnormal{32.3 (4.4)} \\
(\#Classes: 10) & \#Clusters
& \textnormal{34.6 (4.3)} & \textnormal{2.0 (0.0)} & \textnormal{21.3 (4.7)} & \textnormal{5.6 (2.7)} 
& \textnormal{27.7 (2.6)} & \textnormal{1.9 (1.0)} & \textnormal{100.8 (65.4)} & \textnormal{9.2 (2.3)} \\
\hline

Semeion & \#Nodes
& \textnormal{157.4 (3.6)} & \textnormal{--} & \textnormal{160.2 (0.4)} & \textnormal{85.2 (22.2)} 
& \textnormal{--} & \textnormal{117.0 (0.0)} & \textnormal{652.7 (8.7)} & \textnormal{39.2 (21.2)} \\
(\#Classes: 10) & \#Clusters
& \textnormal{138.7 (6.3)} & \textnormal{347.0 (0.0)} & \textnormal{1.0 (0.2)} & \textnormal{44.8 (12.7)} 
& \textnormal{1427.3 (4.8)} & \textnormal{117.0 (0.0)} & \textnormal{498.2 (10.6)} & \textnormal{18.7 (11.7)} \\
\hline

MSRA25 & \#Nodes
& \textnormal{82.1 (4.1)} & \textnormal{--} & \textnormal{86.8 (0.4)} & \textnormal{140.1 (27.6)} 
& \textnormal{--} & \textnormal{8.2 (2.8)} & \textnormal{484.9 (202.1)} & \textnormal{39.4 (10.7)} \\
(\#Classes: 12) & \#Clusters
& \textnormal{69.1 (5.7)} & \textnormal{24.0 (0.0)} & \textnormal{4.1 (1.3)} & \textnormal{43.7 (10.5)} 
& \textnormal{66.5 (4.4)} & \textnormal{2.1 (1.4)} & \textnormal{327.5 (176.6)} & \textnormal{23.4 (7.1)} \\
\hline

Image Seg. & \#Nodes
& \textnormal{85.8 (4.8)} & \textnormal{--} & \textnormal{331.1 (0.8)} & \textnormal{94.3 (31.2)} 
& \textnormal{--} & \textnormal{51.0 (0.0)} & \textnormal{194.0 (62.2)} & \textnormal{25.5 (5.5)} \\
(\#Classes: 7) & \#Clusters
& \textnormal{25.1 (4.4)} & \textnormal{2.0 (0.0)} & \textnormal{1.5 (0.5)} & \textnormal{9.2 (6.0)} 
& \textnormal{14.4 (2.5)} & \textnormal{51.0 (0.0)} & \textnormal{81.8 (48.3)} & \textnormal{10.2 (2.3)} \\
\hline

Rice & \#Nodes
& \textnormal{88.8 (6.8)} & \textnormal{--} & \textnormal{11.8 (5.0)} & \textnormal{129.0 (38.5)} 
& \textnormal{--} & \textnormal{92.0 (0.0)} & \textnormal{1315.1 (424.1)} & \textnormal{17.7 (2.8)} \\
(\#Classes: 2) & \#Clusters
& \textnormal{64.9 (5.4)} & \textnormal{25.0 (0.0)} & \textnormal{4.1 (2.0)} & \textnormal{15.9 (5.0)} 
& \textnormal{44.2 (2.8)} & \textnormal{92.0 (0.0)} & \textnormal{981.9 (498.5)} & \textnormal{5.7 (2.4)} \\
\hline

TUANDROMD & \#Nodes
& \textnormal{97.6 (6.6)} & \textnormal{--} & \textnormal{121.9 (0.3)} & \textnormal{78.0 (22.4)} 
& \textnormal{--} & \textnormal{163.3 (3.5)} & \textnormal{73.0 (75.0)} & \textnormal{7.1 (1.5)} \\
(\#Classes: 2) & \#Clusters
& \textnormal{44.3 (5.5)} & \textnormal{5.0 (0.0)} & \textnormal{4.4 (0.9)} & \textnormal{30.5 (12.9)} 
& \textnormal{16.0 (2.3)} & \textnormal{131.7 (5.0)} & \textnormal{6.9 (2.1)} & \textnormal{3.2 (1.4)} \\
\hline

Phoneme & \#Nodes
& \textnormal{427.5 (11.0)} & \textnormal{--} & \textnormal{2000.0 (0.0)} & \textnormal{150.3 (34.3)} 
& \textnormal{--} & \textnormal{26.9 (6.7)} & \textnormal{245.3 (48.1)} & \textnormal{34.7 (5.8)} \\
(\#Classes: 2) & \#Clusters
& \textnormal{134.8 (12.5)} & \textnormal{37.0 (0.0)} & \textnormal{19.7 (4.8)} & \textnormal{10.2 (4.7)} 
& \textnormal{120.0 (0.0)} & \textnormal{1.1 (0.3)} & \textnormal{78.8 (38.6)} & \textnormal{10.7 (2.9)} \\
\hline

Texture & \#Nodes
& \textnormal{93.0 (4.9)} & \textnormal{--} & \textnormal{28.7 (4.9)} & \textnormal{174.9 (55.7)} 
& \textnormal{--} & \textnormal{30.4 (7.7)} & \textnormal{177.1 (54.2)} & \textnormal{43.9 (5.3)} \\
(\#Classes: 11) & \#Clusters
& \textnormal{76.1 (4.9)} & \textnormal{11.0 (0.0)} & \textnormal{10.2 (1.8)} & \textnormal{19.8 (9.2)} 
& \textnormal{503.5 (3.2)} & \textnormal{3.5 (1.5)} & \textnormal{63.5 (38.7)} & \textnormal{15.8 (2.8)} \\
\hline

OptDigits & \#Nodes
& \textnormal{201.1 (4.5)} & \textnormal{--} & \textnormal{1833.2 (5.0)} & \textnormal{228.8 (49.0)} 
& \textnormal{--} & \textnormal{15.0 (0.0)} & \textnormal{525.8 (177.8)} & \textnormal{70.2 (12.1)} \\
(\#Classes: 10) & \#Clusters
& \textnormal{146.7 (7.1)} & \textnormal{10.0 (0.0)} & \textnormal{100.4 (11.7)} & \textnormal{57.8 (18.0)} 
& \textnormal{9.4 (2.9)} & \textnormal{15.0 (0.0)} & \textnormal{309.1 (148.5)} & \textnormal{27.0 (3.6)} \\
\hline

Statlog & \#Nodes
& \textnormal{133.3 (3.2)} & \textnormal{--} & \textnormal{19.8 (8.0)} & \textnormal{124.7 (42.1)} 
& \textnormal{--} & \textnormal{4.6 (2.6)} & \textnormal{165.1 (81.2)} & \textnormal{36.3 (4.7)} \\
(\#Classes: 6) & \#Clusters
& \textnormal{110.2 (6.2)} & \textnormal{22.0 (0.0)} & \textnormal{2.9 (1.5)} & \textnormal{34.7 (14.8)}
& \textnormal{1310.6 (7.3)} & \textnormal{1.0 (0.2)} & \textnormal{96.4 (63.4)} & \textnormal{16.8 (3.3)} \\
\hline

Anuran Calls & \#Nodes
& \textnormal{120.8 (5.2)} & \textnormal{--} & \textnormal{39.7 (8.8)} & \textnormal{218.7 (63.2)}
& \textnormal{--} & \textnormal{64.7 (5.3)} & \textnormal{254.6 (118.0)} & \textnormal{26.1 (5.5)} \\
(\#Classes: 4) & \#Clusters
& \textnormal{79.1 (5.9)} & \textnormal{5.0 (0.0)} & \textnormal{6.1 (2.0)} & \textnormal{60.7 (20.0)}
& \textnormal{431.9 (8.1)} & \textnormal{19.8 (3.4)} & \textnormal{126.7 (85.6)} & \textnormal{12.9 (2.2)} \\
\hline

Isolet & \#Nodes
& \textnormal{282.0 (0.2)} & \textnormal{--} & \textnormal{41.6 (8.8)} & \textnormal{197.9 (38.6)}
& \textnormal{--} & \textnormal{85.0 (0.0)} & \textnormal{1558.9 (489.1)} & \textnormal{71.5 (17.3)} \\
(\#Classes: 26) & \#Clusters
& \textnormal{282.0 (0.2)} & \textnormal{51.0 (0.0)} & \textnormal{15.2 (3.3)} & \textnormal{102.1 (22.7)}
& \textnormal{2000.0 (0.0)} & \textnormal{85.0 (0.0)} & \textnormal{1187.1 (664.5)} & \textnormal{28.0 (6.1)} \\
\hline

MNIST10K & \#Nodes
& \textnormal{56.4 (4.2)} & \textnormal{--} & \textnormal{2000.0 (0.0)} & \textnormal{197.5 (55.4)}
& \textnormal{--} & \textnormal{17.0 (0.0)} & \textnormal{298.9 (22.6)} & \textnormal{70.2 (10.5)} \\
(\#Classes: 10) & \#Clusters
& \textnormal{27.3 (3.9)} & \textnormal{79.0 (0.0)} & \textnormal{36.1 (4.7)} & \textnormal{30.1 (9.9)}
& \textnormal{8.0 (0.9)} & \textnormal{17.0 (0.0)} & \textnormal{27.9 (20.9)} & \textnormal{37.8 (5.4)} \\
\hline

PenBased & \#Nodes
& \textnormal{300.1 (12.2)} & \textnormal{--} & \textnormal{153.4 (10.2)} & \textnormal{274.8 (53.1)}
& \textnormal{--} & \textnormal{24.0 (0.0)} & \textnormal{371.5 (96.0)} & \textnormal{39.6 (5.3)} \\
(\#Classes: 10) & \#Clusters
& \textnormal{144.8 (9.3)} & \textnormal{12.0 (0.0)} & \textnormal{20.6 (3.4)} & \textnormal{29.1 (6.4)}
& \textnormal{1286.7 (28.0)} & \textnormal{24.0 (0.0)} & \textnormal{124.6 (73.5)} & \textnormal{16.3 (2.6)} \\
\hline

STL10 & \#Nodes
& \textnormal{180.0 (0.0)} & \textnormal{--} & \textnormal{33.4 (5.1)} & \textnormal{94.5 (38.1)}
& \textnormal{--} & \textnormal{28.0 (0.0)} & \textnormal{2000.0 (0.0)} & \textnormal{40.2 (6.2)} \\
(\#Classes: 10) & \#Clusters
& \textnormal{180.0 (0.0)} & \textnormal{2.0 (0.0)} & \textnormal{10.4 (1.9)} & \textnormal{53.5 (23.9)}
& \textnormal{1998.9 (1.2)} & \textnormal{28.0 (0.0)} & \textnormal{2000.0 (0.0)} & \textnormal{12.9 (2.5)} \\
\hline

Letter & \#Nodes
& \textnormal{1964.4 (22.2)} & \textnormal{--} & \textnormal{208.0 (14.1)} & \textnormal{260.2 (55.2)}
& \textnormal{--} & \textnormal{224.4 (18.4)} & \textnormal{657.1 (238.4)} & \textnormal{72.6 (8.0)} \\
(\#Classes: 26) & \#Clusters
& \textnormal{1543.8 (27.5)} & \textnormal{13020.0 (0.0)} & \textnormal{15.8 (3.8)} & \textnormal{12.1 (5.1)}
& \textnormal{2000.0 (0.2)} & \textnormal{39.6 (11.9)} & \textnormal{289.2 (190.9)} & \textnormal{8.9 (2.2)} \\
\hline

Shuttle & \#Nodes
& \textnormal{65.1 (3.8)} & \textnormal{--} & \textnormal{2000.0 (0.0)} & \textnormal{329.3 (66.0)}
& \textnormal{--} & \textnormal{208.0 (0.0)} & \textnormal{468.6 (186.7)} & \textnormal{43.6 (16.1)} \\
(\#Classes: 7) & \#Clusters
& \textnormal{9.1 (1.9)} & \textnormal{2.0 (0.0)} & \textnormal{12.2 (3.8)} & \textnormal{11.1 (3.1)}
& \textnormal{34.8 (1.8)} & \textnormal{208.0 (0.0)} & \textnormal{30.4 (27.8)} & \textnormal{7.2 (2.6)} \\
\hline

Skin & \#Nodes
& \textnormal{N/A} & \textnormal{--} & \textnormal{1999.7 (0.5)} & \textnormal{940.0 (193.7)}
& \textnormal{--} & \textnormal{346.8 (66.1)} & \textnormal{261.2 (55.7)} & \textnormal{25.9 (3.8)} \\
(\#Classes: 2) & \#Clusters
& \textnormal{N/A} & \textnormal{OOM} & \textnormal{225.8 (15.0)} & \textnormal{109.6 (26.8)}
& \textnormal{N/A} & \textnormal{157.5 (29.8)} & \textnormal{14.1 (5.4)} & \textnormal{9.1 (2.5)} \\

\hline\hline
\multicolumn{2}{c|}{Average Cluster Error} & \textnormal{13.8 (17.1)} & \textnormal{25.7 (101.3)} & \textnormal{6.1 (22.2)} & \textnormal{5.6 (10.7)} & \textnormal{52.2 (64.2)} & \textnormal{16.4 (29.4)} & \textnormal{54.8 (101.0)} & \textnormal{1.2 (1.1)} \\
\hline\hline

\end{tabular}}
\\
\vspace{1mm}\footnotesize
\hspace*{7mm}
\begin{minipage}{\linewidth}
The values in parentheses indicate the standard deviation. \\
N/A indicates that no valid hyperparameters were obtained by Bayesian optimization within 12 hours. \\
OOM stands for out-of-memory. \\
Image Seg. stands for Image Segmentation. \\
A node represents a local region of the data, and clusters are connected components of nodes in the topology. \\
TC and DDVFA handle clusters directly without explicitly maintaining node representations. \\
Average Cluster Error shows the relative difference between \#Classes and \#Clusters across 24 real-world datasets. \\
As the error approaches zero, \#Clusters becomes closer to \#Classes. \\
\end{minipage}
\vspace{-3mm}
\end{table*}

\subsubsection{Nonstationary Setting}
\label{sec:nonstationary}

The nonstationary setting assumes class-incremental data arrival, where new classes appear sequentially and the underlying distribution changes over time. In this setting, training samples are presented incrementally in a class-wise manner, and the class order is randomized in each run. Each algorithm is therefore required to learn continually without revisiting past data. As the evaluation metrics, ARI and AMI are adopted to evaluate the final clustering performance of each algorithm under nonstationary settings after learning all classes.

This section compares the proposed IDAT with GNG, SOINN+, DDVFA, CAEA, and CAE, which are capable of incremental learning. Note that all the algorithms in this section can automatically determine the number of clusters without specifying the true number of classes as a hyperparameter.

\stablabel{st:results_nonstationary}
As in Section~\ref{sec:stationary}, the results of ARI and AMI are presented in Table~\stabref{st:results_nonstationary} of the supplementary materials due to page limitations. Accordingly, this section presents the results of statistical significance tests based on the average ARI and AMI reported in Table~\stabref{st:results_nonstationary}. The statistical significance tests are conducted in the same manner as described in Section~\ref{sec:stationary} with the Friedman test and the Nemenyi post-hoc analysis with a significance level of $0.05$.

% Critical difference diagrams based on the average ARI and AMI in the nonstationary setting.
\begin{figure}[htbp]
  \centering
  \subfloat[ARI]{%
    \includegraphics[width=2.4in]{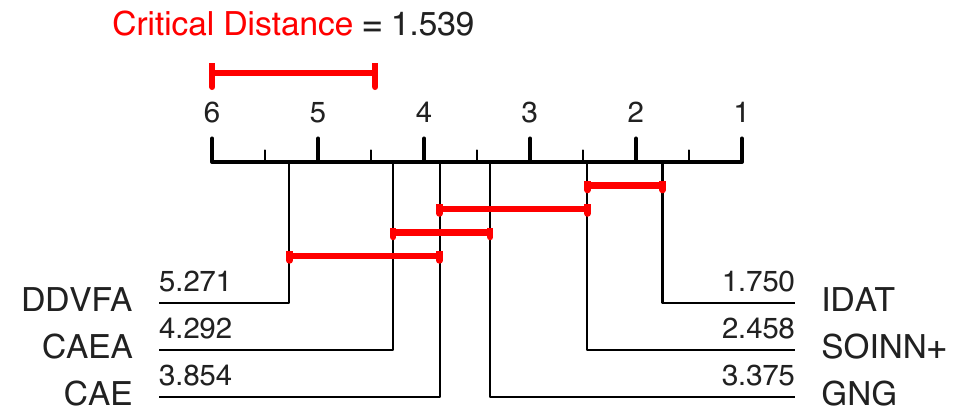}
    \label{fig:CD_ARI_nonstationary_p000}
  } \\
  \subfloat[AMI]{%
    \includegraphics[width=2.4in]{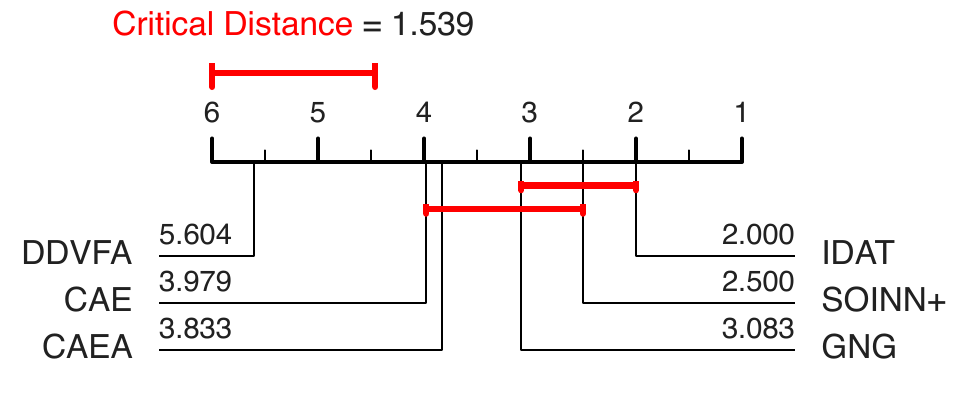}
    \label{fig:CD_AMI_nonstationary_p000}
  }
  % \vspace{2mm}
  \caption{Critical difference diagrams based on the average ARI and AMI in the nonstationary setting.}
  \label{fig:cd_nonstationary_all}
\end{figure}

Fig.~\ref{fig:cd_nonstationary_all} shows critical difference diagrams based on the average ARI and AMI across all datasets in the nonstationary setting. As shown in Fig.~\ref{fig:cd_nonstationary_all}, IDAT achieves the lowest (i.e., best) average ranks for both ARI and AMI under the nonstationary setting. While SOINN+ and GNG show comparable results within the critical distance, ART-based approaches (i.e., CAEA, DDVFA, and CAE) exhibit significantly lower performance. From the viewpoint of the properties measured by ARI and AMI, IDAT maintains high clustering quality under class-incremental conditions by preserving cluster consistency, avoiding over-segmentation, and merging as new classes appear. These results indicate that IDAT effectively adapts to evolving data distributions while maintaining stable and high clustering performance.

Table~\ref{tab:NodesClustersNonstationary} summarizes the average number of nodes and clusters obtained by each algorithm in the nonstationary setting. The cluster error as in (\ref{eq:clusterError}) is also provided and reported as an average over all 24 real-world datasets.

In many datasets, GNG, SOINN+, CAEA, and CAE show a large number of nodes and clusters, which increase model complexity and reduce stability during incremental learning. Their average cluster error becomes large accordingly. DDVFA frequently produces only a very small number of clusters in the nonstationary setting. One possible reason is that the fixed vigilance parameters optimized through Bayesian optimization do not remain effective as the data distribution changes over time. IDAT shows a different tendency. Across a variety of datasets, the number of clusters in IDAT remains close to the true number of classes, and the number of nodes stays compact relative to the training sample size. 

As a consequence, the average cluster error stays small. This stability indicates that IDAT avoids over-segmentation of the data space and continues to represent the data distribution through a topological structure despite changes in the distribution.

% Average number of nodes and clusters in the nonstationary environment
\begin{table*}[htbp]
\centering
\caption{Average number of nodes and clusters in the nonstationary setting}
\label{tab:NodesClustersNonstationary}
\footnotesize
\renewcommand{\arraystretch}{1.1}
\scalebox{0.8}{
\begin{tabular}{ll|*{5}{C{19mm}|}C{19mm}}  \hline\hline
Dataset & Metric & GNG & SOINN+ & DDVFA & CAEA & CAE & IDAT \\ \hline

Iris & \#Nodes
& \textnormal{48.6 (1.0)} & \textnormal{39.1 (6.0)} & \textnormal{--} & \textnormal{13.0 (0.0)} & \textnormal{92.7 (45.5)} & \textnormal{7.7 (1.2)} \\
(\#Classes: 3) & \#Clusters
& \textnormal{4.1 (2.2)} & \textnormal{18.5 (3.9)} & \textnormal{1.3 (0.5)} & \textnormal{13.0 (0.0)} & \textnormal{85.9 (49.7)} & \textnormal{4.1 (1.1)} \\
\hline

Seeds & \#Nodes
& \textnormal{69.5 (1.0)} & \textnormal{59.5 (3.7)} & \textnormal{--} & \textnormal{43.0 (0.0)} & \textnormal{210.0 (0.0)} & \textnormal{12.7 (2.1)} \\
(\#Classes: 3) & \#Clusters
& \textnormal{5.8 (2.2)} & \textnormal{22.5 (5.3)} & \textnormal{3.4 (0.9)} & \textnormal{43.0 (0.0)} & \textnormal{210.0 (0.0)} & \textnormal{6.0 (1.5)} \\
\hline

Dermatology & \#Nodes
& \textnormal{120.5 (1.1)} & \textnormal{110.5 (16.9)} & \textnormal{--} & \textnormal{37.0 (0.0)} & \textnormal{364.7 (3.3)} & \textnormal{14.1 (3.7)} \\
(\#Classes: 6) & \#Clusters
& \textnormal{9.7 (3.2)} & \textnormal{55.9 (13.1)} & \textnormal{1.4 (0.5)} & \textnormal{37.0 (0.0)} & \textnormal{363.8 (5.8)} & \textnormal{7.1 (3.1)} \\
\hline

Pima & \#Nodes
& \textnormal{250.4 (3.1)} & \textnormal{69.4 (17.9)} & \textnormal{--} & \textnormal{129.5 (19.3)} & \textnormal{289.2 (66.3)} & \textnormal{14.0 (5.0)} \\
(\#Classes: 2) & \#Clusters
& \textnormal{15.0 (4.0)} & \textnormal{15.6 (7.0)} & \textnormal{1.8 (1.3)} & \textnormal{70.5 (21.1)} & \textnormal{203.8 (90.7)} & \textnormal{7.4 (3.0)} \\
\hline

Mice Protein & \#Nodes
& \textnormal{350.2 (2.8)} & \textnormal{344.4 (27.7)} & \textnormal{--} & \textnormal{202.0 (0.0)} & \textnormal{1008.8 (91.8)} & \textnormal{31.0 (6.5)} \\
(\#Classes: 8) & \#Clusters
& \textnormal{36.1 (5.8)} & \textnormal{167.3 (20.7)} & \textnormal{1.0 (0.0)} & \textnormal{202.0 (0.0)} & \textnormal{1004.8 (97.0)} & \textnormal{12.8 (2.5)} \\
\hline

Binalpha & \#Nodes
& \textnormal{457.1 (3.5)} & \textnormal{364.7 (28.3)} & \textnormal{--} & \textnormal{65.0 (0.0)} & \textnormal{1120.7 (81.2)} & \textnormal{42.8 (7.3)} \\
(\#Classes: 36) & \#Clusters
& \textnormal{21.2 (4.5)} & \textnormal{226.0 (19.7)} & \textnormal{27.8 (10.7)} & \textnormal{65.0 (0.0)} & \textnormal{1050.7 (108.3)} & \textnormal{18.7 (6.5)} \\
\hline

Yeast & \#Nodes
& \textnormal{480.0 (3.6)} & \textnormal{171.1 (57.9)} & \textnormal{--} & \textnormal{235.0 (0.0)} & \textnormal{335.4 (216.2)} & \textnormal{31.8 (6.4)} \\
(\#Classes: 10) & \#Clusters
& \textnormal{22.1 (8.1)} & \textnormal{36.1 (20.0)} & \textnormal{2.0 (1.0)} & \textnormal{235.0 (0.0)} & \textnormal{176.2 (185.9)} & \textnormal{8.2 (3.0)} \\
\hline

Semeion & \#Nodes
& \textnormal{517.4 (3.3)} & \textnormal{344.3 (40.4)} & \textnormal{--} & \textnormal{87.0 (0.0)} & \textnormal{1272.7 (116.5)} & \textnormal{44.2 (5.6)} \\
(\#Classes: 10) & \#Clusters
& \textnormal{26.2 (6.3)} & \textnormal{180.2 (30.8)} & \textnormal{77.6 (38.0)} & \textnormal{87.0 (0.0)} & \textnormal{1197.0 (150.0)} & \textnormal{25.2 (5.9)} \\
\hline

MSRA25 & \#Nodes
& \textnormal{574.1 (5.0)} & \textnormal{543.2 (42.3)} & \textnormal{--} & \textnormal{202.0 (0.0)} & \textnormal{1039.6 (505.1)} & \textnormal{41.0 (5.3)} \\
(\#Classes: 12) & \#Clusters
& \textnormal{59.2 (7.8)} & \textnormal{255.7 (29.0)} & \textnormal{1.7 (0.6)} & \textnormal{202.0 (0.0)} & \textnormal{950.2 (518.5)} & \textnormal{19.0 (3.4)} \\
\hline

Image Seg. & \#Nodes
& \textnormal{747.7 (5.7)} & \textnormal{608.5 (54.2)} & \textnormal{--} & \textnormal{202.0 (0.0)} & \textnormal{204.6 (64.3)} & \textnormal{36.7 (9.3)} \\
(\#Classes: 7) & \#Clusters
& \textnormal{58.2 (11.4)} & \textnormal{223.7 (40.2)} & \textnormal{1.0 (0.0)} & \textnormal{202.0 (0.0)} & \textnormal{63.0 (43.3)} & \textnormal{14.5 (3.6)} \\
\hline

Rice & \#Nodes
& \textnormal{771.3 (0.4)} & \textnormal{216.3 (22.5)} & \textnormal{--} & \textnormal{57.2 (66.1)} & \textnormal{2000.0 (0.0)} & \textnormal{23.4 (0.5)} \\
(\#Classes: 2) & \#Clusters
& \textnormal{6.7 (0.4)} & \textnormal{20.7 (6.3)} & \textnormal{1.0 (0.0)} & \textnormal{57.2 (66.1)} & \textnormal{2000.0 (0.0)} & \textnormal{7.2 (3.5)} \\
\hline

TUANDROMD & \#Nodes
& \textnormal{213.8 (284.7)} & \textnormal{147.3 (33.3)} & \textnormal{--} & \textnormal{16.9 (4.9)} & \textnormal{88.5 (93.6)} & \textnormal{9.4 (3.0)} \\
(\#Classes: 2) & \#Clusters
& \textnormal{24.9 (36.9)} & \textnormal{66.6 (24.3)} & \textnormal{2.3 (0.4)} & \textnormal{16.9 (4.9)} & \textnormal{19.3 (6.3)} & \textnormal{5.2 (1.5)} \\
\hline

Phoneme & \#Nodes
& \textnormal{1757.0 (7.5)} & \textnormal{271.0 (84.7)} & \textnormal{--} & \textnormal{207.1 (72.4)} & \textnormal{199.0 (34.9)} & \textnormal{35.6 (3.0)} \\
(\#Classes: 2) & \#Clusters
& \textnormal{97.9 (10.2)} & \textnormal{30.8 (15.9)} & \textnormal{3.5 (0.9)} & \textnormal{207.1 (72.4)} & \textnormal{58.6 (39.4)} & \textnormal{12.4 (0.5)} \\
\hline

Texture & \#Nodes
& \textnormal{1773.3 (7.5)} & \textnormal{1336.6 (76.6)} & \textnormal{--} & \textnormal{350.1 (72.9)} & \textnormal{374.9 (144.1)} & \textnormal{128.6 (16.3)} \\
(\#Classes: 11) & \#Clusters
& \textnormal{126.3 (12.4)} & \textnormal{451.8 (53.2)} & \textnormal{1.0 (0.0)} & \textnormal{99.2 (35.2)} & \textnormal{82.6 (84.9)} & \textnormal{34.3 (4.7)} \\
\hline

OptDigits & \#Nodes
& \textnormal{1833.5 (5.1)} & \textnormal{1180.7 (81.0)} & \textnormal{--} & \textnormal{155.4 (53.9)} & \textnormal{1652.9 (648.9)} & \textnormal{129.3 (15.7)} \\
(\#Classes: 10) & \#Clusters
& \textnormal{100.4 (12.6)} & \textnormal{385.8 (55.4)} & \textnormal{1.0 (0.0)} & \textnormal{30.1 (10.3)} & \textnormal{1183.7 (542.6)} & \textnormal{31.2 (5.9)} \\
\hline

Statlog & \#Nodes
& \textnormal{2000.0 (0.0)} & \textnormal{639.3 (123.4)} & \textnormal{--} & \textnormal{285.3 (50.9)} & \textnormal{541.1 (565.6)} & \textnormal{45.8 (8.5)} \\
(\#Classes: 6) & \#Clusters
& \textnormal{91.2 (11.7)} & \textnormal{252.4 (53.2)} & \textnormal{12.8 (19.9)} & \textnormal{146.2 (47.1)} & \textnormal{382.0 (480.1)} & \textnormal{17.0 (3.8)} \\
\hline

Anuran Calls & \#Nodes
& \textnormal{2000.0 (0.0)} & \textnormal{546.7 (131.8)} & \textnormal{--} & \textnormal{174.8 (29.8)} & \textnormal{297.7 (125.9)} & \textnormal{27.5 (7.2)} \\
(\#Classes: 4) & \#Clusters
& \textnormal{243.0 (27.9)} & \textnormal{188.1 (54.9)} & \textnormal{2.8 (1.6)} & \textnormal{174.8 (29.8)} & \textnormal{170.1 (89.3)} & \textnormal{11.1 (4.4)} \\
\hline

Isolet & \#Nodes
& \textnormal{2000.0 (0.0)} & \textnormal{1119.5 (70.1)} & \textnormal{--} & \textnormal{202.0 (0.0)} & \textnormal{2000.0 (0.0)} & \textnormal{76.2 (8.9)} \\
(\#Classes: 26) & \#Clusters
& \textnormal{132.0 (14.6)} & \textnormal{556.5 (54.6)} & \textnormal{3.9 (6.8)} & \textnormal{202.0 (0.0)} & \textnormal{2000.0 (0.0)} & \textnormal{24.4 (3.4)} \\
\hline

MNIST10K & \#Nodes
& \textnormal{2000.0 (0.0)} & \textnormal{1773.1 (288.1)} & \textnormal{--} & \textnormal{1200.7 (448.8)} & \textnormal{1564.8 (461.7)} & \textnormal{124.5 (32.8)} \\
(\#Classes: 10) & \#Clusters
& \textnormal{123.9 (11.8)} & \textnormal{568.1 (136.8)} & \textnormal{1.0 (0.0)} & \textnormal{144.8 (59.7)} & \textnormal{146.2 (51.9)} & \textnormal{32.3 (7.6)} \\
\hline

PenBased & \#Nodes
& \textnormal{2000.0 (0.0)} & \textnormal{2420.3 (197.0)} & \textnormal{--} & \textnormal{936.7 (235.8)} & \textnormal{898.6 (509.5)} & \textnormal{132.1 (29.2)} \\
(\#Classes: 10) & \#Clusters
& \textnormal{156.2 (13.6)} & \textnormal{798.4 (143.0)} & \textnormal{2.1 (1.3)} & \textnormal{188.1 (36.3)} & \textnormal{276.4 (371.1)} & \textnormal{32.5 (5.8)} \\
\hline

STL10 & \#Nodes
& \textnormal{2000.0 (0.2)} & \textnormal{359.9 (63.5)} & \textnormal{--} & \textnormal{244.1 (8.0)} & \textnormal{2000.0 (0.0)} & \textnormal{37.7 (8.4)} \\
(\#Classes: 10) & \#Clusters
& \textnormal{72.1 (11.5)} & \textnormal{166.5 (29.7)} & \textnormal{1.0 (0.0)} & \textnormal{174.5 (6.7)} & \textnormal{2000.0 (0.0)} & \textnormal{10.1 (3.9)} \\
\hline

Letter & \#Nodes
& \textnormal{1999.9 (0.3)} & \textnormal{6251.3 (353.9)} & \textnormal{--} & \textnormal{567.0 (193.1)} & \textnormal{882.5 (314.4)} & \textnormal{189.0 (20.4)} \\
(\#Classes: 26) & \#Clusters
& \textnormal{174.5 (18.6)} & \textnormal{2561.1 (306.4)} & \textnormal{1.4 (0.7)} & \textnormal{166.2 (41.1)} & \textnormal{228.5 (108.4)} & \textnormal{51.0 (9.0)} \\
\hline

Shuttle & \#Nodes
& \textnormal{2000.0 (0.0)} & \textnormal{1683.6 (1064.1)} & \textnormal{--} & \textnormal{389.2 (70.7)} & \textnormal{503.1 (191.0)} & \textnormal{131.6 (46.5)} \\
(\#Classes: 7) & \#Clusters
& \textnormal{79.2 (25.4)} & \textnormal{110.0 (84.8)} & \textnormal{2.5 (1.8)} & \textnormal{226.1 (83.4)} & \textnormal{55.9 (22.8)} & \textnormal{20.3 (7.5)} \\
\hline

Skin & \#Nodes
& N/A & \textnormal{1187.6 (74.2)} & \textnormal{--} & \textnormal{104.6 (36.4)} & \textnormal{407.4 (130.0)} & \textnormal{42.2 (11.5)} \\
(\#Classes: 2) & \#Clusters
& N/A & \textnormal{111.4 (27.9)} & N/A & \textnormal{3.6 (2.7)} & \textnormal{21.0 (7.5)} & \textnormal{17.0 (2.5)} \\

\hline\hline
\multicolumn{2}{c|}{Average Cluster Error} & \textnormal{10.2 (14.3)} & \textnormal{28.4 (24.0)} & \textnormal{0.9 (1.3)} & \textnormal{19.2 (20.7)} & \textnormal{93.1 (195.0)} & \textnormal{1.7 (1.7)} \\
\hline\hline

\end{tabular}}
\\
\vspace{1mm}\footnotesize
\hspace*{19mm}
\begin{minipage}{\linewidth}
The values in parentheses indicate the standard deviation. \\
N/A indicates that no valid hyperparameters were obtained by Bayesian optimization within 12 hours. \\
Image Seg. stands for Image Segmentation. \\
A node represents a local region of the data, and clusters are connected components of nodes in the topology. \\
DDVFA handles clusters directly without explicitly maintaining node representations. \\
Average Cluster Error shows the relative difference between \#Classes and \#Clusters across 24 real-world datasets. \\
As the error approaches zero, \#Clusters becomes closer to \#Classes. \\
\end{minipage}
\vspace{-3mm}
\end{table*}

\subsubsection{Continual Learning Performance}
\label{sec:continual}

\begin{figure}[htbp]
  \centering
  \subfloat[AI-ARI]{%
    \includegraphics[width=0.485\linewidth]{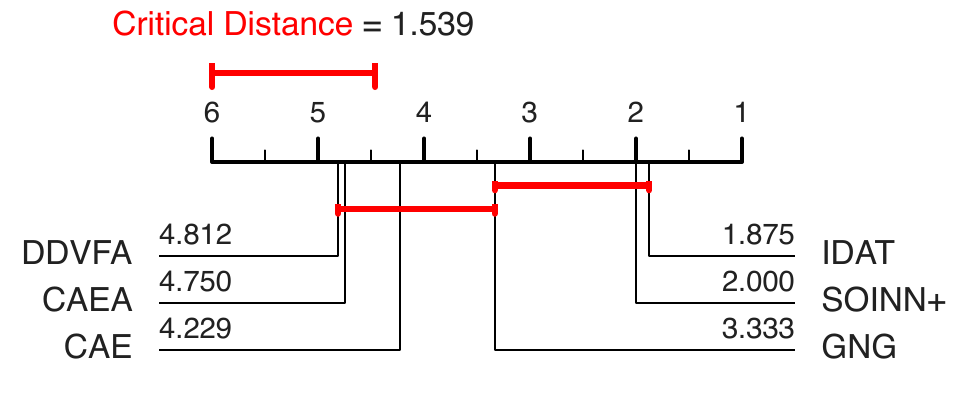}
    \label{fig:CD_IncARI_nonstationary_p000}
  }\hfill
  \subfloat[AI-AMI]{%
    \includegraphics[width=0.485\linewidth]{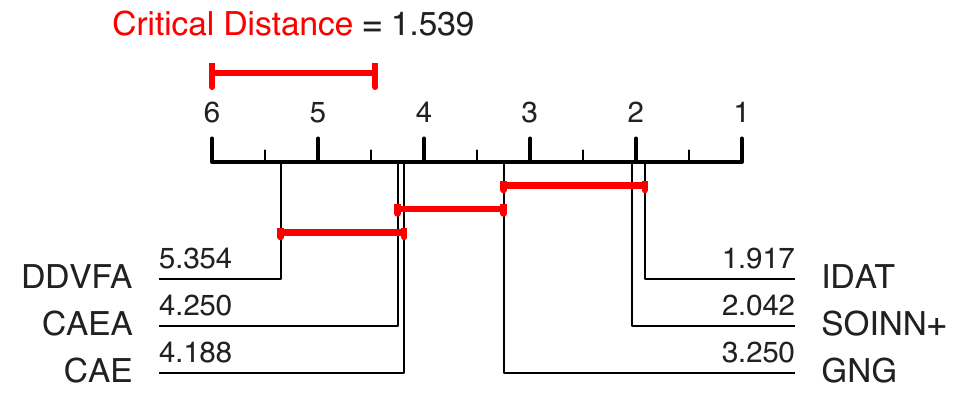}
    \label{fig:CD_IncAMI_nonstationary_p000}
  }\\
  \subfloat[BWT-ARI]{%
    \includegraphics[width=0.485\linewidth]{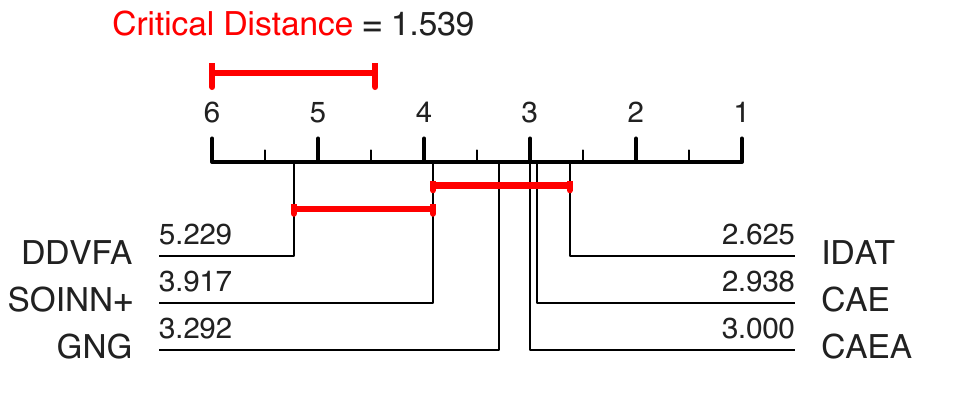}
    \label{fig:CD_BwtARI_nonstationary_p000}
  }\hfill
  \subfloat[BWT-AMI]{%
    \includegraphics[width=0.485\linewidth]{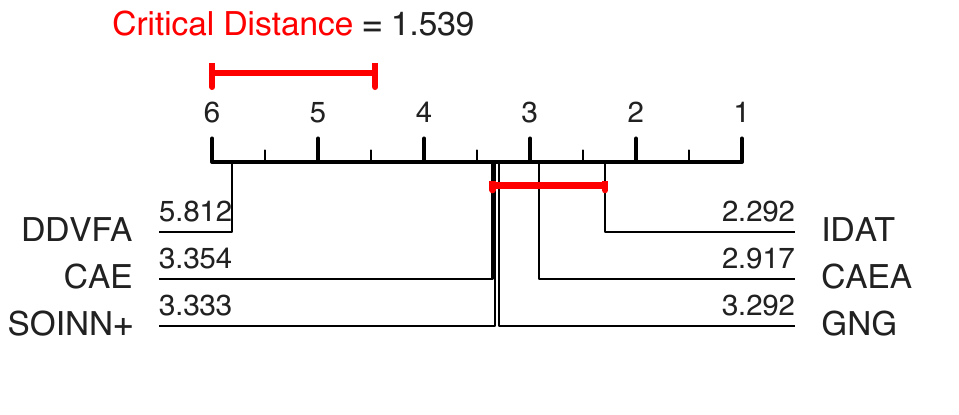}
    \label{fig:CD_BwtAMI_nonstationary_p000}
  }
  \caption{Critical difference diagrams based on AI-ARI, AI-AMI, BWT-ARI, and BWT-AMI in the nonstationary setting.}
  \label{fig:cd_continual}
\end{figure}

While Sections~\ref{sec:stationary} and \ref{sec:nonstationary} focused on the final clustering performance, a significant aspect of continual learning is to preserve past knowledge and prevent forgetting. In this section, AI and BWT (see Section~\ref{sec:metrics} for details) are adopted to evaluate the ability to retain previously acquired knowledge and minimize forgetting as new classes are learned. Here, the ARI and AMI values obtained after learning each class in Section~\ref{sec:nonstationary} are used as the base performance measures for computing AI and BWT. The results are therefore reported as AI-ARI, AI-AMI, BWT-ARI, and BWT-AMI.

\stablabel{st:results_incbwt}
As in the previous section, the statistical significance tests are conducted in the same manner as described in Section~\ref{sec:stationary} with the Friedman test and the Nemenyi post-hoc analysis with a significance level of $0.05$. The detailed numerical results of this experiment are provided in Table~\stabref{st:results_incbwt} of the supplementary materials.

Fig. \ref{fig:cd_continual} shows critical difference diagrams based on AI-ARI, AI-AMI, BWT-ARI, and BWT-AMI in the nonstationary setting. Note that AI reflects both the acquisition of new classes and the retention of previously learned ones, and BWT quantifies the extent of forgetting or improvement caused by later updates. As shown in Fig.~\ref{fig:cd_continual}, IDAT achieves the lowest (i.e., best) average ranks across all metrics for continual learning performance. This result indicates that IDAT effectively retains previously acquired knowledge while incorporating new classes, thereby minimizing catastrophic forgetting. In contrast, DDVFA and CAEA exhibit higher (i.e., worse) ranks, suggesting that their fixed vigilance and structural parameters limit adaptability to distributional shifts in the nonstationary setting. GNG and SOINN+ show inconsistent performance across evaluation metrics. They achieve relatively low ranks in AI but high ranks in BWT, suggesting difficulty in maintaining an appropriate balance between stability and plasticity. These results clearly highlight the superiority of IDAT in continual learning performance.

\subsection{Ablation Study}
\label{sec:ablation}

This section evaluates the contribution of the diversity-driven adaptation for the recalculation interval $\Lambda$ and the vigilance threshold $V_{\text{threshold}}$ in IDAT to clustering and continual learning performance. The ablation study focuses on the nonstationary setting because adaptive adjustments are essential and effective for continual learning. Ablation variants disable the decremental direction that shortens $\Lambda$ when the diversity condition in (\ref{eq:diversity_condition}) fails, or the incremental direction that enlarges $\Lambda$ when it holds. Here, these variants are denoted as \textit{w/o Dec.} and \textit{w/o Inc.}, respectively. Each variant still recomputes $V_{\text{threshold}}$ under the remaining mechanism. In addition, another variant without any recalculation of $\Lambda$ or $V_{\text{threshold}}$ is also considered and is denoted as \textit{w/o All}. Throughout the ablation study, two initialization settings for $\Lambda_{\text{init}}$ are considered. The setting $\Lambda_{\text{init}} = 2$ is used to remain consistent with the main evaluation protocol. In contrast, the setting $\Lambda_{\text{init}} = 500$ is used to clarify the necessity of the bidirectional adjustment mechanism for $\Lambda$.

\stablabel{st:AblationARIAMINonstationary}
\stablabel{st:AblationContinualNonstationary}
\stablabel{st:AblationClustersNonstationary}

As in Section~\ref{sec:nonstationary}, due to page limitations, the results on the clustering performance of IDAT and its three variants are provided in the supplementary materials. Table~\stabref{st:AblationARIAMINonstationary} shows the average ARI and AMI after learning all classes as the final clustering performance in the nonstationary setting, and Table~\stabref{st:AblationContinualNonstationary} reports the average AI-ARI, AI-AMI, BWT-ARI, and BWT-AMI. In addition, Table~\stabref{st:AblationClustersNonstationary} summarizes the numbers of nodes and clusters after learning all classes.

Accordingly, this section presents the results of statistical significance tests based on the average ARI and AMI in Table~\stabref{st:results_nonstationary}. The tests are conducted in the same manner as described in Section~\ref{sec:stationary}, using the Friedman test and the Nemenyi post-hoc analysis with a significance level of $0.05$.

\begin{figure}[htbp]
  \centering
  \subfloat[ARI]{%
    \includegraphics[width=0.485\linewidth]{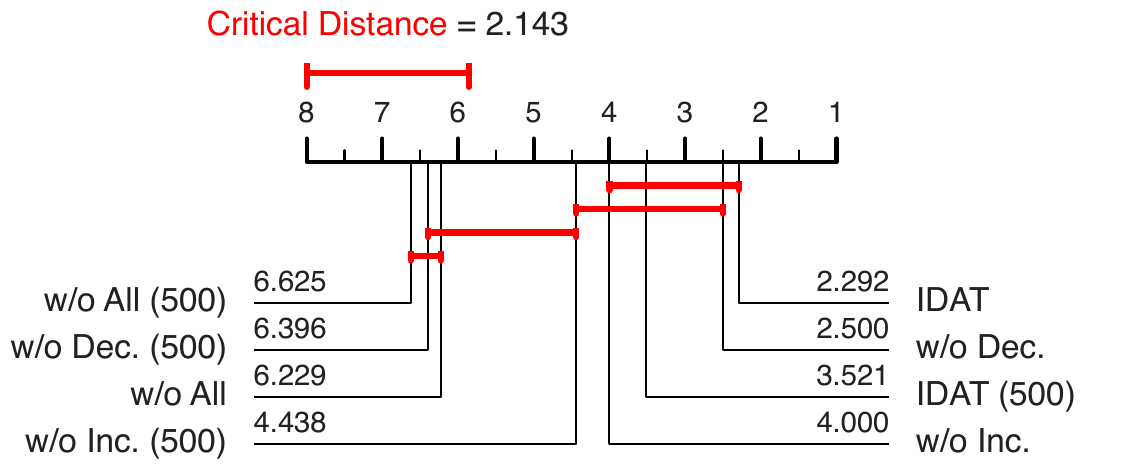}
    \label{fig:Ablation_CD_ARI_nonstationary_p000}
  }\hfill
  \subfloat[AMI]{%
    \includegraphics[width=0.485\linewidth]{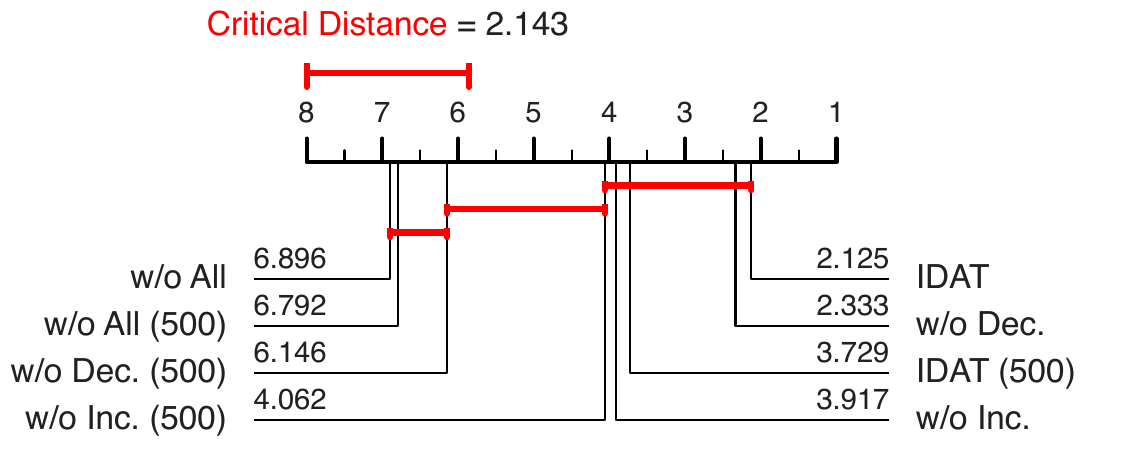}
    \label{fig:Ablation_CD_AMI_nonstationary_p000}
  } \\
  \subfloat[AI-ARI]{%
    \includegraphics[width=0.485\linewidth]{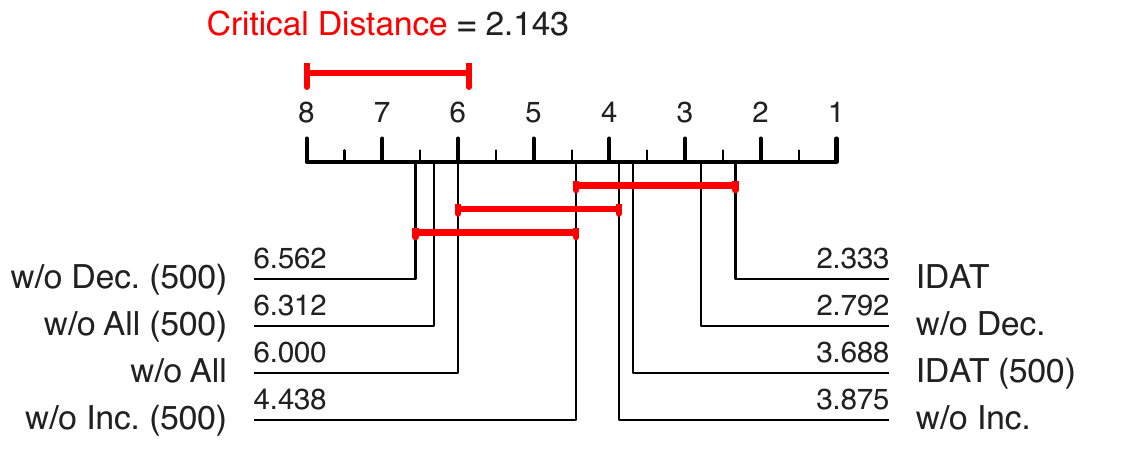}
    \label{fig:Ablation_CD_IncARI_nonstationary_p000}
  }\hfill
  \subfloat[AI-AMI]{%
    \includegraphics[width=0.485\linewidth]{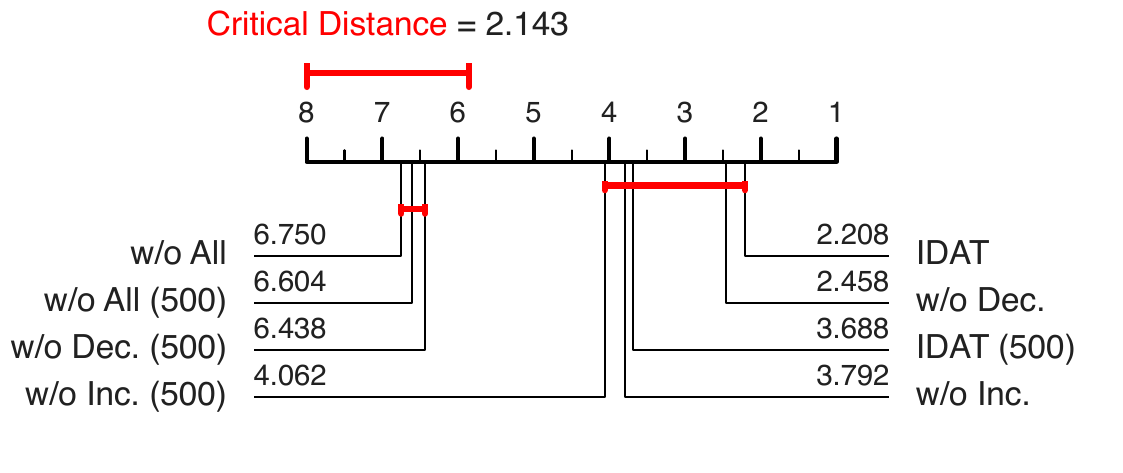}
    \label{fig:Ablation_CD_IncAMI_nonstationary_p000}
  } \\
  \subfloat[BWT-ARI]{%
    \includegraphics[width=0.485\linewidth]{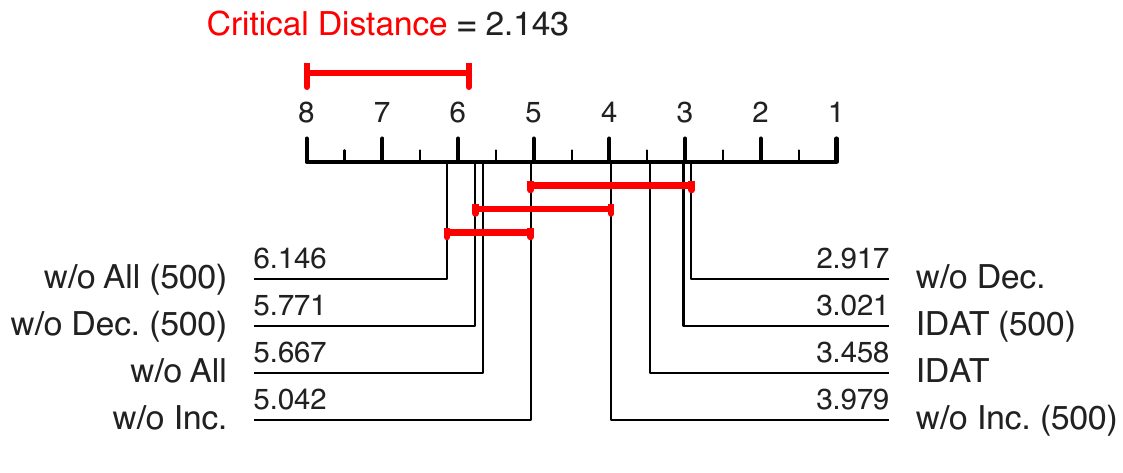}
    \label{fig:Ablation_CD_BwtARI_nonstationary_p000}
  }\hfill
  \subfloat[BWT-AMI]{%
    \includegraphics[width=0.485\linewidth]{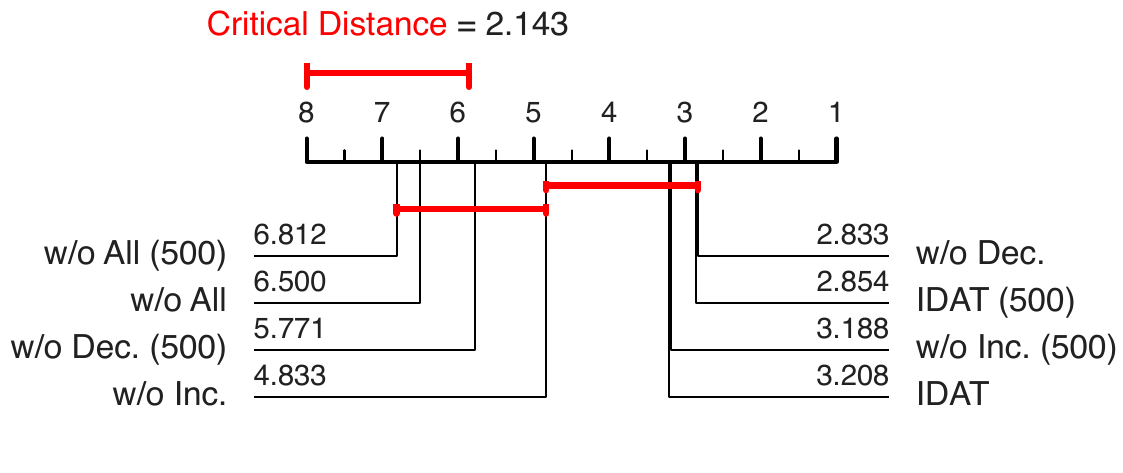}
    \label{fig:Ablation_CD_BwtAMI_nonstationary_p000}
  }
  \caption{
  Critical difference diagrams based on ARI, AMI, and continual performance metrics in the nonstationary setting for the ablation study.
  The notation (500) indicates $\Lambda_{\text{init}} = 500$, and its absence implies $\Lambda_{\text{init}} = 2$.
  }
  \label{fig:cd_ablation_continual}
  \vspace{-3mm}
\end{figure}

Fig.~\ref{fig:cd_ablation_continual} presents critical difference diagrams for the ablation study based on ARI, AMI, and continual learning metrics in the nonstationary setting. The w/o Inc. and w/o All variants record consistently higher (i.e., worse) ranks in all metrics, and these results show that the incremental direction is essential and that performance declines further when both directions are disabled. The w/o Dec. variant achieves lower (i.e., better) ranks than these two variants, but its performance remains below that of IDAT. This difference indicates that the decremental direction also provides a clear benefit for high and stable clustering performance.

Moreover, Fig.~\ref{fig:cd_ablation_continual} shows that IDAT demonstrates smaller differences in performance between $\Lambda_{\text{init}} = 2$ and $\Lambda_{\text{init}} = 500$ compared to the other variants. This suggests that the adaptive adjustments for $\Lambda$ maintain consistent performance regardless of the initial setting. Table~\ref{st:AblationClustersNonstationary} in the supplementary materials further supports this by showing that IDAT generates similar numbers of nodes and clusters in both settings. In contrast, the other variants exhibit larger differences in the number of generated nodes and clusters depending on the initial setting. These results emphasize the robustness of the adaptive mechanisms in ensuring consistent clustering performance across varying initial intervals and stabilizing the clustering process.

\sfiglabel{sf:AblationHistoryNonstationary}
\sfiglabel{sf:AblationHistoryNonstationary_500}
\sfiglabel{sf:AblationHistoryDecNonstationary}
\sfiglabel{sf:AblationHistoryDecNonstationary_500}
\sfiglabel{sf:AblationHistoryIncNonstationary}
\sfiglabel{sf:AblationHistoryIncNonstationary_500}

\begin{figure}[htbp]
  \centering
  \subfloat[IDAT ($\Lambda_{\text{init}} = 2$)]{%
    \includegraphics[width=0.391\linewidth]{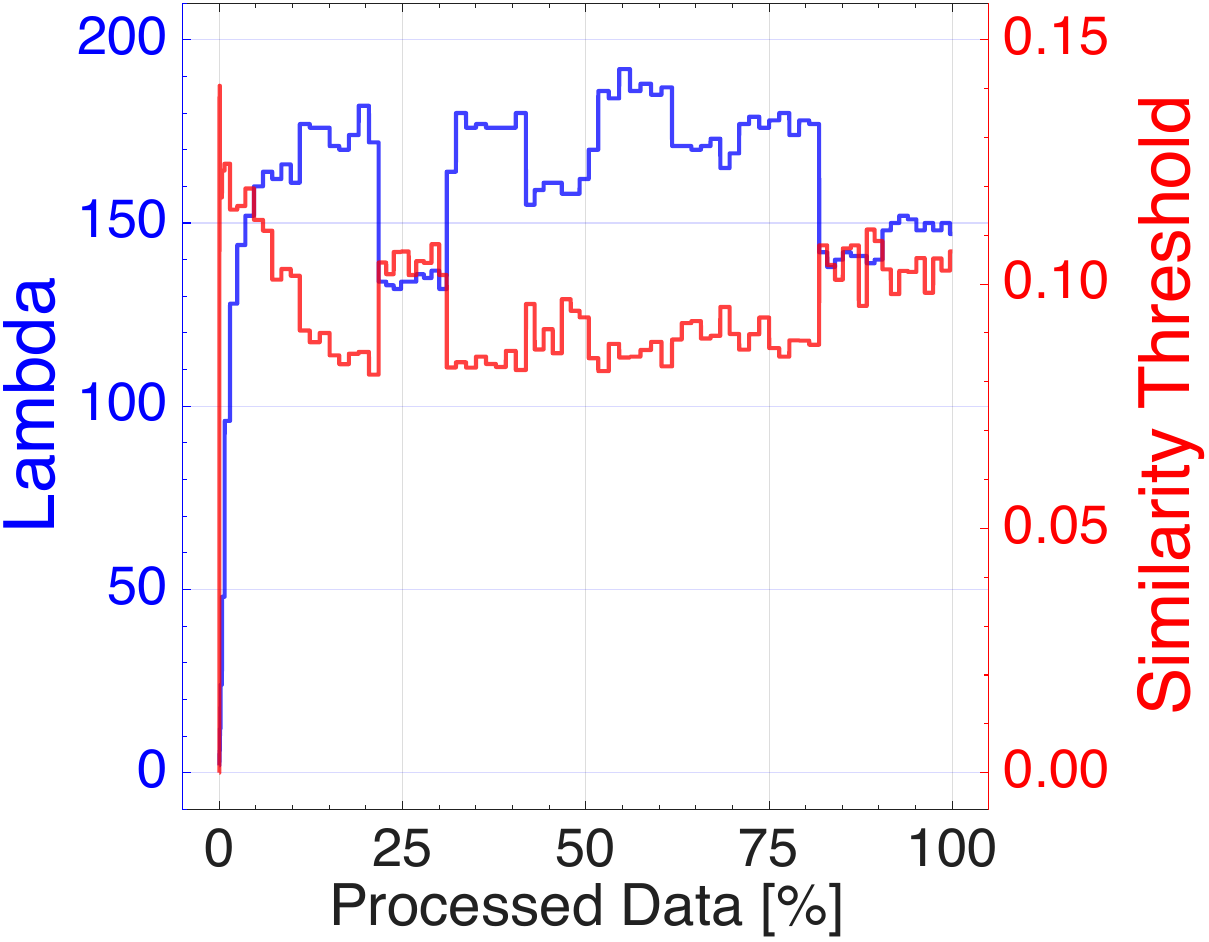}
    \label{fig:ablation_lambda_history_idat_nonstationary}
  }\hspace{3mm}
  \subfloat[IDAT ($\Lambda_{\text{init}} = 500$)]{%
    \includegraphics[width=0.391\linewidth]{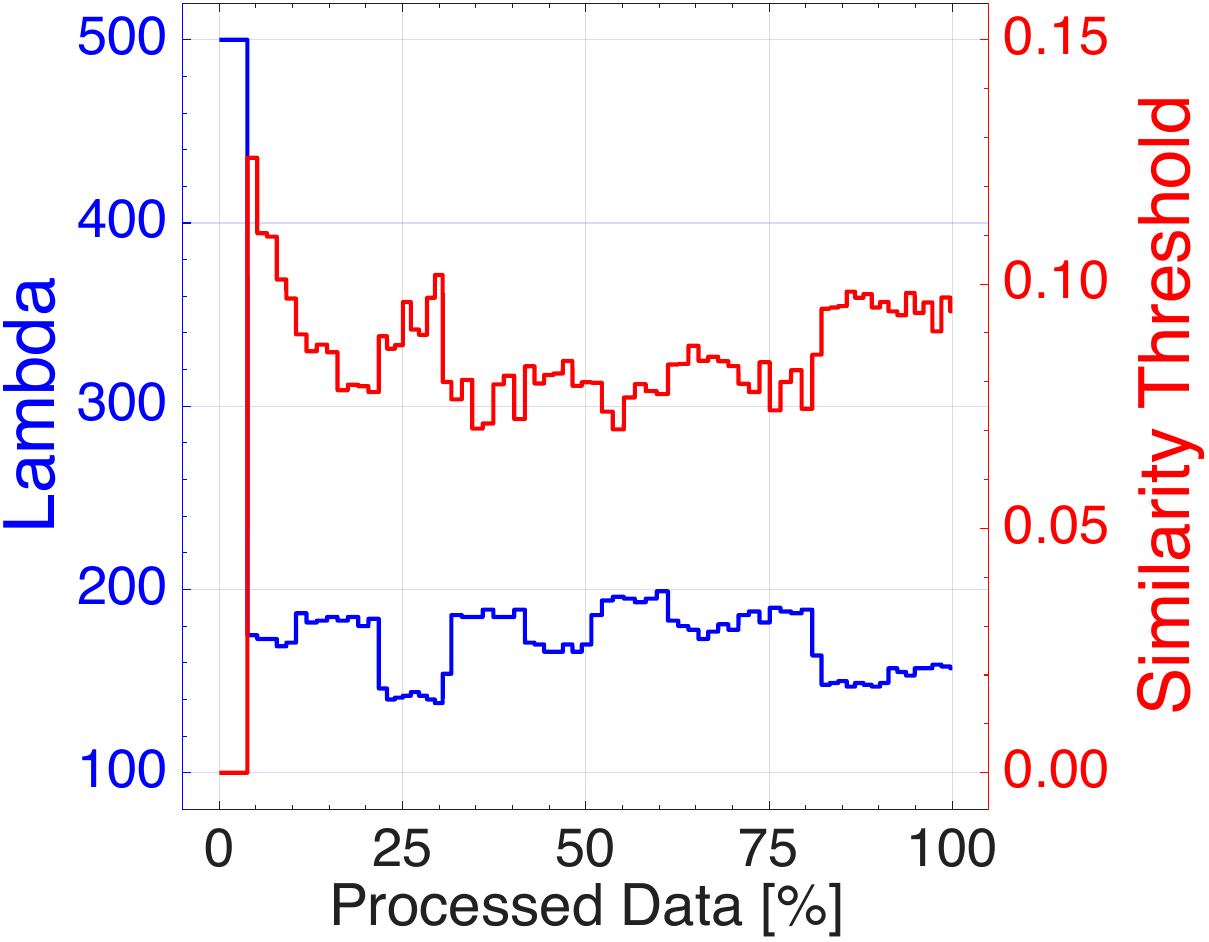}
    \label{fig:ablation_lambda_history_idat_500_nonstationary}
  }\\
  \subfloat[w/o Dec. ($\Lambda_{\text{init}} = 2$)]{%
    \includegraphics[width=0.391\linewidth]{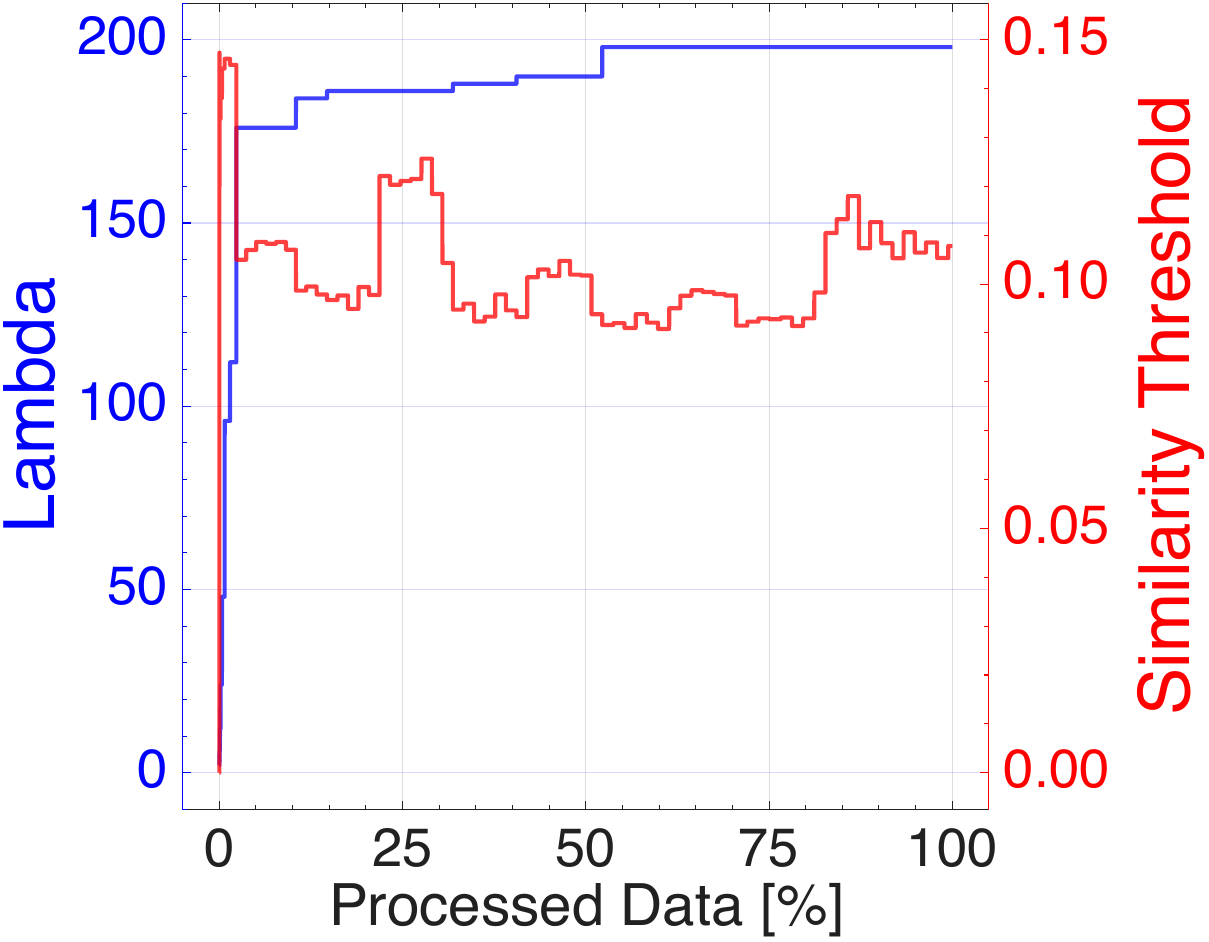}
    \label{fig:ablation_lambda_history_wodec_nonstationary}
  }\hspace{3mm}
  \subfloat[w/o Dec. ($\Lambda_{\text{init}} = 500$)]{%
    \includegraphics[width=0.391\linewidth]{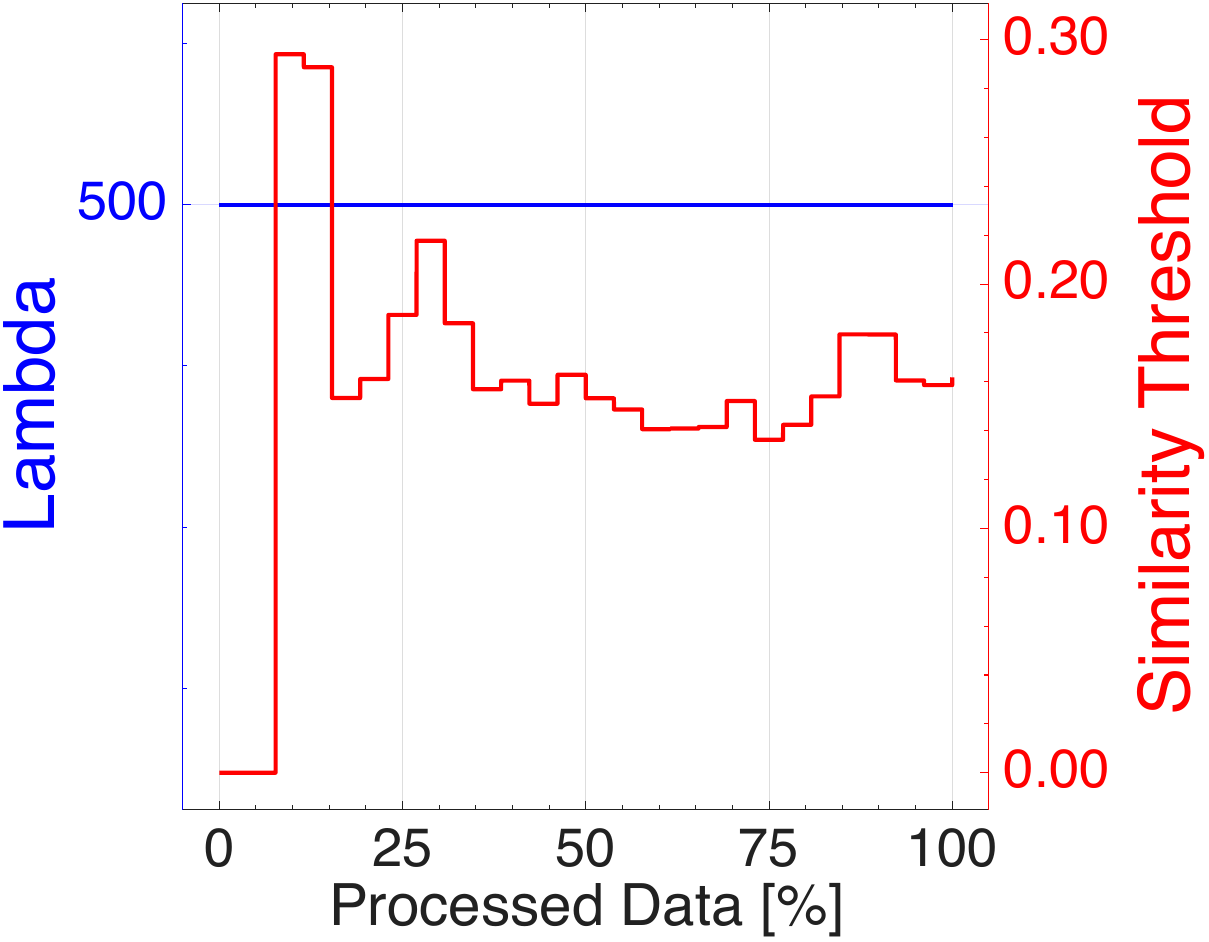}
    \label{fig:ablation_lambda_history_wodec_500_nonstationary}
  }\\
  \subfloat[w/o Inc. ($\Lambda_{\text{init}} = 2$)]{%
    \includegraphics[width=0.391\linewidth]{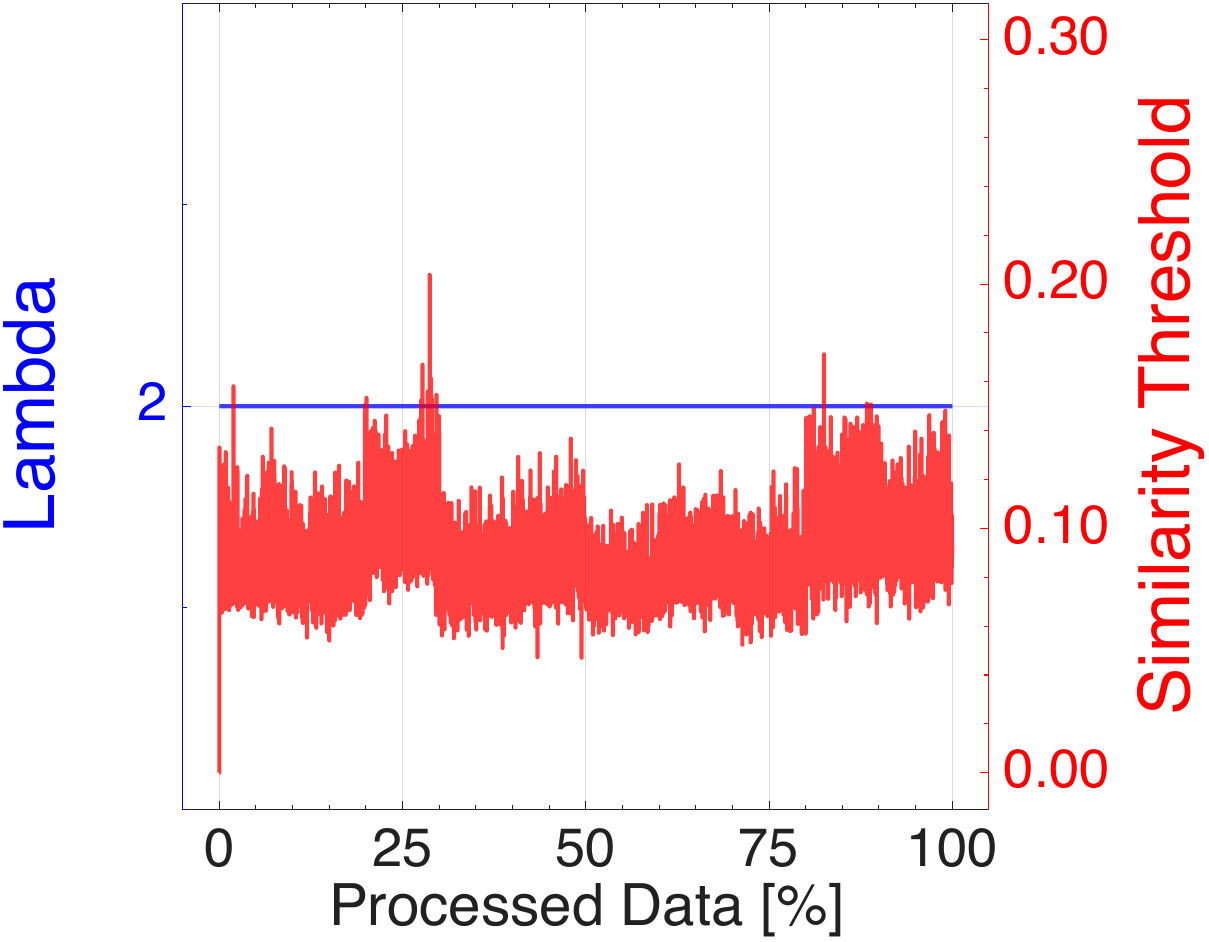}
    \label{fig:ablation_lambda_history_woinc_nonstationary}
  }\hspace{3mm}
  \subfloat[w/o Inc. ($\Lambda_{\text{init}} = 500$)]{%
    \includegraphics[width=0.391\linewidth]{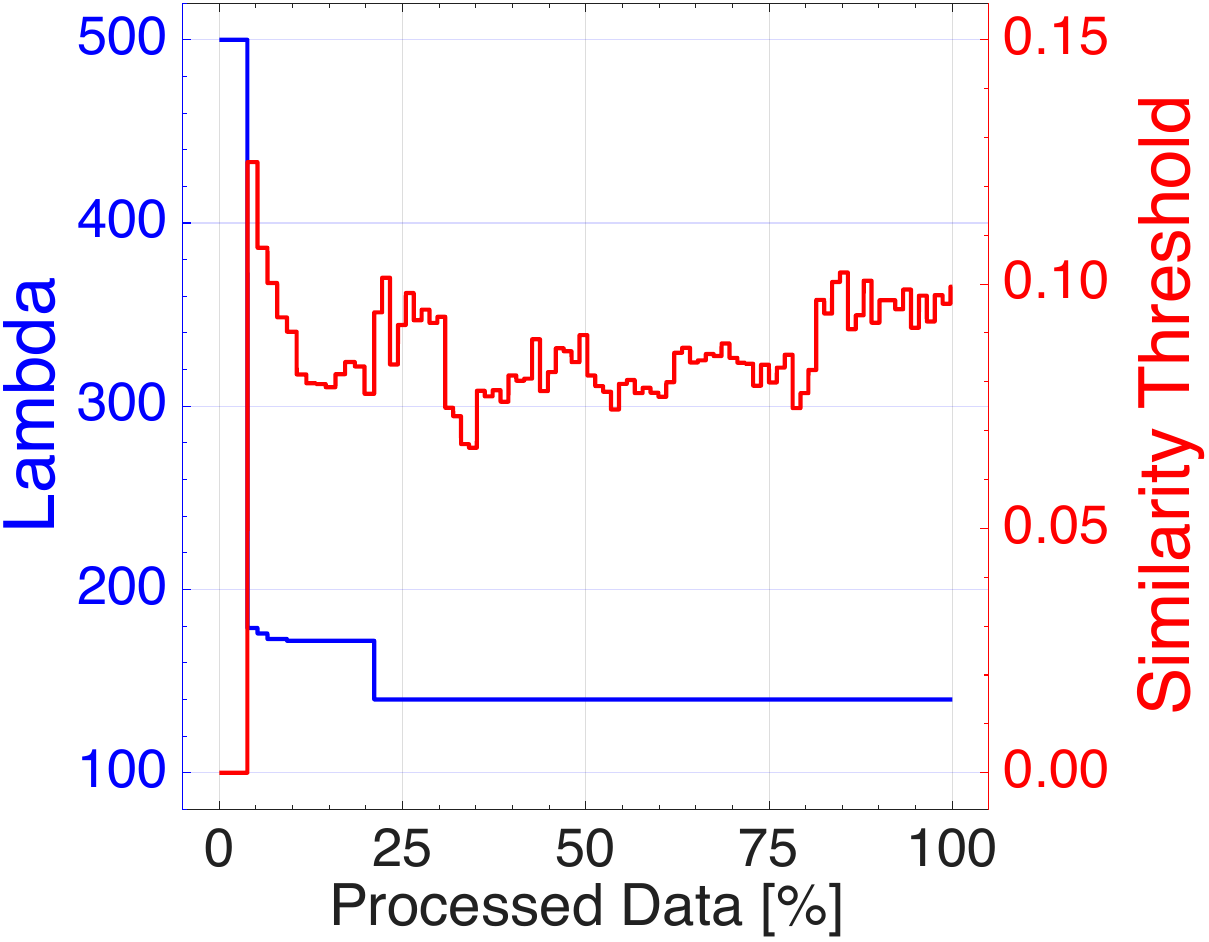}
    \label{fig:ablation_lambda_history_woinc_500_nonstationary}
  }\\
  \subfloat[w/o All ($\Lambda_{\text{init}} = 2$)]{%
    \includegraphics[width=0.391\linewidth]{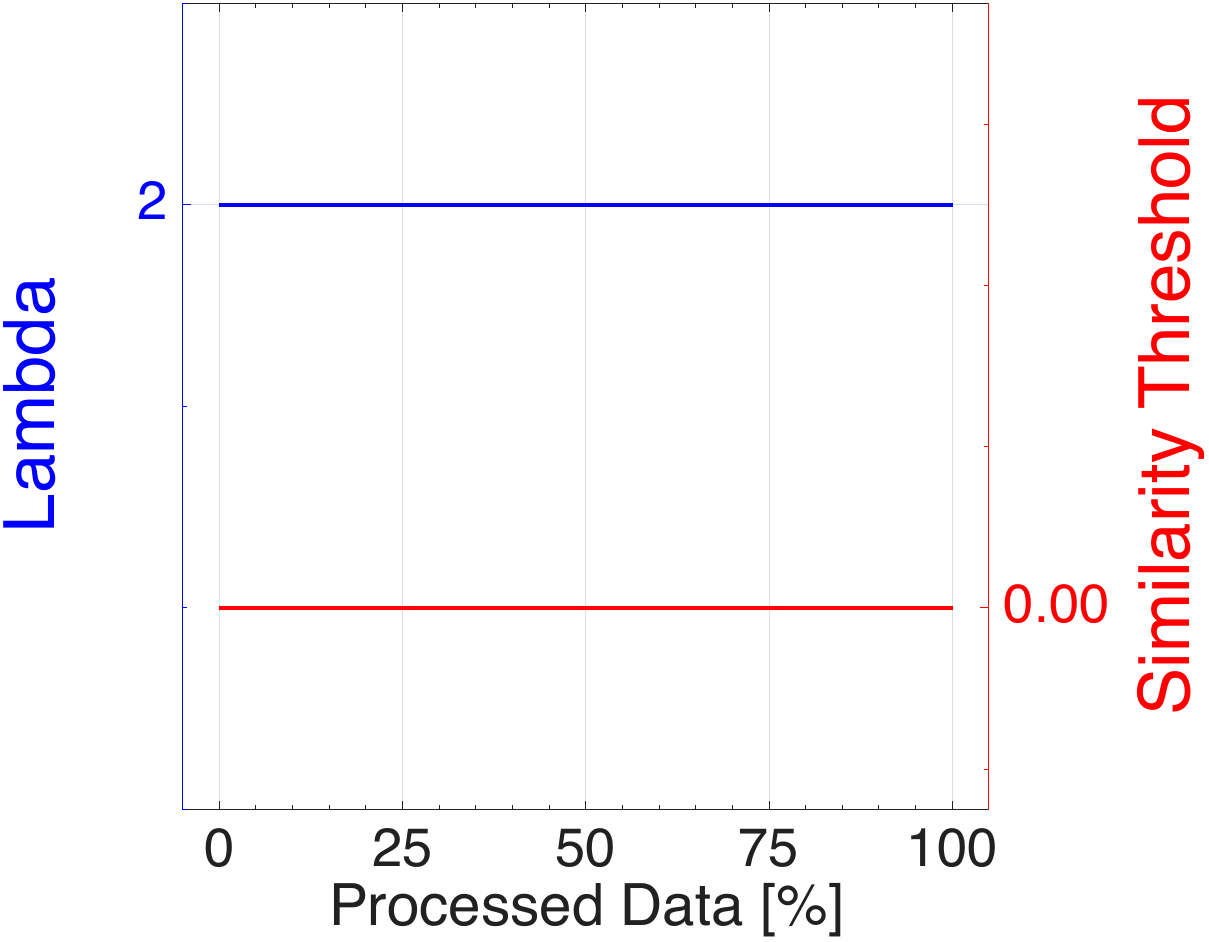}
    \label{fig:ablation_lambda_history_woboth_nonstationary}
  }\hspace{3mm}
  \subfloat[w/o All ($\Lambda_{\text{init}} = 500$)]{%
    \includegraphics[width=0.391\linewidth]{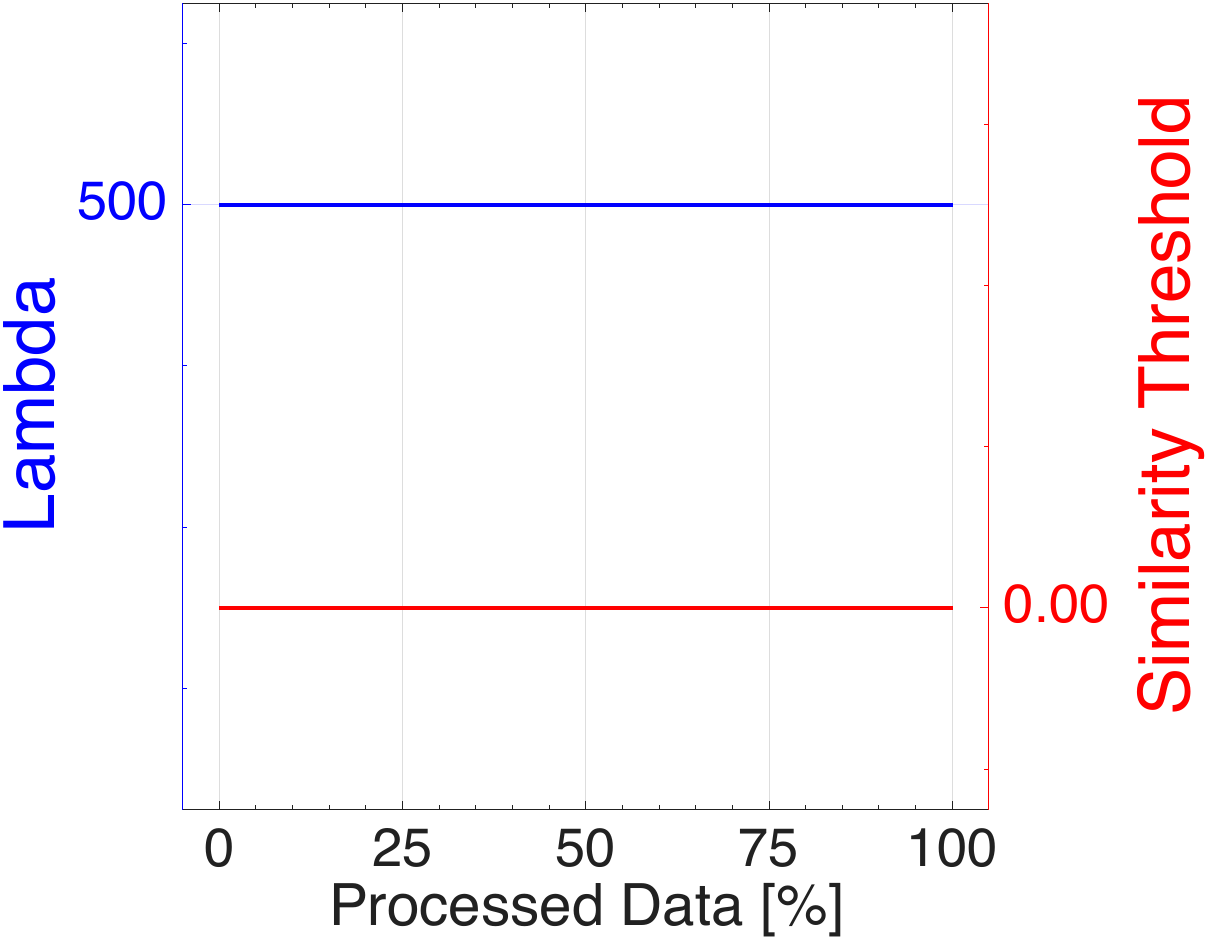}
    \label{fig:ablation_lambda_history_woboth_500_nonstationary}
  }
  \caption{Histories of $\Lambda$ and $V_{\text{threshold}}$ for STL10 in the nonstationary setting. In (g) and (h), the two lines are constant and would overlap. Thus, an offset is applied to improve visibility.}
  \label{fig:cd_ablation_history}
  \vspace{-2mm}
\end{figure}

Additionally, the histories of the recalculation interval $\Lambda$ and the vigilance threshold $V_{\text{threshold}}$ during learning are visualized to provide a better understanding of their adjustment behavior. Due to page limitations, Fig.~\ref{fig:cd_ablation_history} shows the result only for the STL10 dataset. The results for all datasets are provided in Figs.~\sfigref{sf:AblationHistoryNonstationary}-\sfigref{sf:AblationHistoryIncNonstationary_500} in the supplementary materials. All figures present the trial whose ARI is the median among the 11 runs.

% % (Ablation Study) History Lambda and V in IDAT
% \sfiglabel{sf:AblationHistoryNonstationary}
% Fig.~\sfigref{sf:AblationHistoryNonstationary}

% % (Ablation Study) History Lambda and V in w/o Decremental
% \sfiglabel{sf:AblationHistoryDecNonstationary}
% Fig.~\sfigref{sf:AblationHistoryDecNonstationary}

% % (Ablation Study) History Lambda and V in w/o Incremental
% \sfiglabel{sf:AblationHistoryIncNonstationary}
% Fig.~\sfigref{sf:AblationHistoryIncNonstationary}

As shown in Fig.~\ref{fig:cd_ablation_history}, the w/o All variant disables both adjustments of $\Lambda$ and $V_{\text{threshold}}$. As a result, these values remain fixed throughout learning. The w/o Inc. variant also keeps $\Lambda$ fixed at its initial value because the incremental adjustment is not applied. The w/o Dec. variant shows a continuous increase in $\Lambda$ because the decremental adjustment is disabled. This leads to an expanding buffer size during learning. A large $\Lambda$ makes the recalculation of $V_{\text{threshold}}$ infrequent and reduces its responsiveness to distributional changes. These results emphasize the essential role of adaptive adjustments in ensuring effective and stable clustering performance.

In contrast, IDAT adjusts $\Lambda$ adaptively based on the incoming samples. This adjustment prevents unnecessary buffer growth and facilitates the recalculation of $V_{\text{threshold}}$ as needed. The coordination between $\Lambda$ and $V_{\text{threshold}}$ is key to the robustness of IDAT. It prevents the limitations observed in the variants with fixed or one-directional adjustments. Moreover, Fig.~\ref{fig:cd_ablation_continual} (and also in Figs. \sfigref{sf:AblationHistoryNonstationary}-\sfigref{sf:AblationHistoryNonstationary_500} in the supplementary materials) shows that IDAT generally exhibits smaller differences between $\Lambda_{\text{init}} = 2$ and $\Lambda_{\text{init}} = 500$ compared to the other variants. This consistency indicates that the adaptive adjustments perform effectively across different initial settings. The other variants exhibit larger differences in $\Lambda$ and $V_{\text{threshold}}$. These observations further highlight the critical role of adaptive adjustments in maintaining stable and reliable clustering performance.

Note that although both initializations lead to stable clustering performance, $\Lambda_{\text{init}} = 2$ is preferred in practice. This initialization allows for more flexible adjustments in the early stages of learning, whereas $\Lambda_{\text{init}} = 500$ introduces a larger recalculation interval that delays necessary adjustments. Therefore, the results suggest that $\Lambda_{\text{init}} = 2$ enables more granular adaptation, which contributes to stable and high clustering performance.

\section{Concluding Remarks}
\label{sec:conclusion}

This paper presented IDAT, an ART-based topological clustering algorithm with diversity-driven adaptation of its recalculation interval $\Lambda$ and vigilance threshold $V_{\text{threshold}}$. Through extensive experiments on 24 real-world datasets, IDAT demonstrated superior clustering performance, continual learning capability, and robustness to distributional changes compared with a wide range of state-of-the-art clustering algorithms. These results indicate that the adaptive adjustment of $\Lambda$ and $V_{\text{threshold}}$ effectively balances stability and plasticity, enabling the model to maintain consistent clusters without manual tuning. Furthermore, the self-adjusting mechanism contributes to mitigating catastrophic forgetting, which is essential for long-term online learning in dynamic environments. In addition, the parameter-free adaptation of $\Lambda$ and $V_{\text{threshold}}$ enhances the practical usefulness of IDAT by removing the need for manual hyperparameter tuning in real-world streaming scenarios. Overall, these findings highlight the effectiveness of the proposed adaptation mechanisms in maintaining cluster consistency and ensuring robust continual learning in dynamic data streams.

As future work, the adaptive learning mechanism will be extended to concept drift-aware learning to better regulate cluster growth under shifting data distributions. Another direction is to evaluate the algorithm on more challenging datasets, including many-class and class-imbalance settings, to clarify its robustness in diverse scenarios.

% or


%\appendices
%% you can choose not to have a title for an appendix
%% if you want by leaving the argument blank
%\section{}

%\ifCLASSOPTIONcompsoc
%  % The Computer Society usually uses the plural form
%\else
%  % regular IEEE prefers the singular form
  \section*{Acknowledgment}
%\fi
% This work was supported by the Japan Society for the Promotion of Science (JSPS) KAKENHI Grant Number JP25K03187.

The authors used an AI system (OpenAI ChatGPT, version GPT-5.1) to assist in the refinement of English expression and the clarification of several explanations in this manuscript. All technical content, experimental design, and conclusions were created and verified by the authors.

\ifCLASSOPTIONcaptionsoff
  \newpage
\fi



\bibliographystyle{IEEEtran}
\bibliography{myref}

\clearpage

% Reset section counter
\setcounter{section}{0}     % Restart numbering from 1
\renewcommand{\thesection}{S\arabic{section}} % Optional prefix “S” (e.g., S1, S2)

% Optional: switch to one column mode for supplementary
% \onecolumn

% Supplementary title
\section*{Supplementary Materials}

% If you want the first numbered section to start right after this:
\addcontentsline{toc}{section}{Supplementary Materials}

% Supplementary: use S-prefix for tables/figures/equations
\setcounter{table}{0}
\setcounter{figure}{0}
\setcounter{equation}{0}
\renewcommand{\thetable}{S\arabic{table}}
\renewcommand{\thefigure}{S\arabic{figure}}
\renewcommand{\theequation}{S\arabic{equation}}

This supplementary material provides additional implementation details and extended experimental results that complement the main paper. It describes the deep clustering baselines (TableDC and G-CEALS), the optimized hyperparameters, and the full quantitative comparisons under both stationary and nonstationary settings. The ablation study of the proposed IDAT algorithm is also included.

\section{Implementation Details of TableDC and G-CEALS}
\label{sec:deepImplementation}

TableDC~\cite{rauf25} and G-CEALS~\cite{rabbani25} are deep clustering methods designed for tabular data. Both methods employ autoencoders to learn latent feature representations that improve clustering performance. TableDC incorporates Mahalanobis distance and a Cauchy distribution for noise-robust feature correlation. G-CEALS employs Gaussian mixture regularization in the latent space to enhance cluster separability. 

In TableDC, we use a symmetric fully connected autoencoder of $d$-500-500-2000-10-2000-500-500-$d$ with ReLU activations, a latent dimension of 10, and the Adam optimizer (learning rate 0.001, batch size 256). The autoencoder is pretrained on raw features for up to 500 epochs with early stopping (patience 20, improvement threshold $10^{-5}$). The clustering module is then trained for 200 epochs with Adam (learning rate 0.001). Latent-space centers are initialized through an adaptive Birch-based threshold search that adjusts the clustering threshold until the number of subclusters falls within $[k, 2k)$, where $k$ is the number of classes in the dataset, and the resulting subcluster means are used as initial centers. The model minimizes a joint objective that combines the reconstruction mean squared error and the Kullback-Leibler divergence between the predicted and target distributions.

For G-CEALS, we adopt the same autoencoder architecture and a latent dimension of 10, using Adam with a learning rate of 0.001 and a batch size of 256. The autoencoder is pretrained for 1,000 epochs and then fine-tuned jointly with the clustering loss for another 1,000 epochs (update interval 1, clustering weight $\gamma=0.1$), with cluster centroids initialized by $k$-means in the latent space and the number of clusters fixed to the number of classes in the dataset.

\section{Additional Results for Hyperparameter Settings}
\label{sec:hyperparams}
The hyperparameters of SR-PCM-HDP\cite{hu25}, TBM-P~\cite{yelugam23}, GNG~\cite{fritzke95}, DDVFA~\cite{da20}, and CAEA~\cite{masuyama22a} are optimized using the \texttt{bayesopt} function provided in the Statistics and Machine Learning Toolbox of MATLAB\footnote{\url{https://www.mathworks.com/help/stats/bayesopt.html}}. In both stationary and nonstationary settings, Bayesian optimization is performed for 25 iterations to maximize the average Adjusted Rand Index (ARI) obtained over ten independent runs with different random seeds. In each run, the model is trained and evaluated using the parameter set proposed by the optimizer, and the mean ARI across runs served as the objective value. Other aspects of the experimental procedure, including the optimization protocol and evaluation criteria, are kept identical across the two settings.

All optimization experiments are conducted in MATLAB~R2025b on an Apple M2 Ultra processor with 128~GB of RAM running macOS.

Tables~\ref{tab:srpcmdp-param-def}-\ref{tab:caea-param-def} list the hyperparameter definitions for SR-PCM-HDP, TBM-P, GNG, DDVFA, and CAEA. Tables~\ref{tab:srpcmdp-stationary}-\ref{tab:caea-stationary} present the optimized hyperparameters of these algorithms under the stationary setting, and Tables~\ref{tab:gng-nonstationary}-\ref{tab:caea-nonstationary} show those obtained under the nonstationary setting.

The optimized parameter values vary greatly across datasets under both stationary and nonstationary settings. These results suggest that the performance of these algorithms strongly depends on how the parameters are chosen. Therefore, algorithms that work well without careful parameter tuning have a clear advantage because they can maintain high performance without additional optimization.

\section{Additional Results for Clustering Performance in the Stationary Setting}
\label{sec:comparison_stationary}

This section presents the results of quantitative comparisons in the stationary setting across 12 algorithms. Since the clustering process is unsupervised, the same dataset is used for both training and testing, and class labels are used only for evaluating clustering performance. The results are averaged over 30 independent runs with different random seeds to ensure fair and reliable comparisons.

Table \ref{tab:ResultsClusteringStationary} summarizes the average ARI and AMI. In the table, the values in parentheses indicate the standard deviation, a number to the right of a metric value is the average rank of an algorithm, and the smaller the rank, the better the metric score.

\section{Additional Results for Clustering Performance in the Nonstationary Setting}
\label{sec:comparison_nonstationary}

This section presents the results of quantitative comparisons in the nonstationary setting across six algorithms. Similar to the stationary setting, the same dataset is used for both training and testing, and class labels are used only for evaluating clustering performance. The results are averaged over 30 independent runs with different random seeds to ensure fair and reliable comparisons.

Table \ref{tab:ResultsClusteringNonstationary} summarizes the average ARI and AMI. In the table, the values in parentheses indicate the standard deviation, a number to the right of a metric value is the average rank of an algorithm, and the smaller the rank, the better the metric score.

\section{Additional Results for Continual Learning Performance}
\label{sec:comparison_continual}

This section focuses on the continual learning performance, i.e., Average Incremental (AI) and Backward Transfer (BWT). While Sections~\ref{sec:comparison_stationary} and \ref{sec:comparison_nonstationary} focused on the final clustering performance, an important aspect of continual learning is to preserve past knowledge and prevent forgetting. Here, the ARI and Adjusted Mutual Information (AMI) values obtained after learning each class in Section~\ref{sec:comparison_nonstationary} are used as the base performance measures for computing AI and BWT. The results are therefore reported as AI-ARI, AI-AMI, BWT-ARI, and BWT-AMI.

Table \ref{tab:IncBwtNonstationary} summarizes the average AI-ARI, AI-AMI, BWT-ARI, and BWT-AMI. In the table, the values in parentheses indicate the standard deviation, a number to the right of a metric value is the average rank of an algorithm, and the smaller the rank, the better the metric score.

\section{Additional Results for Ablation Study}
\label{sec:ablation}

This section presents additional results for ablation study, which include the numerical values of ARI, AMI, AI-ARI, AI-AMI, BWT-ARI, BWT-AMI, and the number of nodes and clusters in the nonstationary setting.

The ablation study evaluates the contribution of the diversity-driven adaptation of the recalculation interval $\Lambda$ and the vigilance threshold $V_{\text{threshold}}$ in IDAT to clustering and continual learning performance. Ablation variants disable the decremental direction that shortens $\Lambda$ when the diversity condition fails, the incremental direction that enlarges $\Lambda$ when it holds, or both. Each variant still recomputes $V_{\text{threshold}}$ under the remaining mechanism.

% Table \ref{tab:AblationStationary}

Table \ref{tab:AblationNonstationary} shows the final clustering performance, i.e., ARI and AMI. Table \ref{tab:AblationNonstationary_IncBWT} shows the continual learning performance, i.e., AI-ARI, AI-AMI, BWT-ARI, BWT-AMI. Table \ref{tab:AblationNonstationary_NodesClusters} shows the number of nodes and clusters.

\begin{figure*}[htbp]
  \centering
  \subfloat[Iris]{%
    \includegraphics[width=0.22\linewidth]{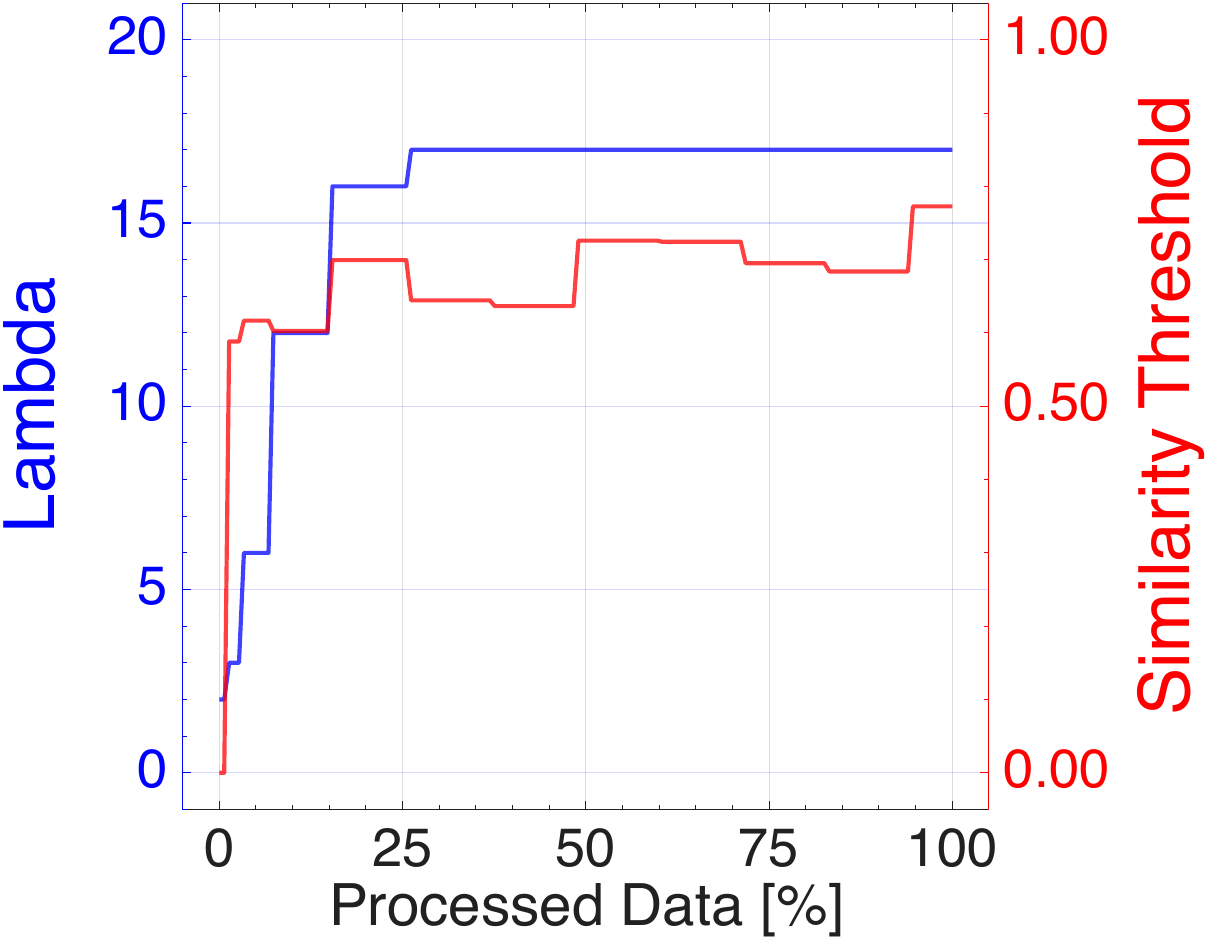}
  }\hfill
  \subfloat[Seeds]{%
    \includegraphics[width=0.22\linewidth]{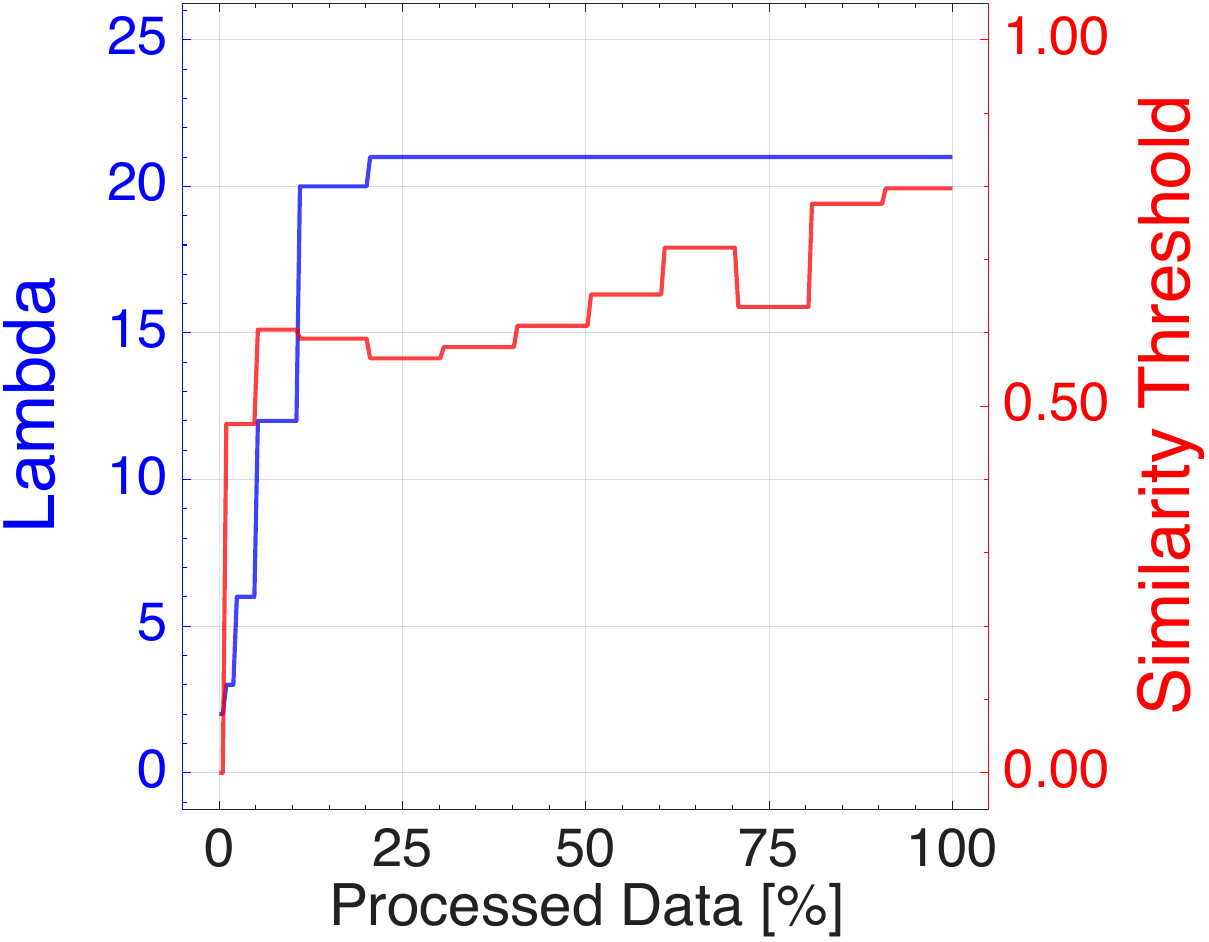}
  }\hfill
  \subfloat[Dermatology]{%
    \includegraphics[width=0.22\linewidth]{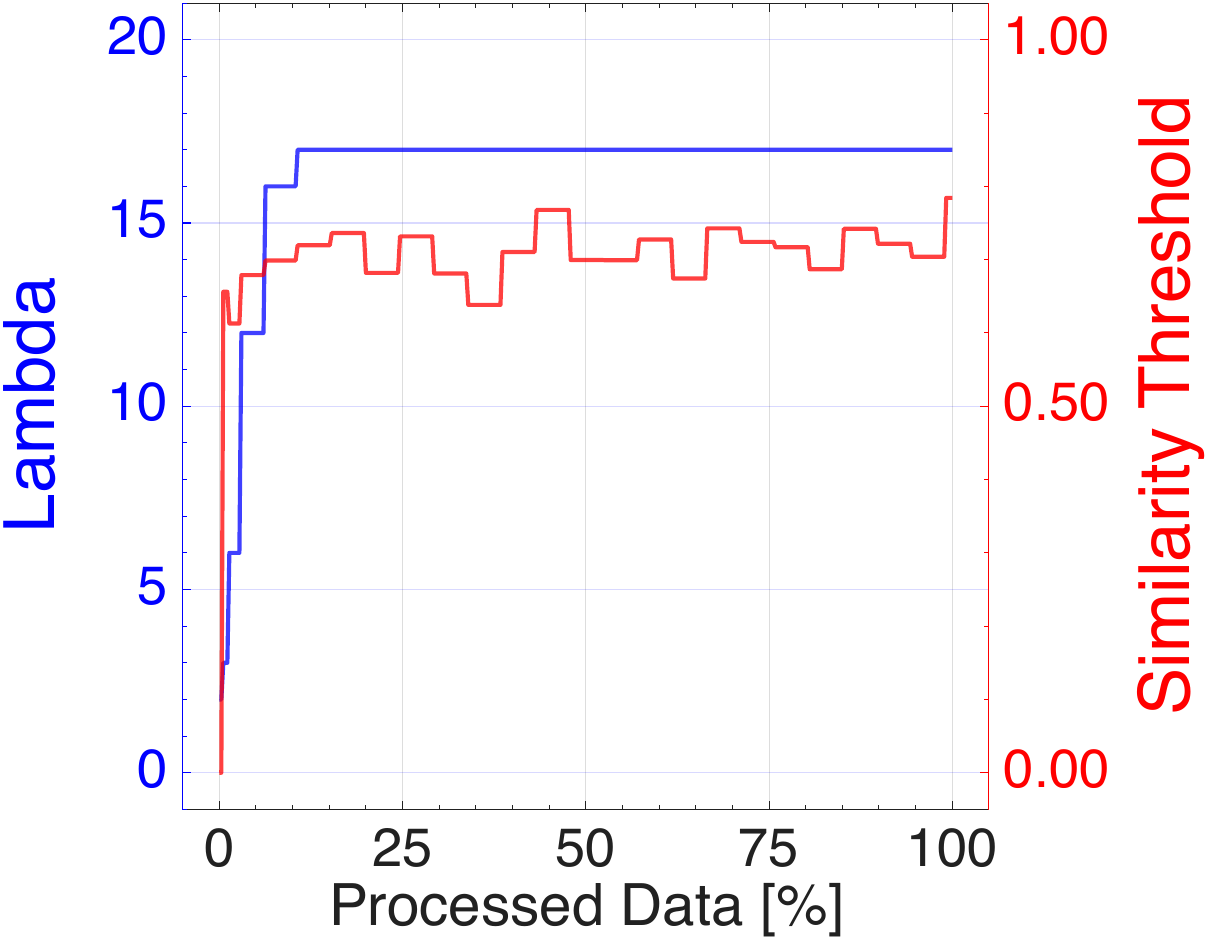}
  }\hfill
  \subfloat[Pima]{%
    \includegraphics[width=0.22\linewidth]{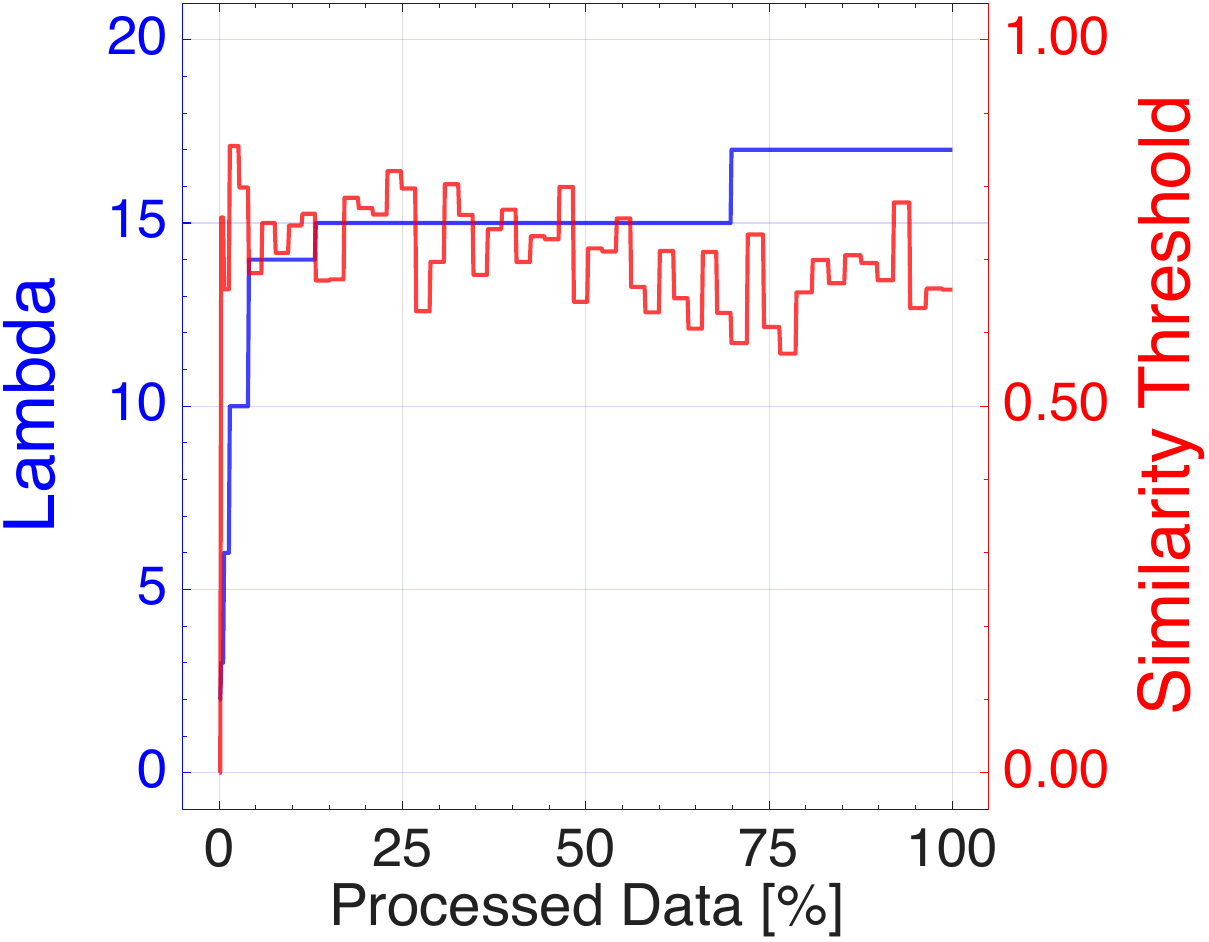}
  }\\
  \subfloat[Mice Protein]{%
    \includegraphics[width=0.22\linewidth]{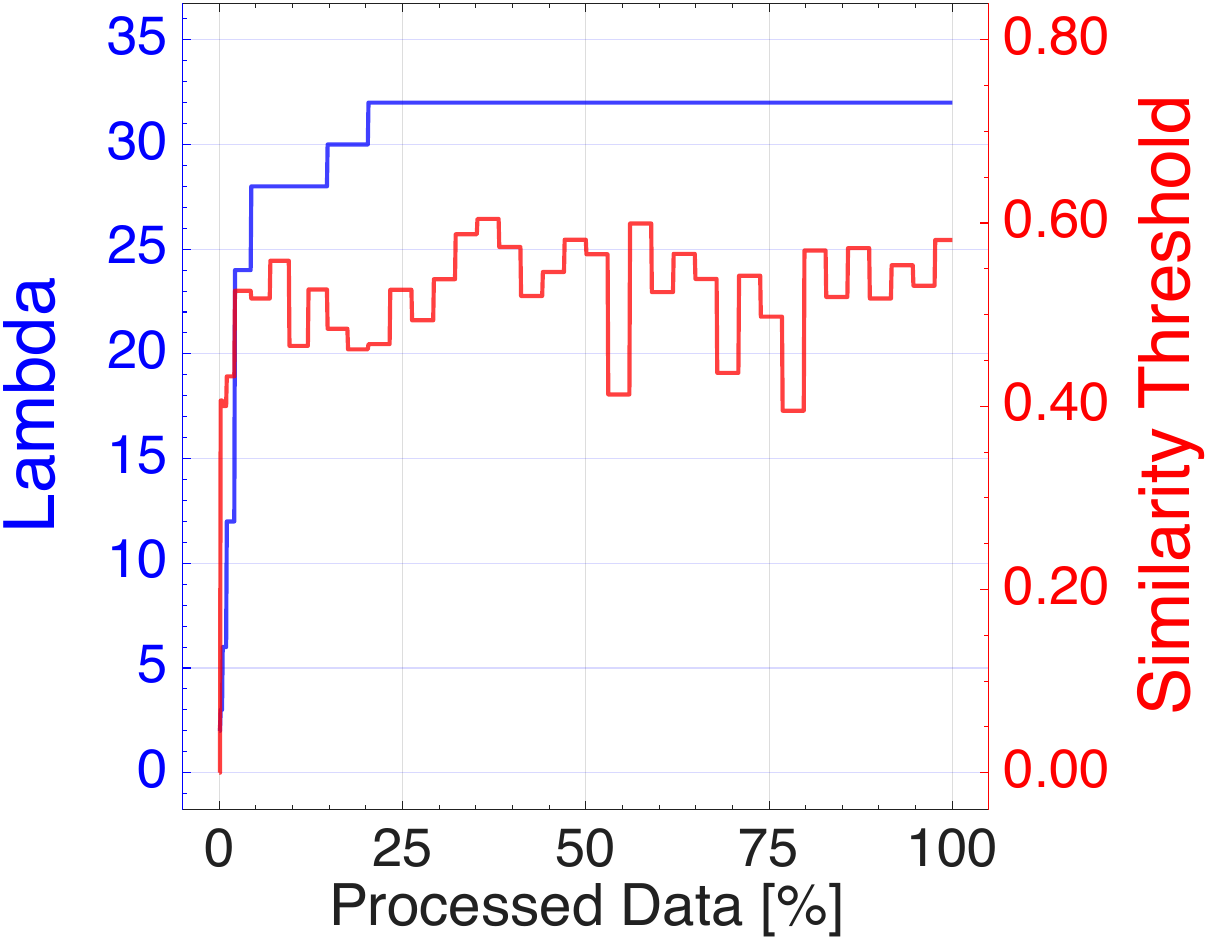}
  }\hfill
  \subfloat[Binalpha]{%
    \includegraphics[width=0.22\linewidth]{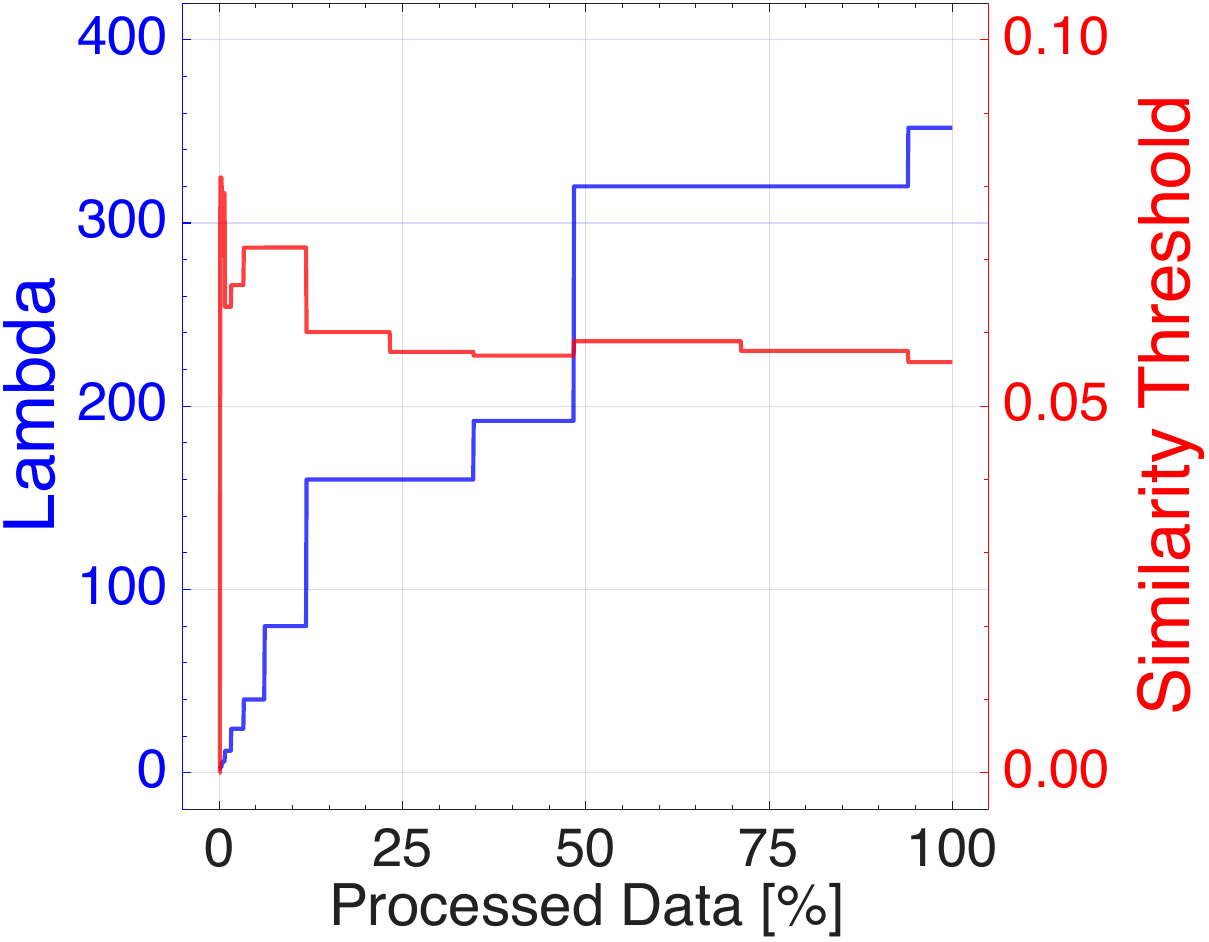}
  }\hfill
  \subfloat[Yeast]{%
    \includegraphics[width=0.22\linewidth]{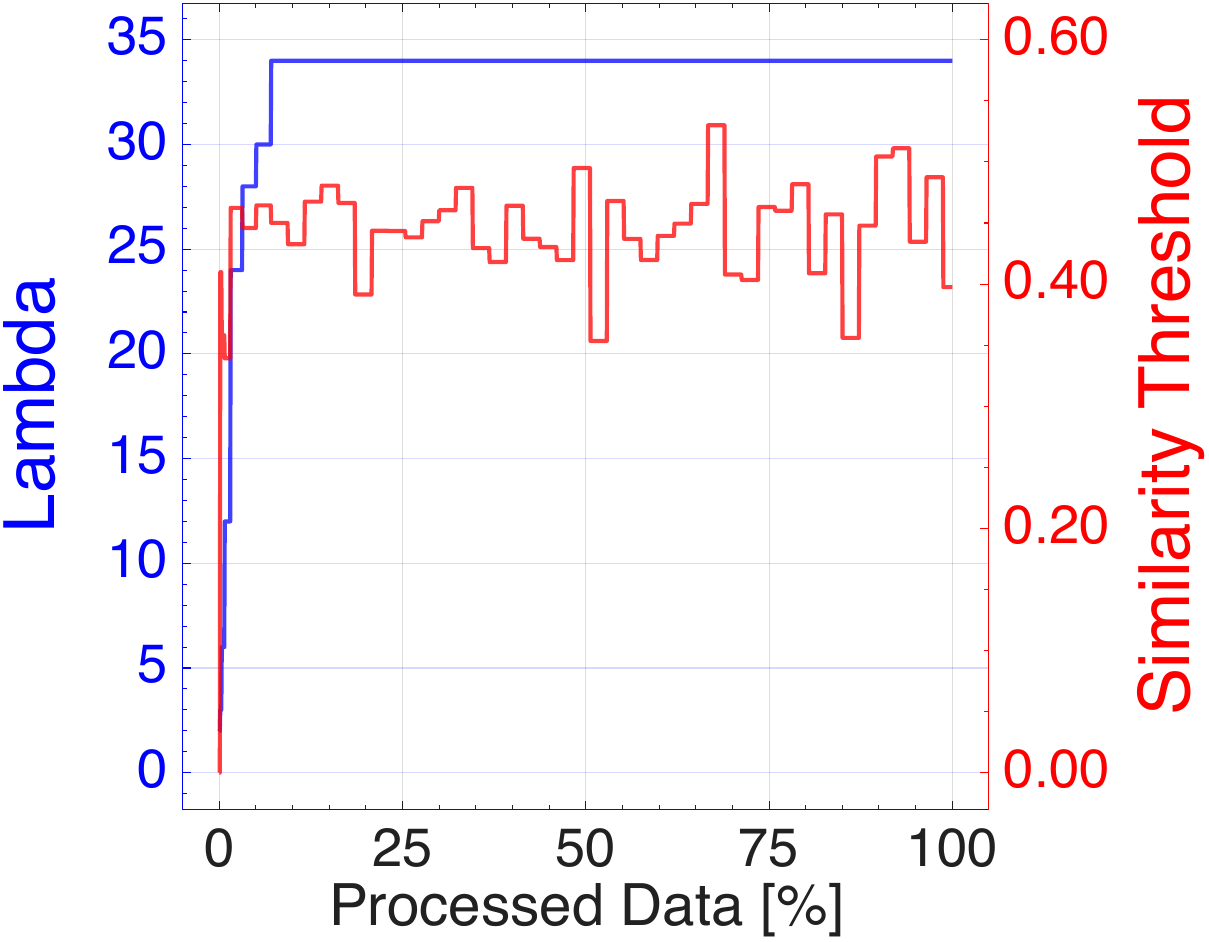}
  }\hfill
  \subfloat[Semeion]{%
    \includegraphics[width=0.22\linewidth]{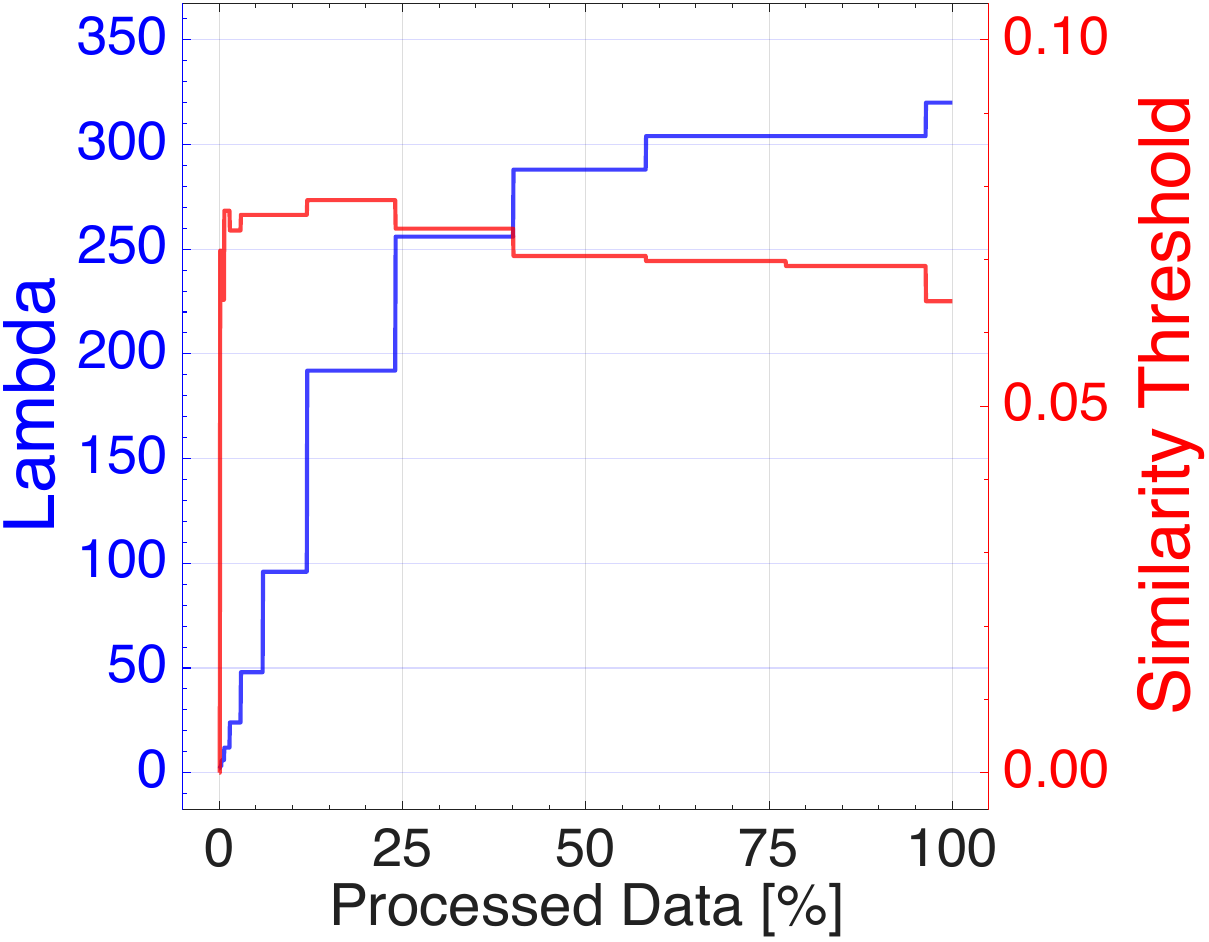}
  }\\
  \subfloat[MSRA25]{%
    \includegraphics[width=0.22\linewidth]{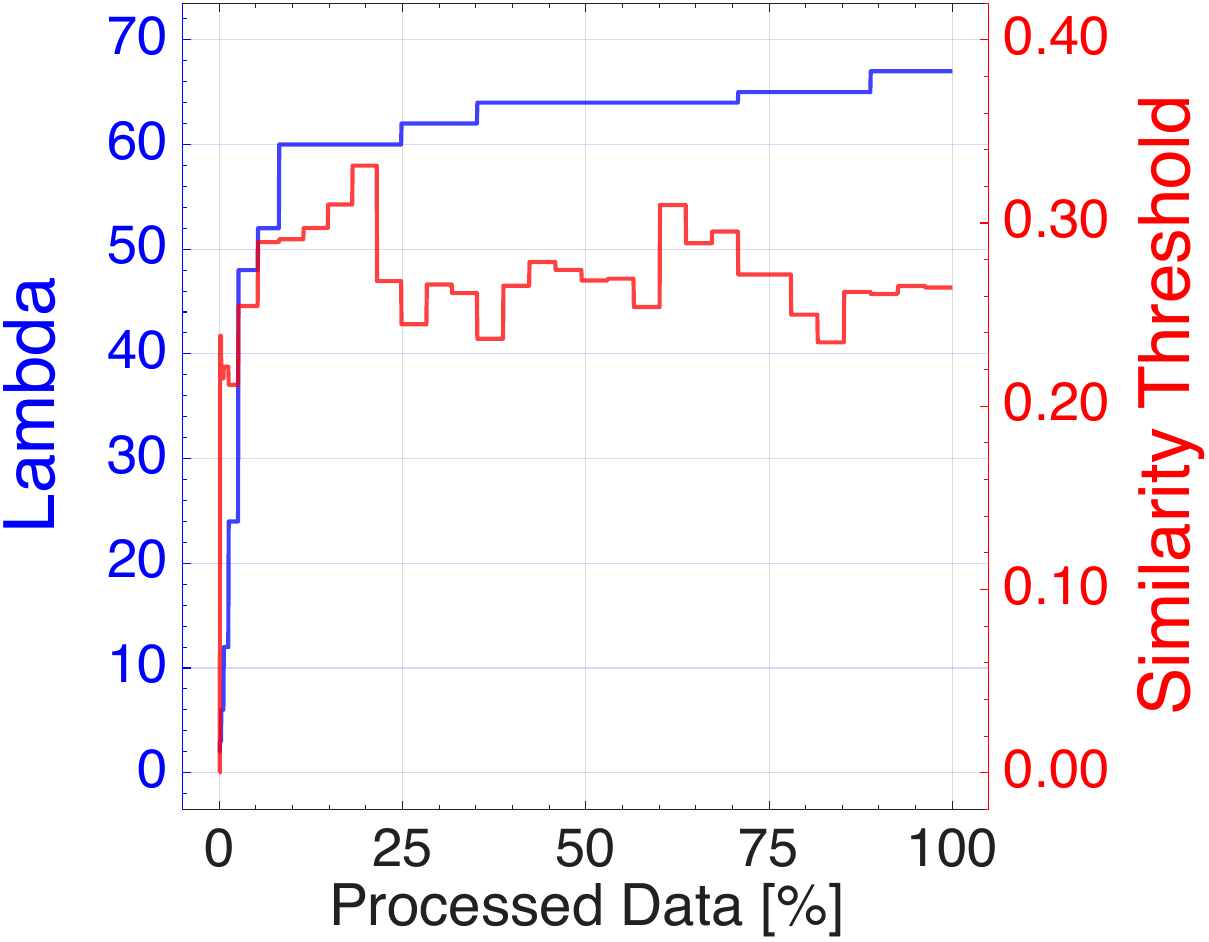}
  }\hfill
  \subfloat[Image Segmentation]{%
    \includegraphics[width=0.22\linewidth]{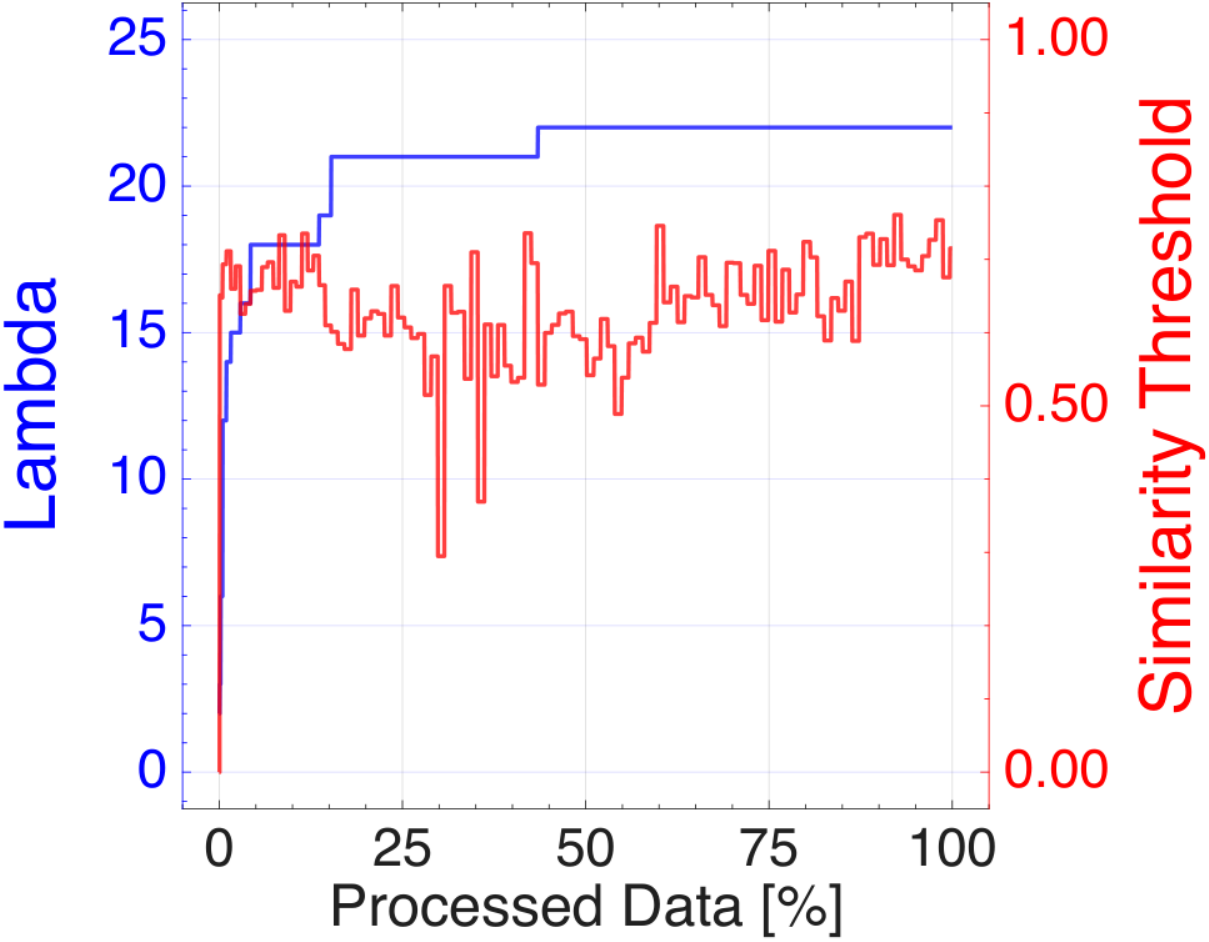}
  }\hfill
  \subfloat[Rice]{%
    \includegraphics[width=0.22\linewidth]{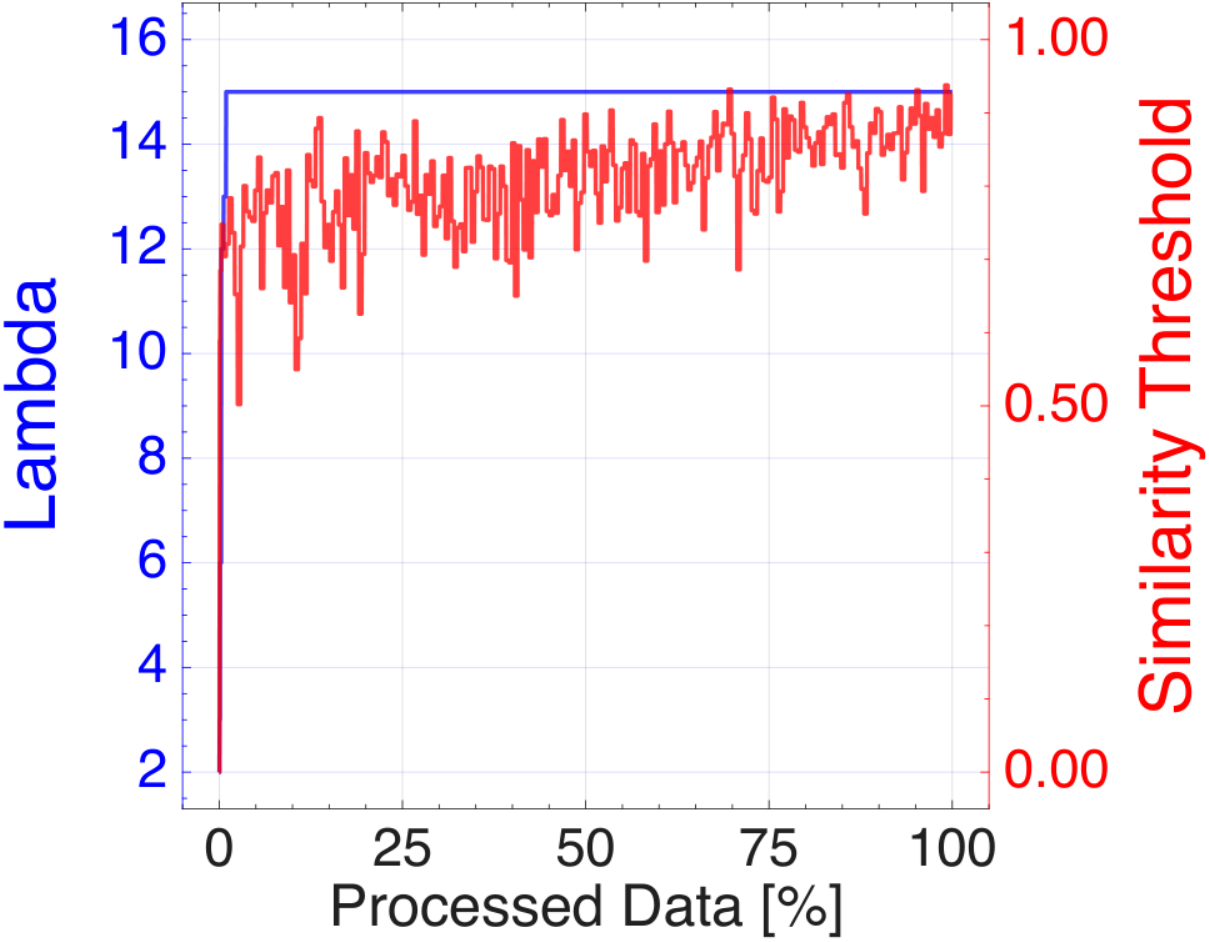}
  }\hfill
  \subfloat[TUANDROMD]{%
    \includegraphics[width=0.22\linewidth]{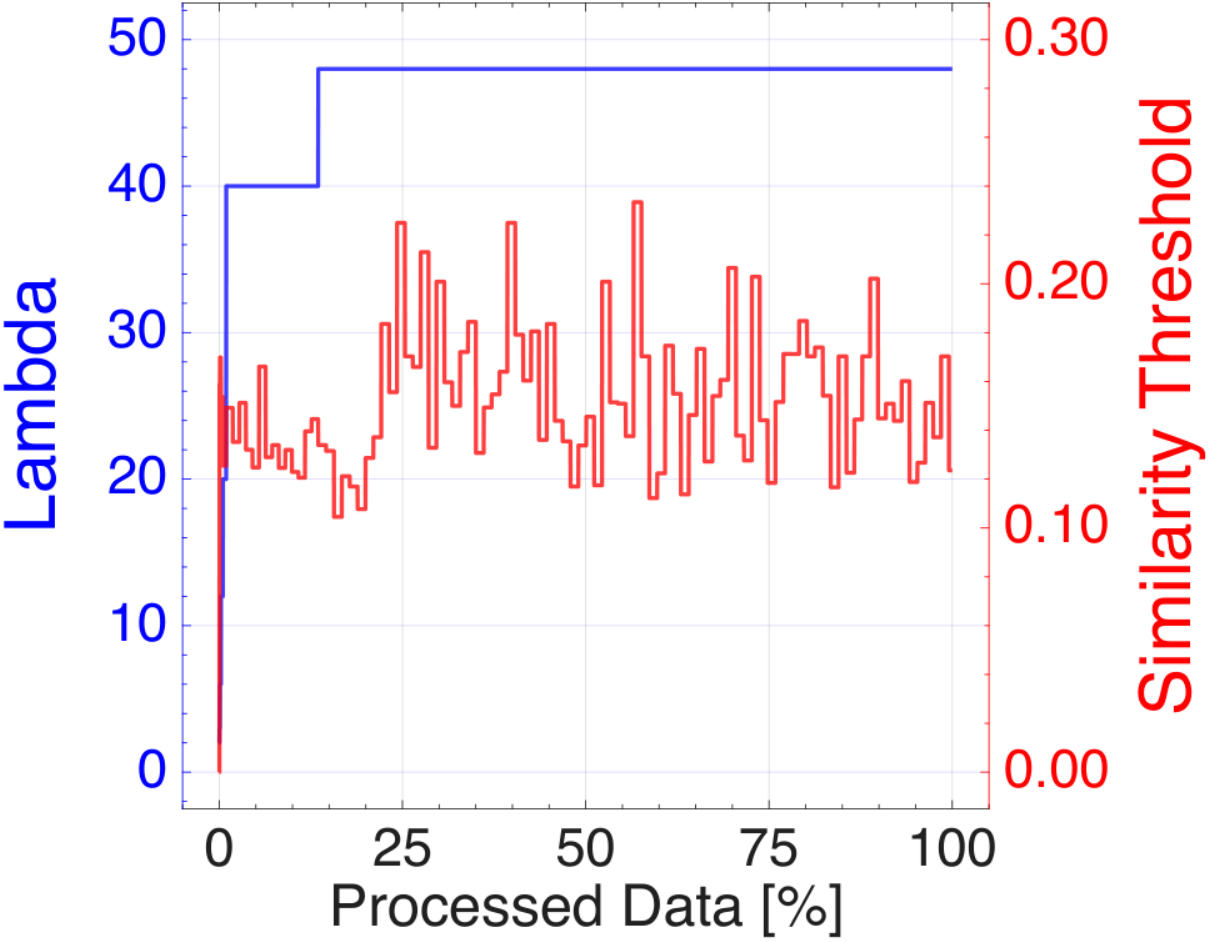}
  }\\
  \subfloat[Phoneme]{%
    \includegraphics[width=0.22\linewidth]{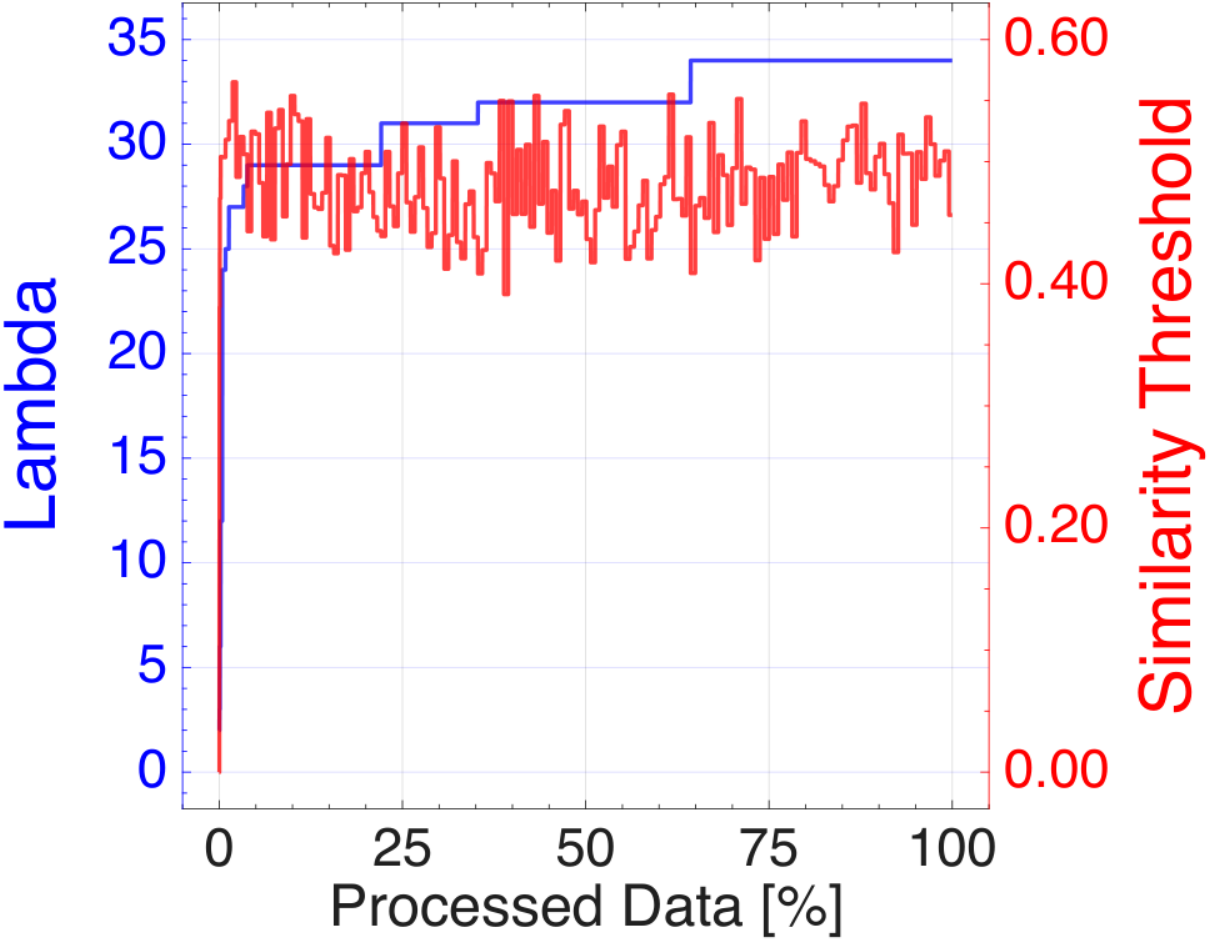}
  }\hfill
  \subfloat[Texture]{%
    \includegraphics[width=0.22\linewidth]{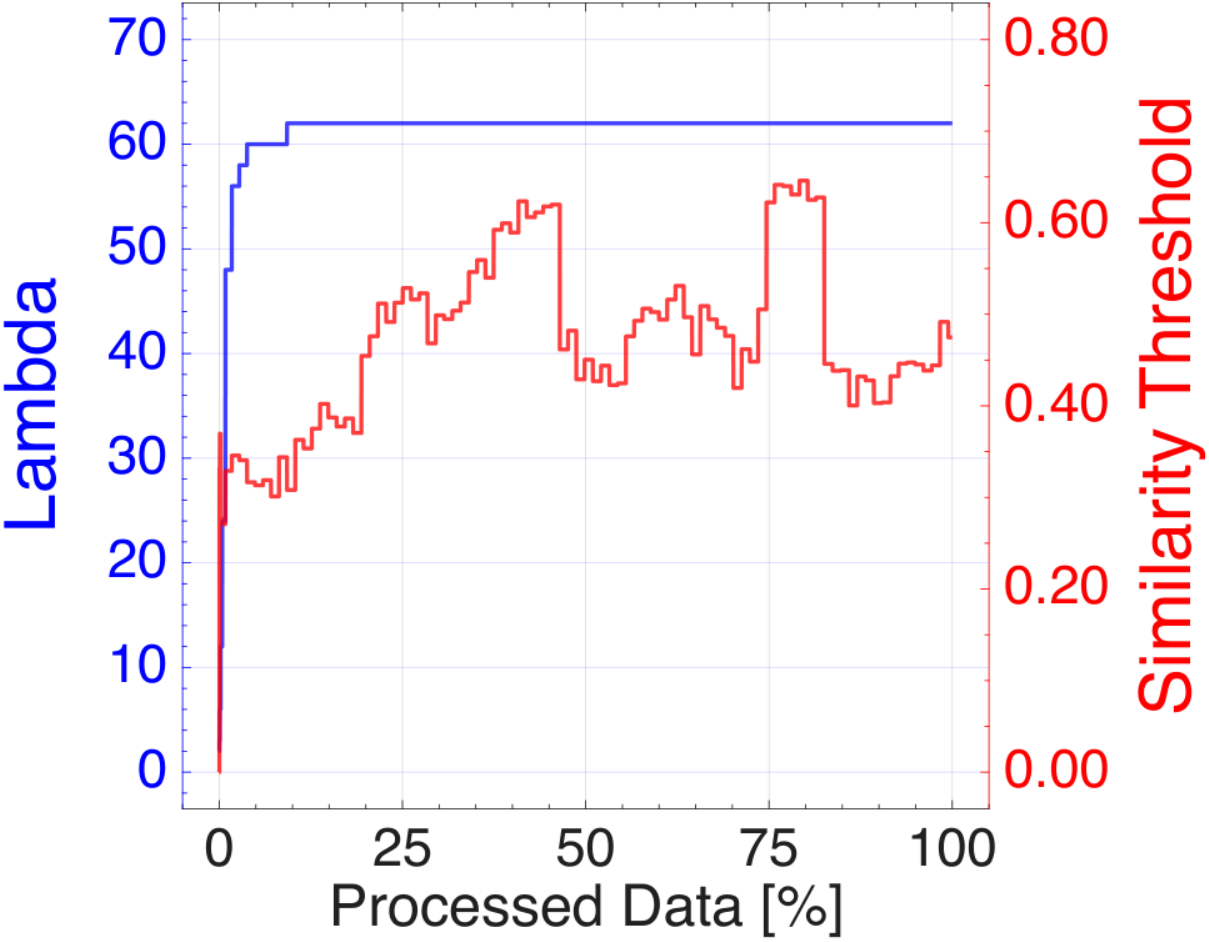}
  }\hfill
  \subfloat[OptDigits]{%
    \includegraphics[width=0.22\linewidth]{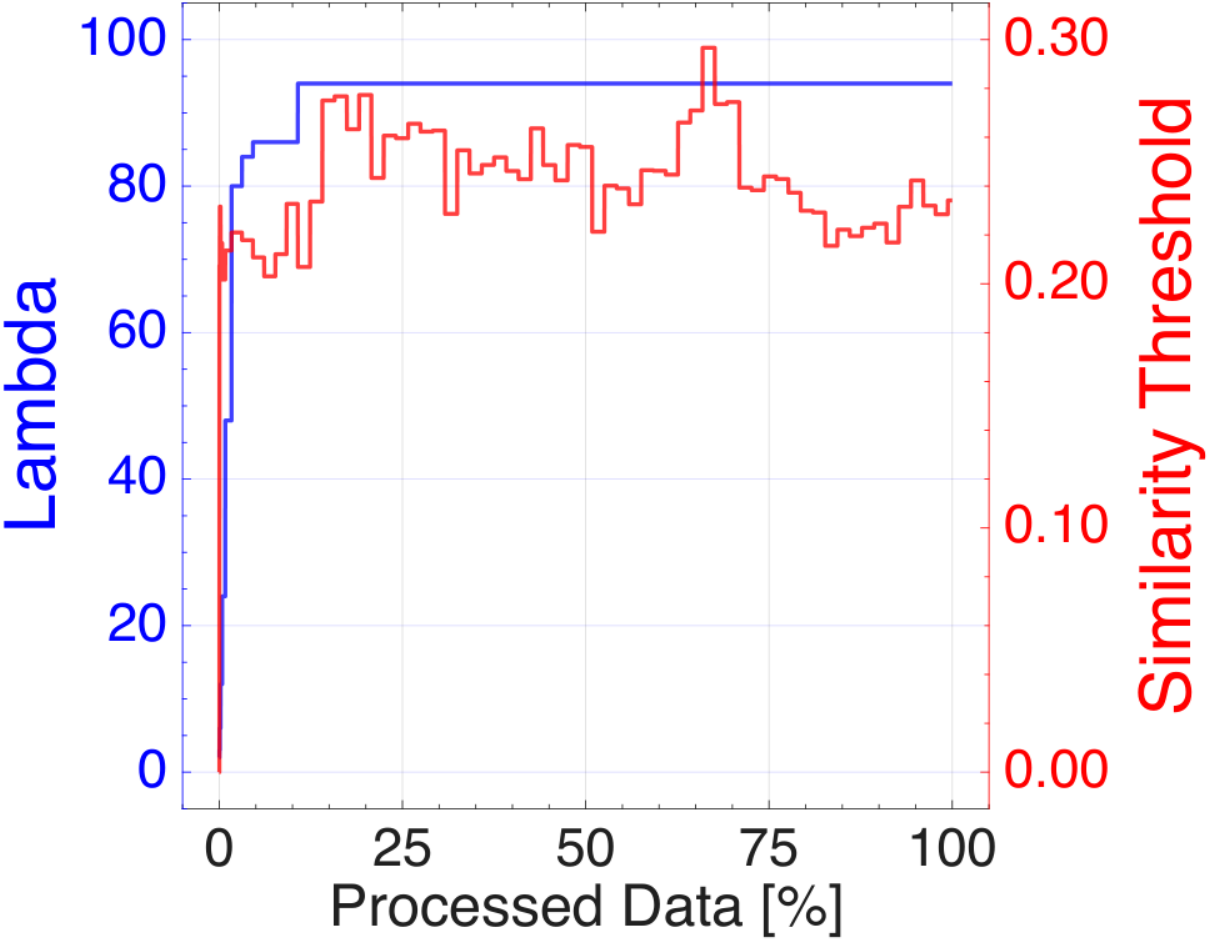}
  }\hfill
  \subfloat[Statlog]{%
    \includegraphics[width=0.22\linewidth]{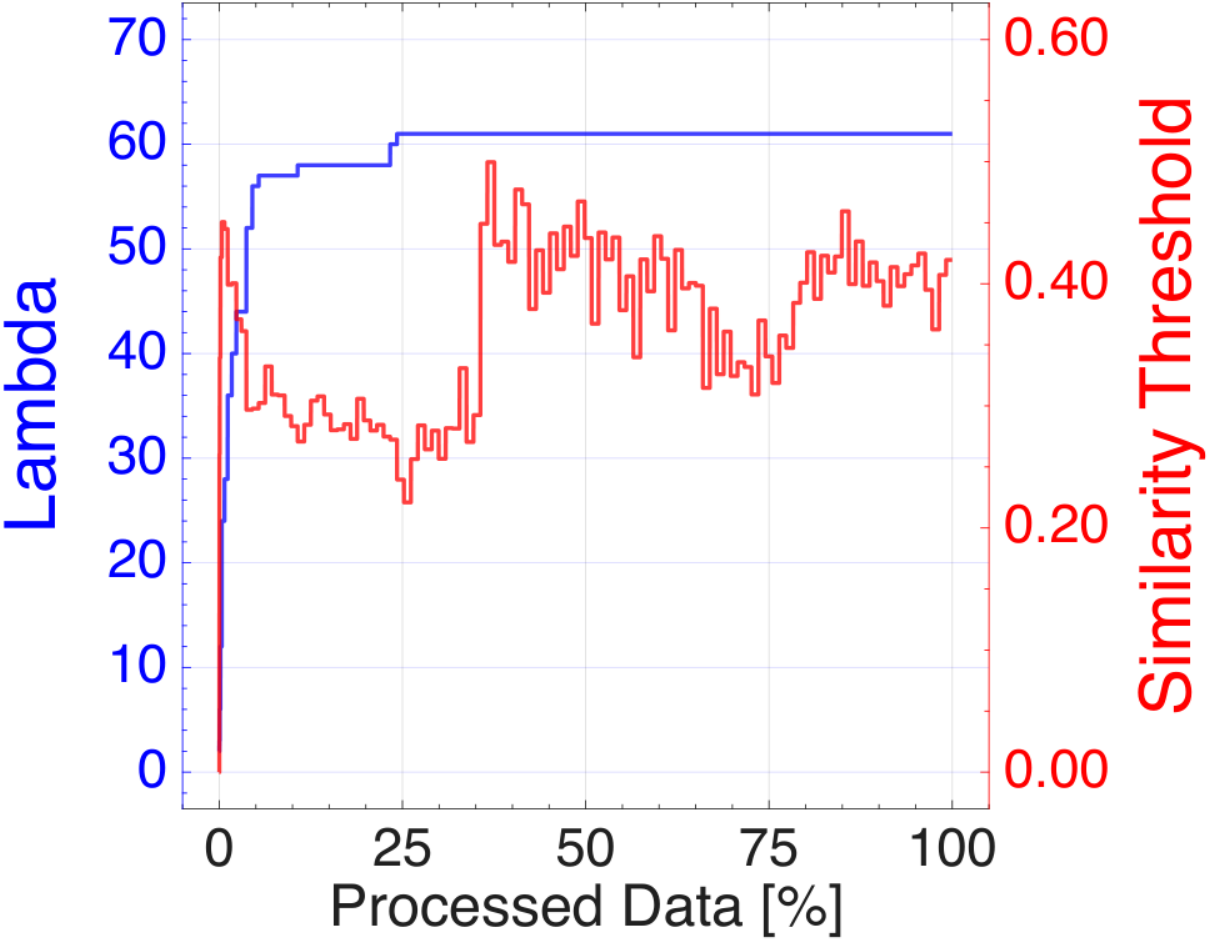}
  }\\
  \subfloat[Anuran Calls]{%
    \includegraphics[width=0.22\linewidth]{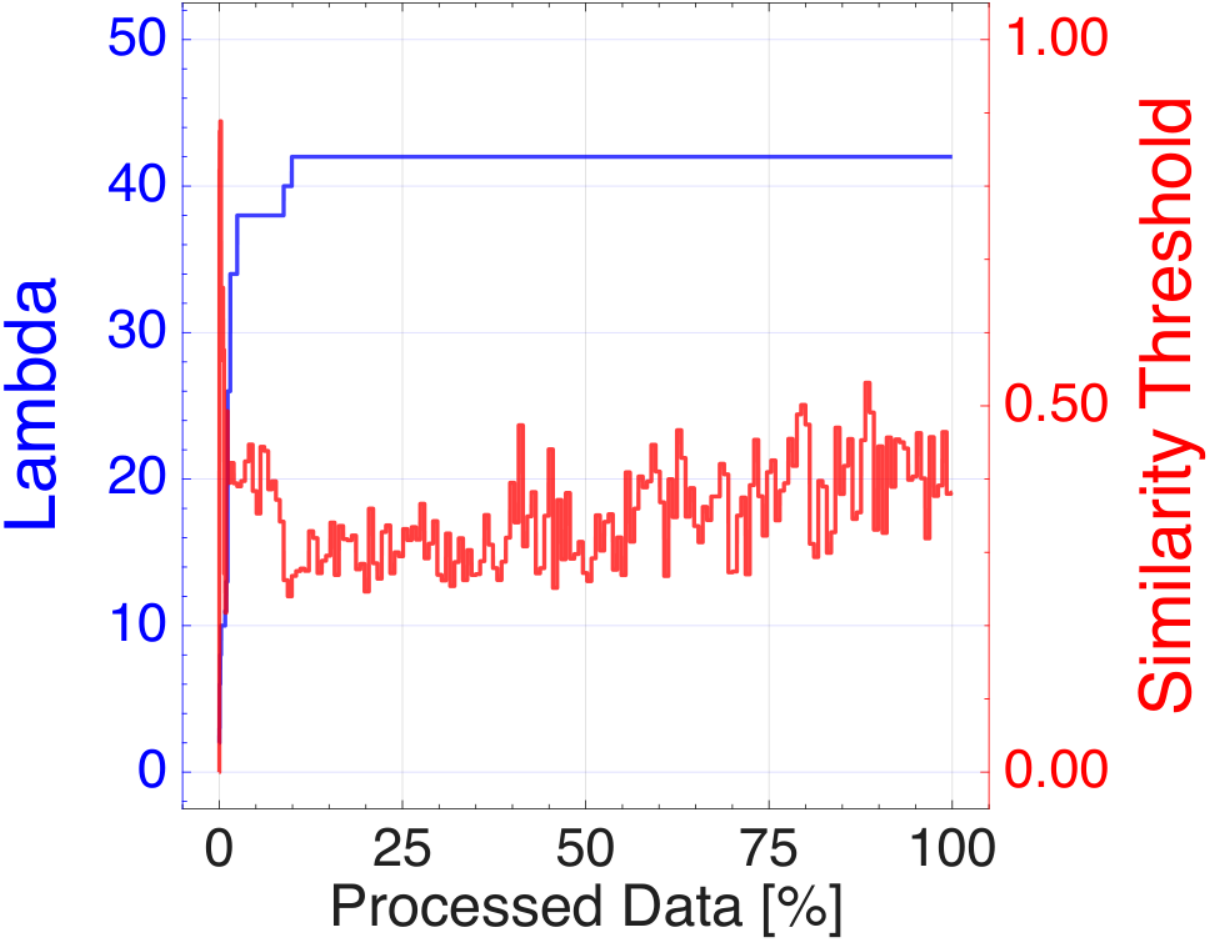}
  }\hfill
  \subfloat[Isolet]{%
    \includegraphics[width=0.22\linewidth]{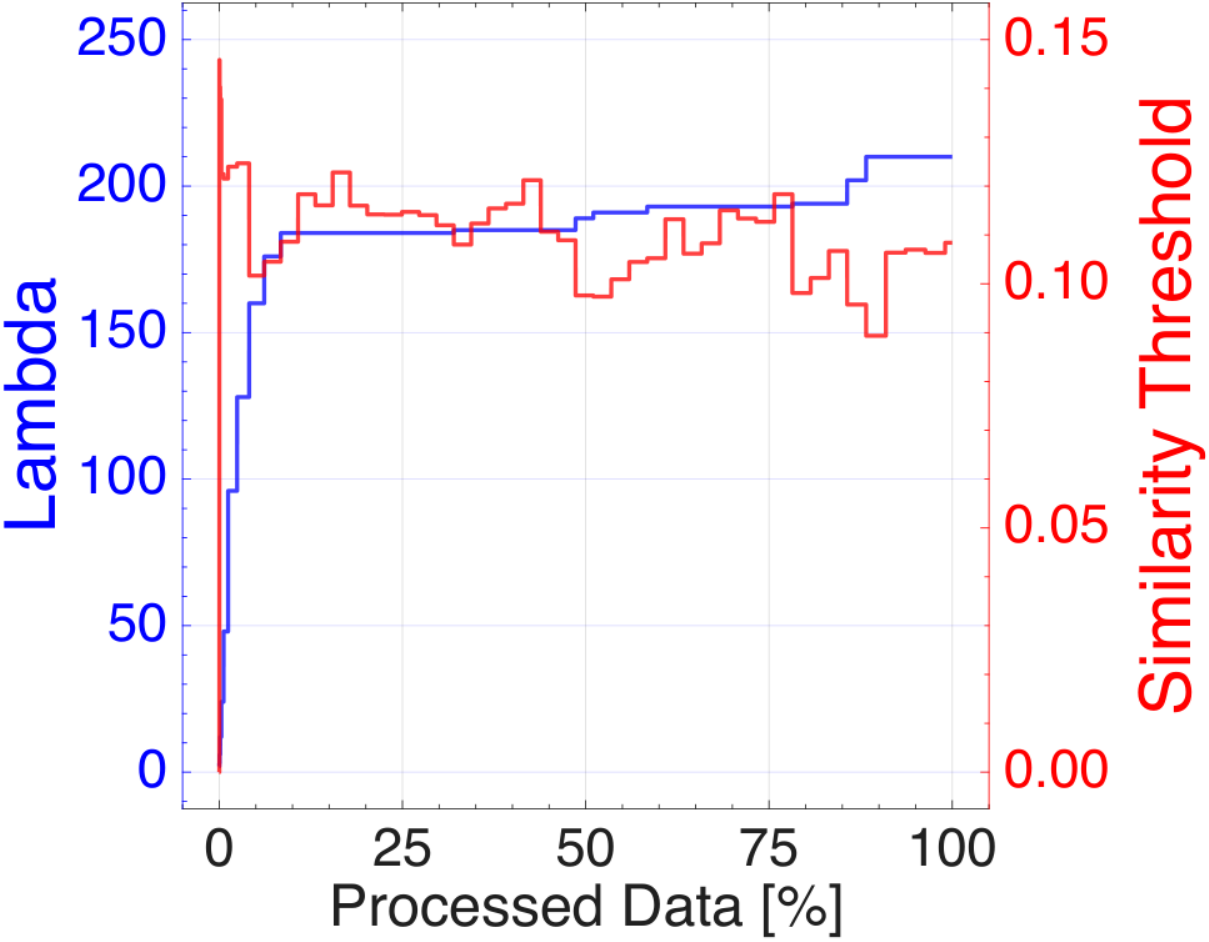}
  }\hfill
  \subfloat[MNIST10K]{%
    \includegraphics[width=0.22\linewidth]{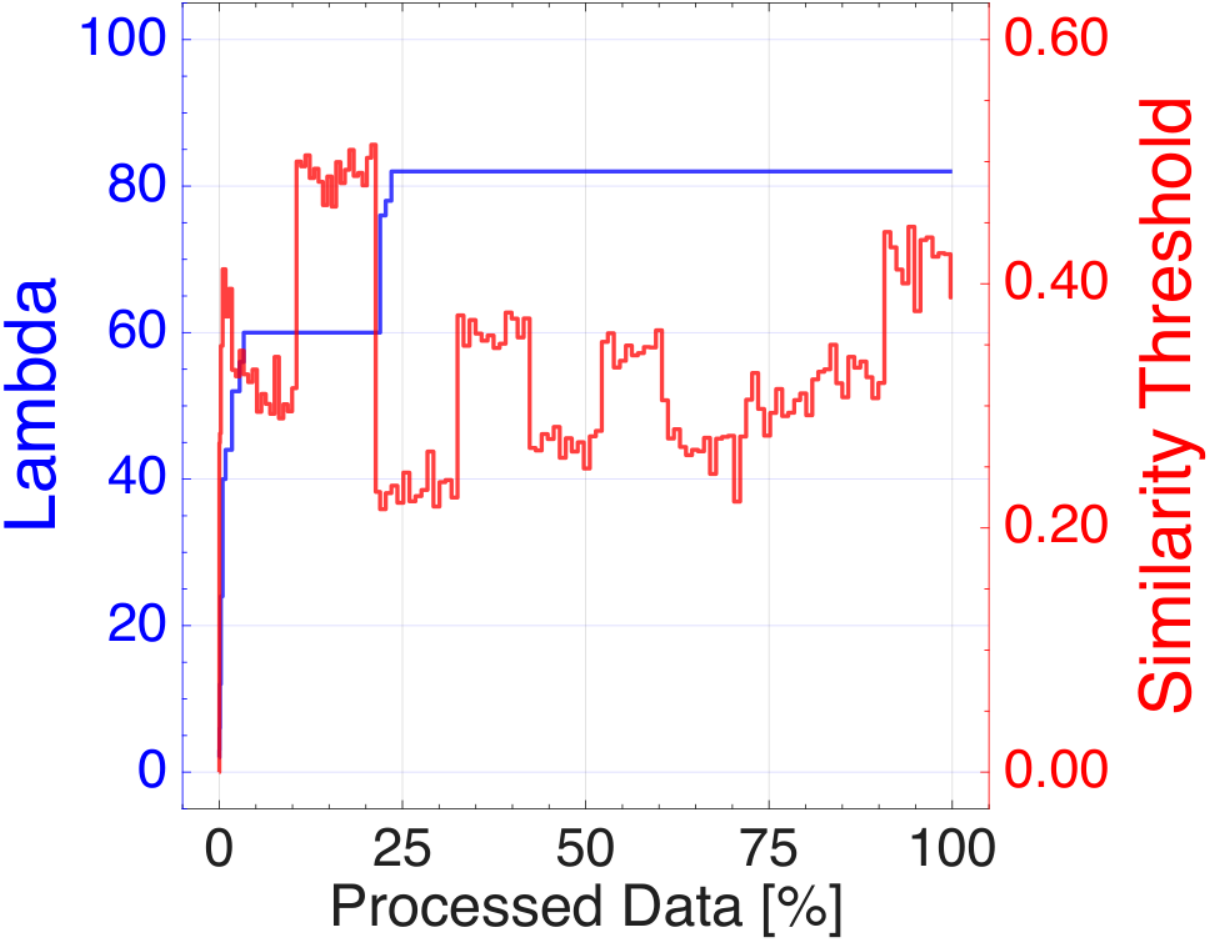}
  }\hfill
  \subfloat[PenBased]{%
    \includegraphics[width=0.22\linewidth]{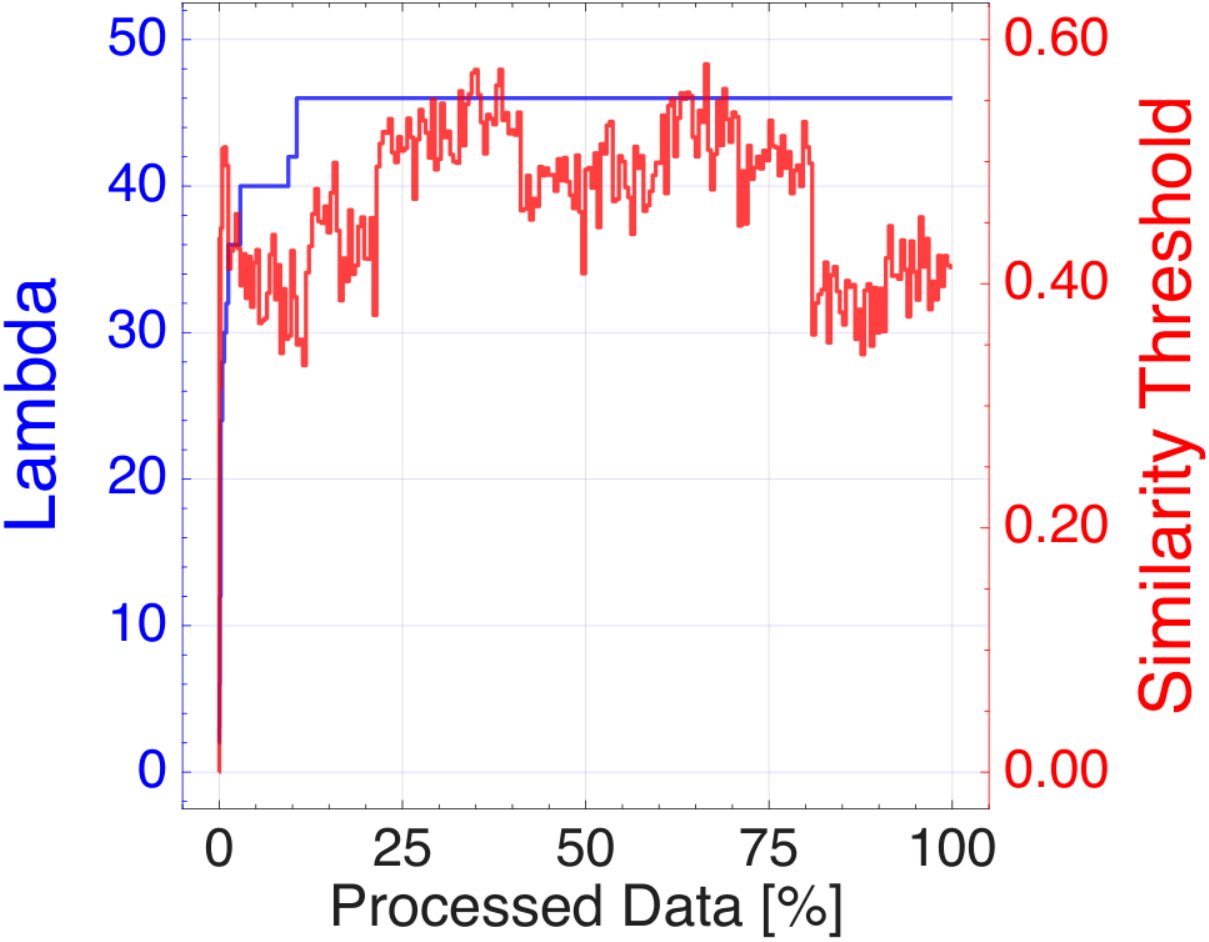}
  }\\
  \subfloat[STL10]{%
    \includegraphics[width=0.22\linewidth]{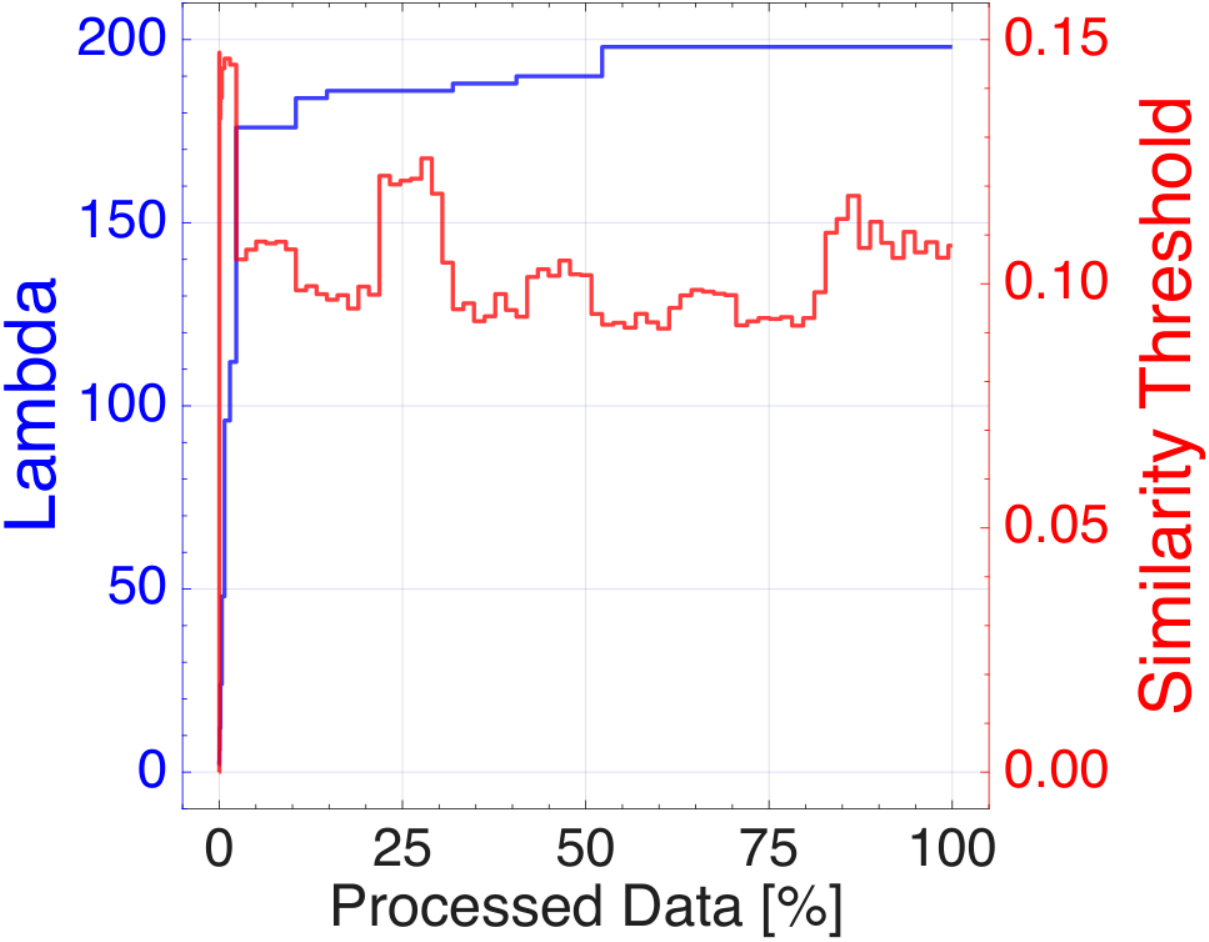}
  }\hfill
  \subfloat[Letter]{%
    \includegraphics[width=0.22\linewidth]{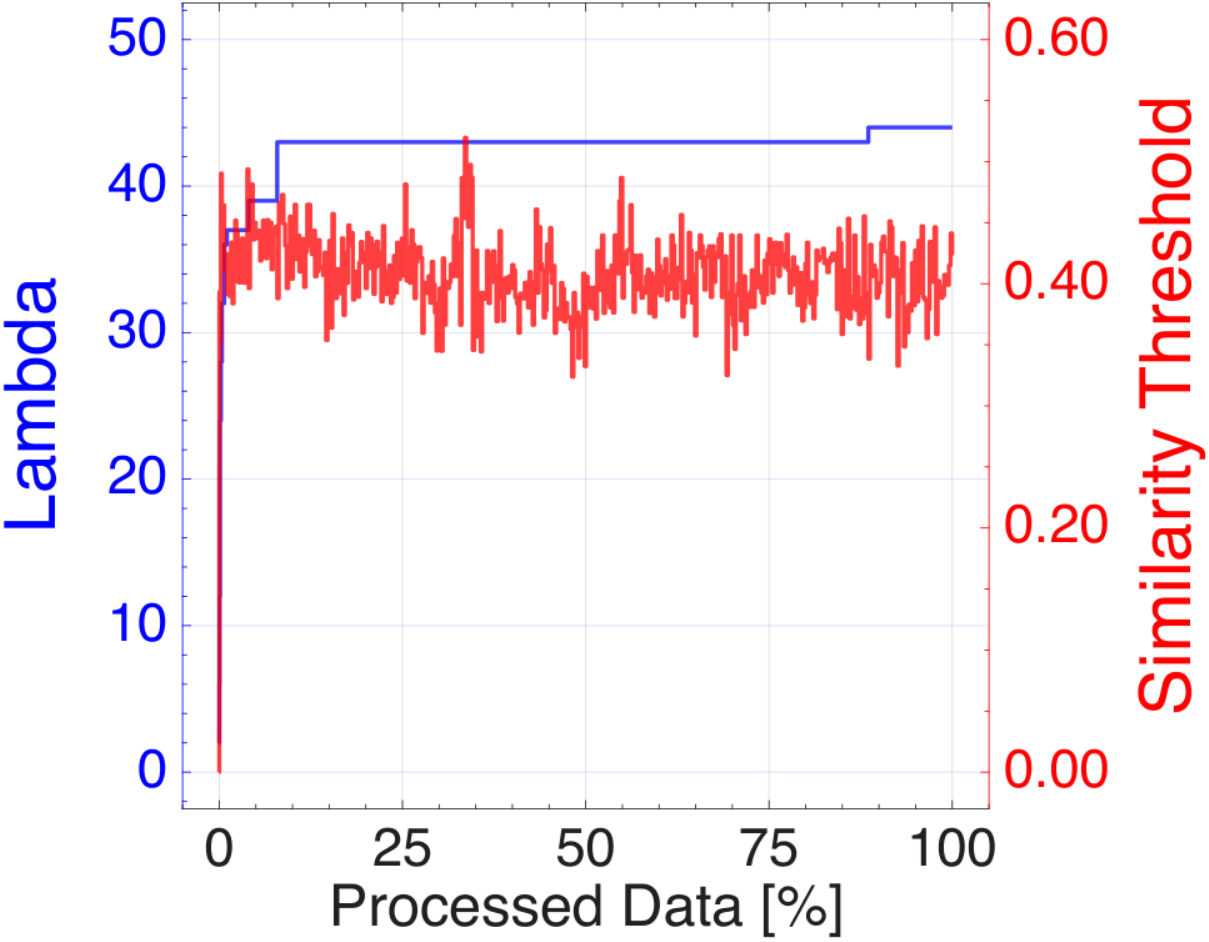}
  }\hfill
  \subfloat[Shuttle]{%
    \includegraphics[width=0.22\linewidth]{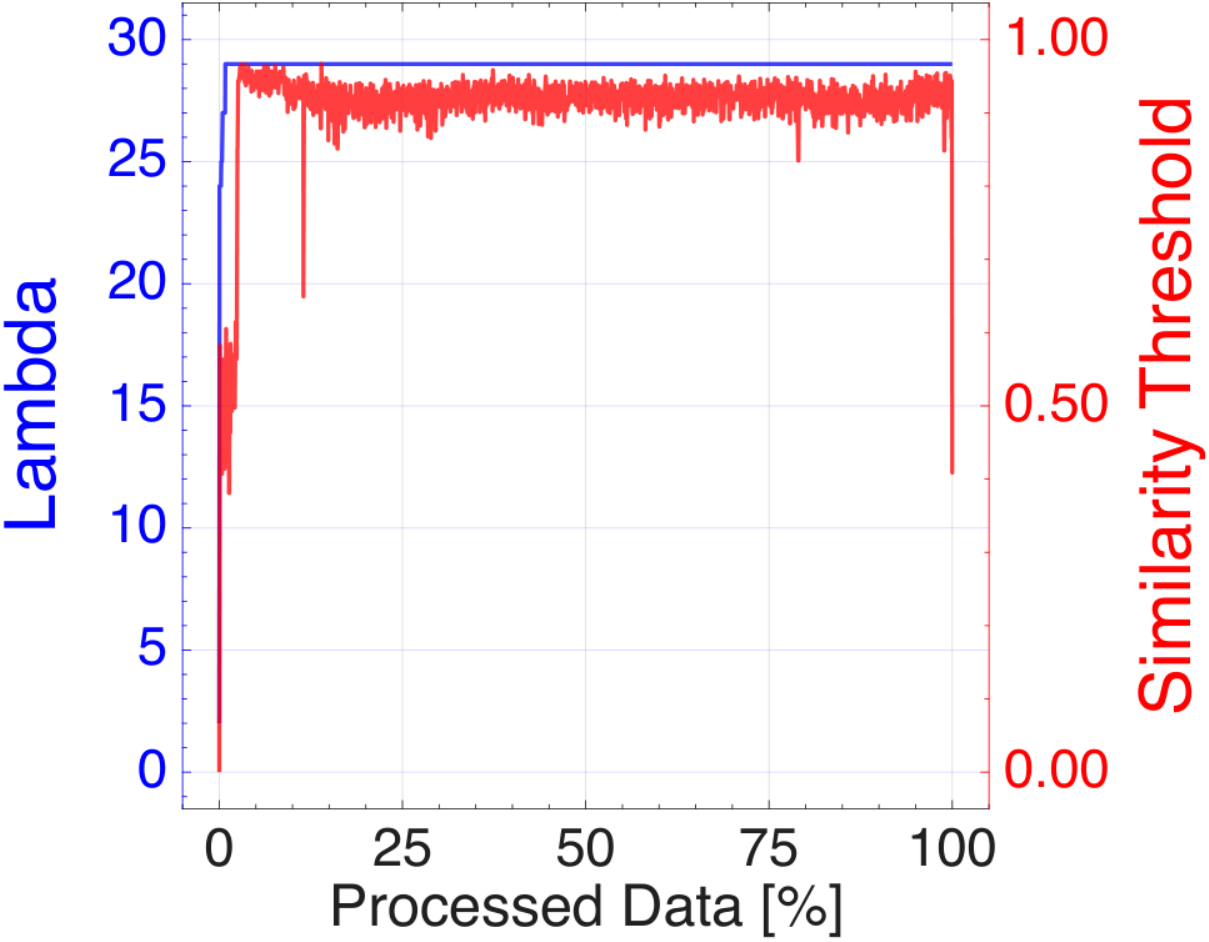}
  }\hfill
  \subfloat[Skin]{%
    \includegraphics[width=0.22\linewidth]{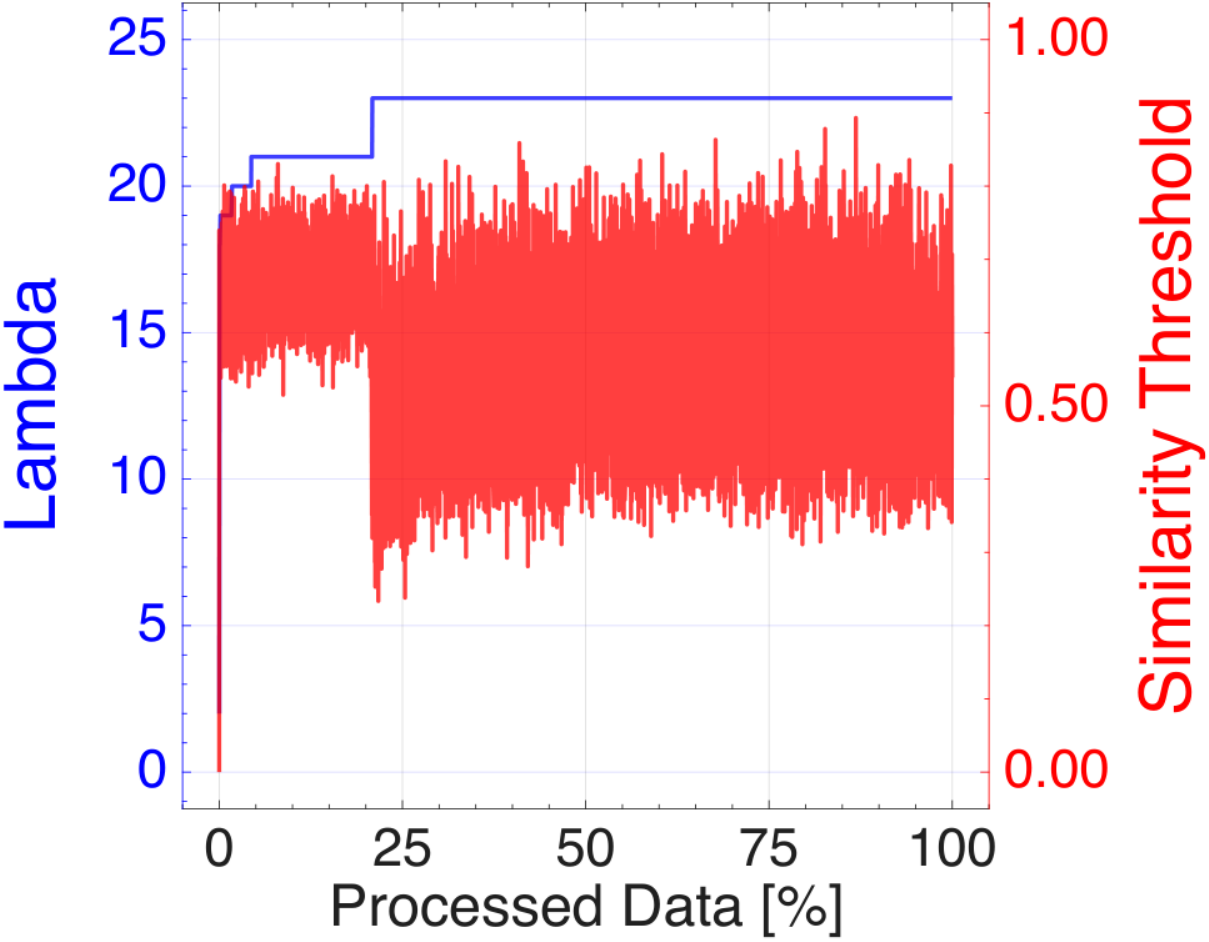}
  }
  \caption{Histories of $\Lambda$ and $V_{\text{threshold}}$ for the w/o Dec. variant in the nonstationary setting ($\Lambda_{\text{init}} = 2$).}
  \label{fig:ablation_lambda_history_nodecrease_nonstationary}
\end{figure*}

\begin{figure*}[htbp]
  \centering
  \subfloat[Iris]{%
    \includegraphics[width=0.22\linewidth]{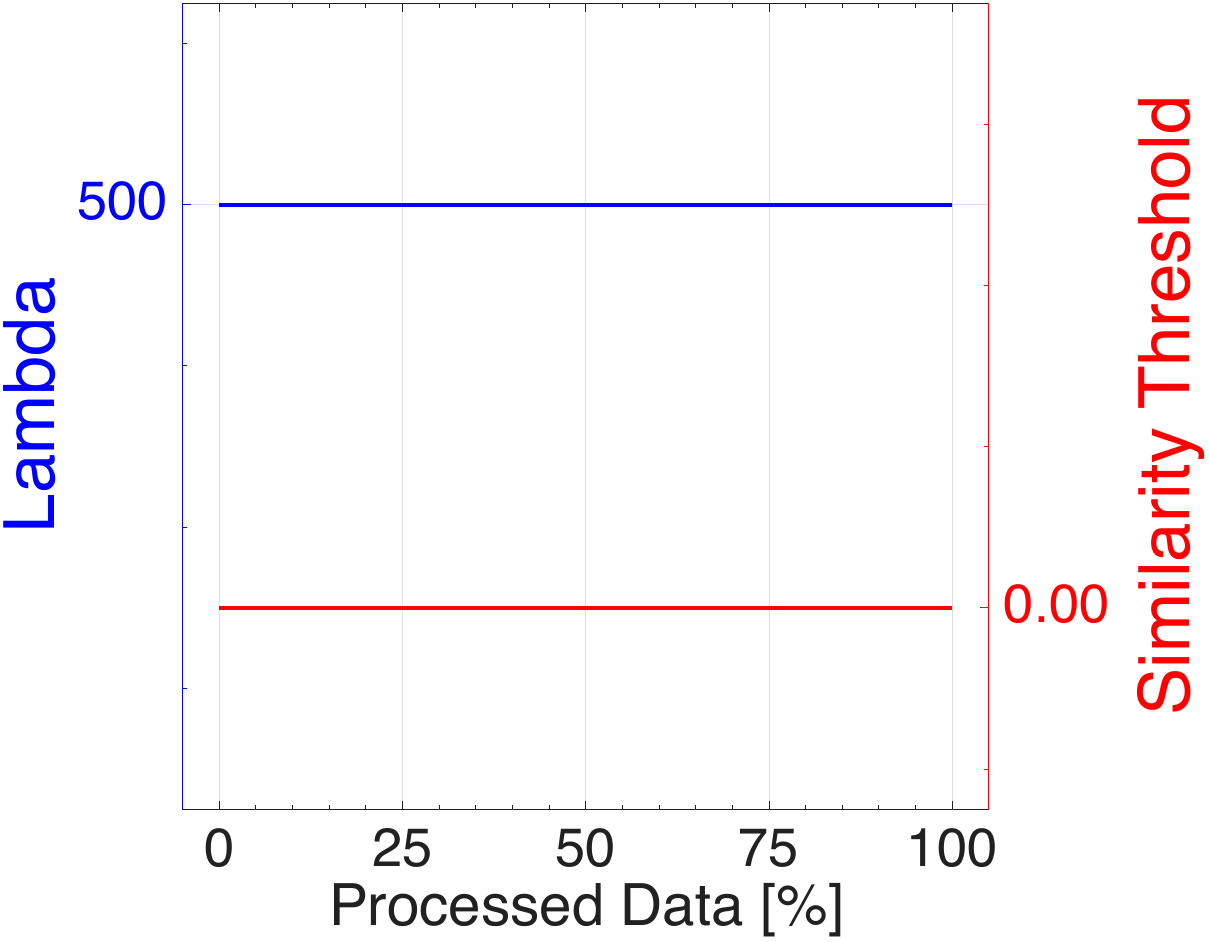}
  }\hfill
  \subfloat[Seeds]{%
    \includegraphics[width=0.22\linewidth]{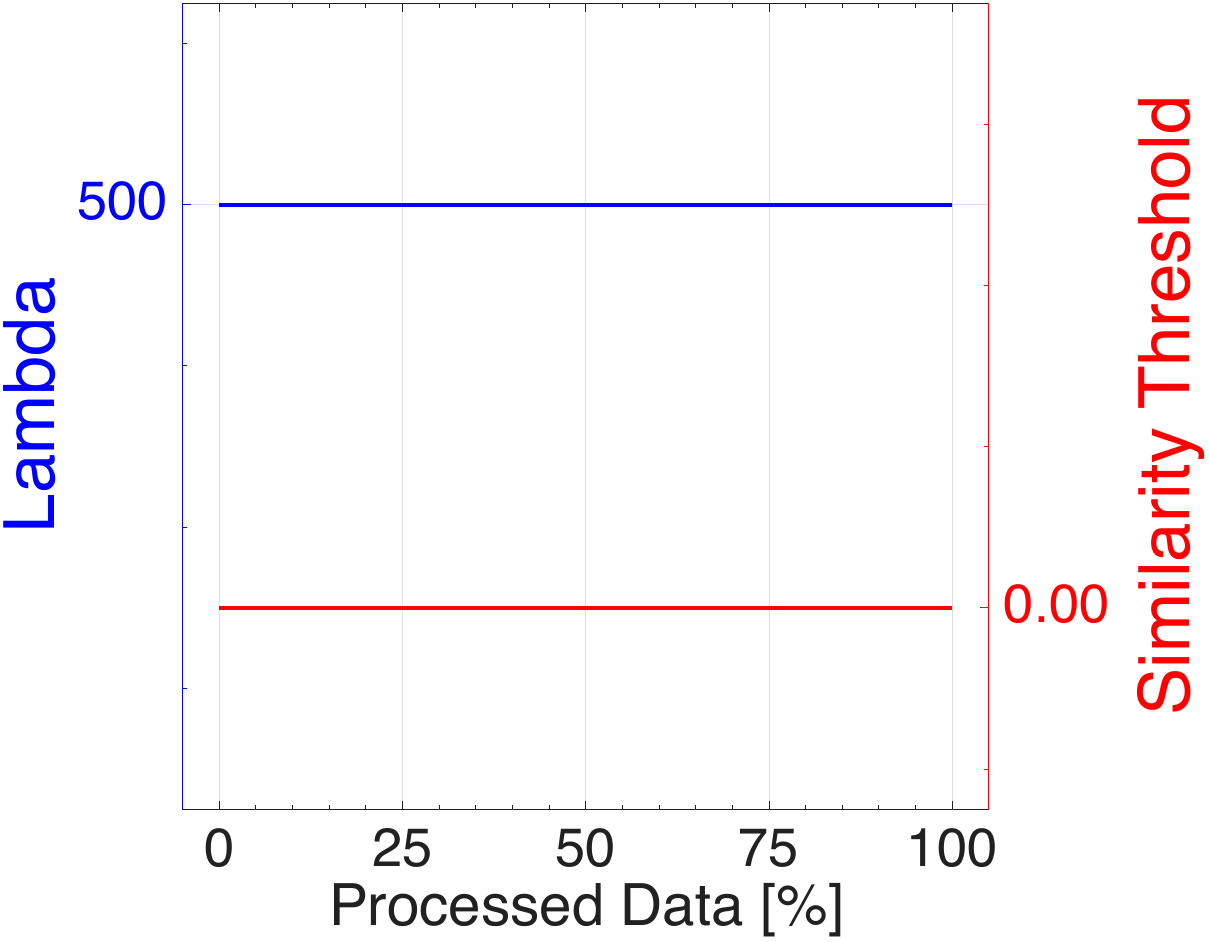}
  }\hfill
  \subfloat[Dermatology]{%
    \includegraphics[width=0.22\linewidth]{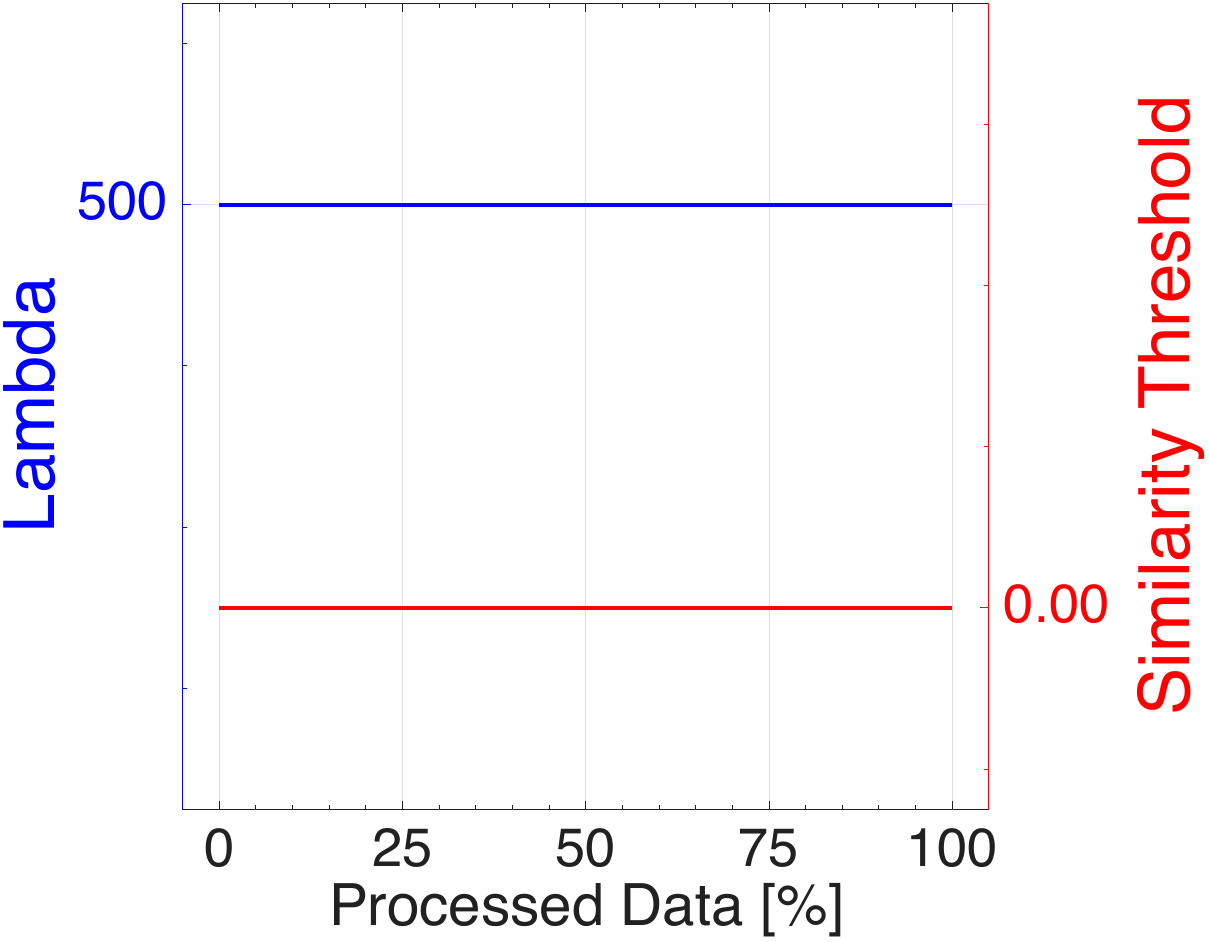}
  }\hfill
  \subfloat[Pima]{%
    \includegraphics[width=0.22\linewidth]{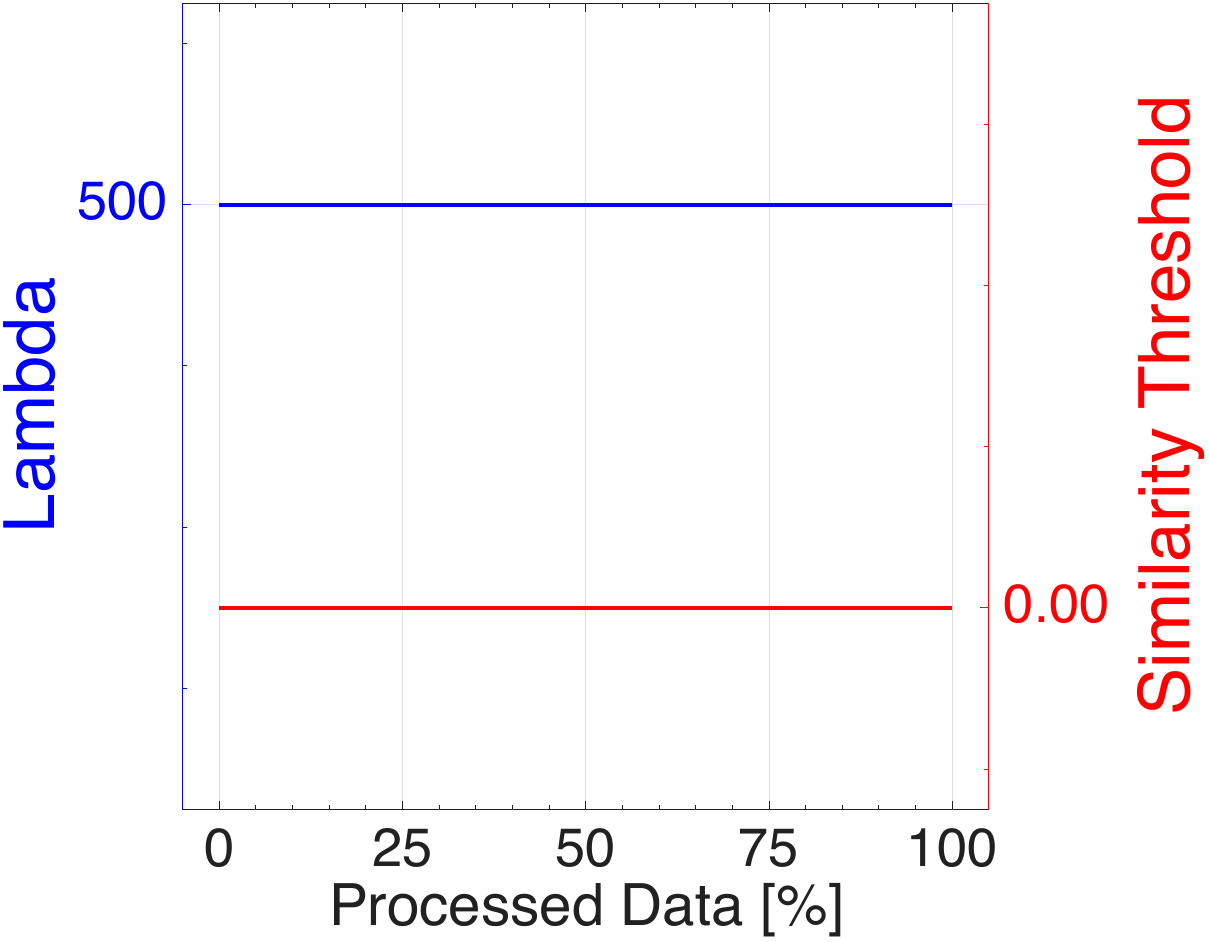}
  }\\
  \subfloat[Mice Protein]{%
    \includegraphics[width=0.22\linewidth]{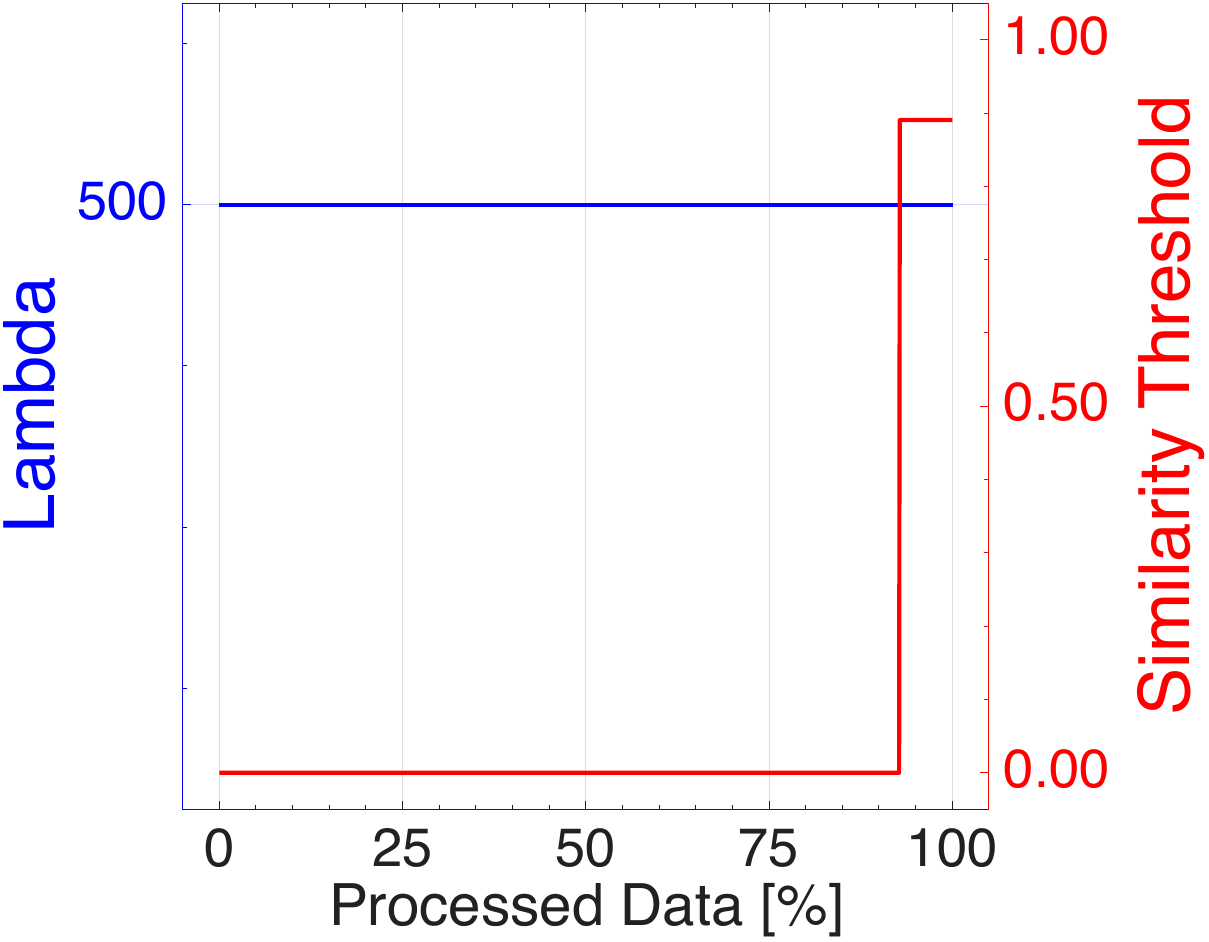}
  }\hfill
  \subfloat[Binalpha]{%
    \includegraphics[width=0.22\linewidth]{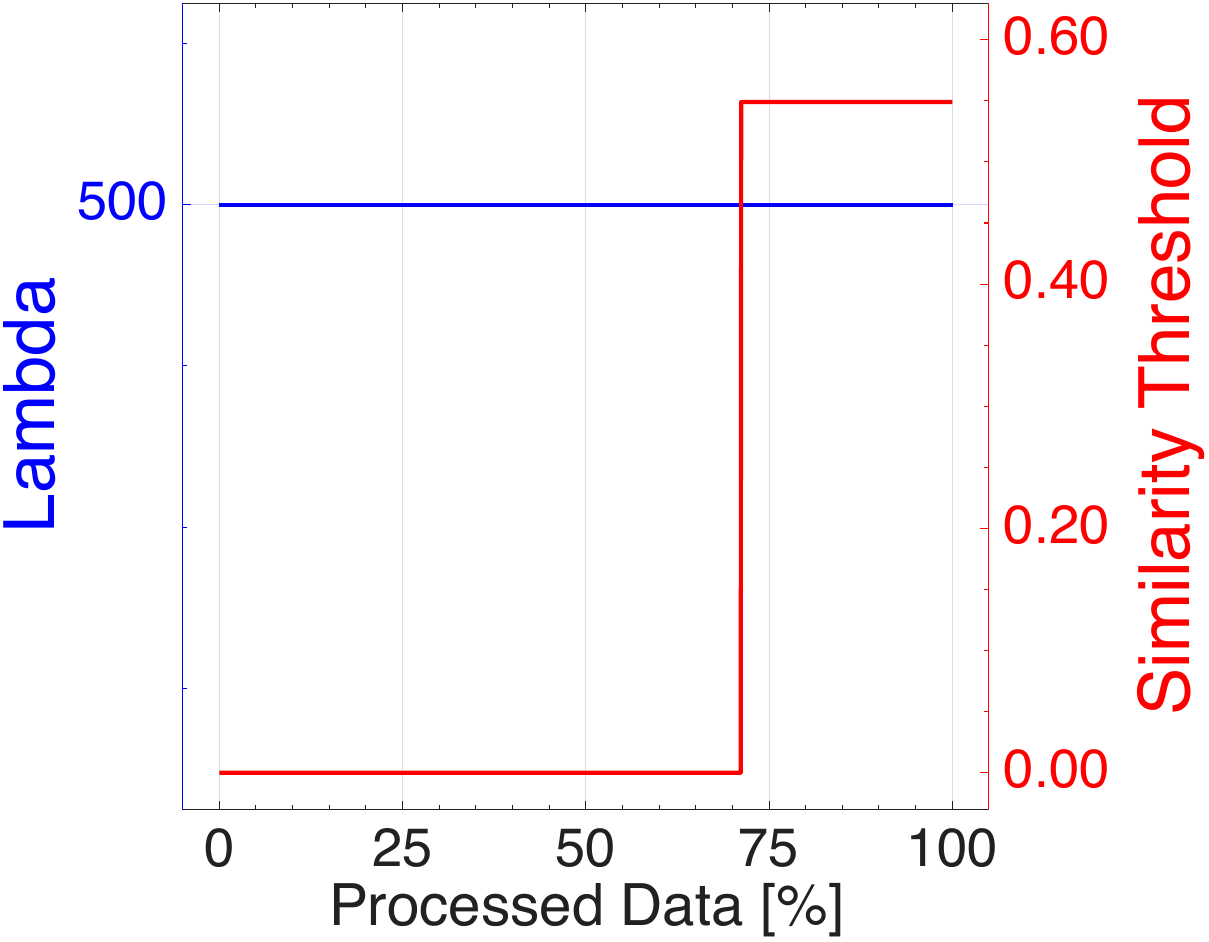}
  }\hfill
  \subfloat[Yeast]{%
    \includegraphics[width=0.22\linewidth]{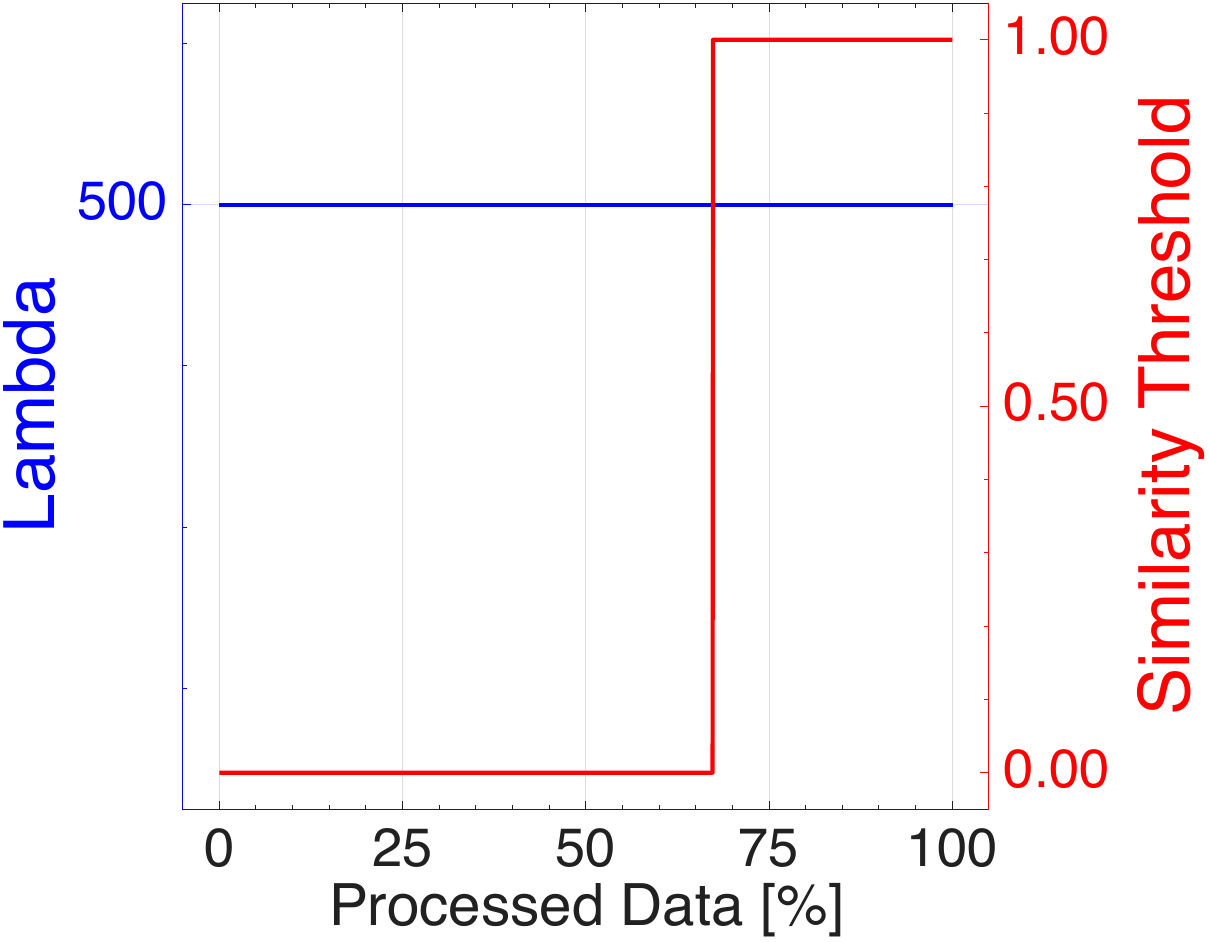}
  }\hfill
  \subfloat[Semeion]{%
    \includegraphics[width=0.22\linewidth]{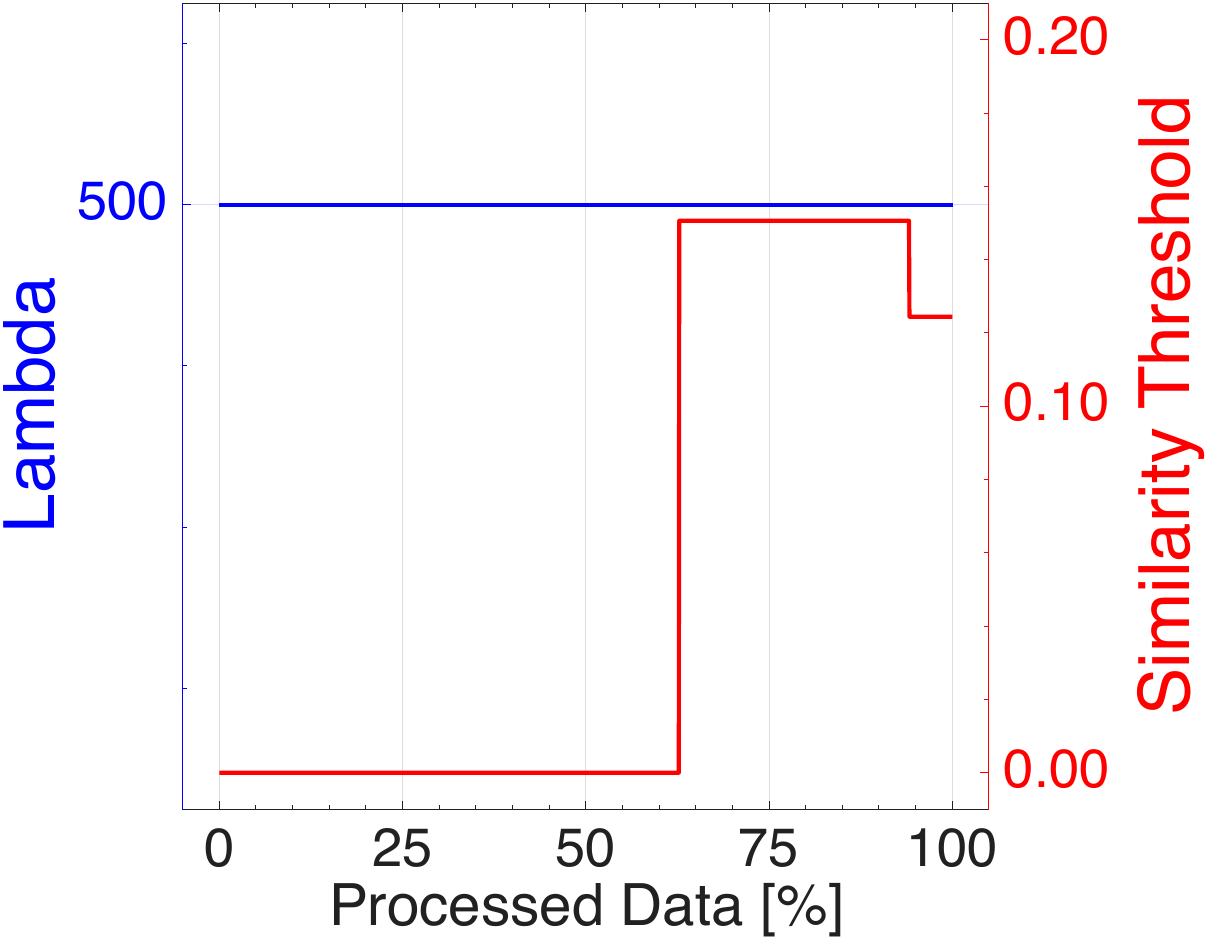}
  }\\
  \subfloat[MSRA25]{%
    \includegraphics[width=0.22\linewidth]{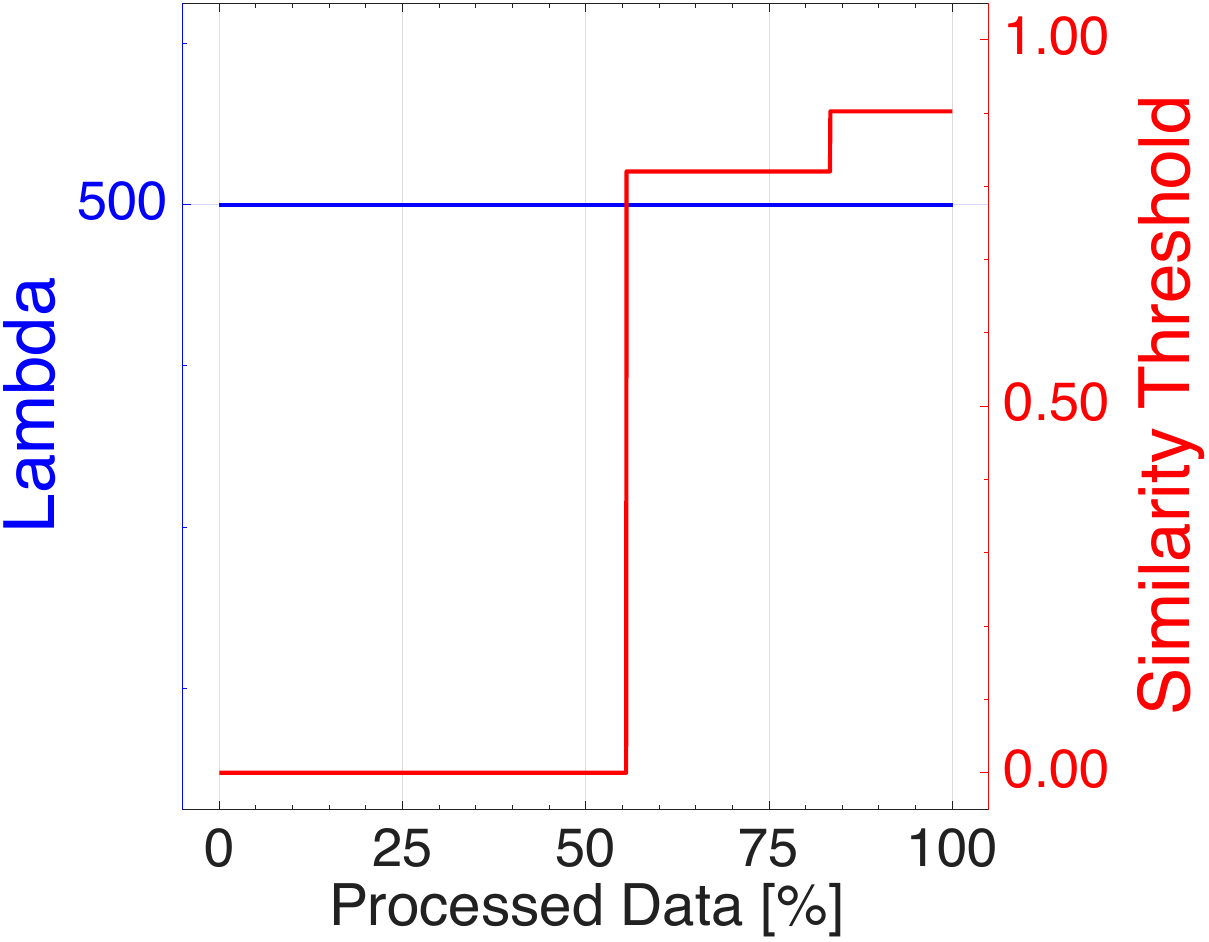}
  }\hfill
  \subfloat[Image Segmentation]{%
    \includegraphics[width=0.22\linewidth]{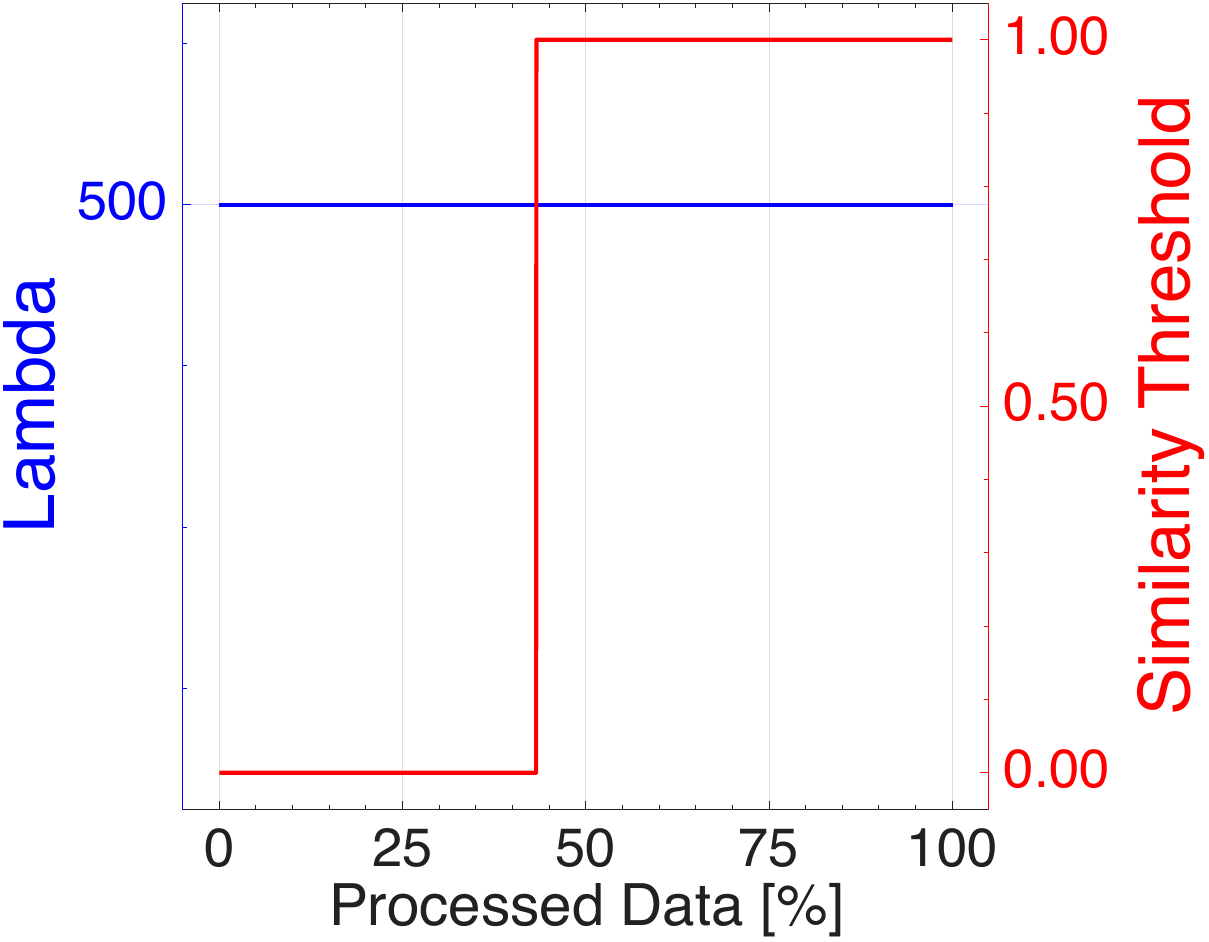}
  }\hfill
  \subfloat[Rice]{%
    \includegraphics[width=0.22\linewidth]{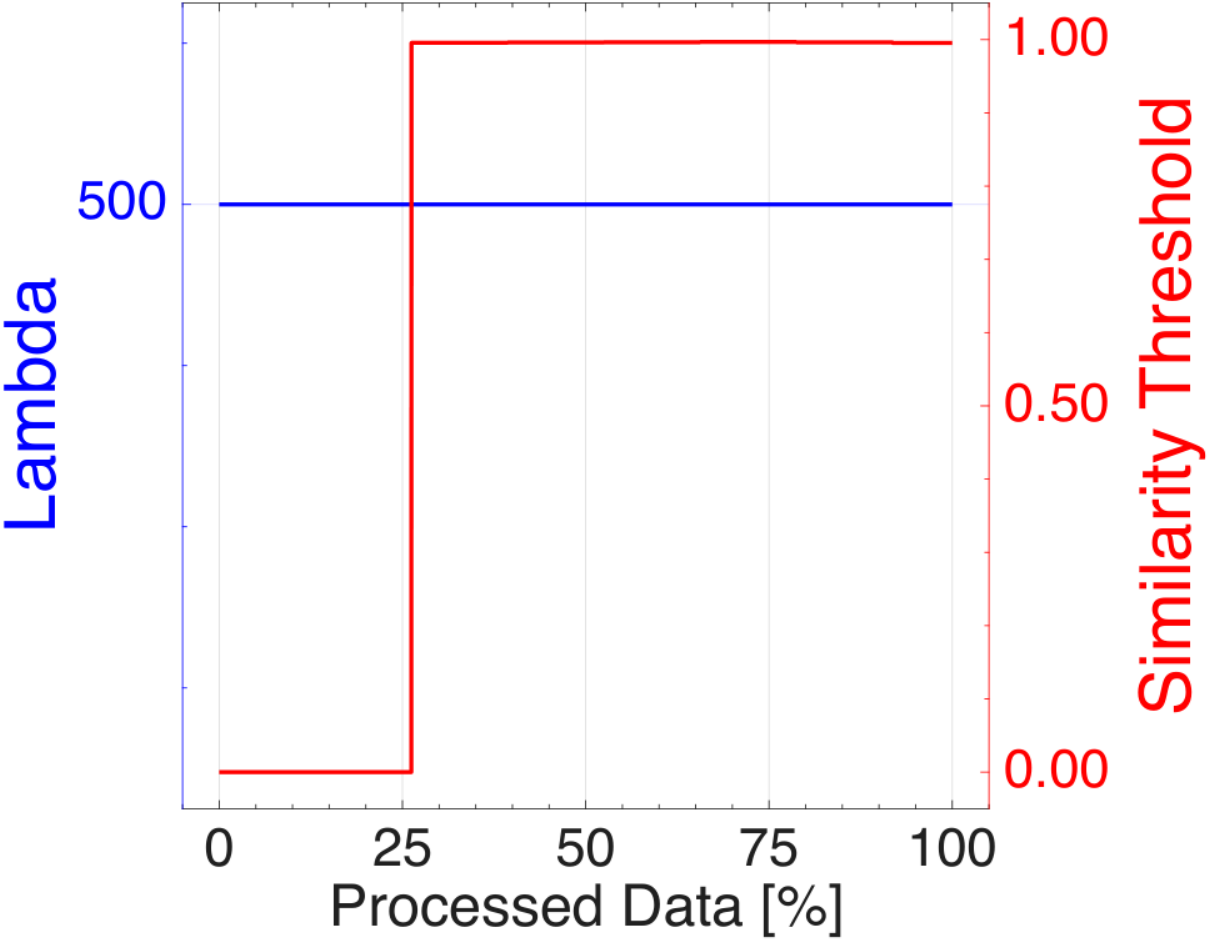}
  }\hfill
  \subfloat[TUANDROMD]{%
    \includegraphics[width=0.22\linewidth]{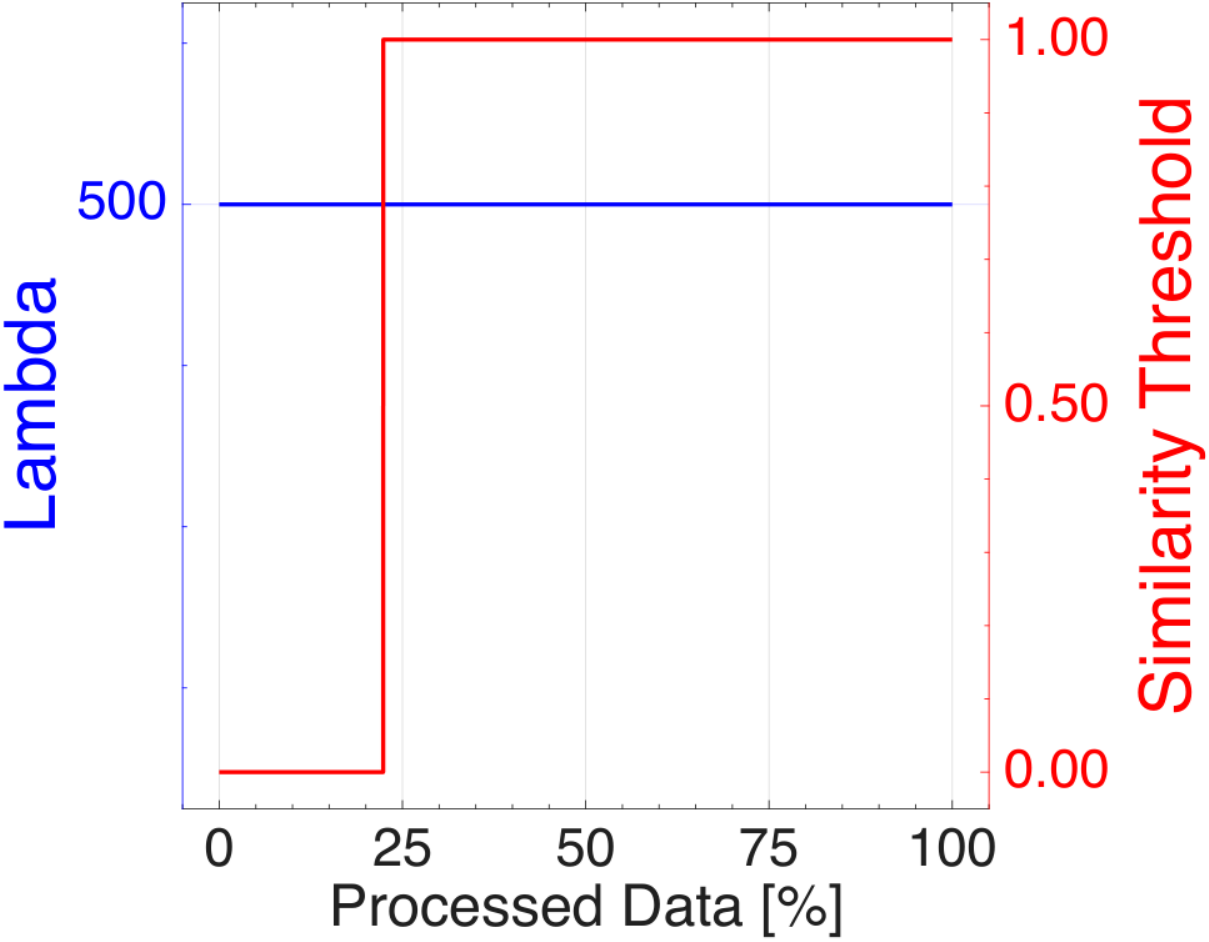}
  }\\
  \subfloat[Phoneme]{%
    \includegraphics[width=0.22\linewidth]{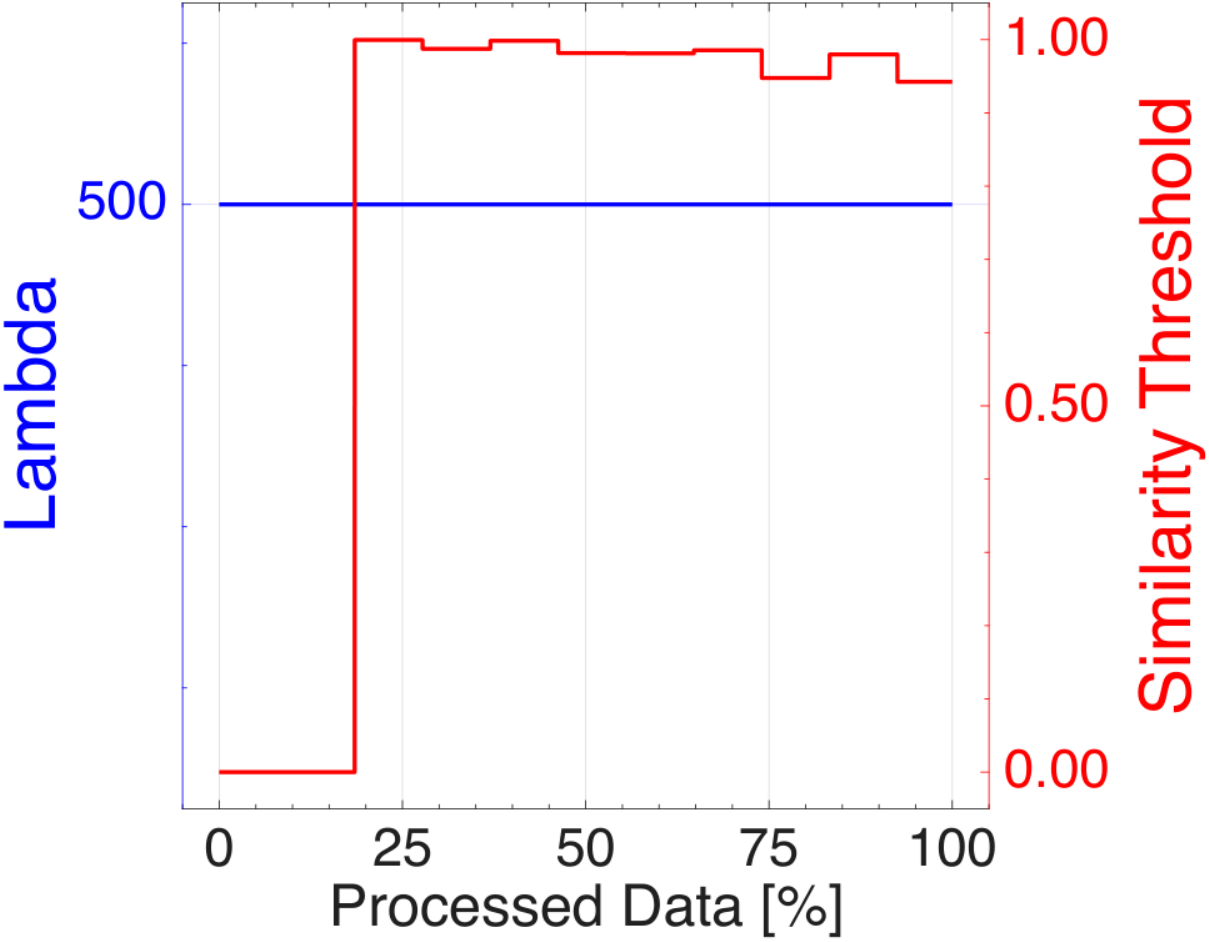}
  }\hfill
  \subfloat[Texture]{%
    \includegraphics[width=0.22\linewidth]{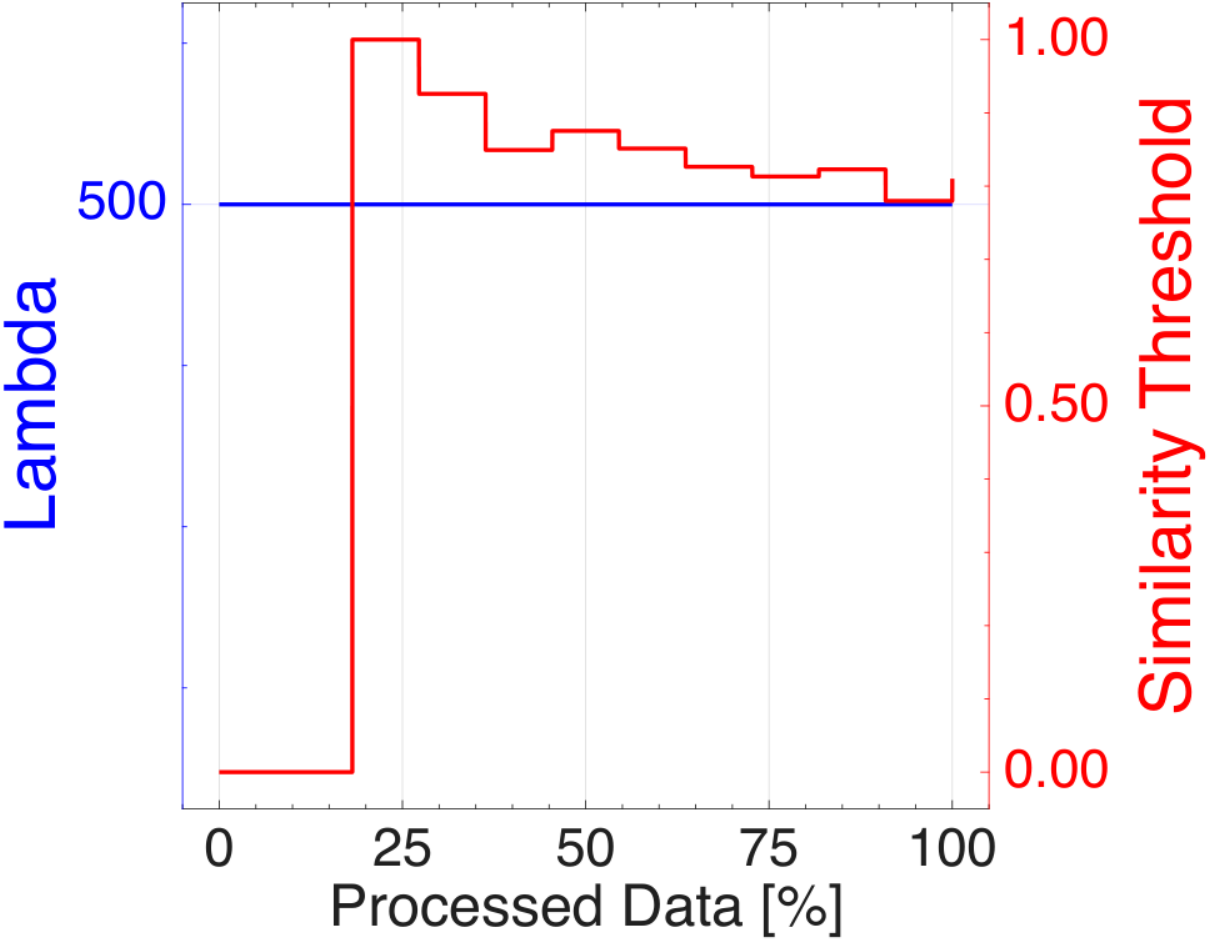}
  }\hfill
  \subfloat[OptDigits]{%
    \includegraphics[width=0.22\linewidth]{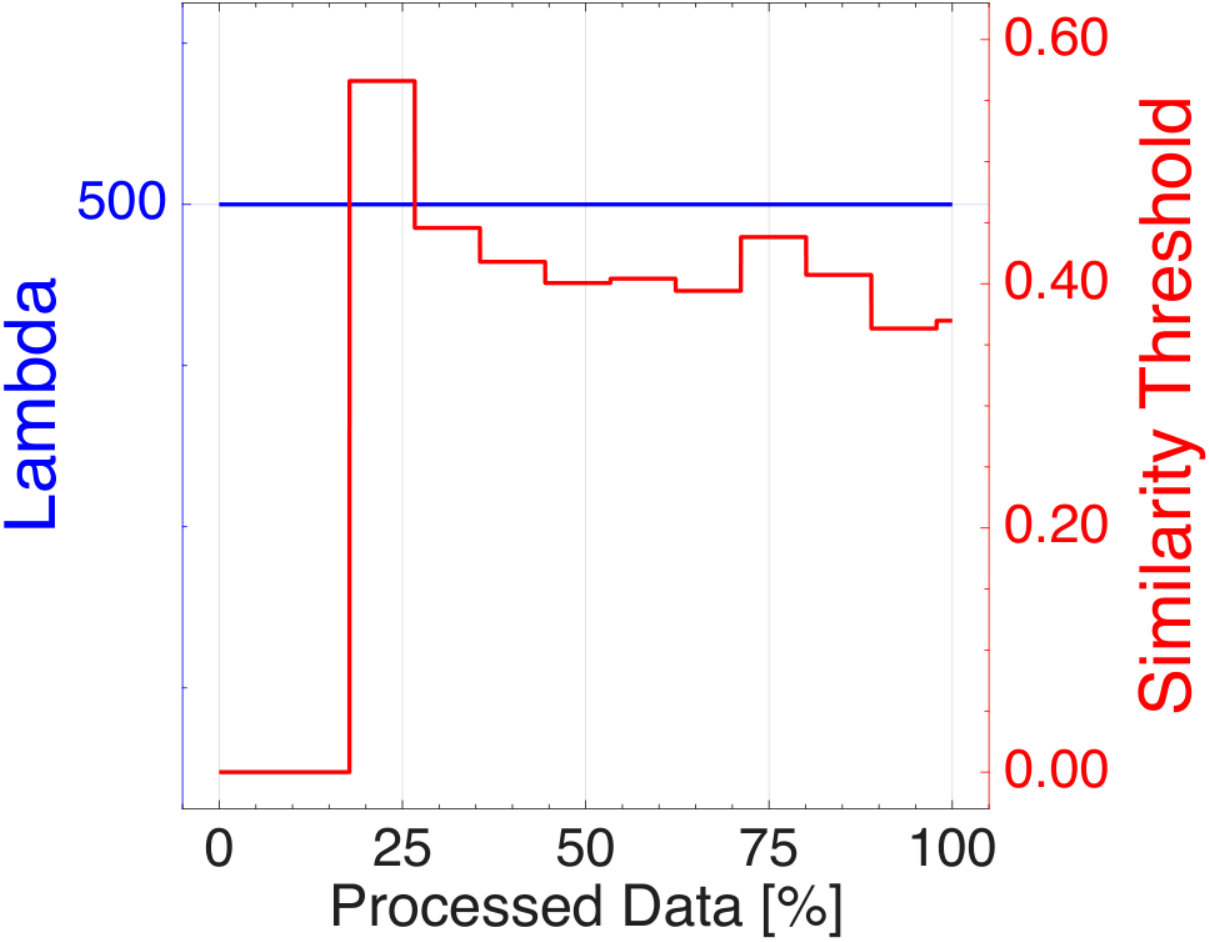}
  }\hfill
  \subfloat[Statlog]{%
    \includegraphics[width=0.22\linewidth]{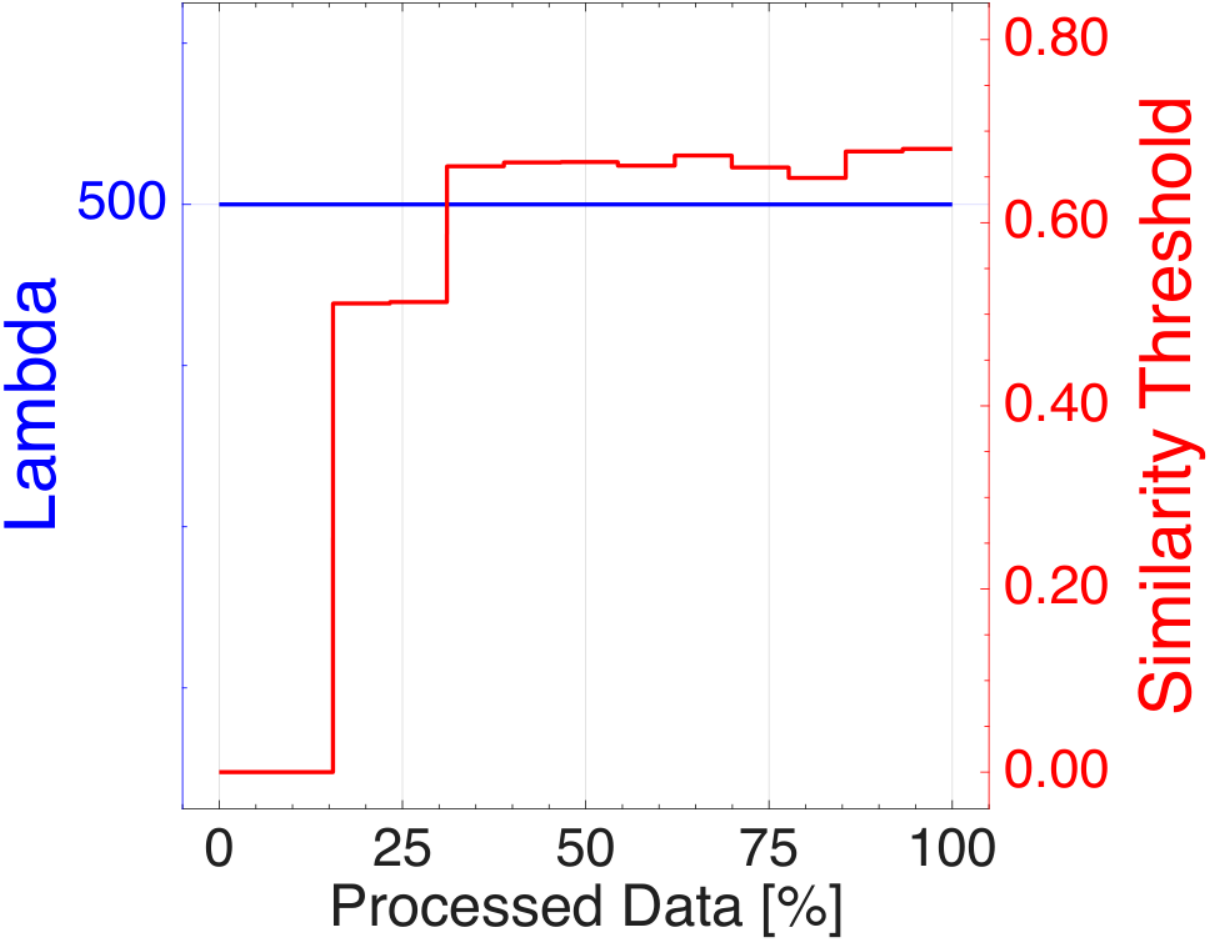}
  }\\
  \subfloat[Anuran Calls]{%
    \includegraphics[width=0.22\linewidth]{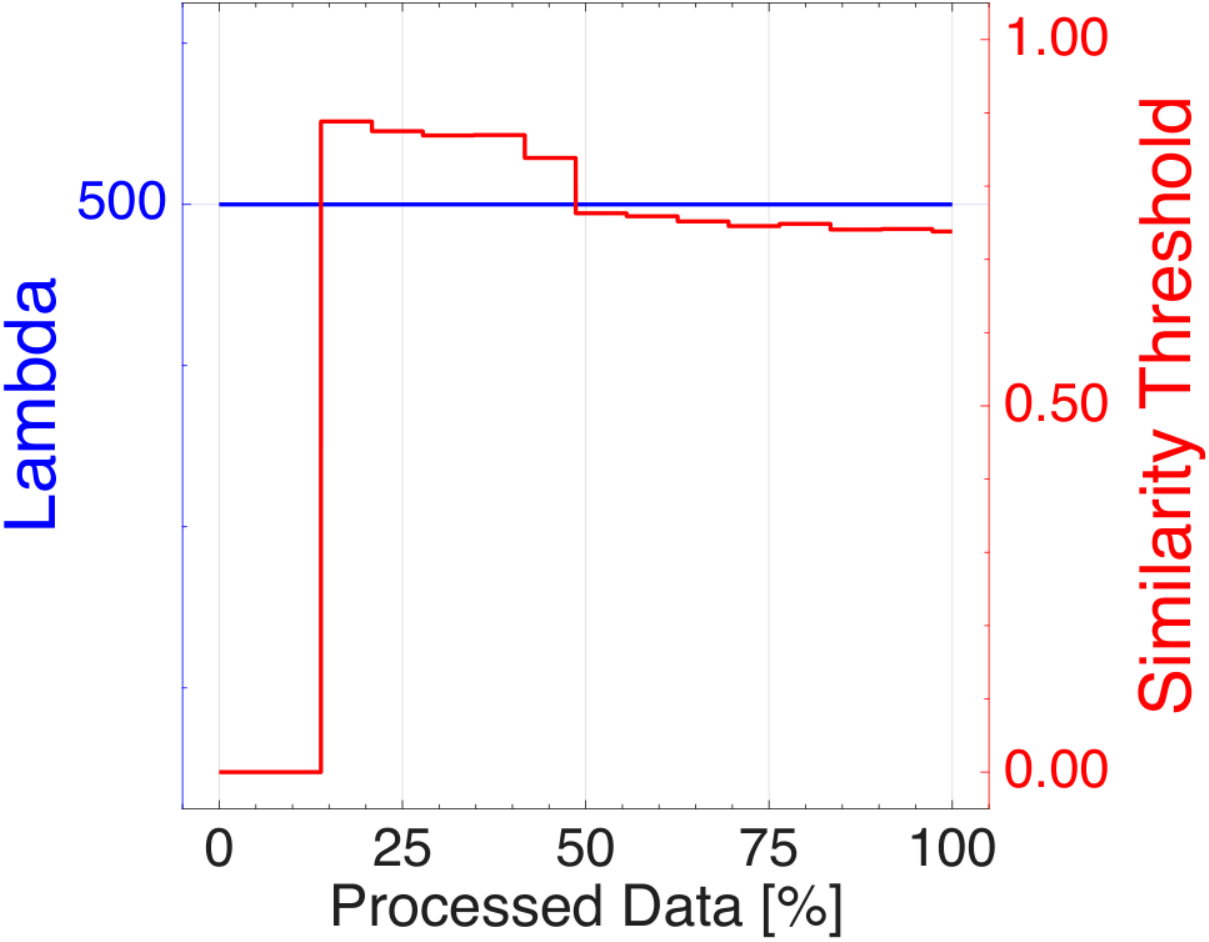}
  }\hfill
  \subfloat[Isolet]{%
    \includegraphics[width=0.22\linewidth]{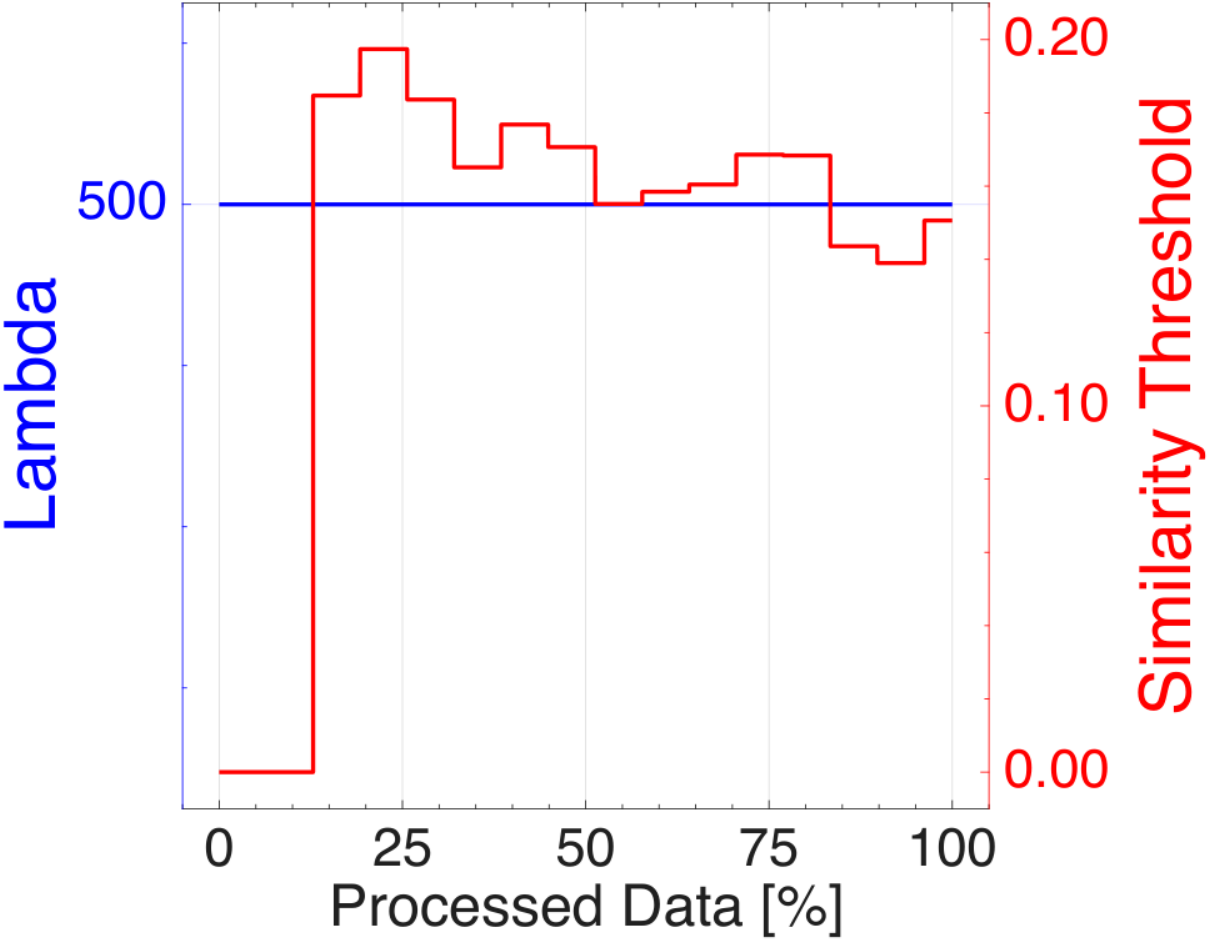}
  }\hfill
  \subfloat[MNIST10K]{%
    \includegraphics[width=0.22\linewidth]{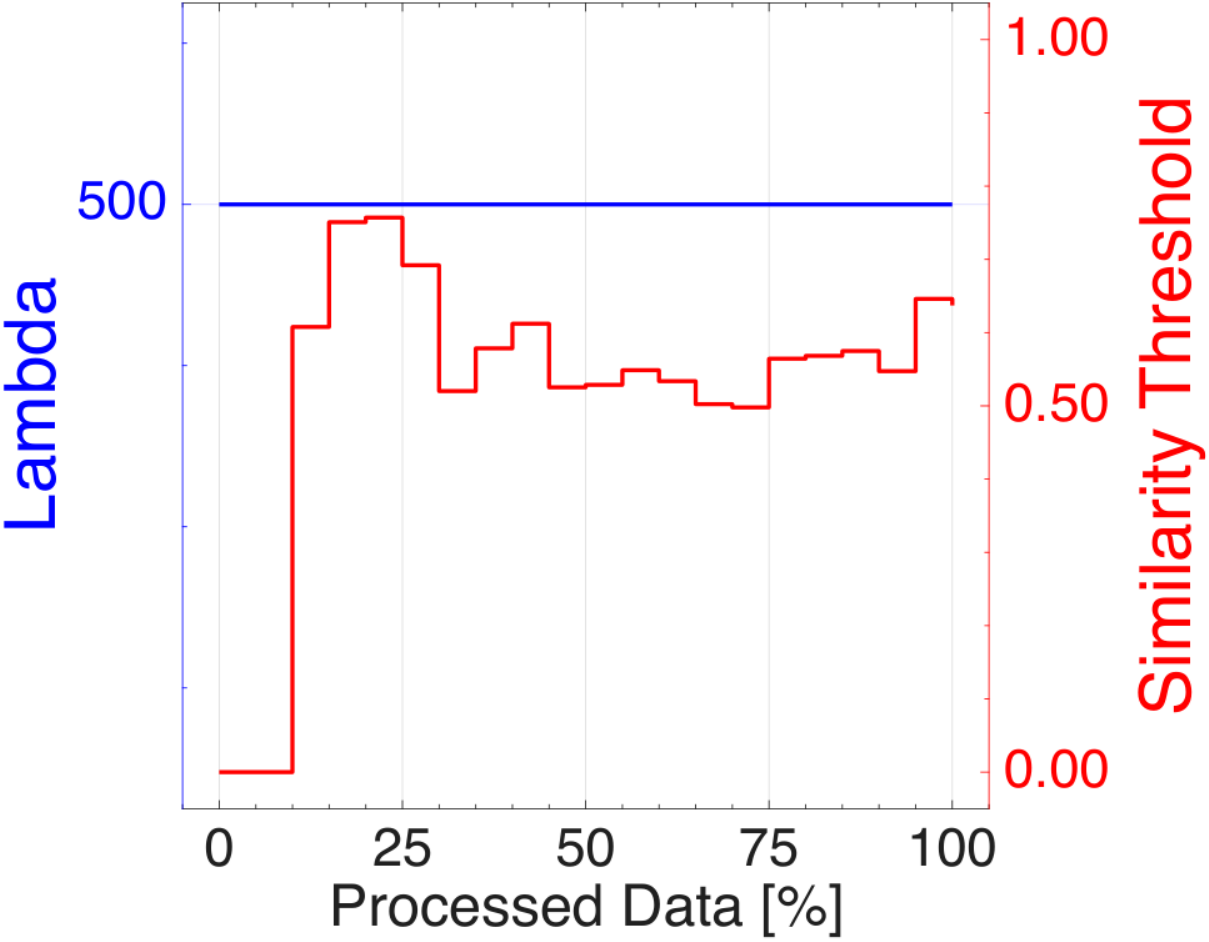}
  }\hfill
  \subfloat[PenBased]{%
    \includegraphics[width=0.22\linewidth]{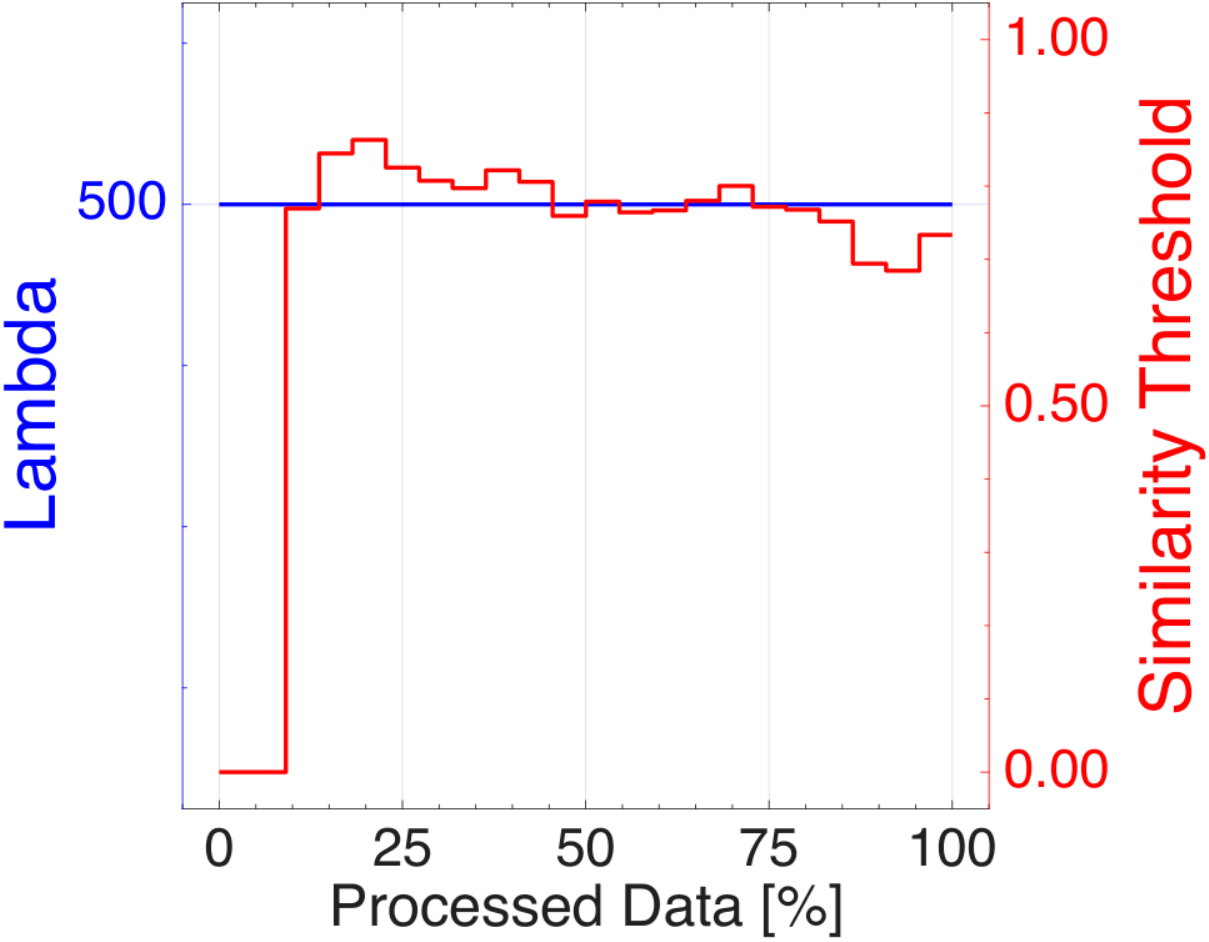}
  }\\
  \subfloat[STL10]{%
    \includegraphics[width=0.22\linewidth]{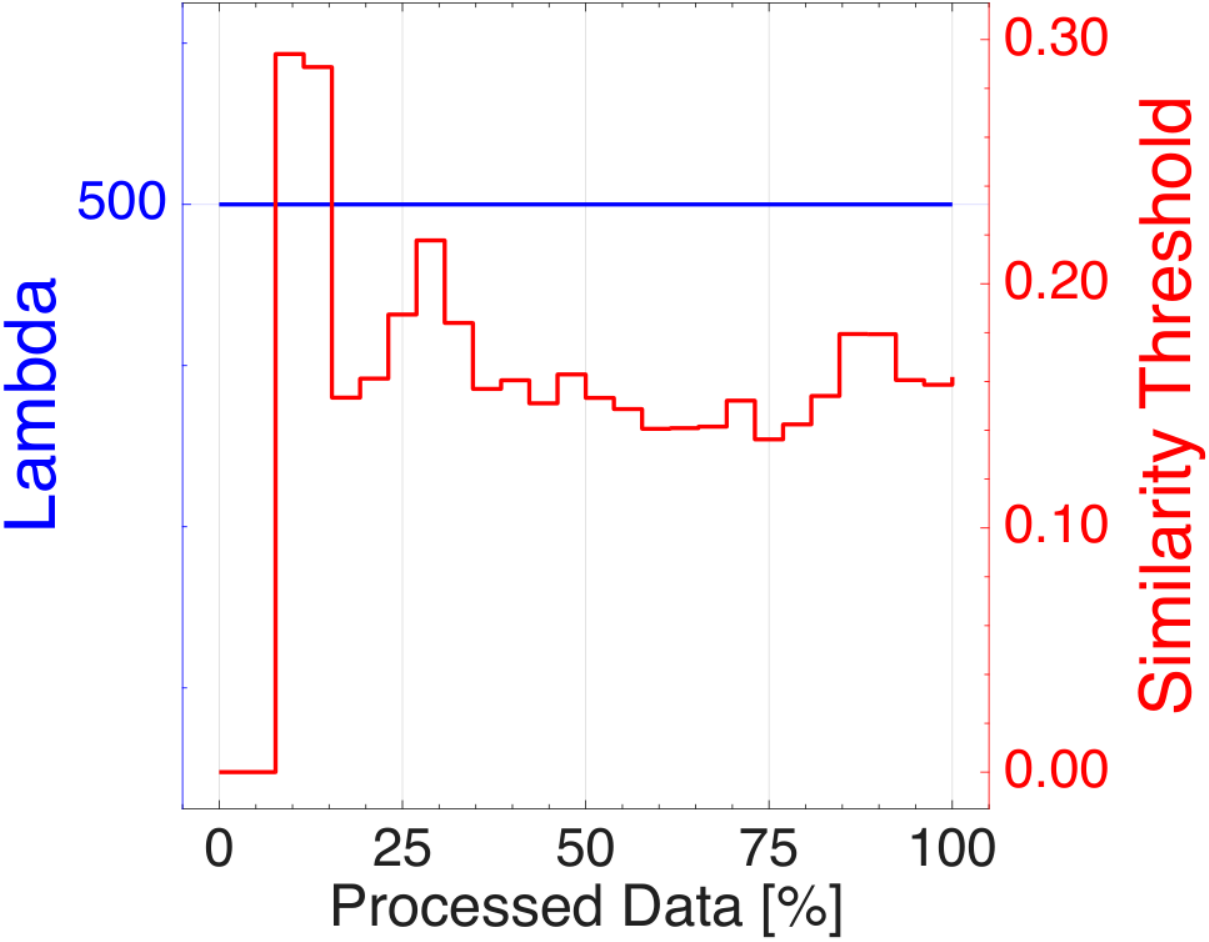}
  }\hfill
  \subfloat[Letter]{%
    \includegraphics[width=0.22\linewidth]{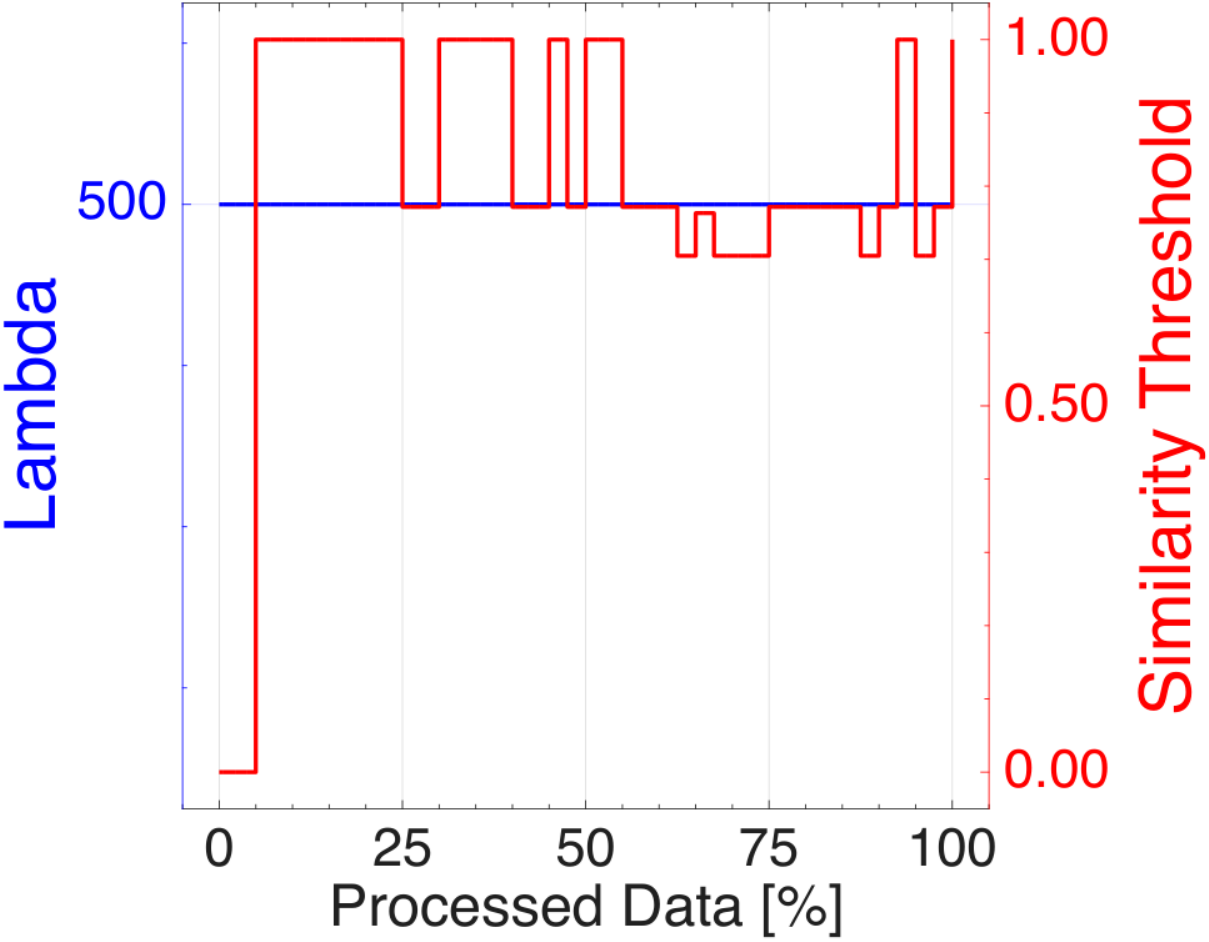}
  }\hfill
  \subfloat[Shuttle]{%
    \includegraphics[width=0.22\linewidth]{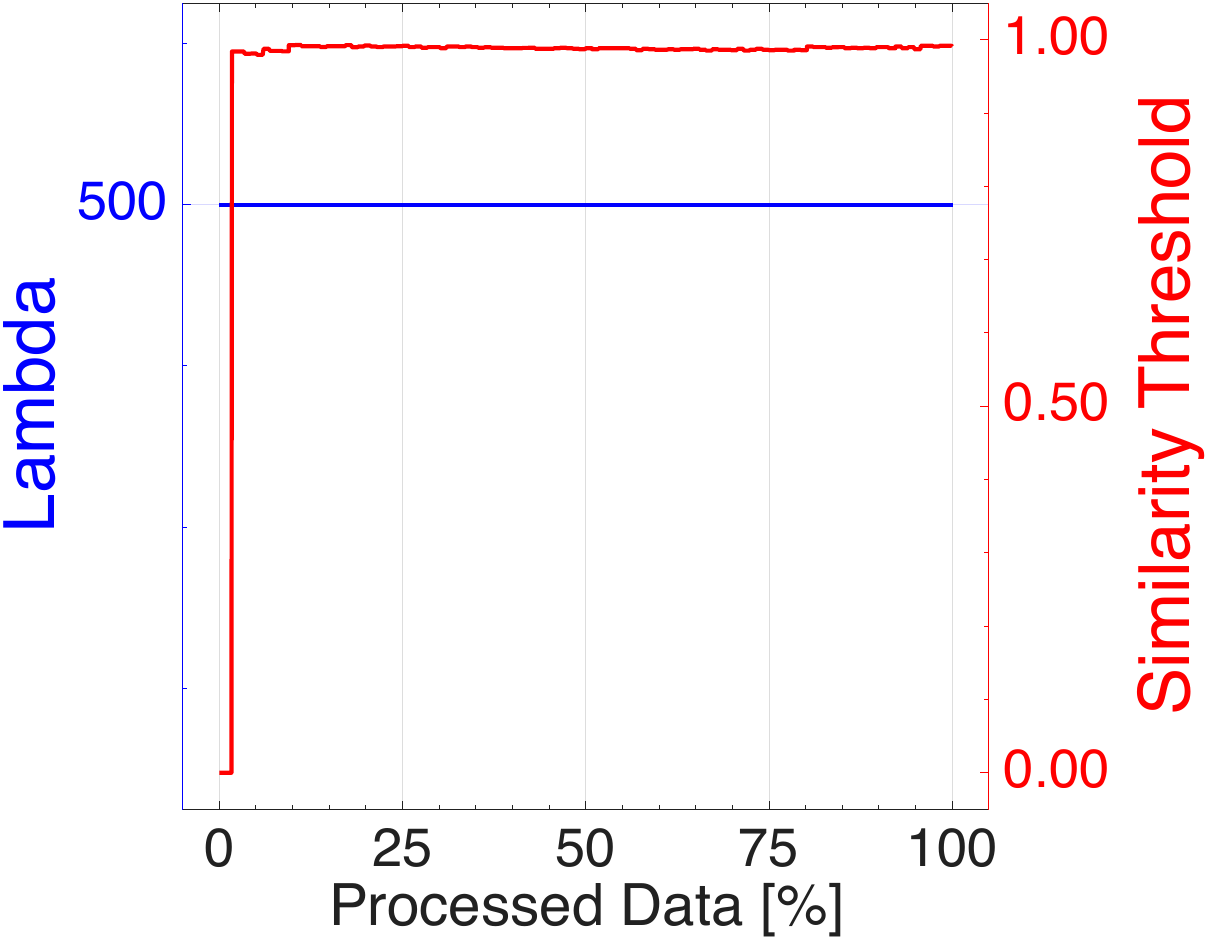}
  }\hfill
  \subfloat[Skin]{%
    \includegraphics[width=0.22\linewidth]{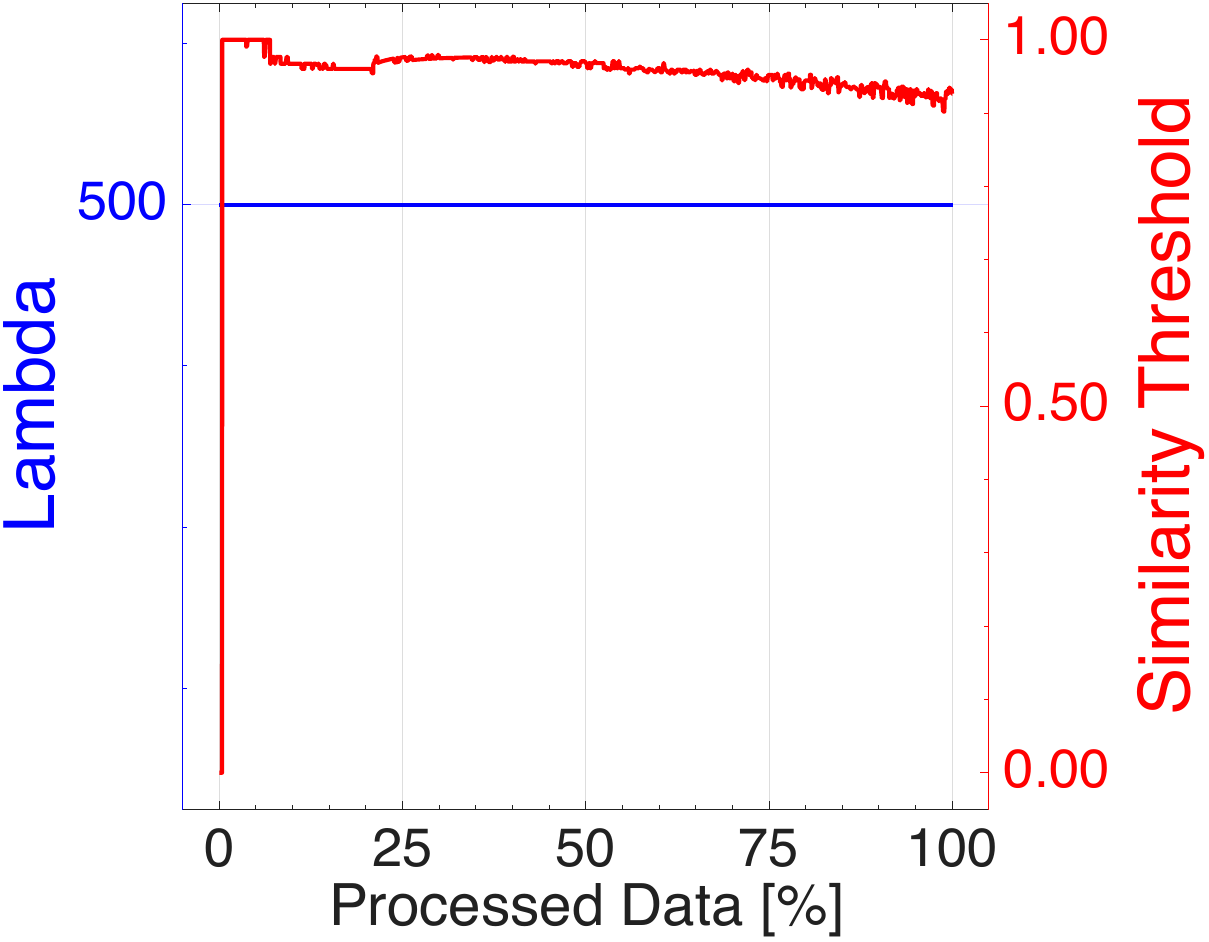}
  }
  \caption{Histories of $\Lambda$ and $V_{\text{threshold}}$ for the w/o Dec. variant in the nonstationary setting ($\Lambda_{\text{init}} = 500$).}
  \label{fig:ablation_lambda_history_nodecrease_500_nonstationary}
\end{figure*}

\begin{figure*}[htbp]
  \centering
  \subfloat[Iris]{%
    \includegraphics[width=0.22\linewidth]{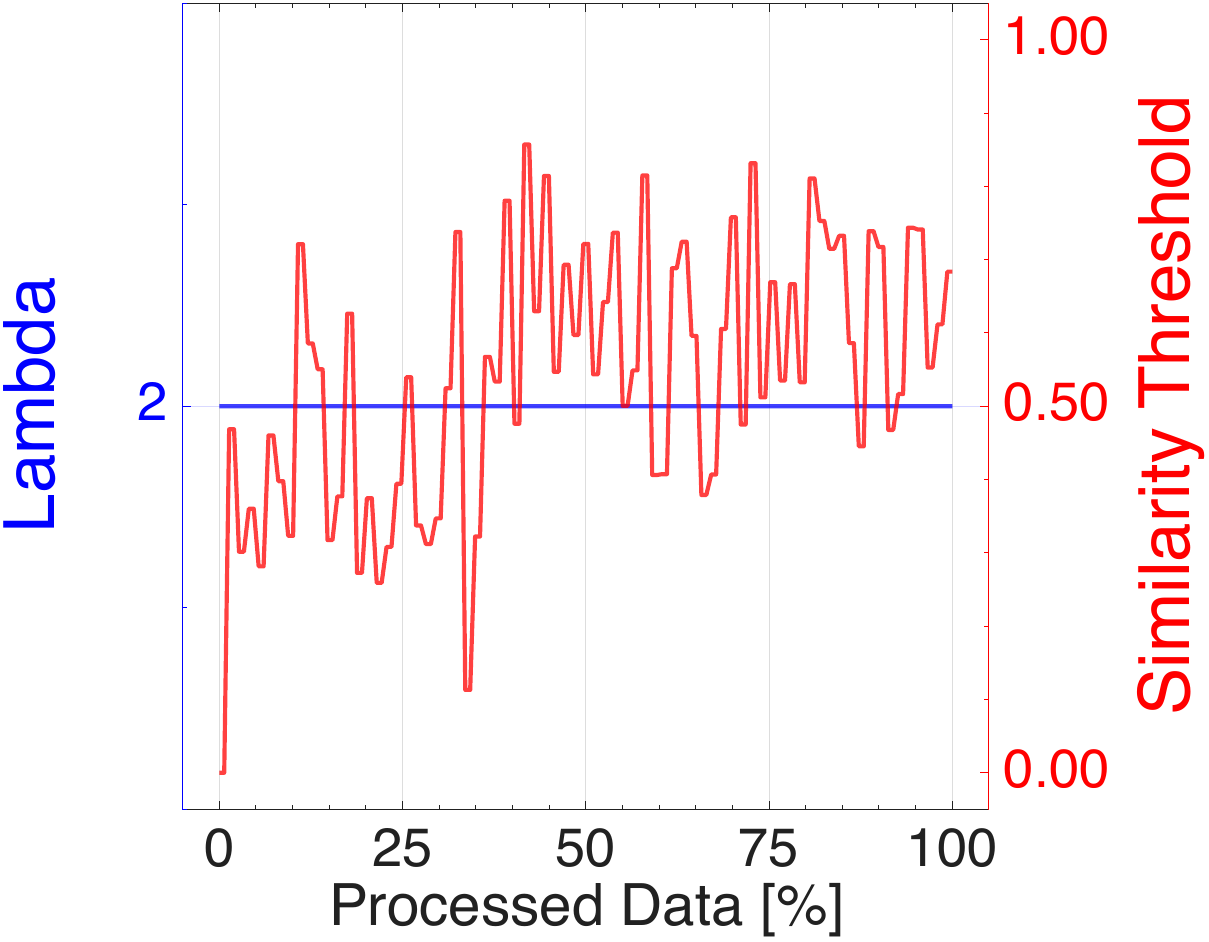}
  }\hfill
  \subfloat[Seeds]{%
    \includegraphics[width=0.22\linewidth]{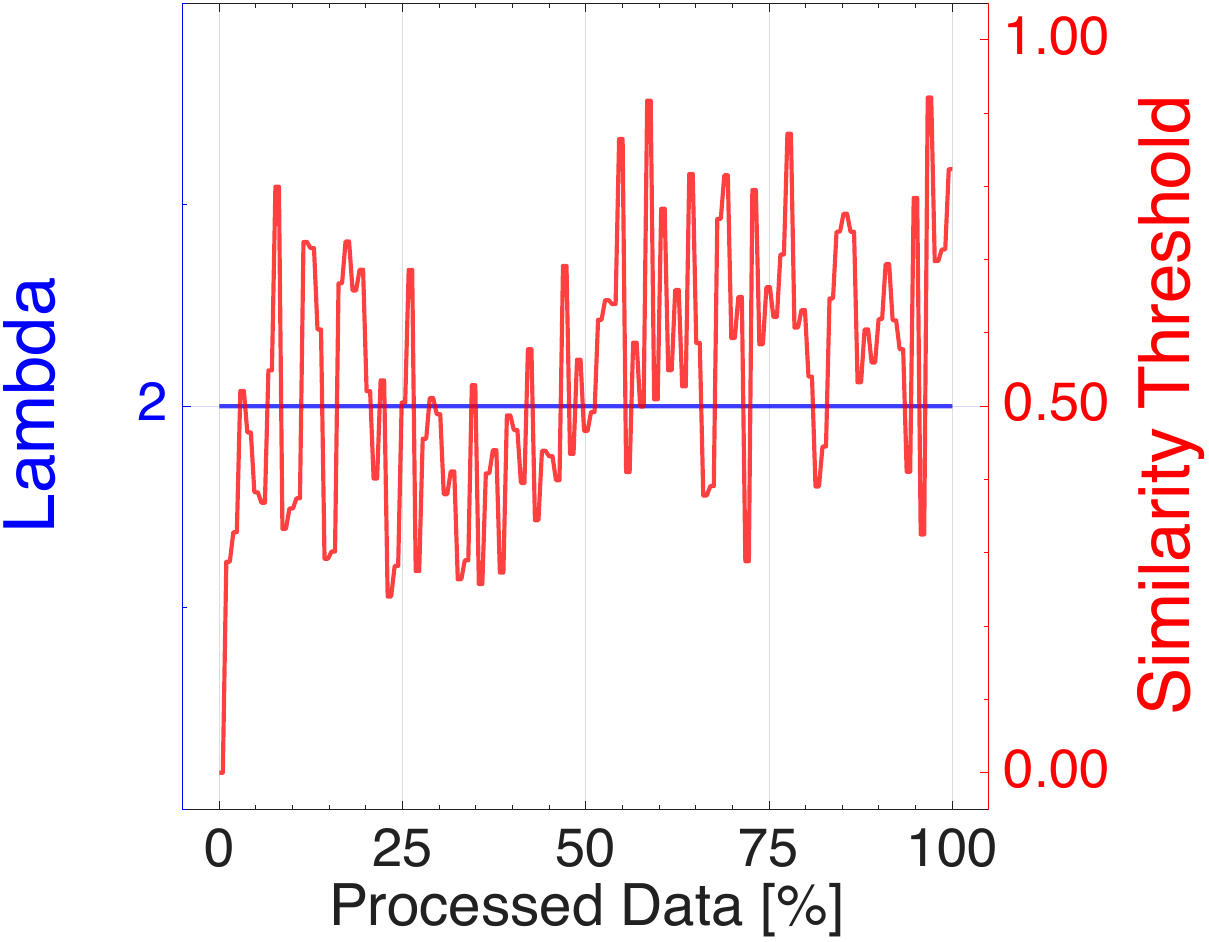}
  }\hfill
  \subfloat[Dermatology]{%
    \includegraphics[width=0.22\linewidth]{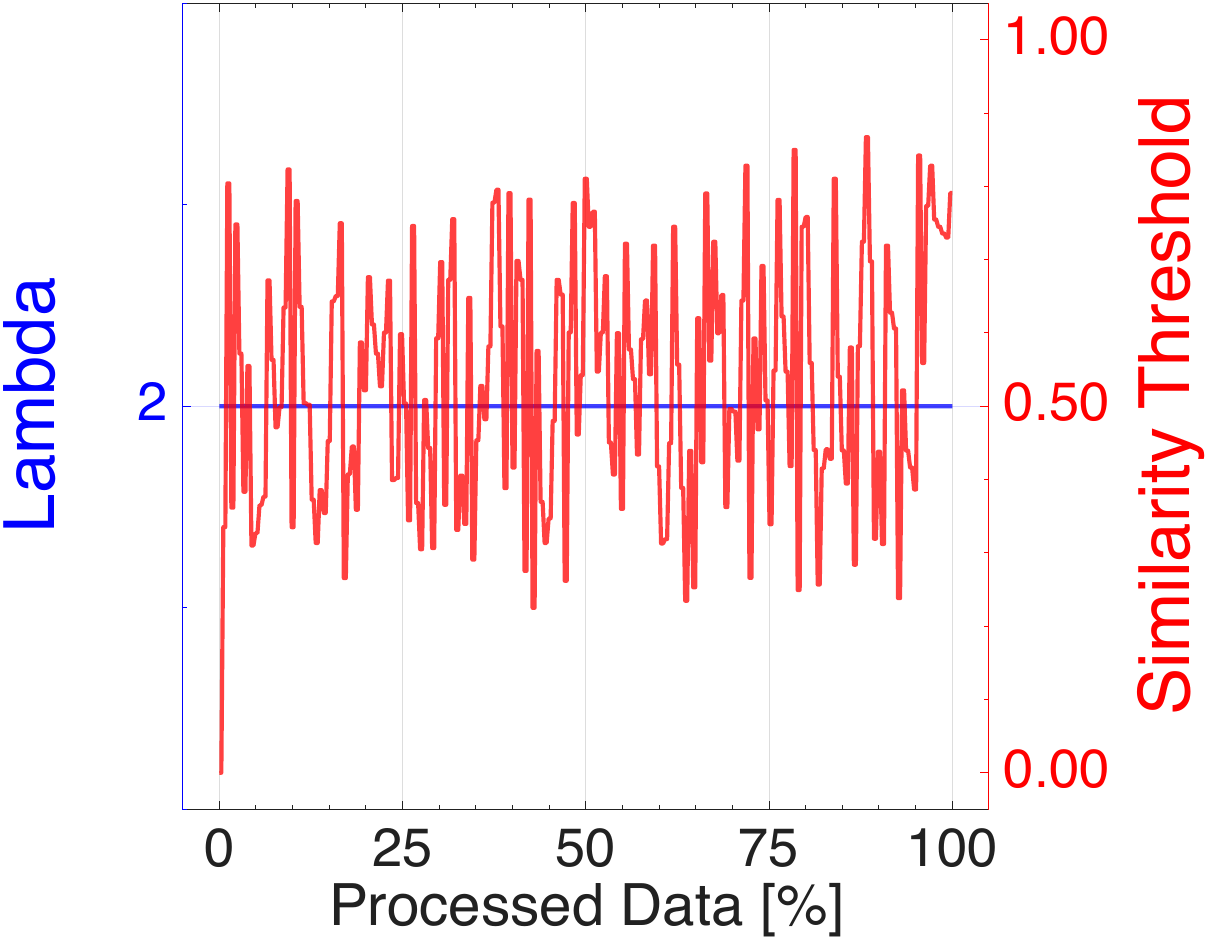}
  }\hfill
  \subfloat[Pima]{%
    \includegraphics[width=0.22\linewidth]{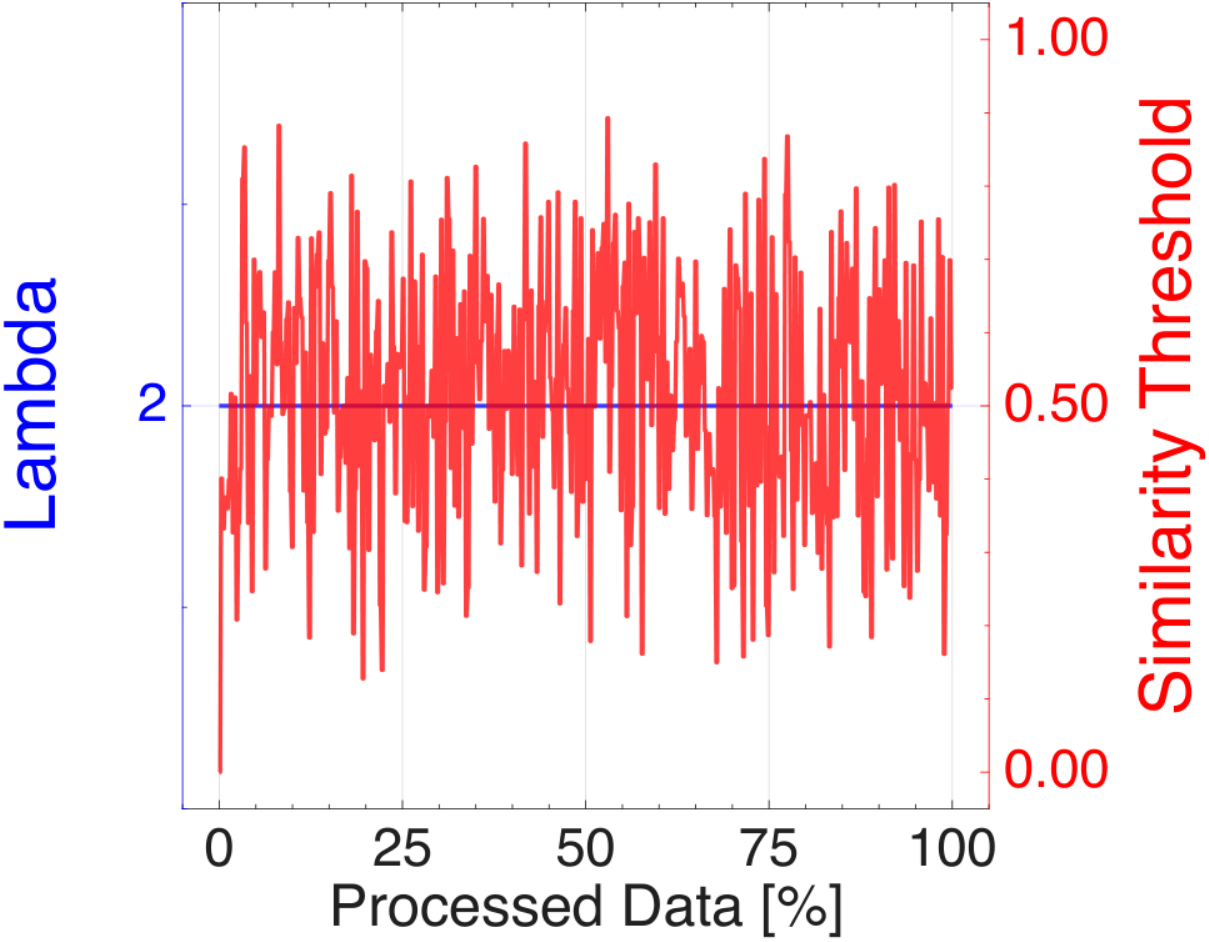}
  }\\
  \subfloat[Mice Protein]{%
    \includegraphics[width=0.22\linewidth]{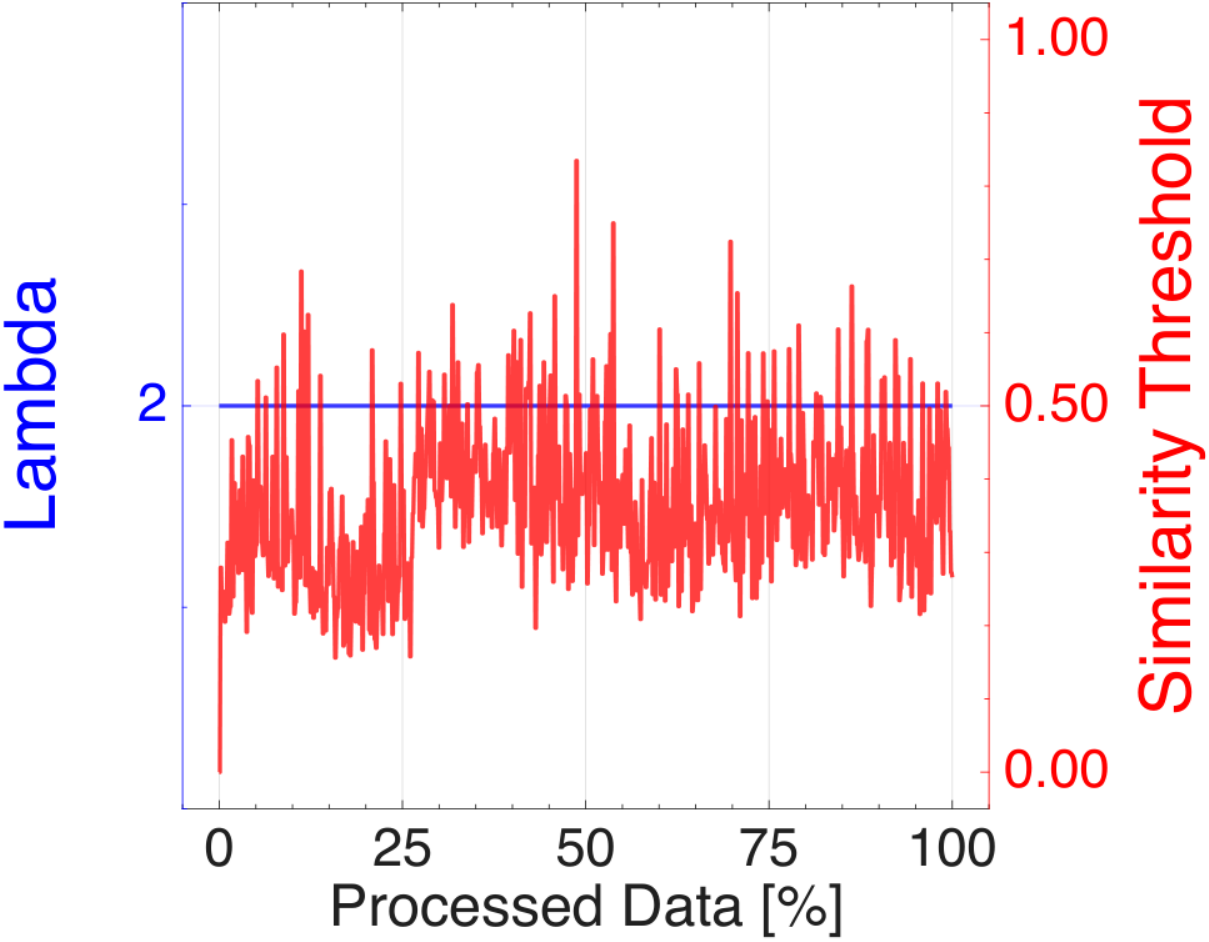}
  }\hfill
  \subfloat[Binalpha]{%
    \includegraphics[width=0.22\linewidth]{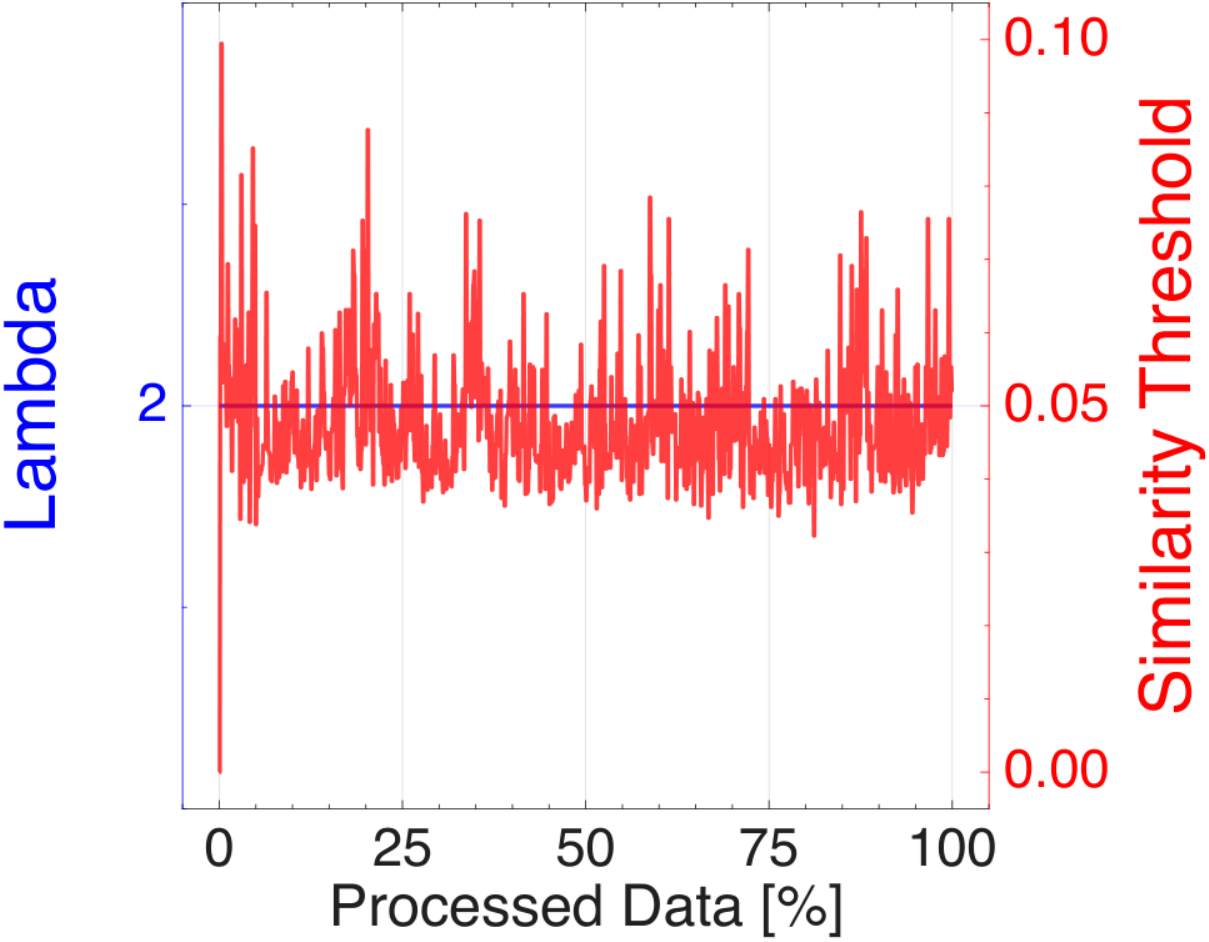}
  }\hfill
  \subfloat[Yeast]{%
    \includegraphics[width=0.22\linewidth]{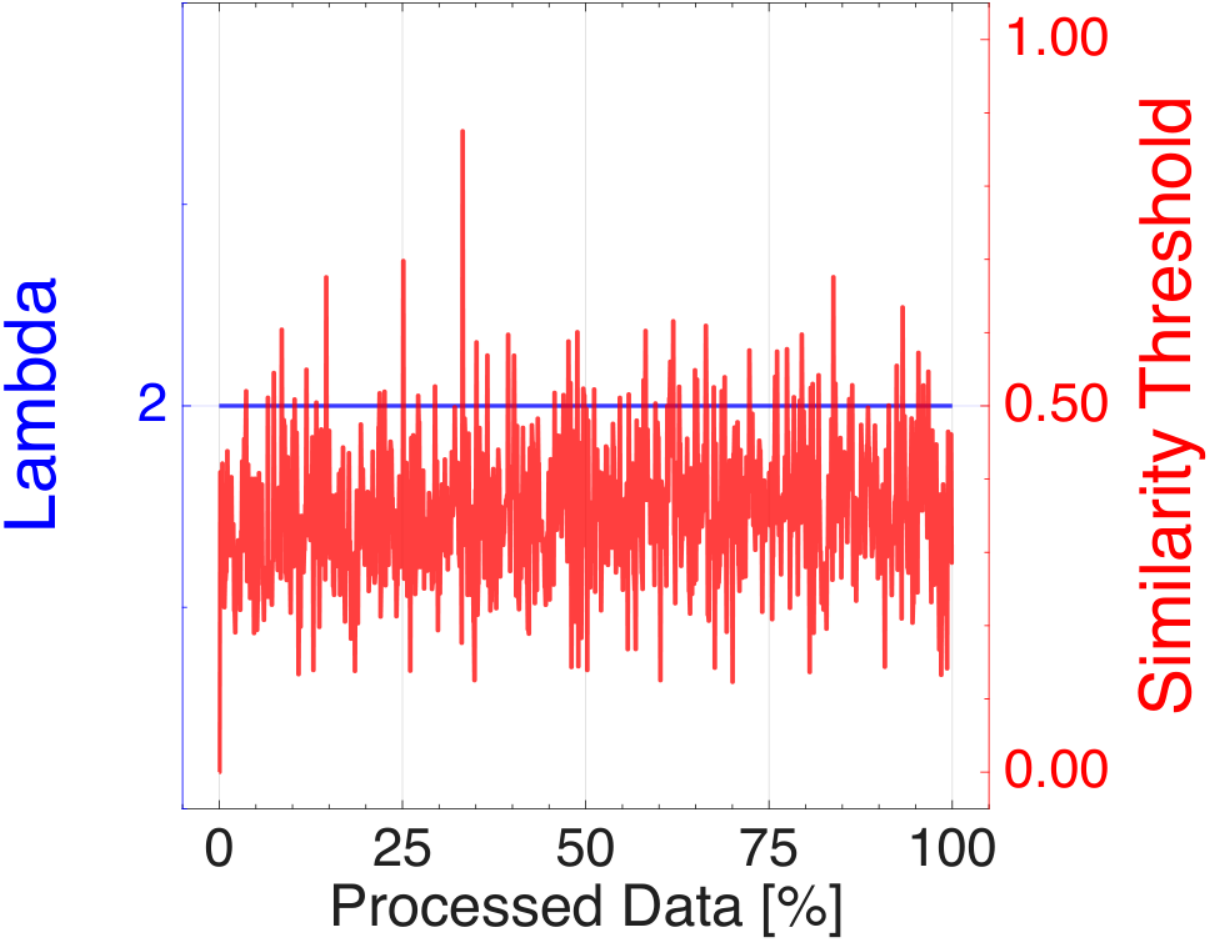}
  }\hfill
  \subfloat[Semeion]{%
    \includegraphics[width=0.22\linewidth]{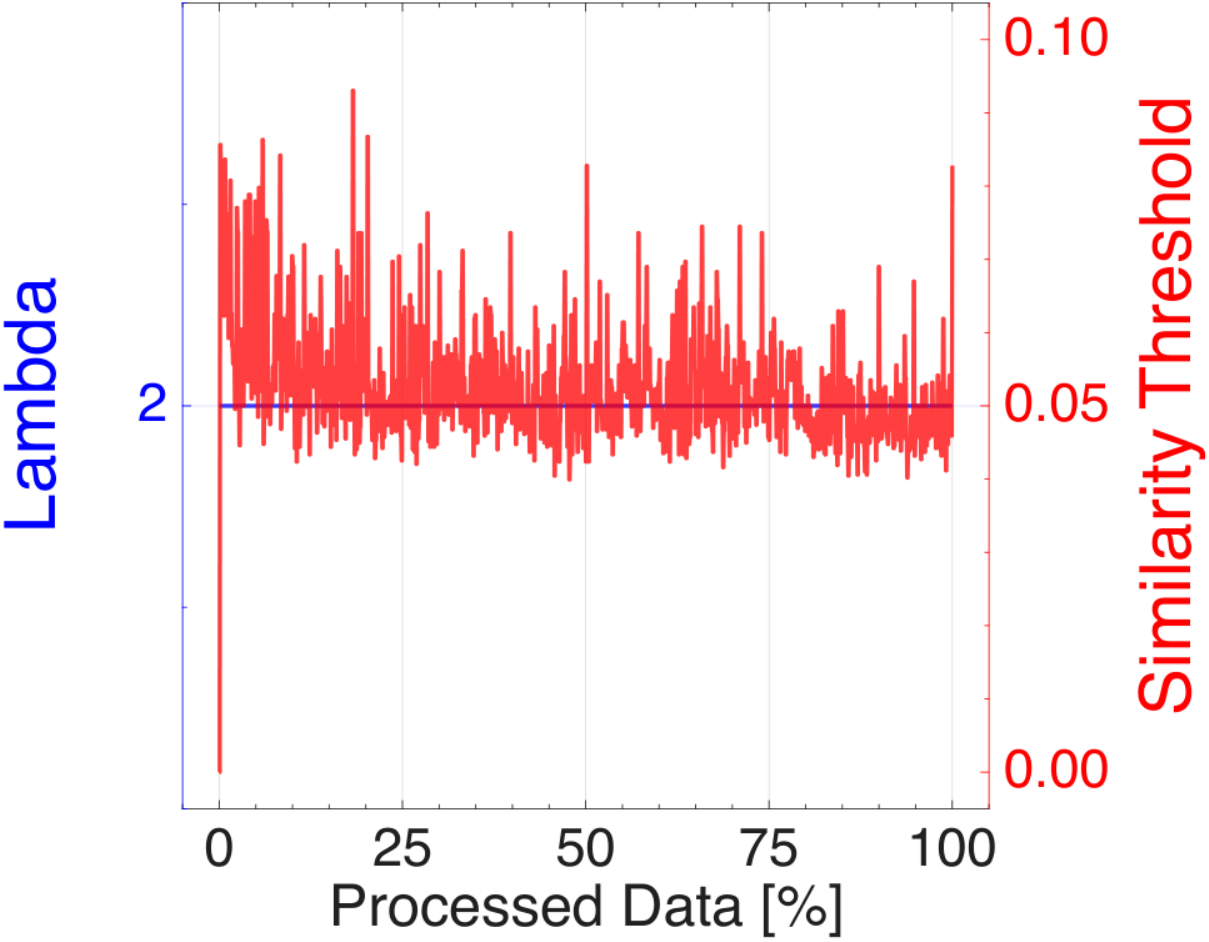}
  }\\
  \subfloat[MSRA25]{%
    \includegraphics[width=0.22\linewidth]{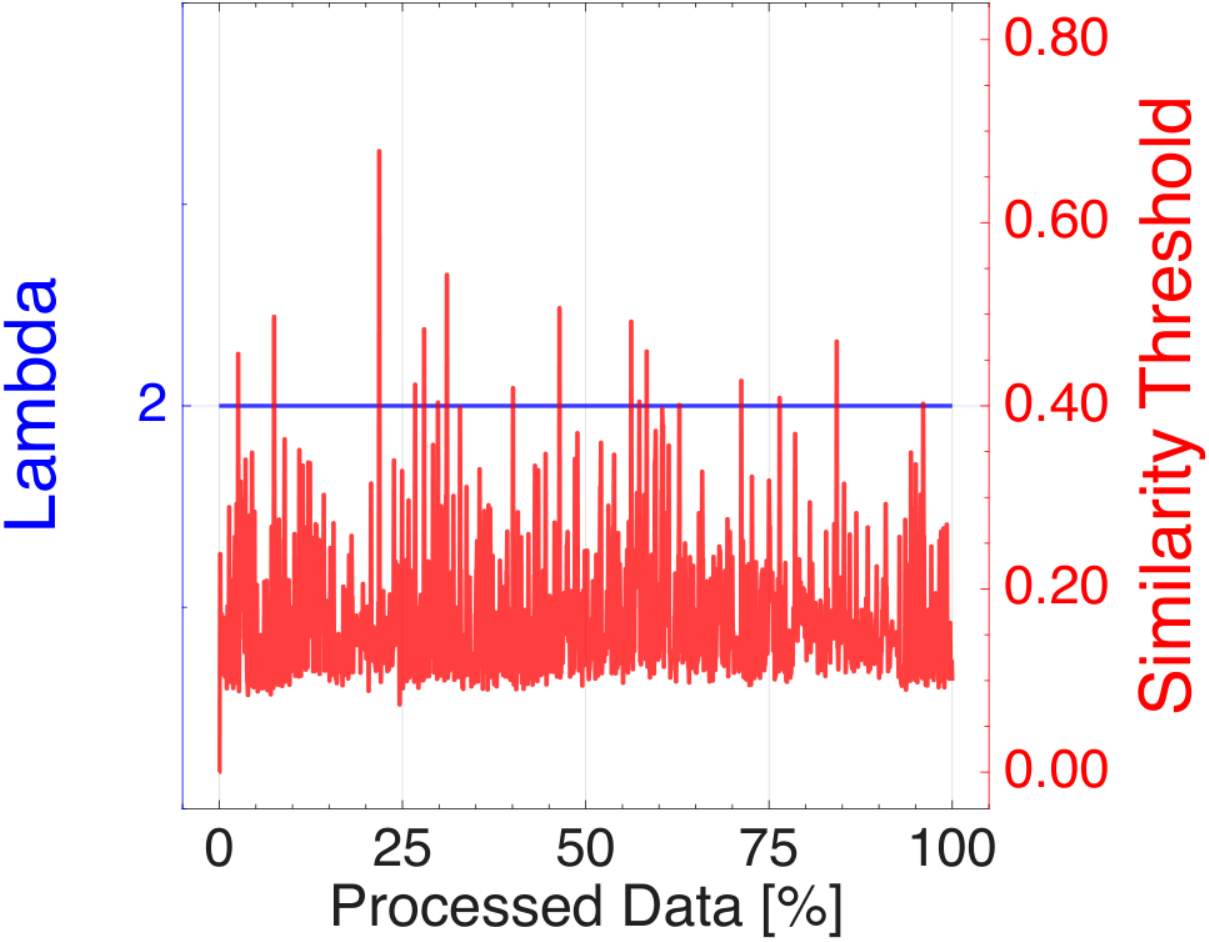}
  }\hfill
  \subfloat[Image Segmentation]{%
    \includegraphics[width=0.22\linewidth]{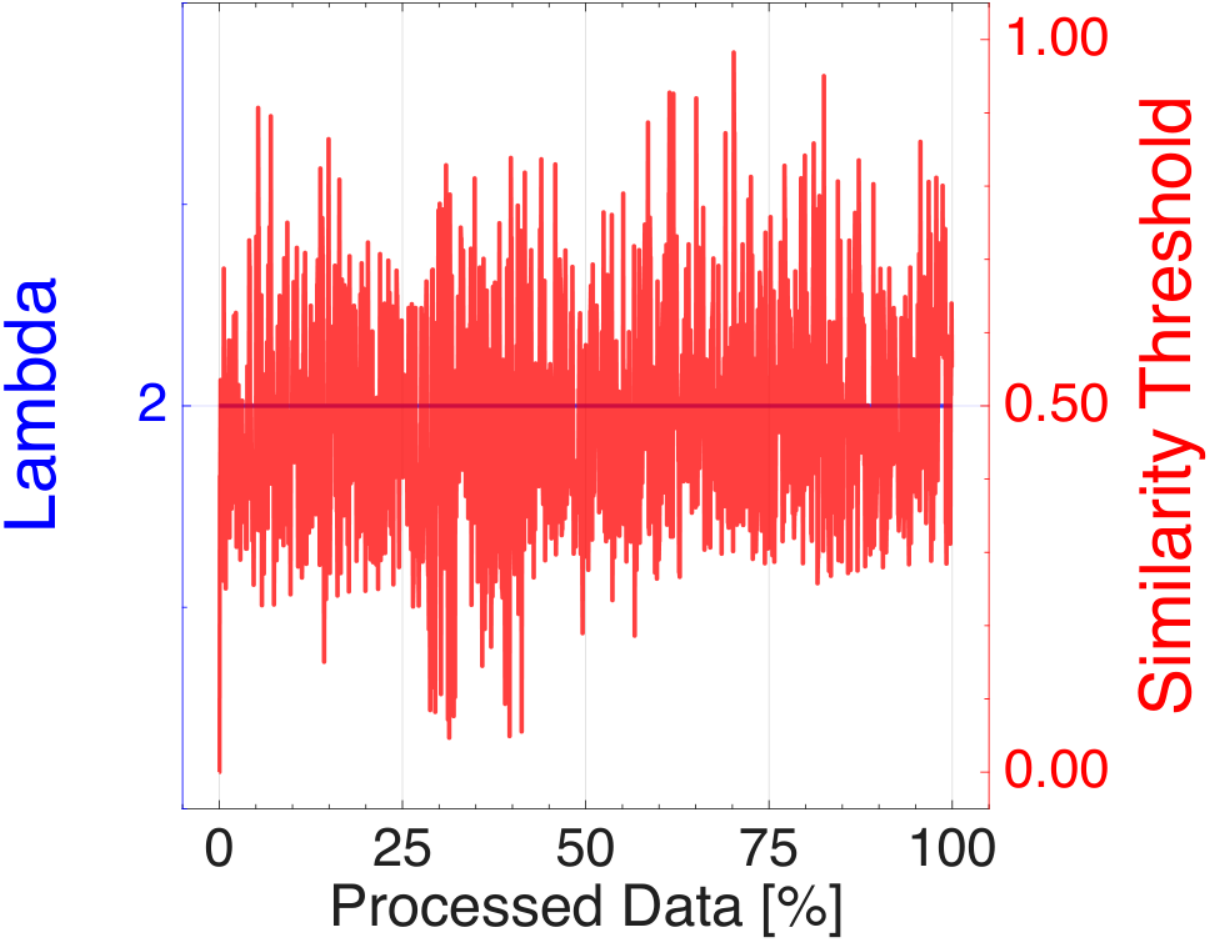}
  }\hfill
  \subfloat[Rice]{%
    \includegraphics[width=0.22\linewidth]{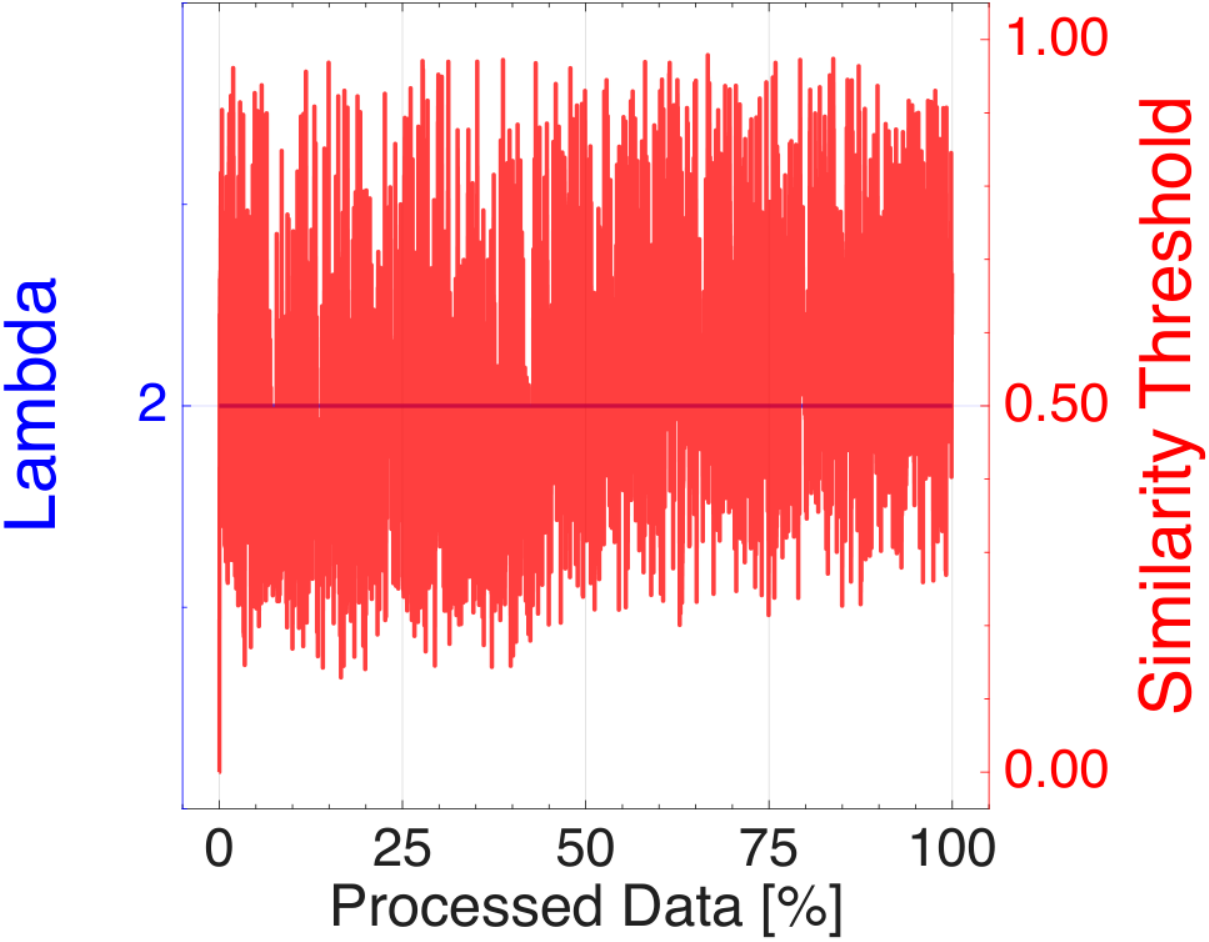}
  }\hfill
  \subfloat[TUANDROMD]{%
    \includegraphics[width=0.22\linewidth]{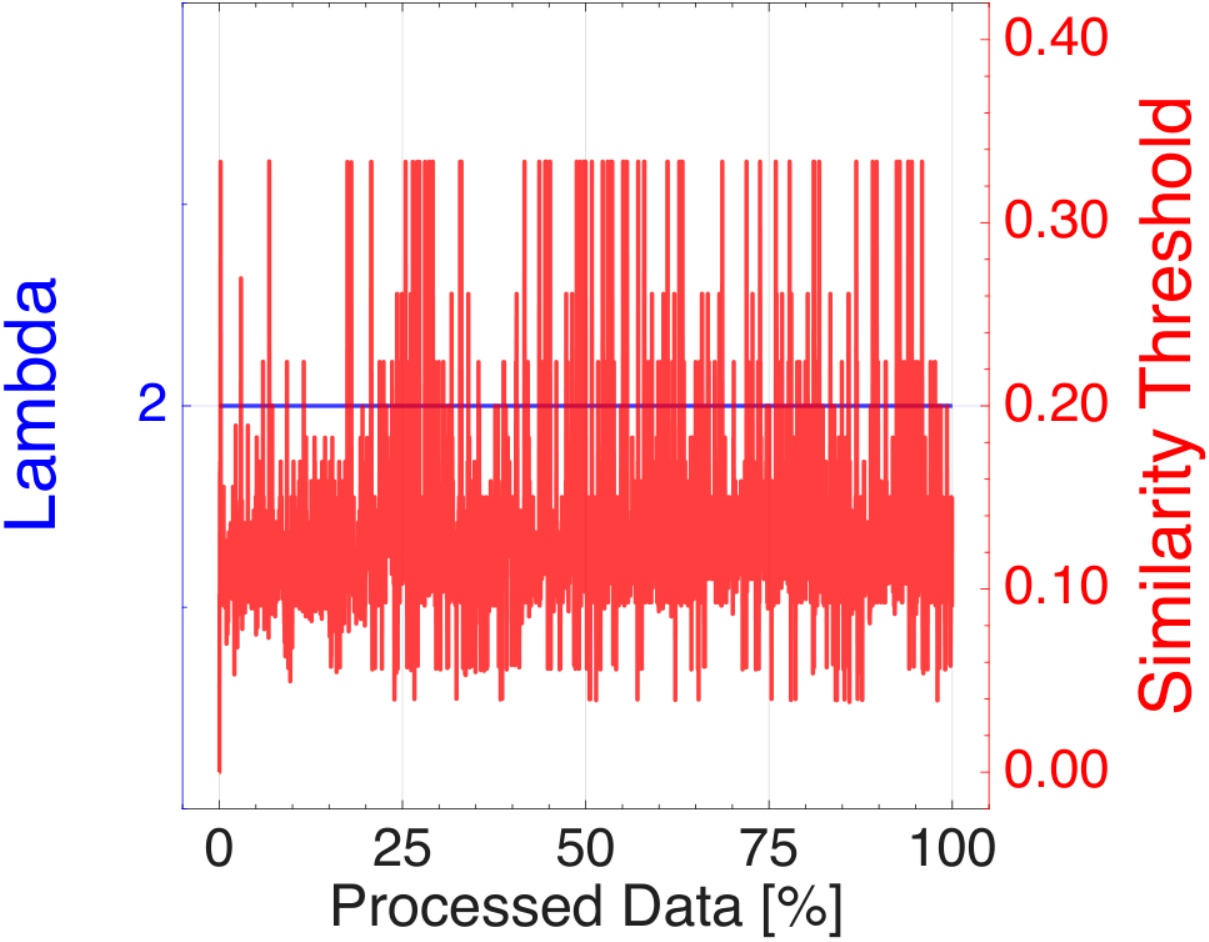}
  }\\
  \subfloat[Phoneme]{%
    \includegraphics[width=0.22\linewidth]{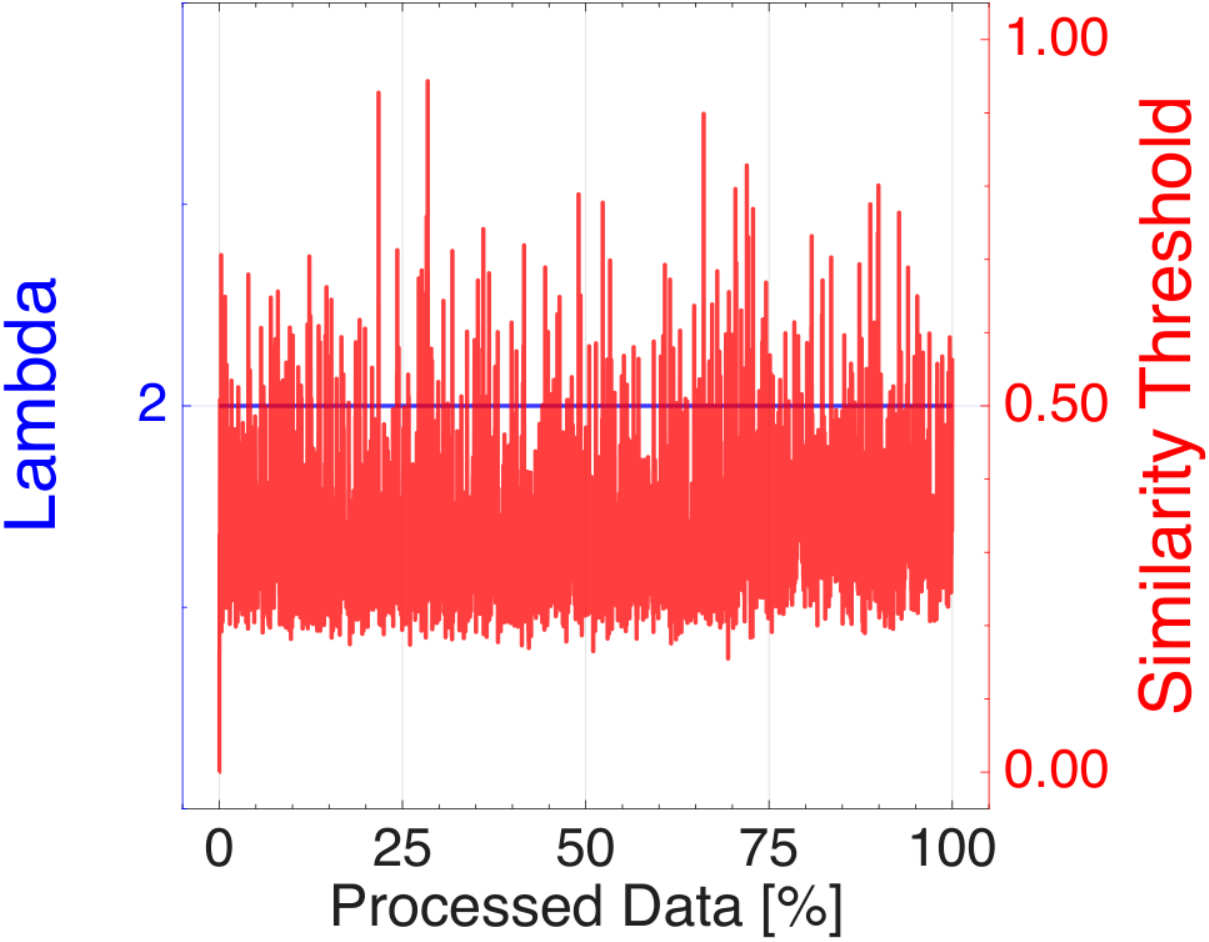}
  }\hfill
  \subfloat[Texture]{%
    \includegraphics[width=0.22\linewidth]{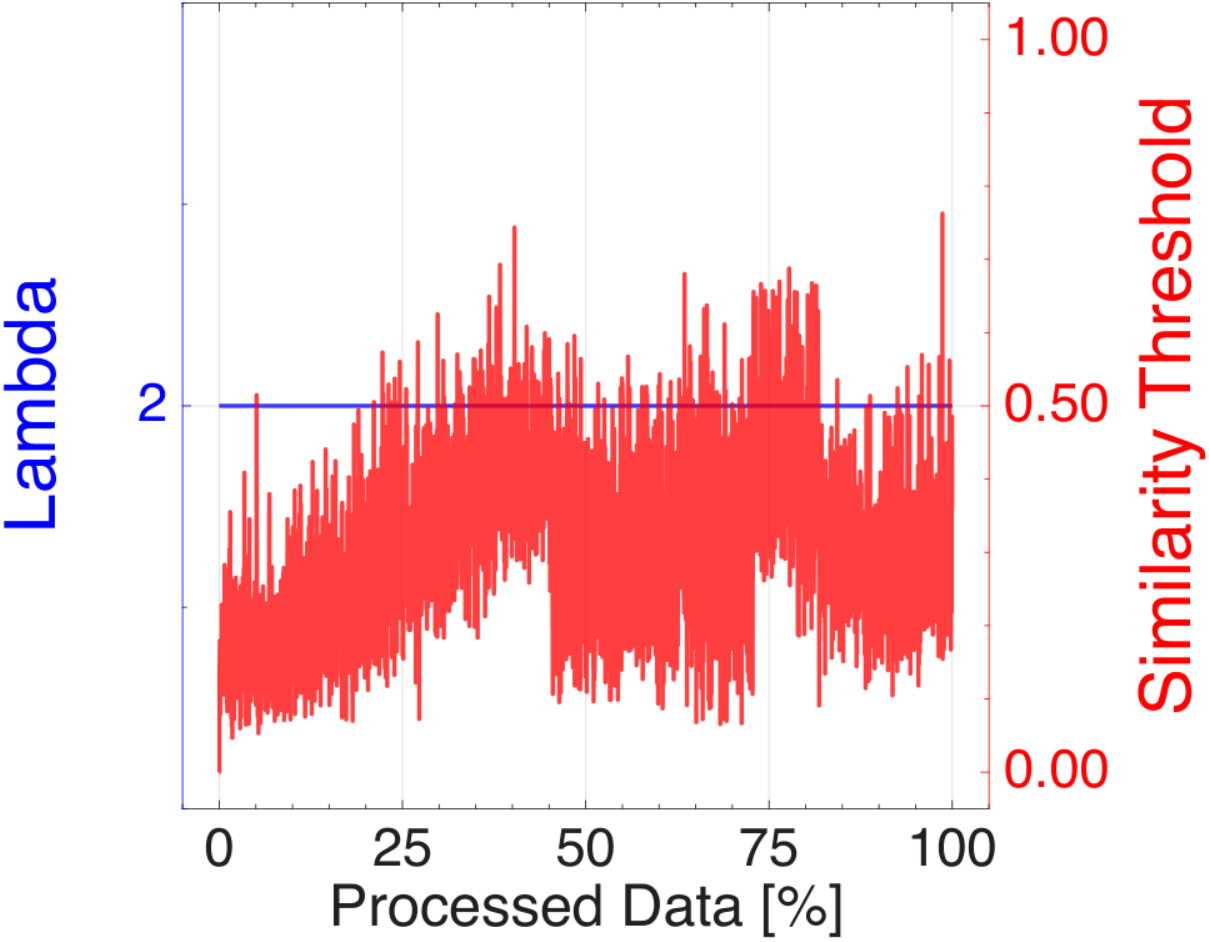}
  }\hfill
  \subfloat[OptDigits]{%
    \includegraphics[width=0.22\linewidth]{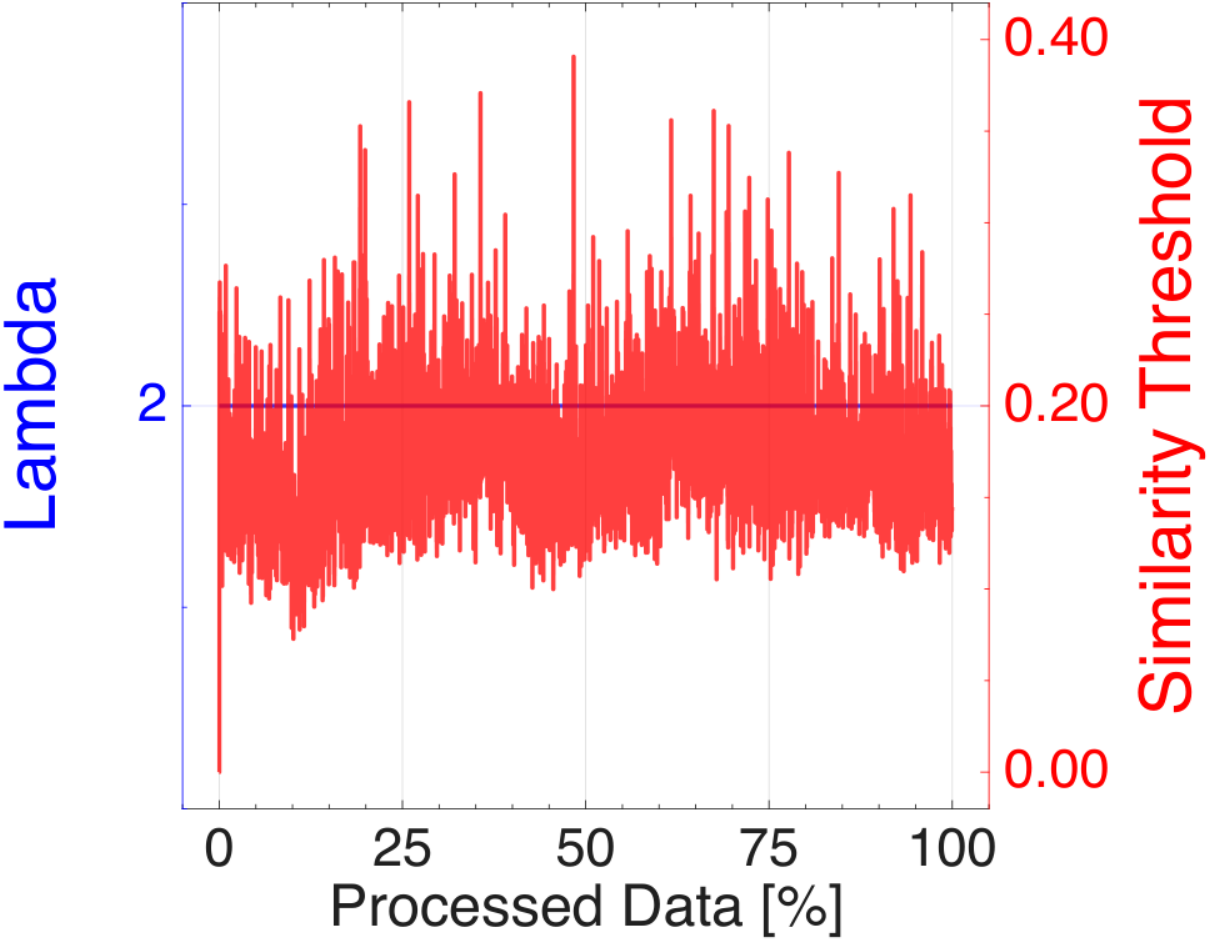}
  }\hfill
  \subfloat[Statlog]{%
    \includegraphics[width=0.22\linewidth]{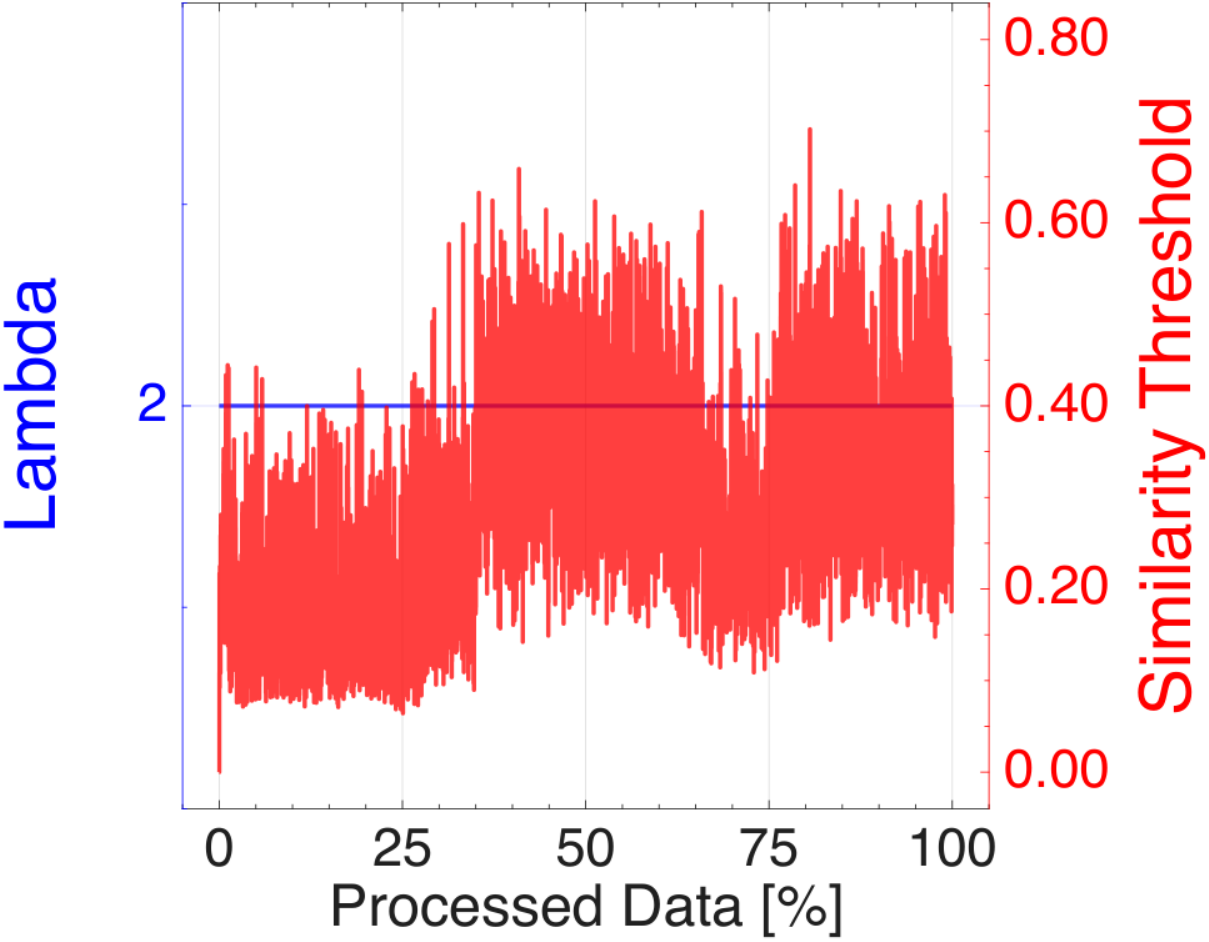}
  }\\
  \subfloat[Anuran Calls]{%
    \includegraphics[width=0.22\linewidth]{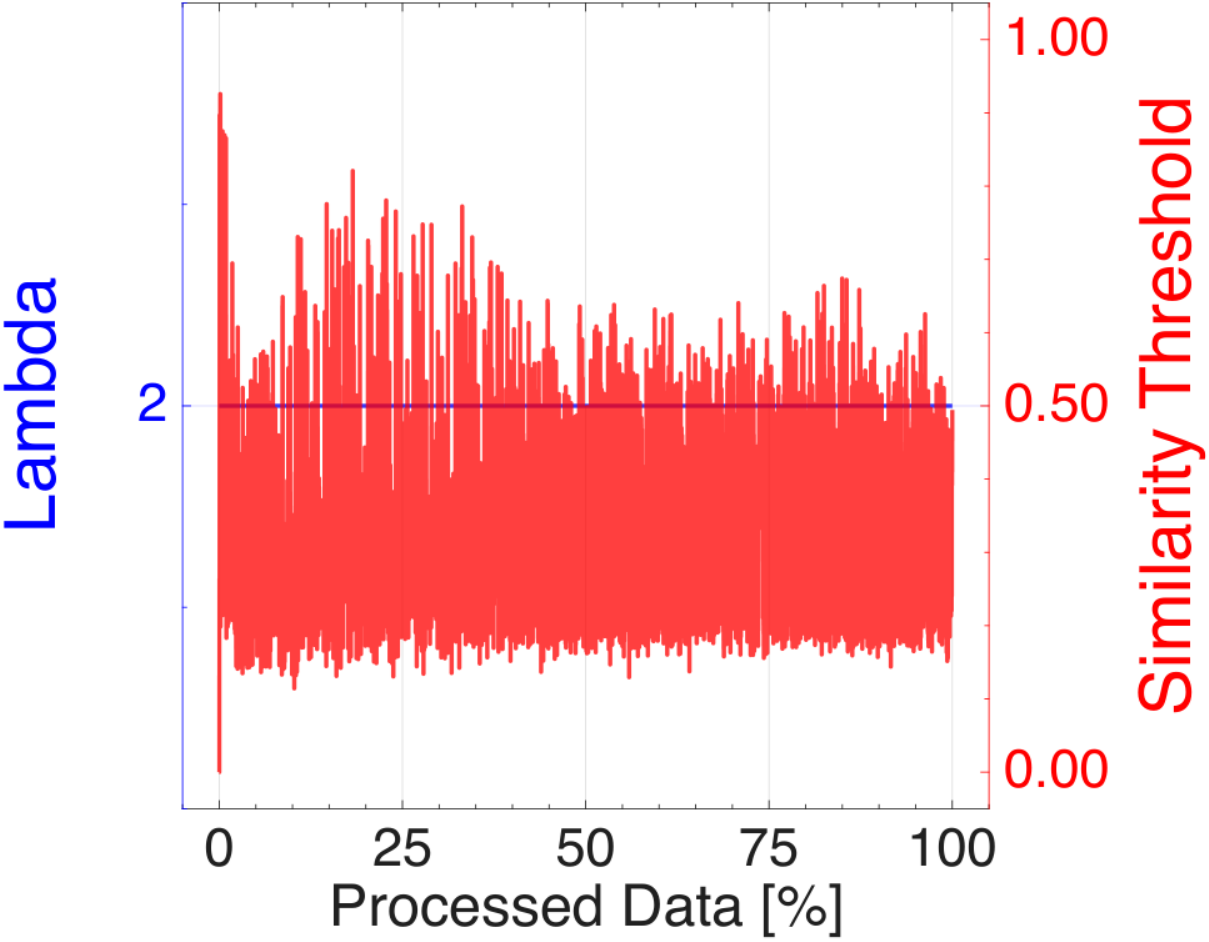}
  }\hfill
  \subfloat[Isolet]{%
    \includegraphics[width=0.22\linewidth]{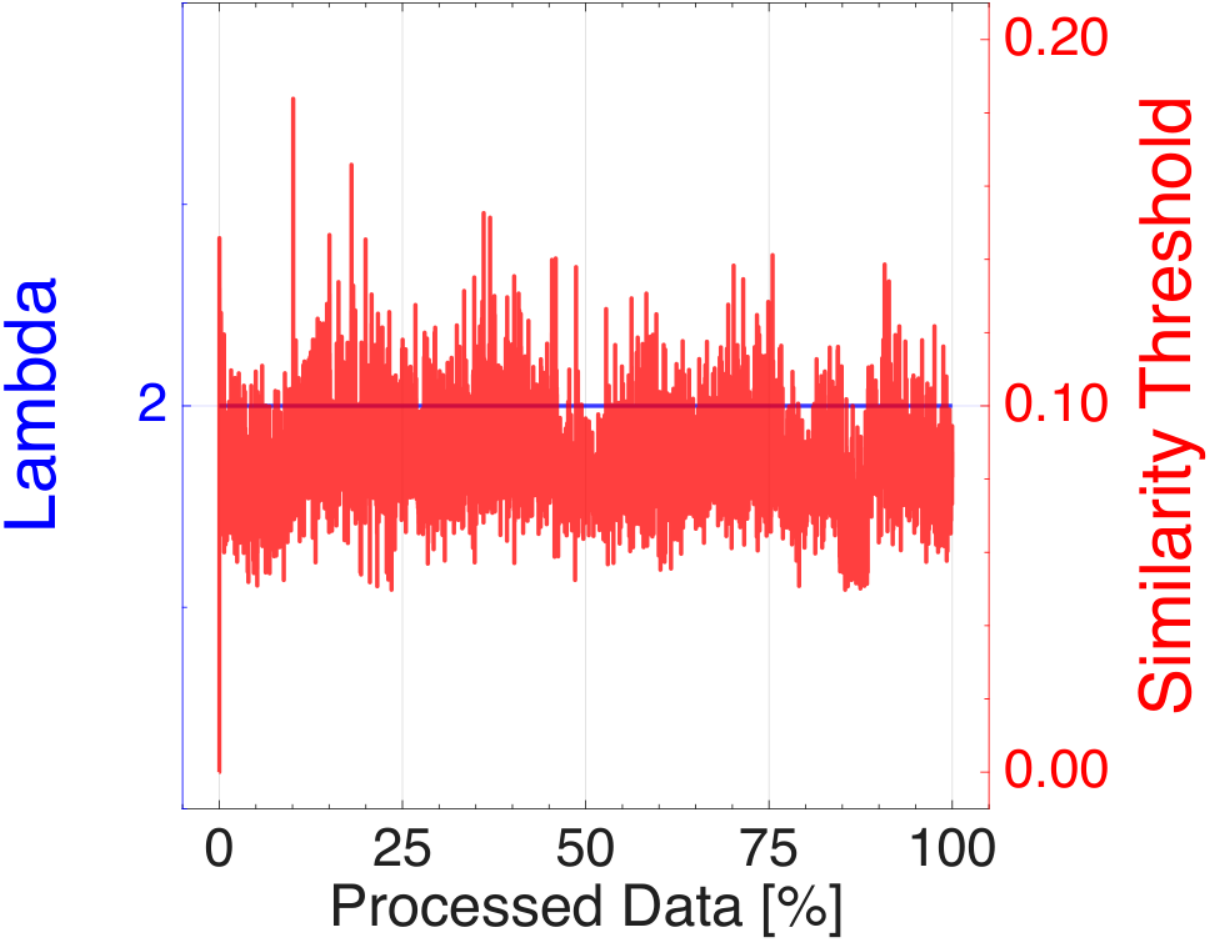}
  }\hfill
  \subfloat[MNIST10K]{%
    \includegraphics[width=0.22\linewidth]{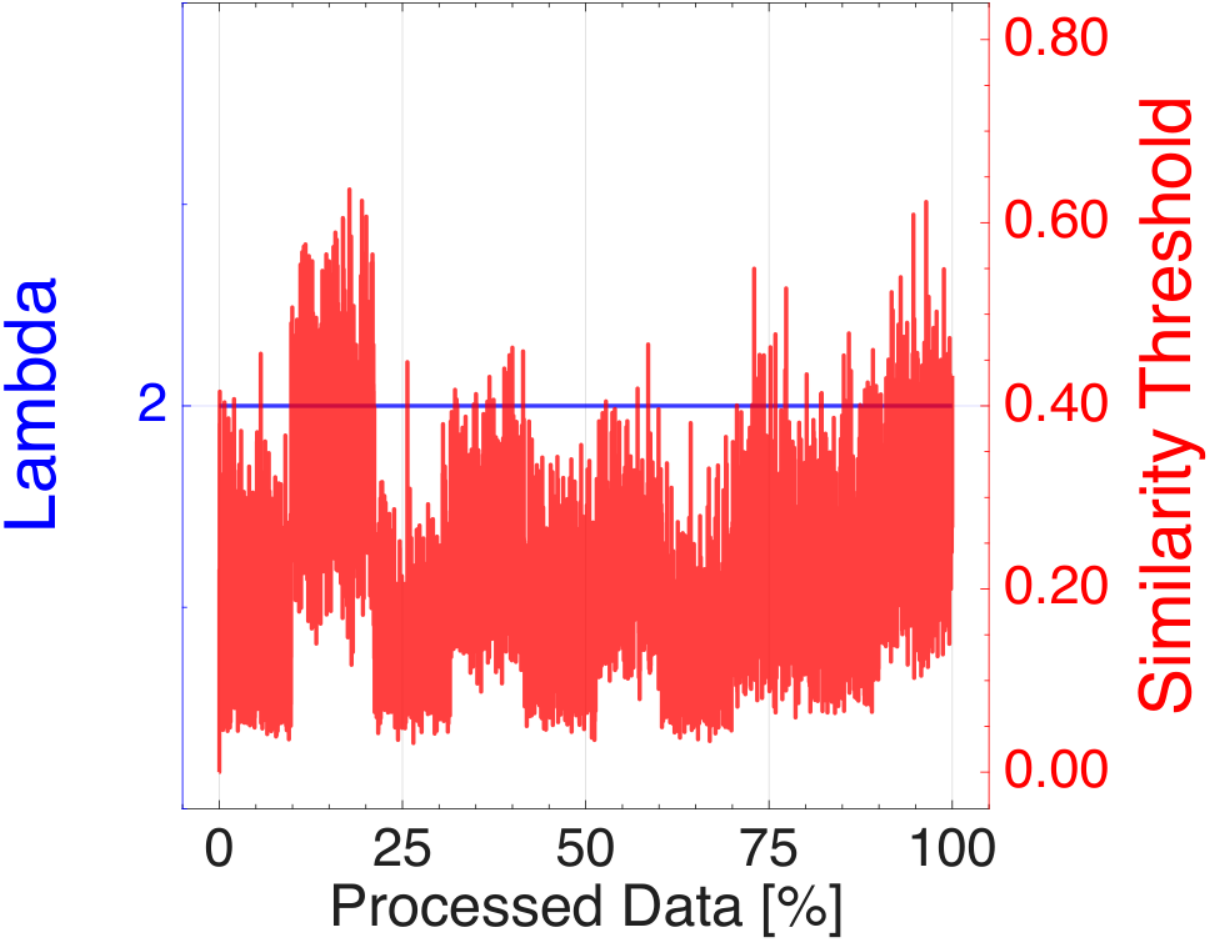}
  }\hfill
  \subfloat[PenBased]{%
    \includegraphics[width=0.22\linewidth]{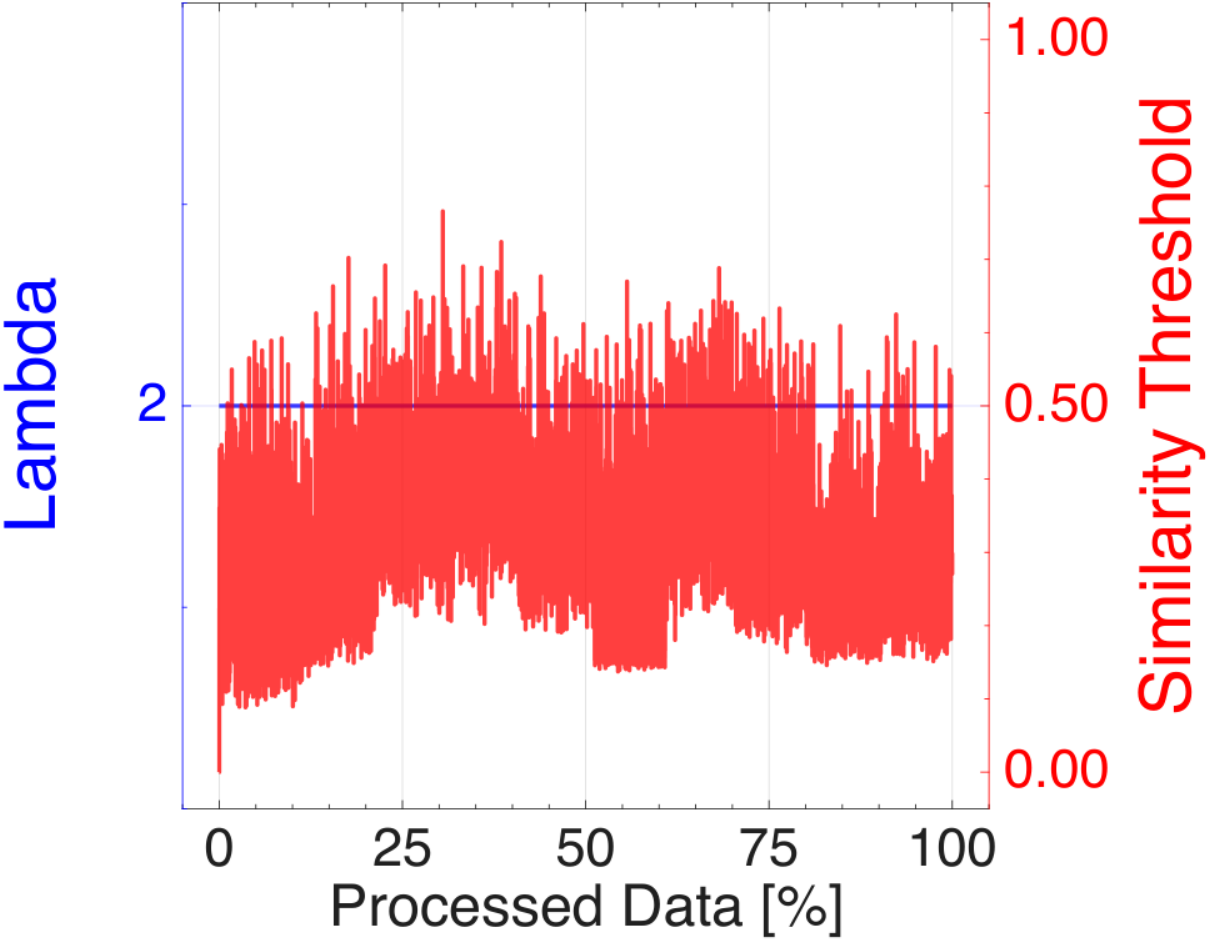}
  }\\
  \subfloat[STL10]{%
    \includegraphics[width=0.22\linewidth]{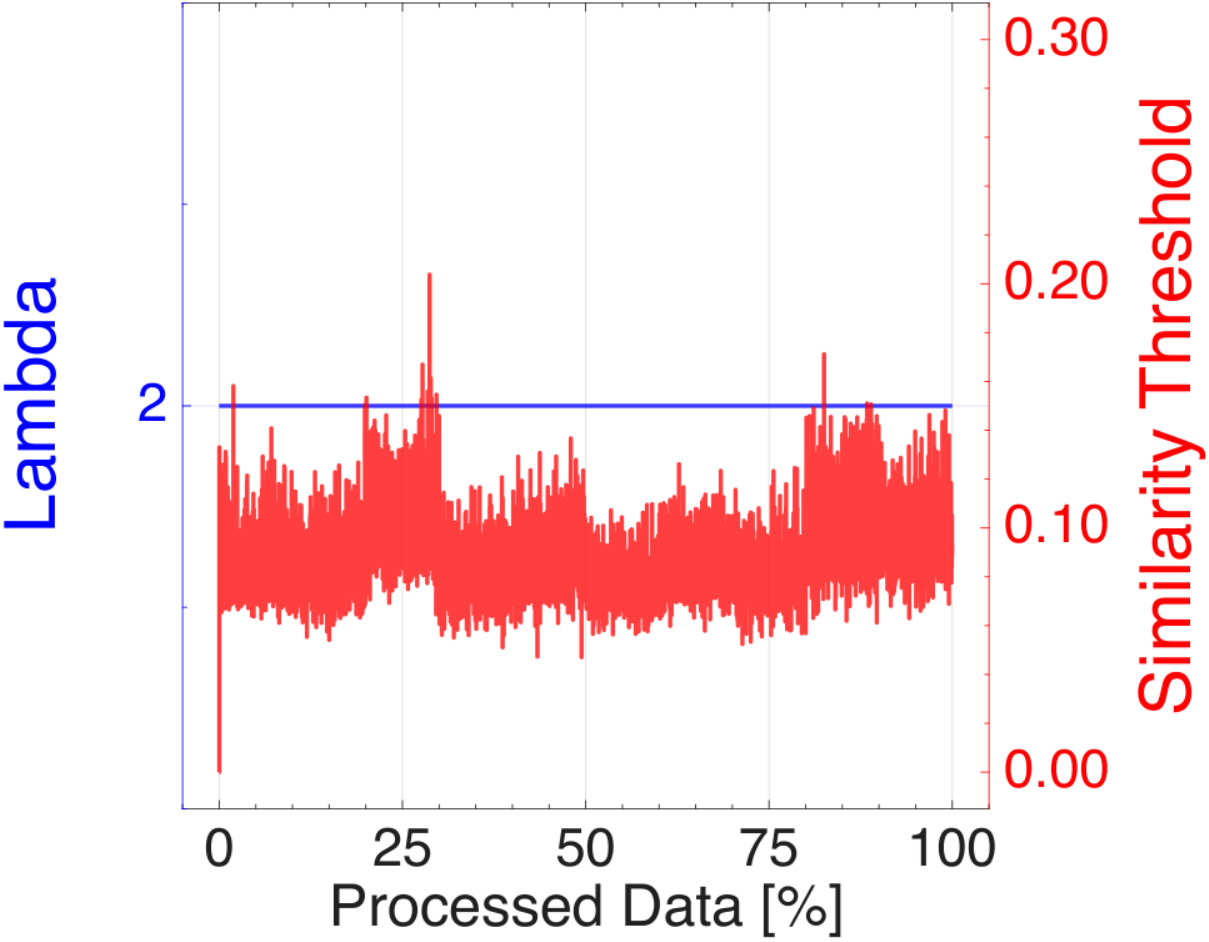}
  }\hfill
  \subfloat[Letter]{%
    \includegraphics[width=0.22\linewidth]{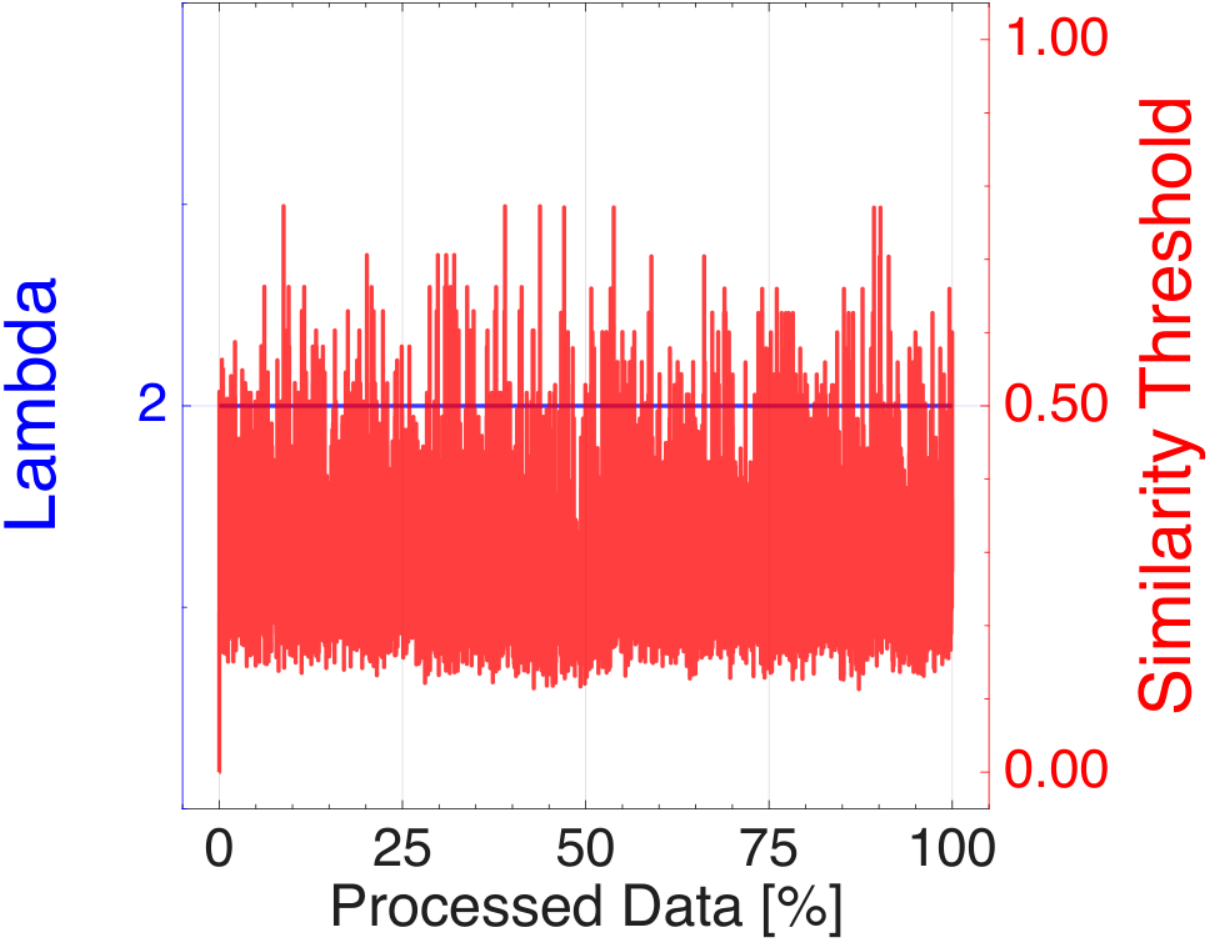}
  }\hfill
  \subfloat[Shuttle]{%
    \includegraphics[width=0.22\linewidth]{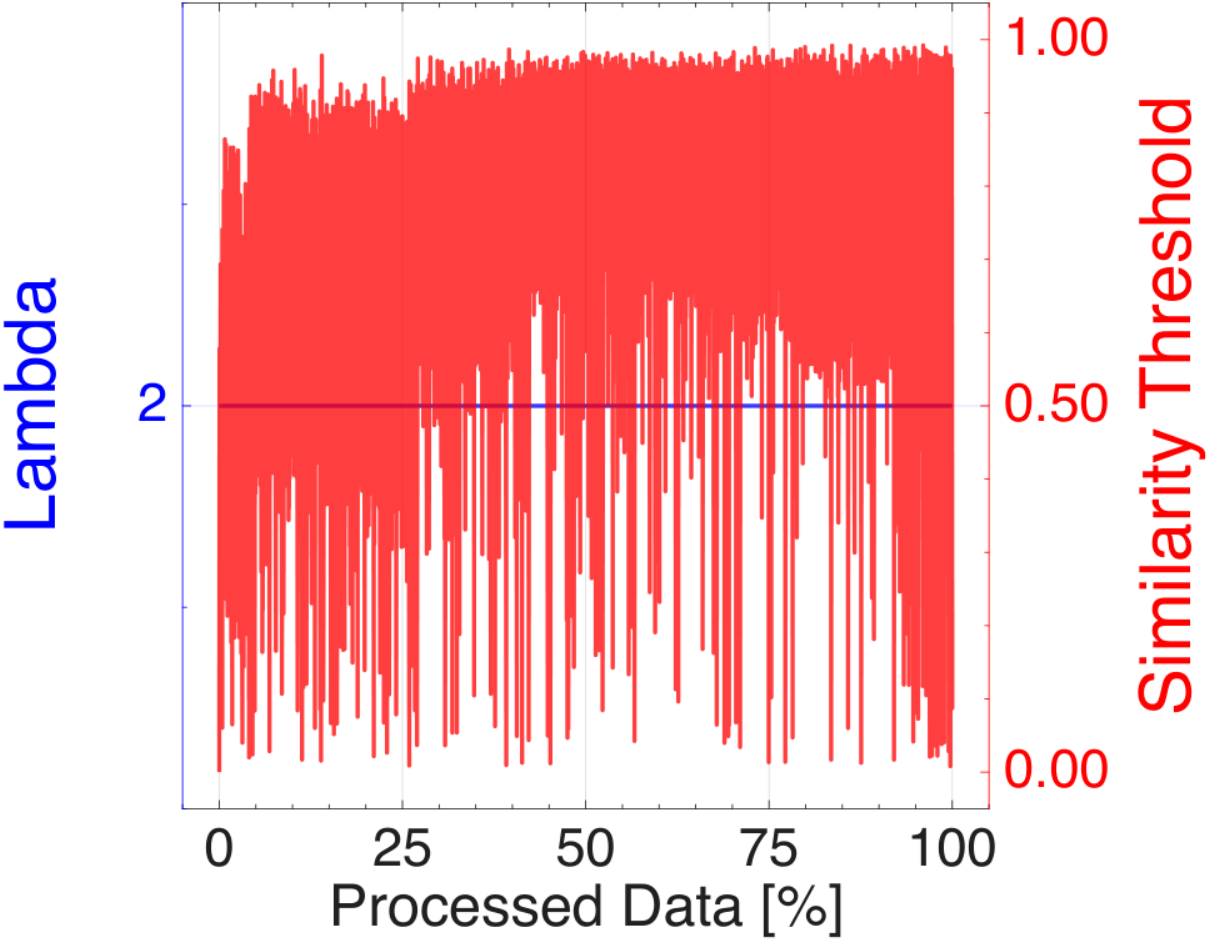}
  }\hfill
  \subfloat[Skin]{%
    \includegraphics[width=0.22\linewidth]{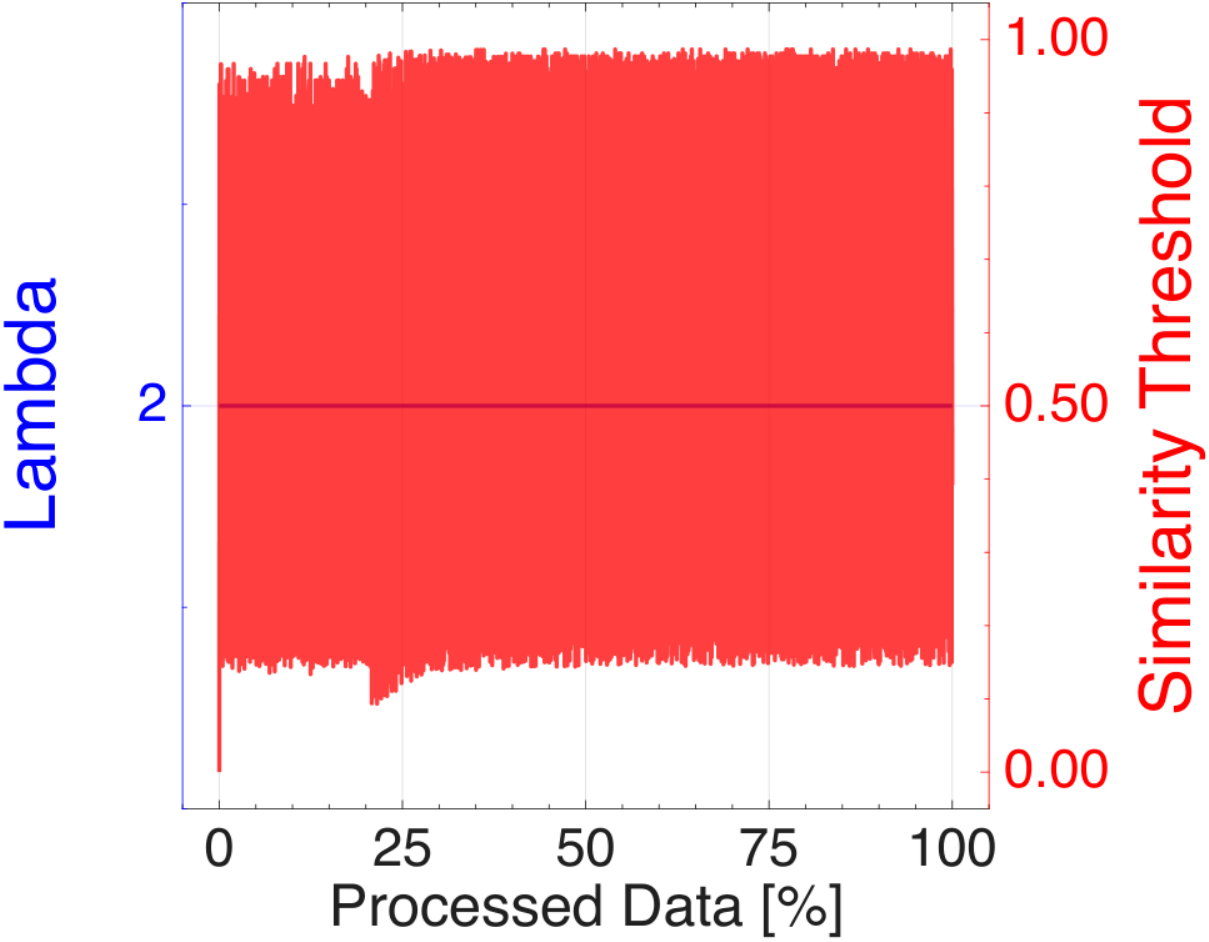}
  }
  \caption{Histories of $\Lambda$ and $V_{\text{threshold}}$ for the w/o Inc. variant in the nonstationary setting ($\Lambda_{\text{init}} = 2$).}
  \label{fig:ablation_lambda_history_noincrease_nonstationary}
\end{figure*}

\begin{figure*}[htbp]
  \centering
  \subfloat[Iris]{%
    \includegraphics[width=0.22\linewidth]{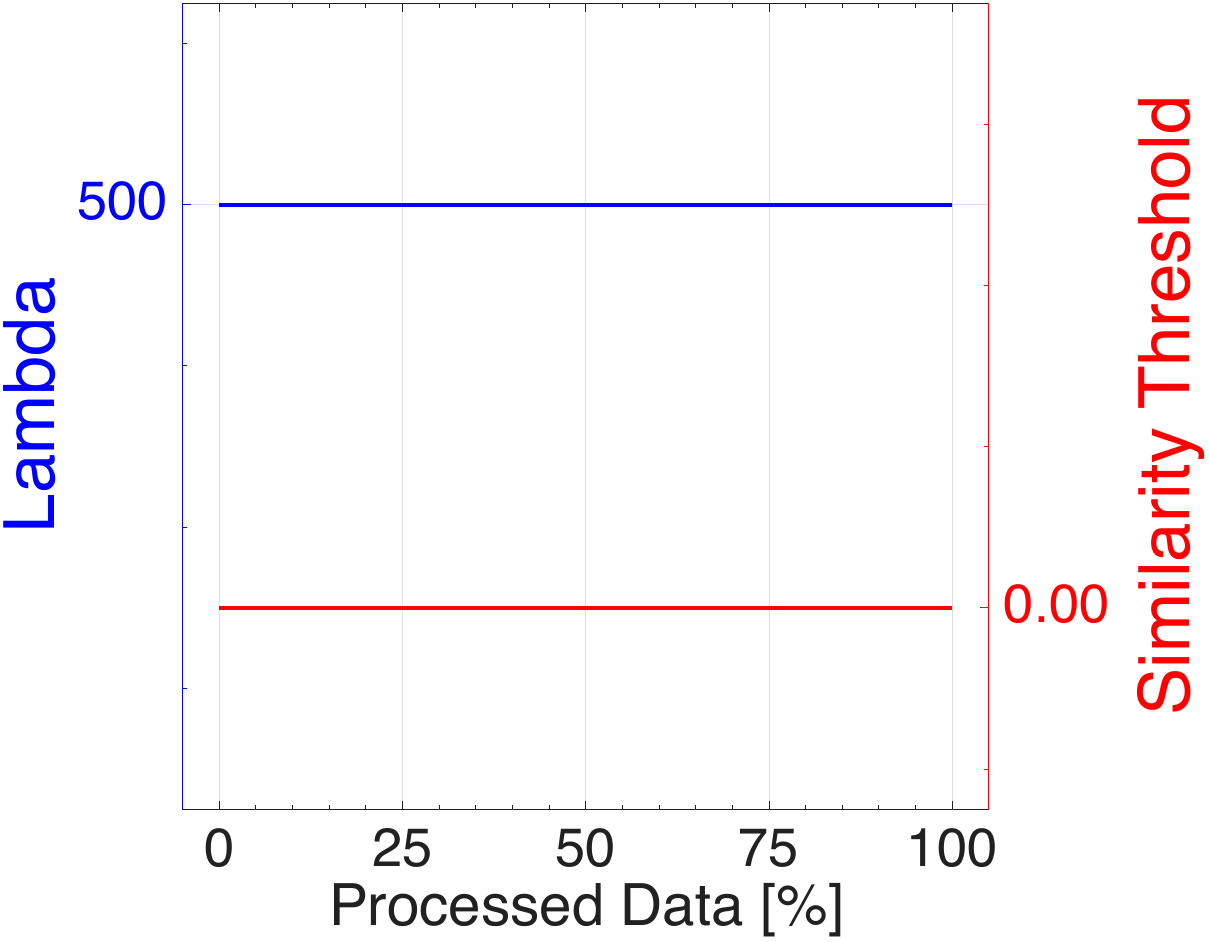}
  }\hfill
  \subfloat[Seeds]{%
    \includegraphics[width=0.22\linewidth]{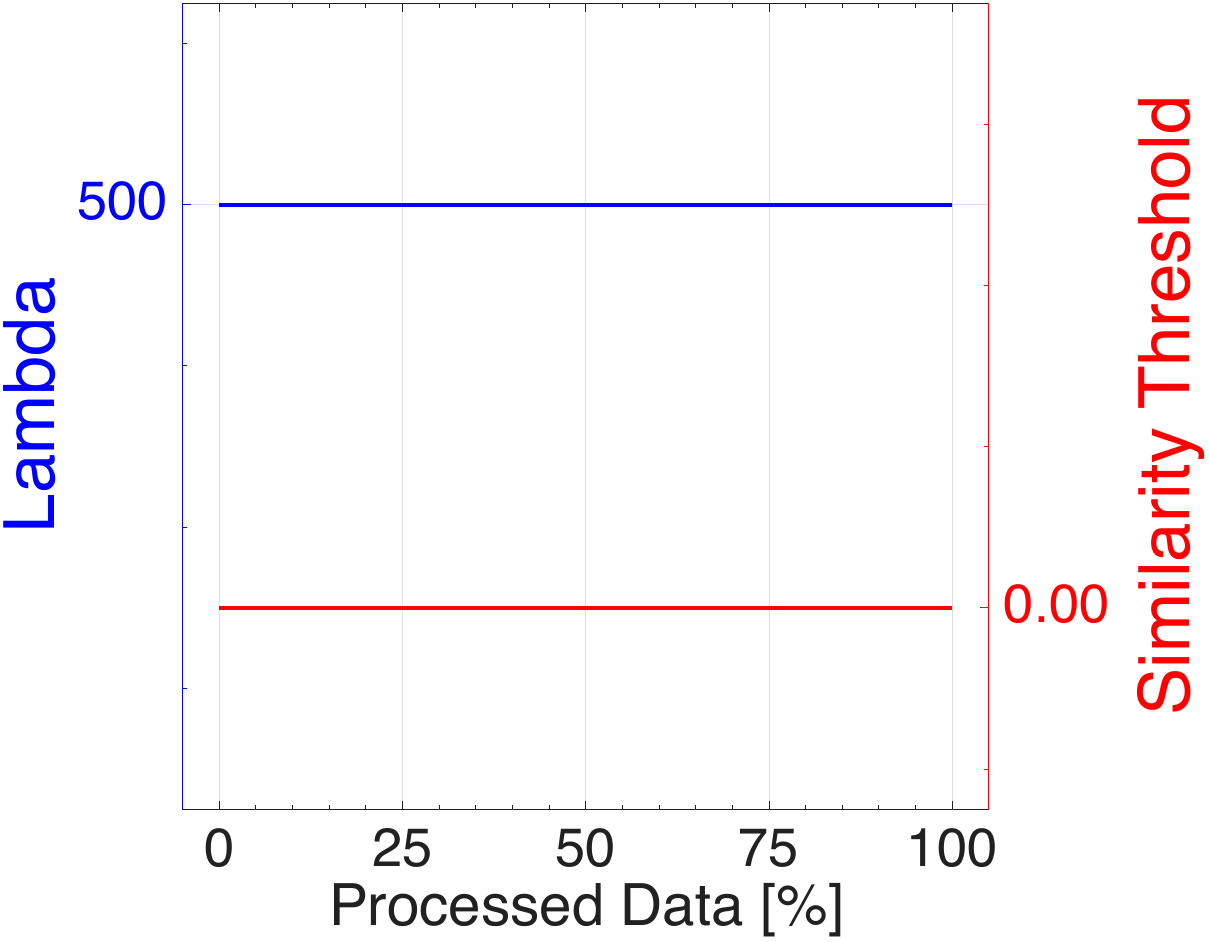}
  }\hfill
  \subfloat[Dermatology]{%
    \includegraphics[width=0.22\linewidth]{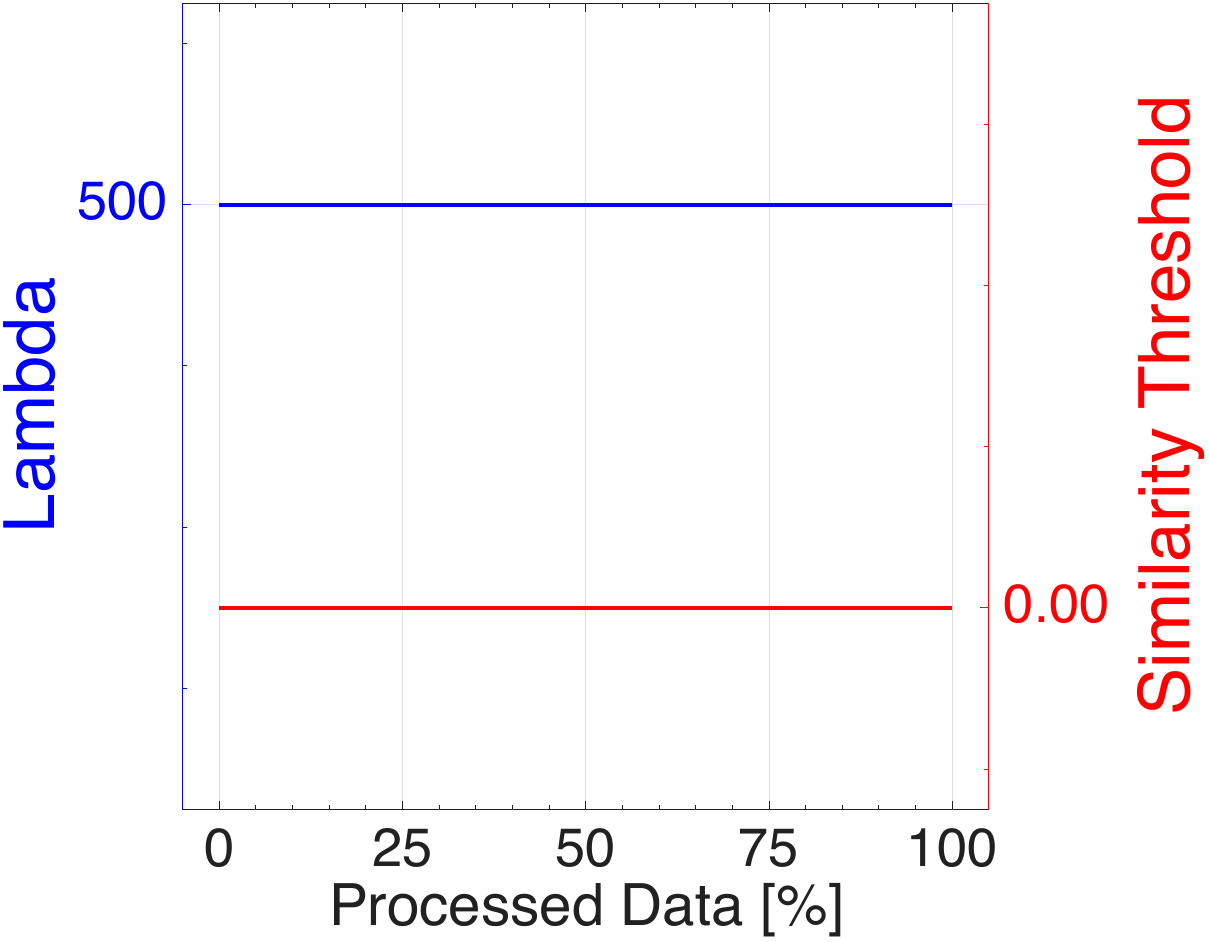}
  }\hfill
  \subfloat[Pima]{%
    \includegraphics[width=0.22\linewidth]{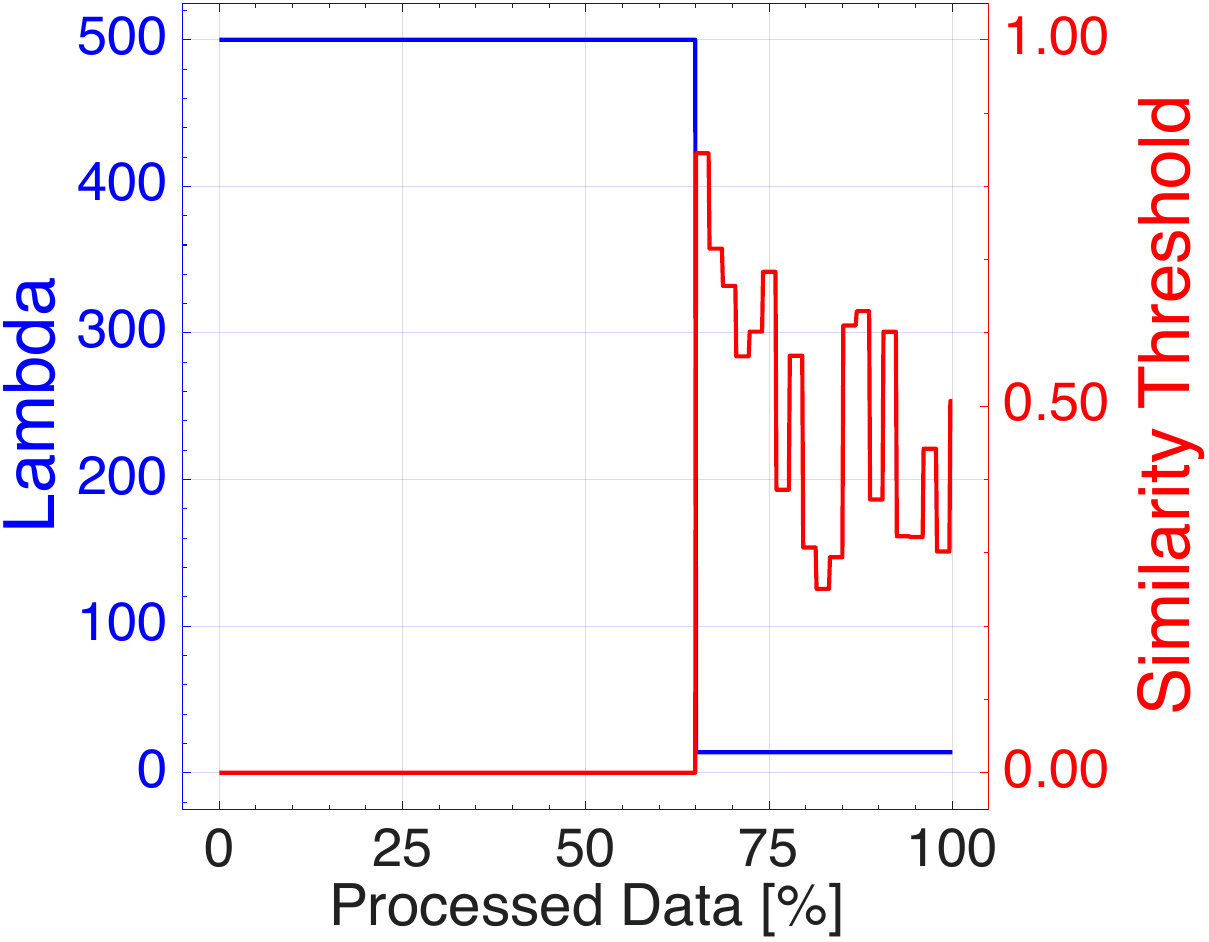}
  }\\
  \subfloat[Mice Protein]{%
    \includegraphics[width=0.22\linewidth]{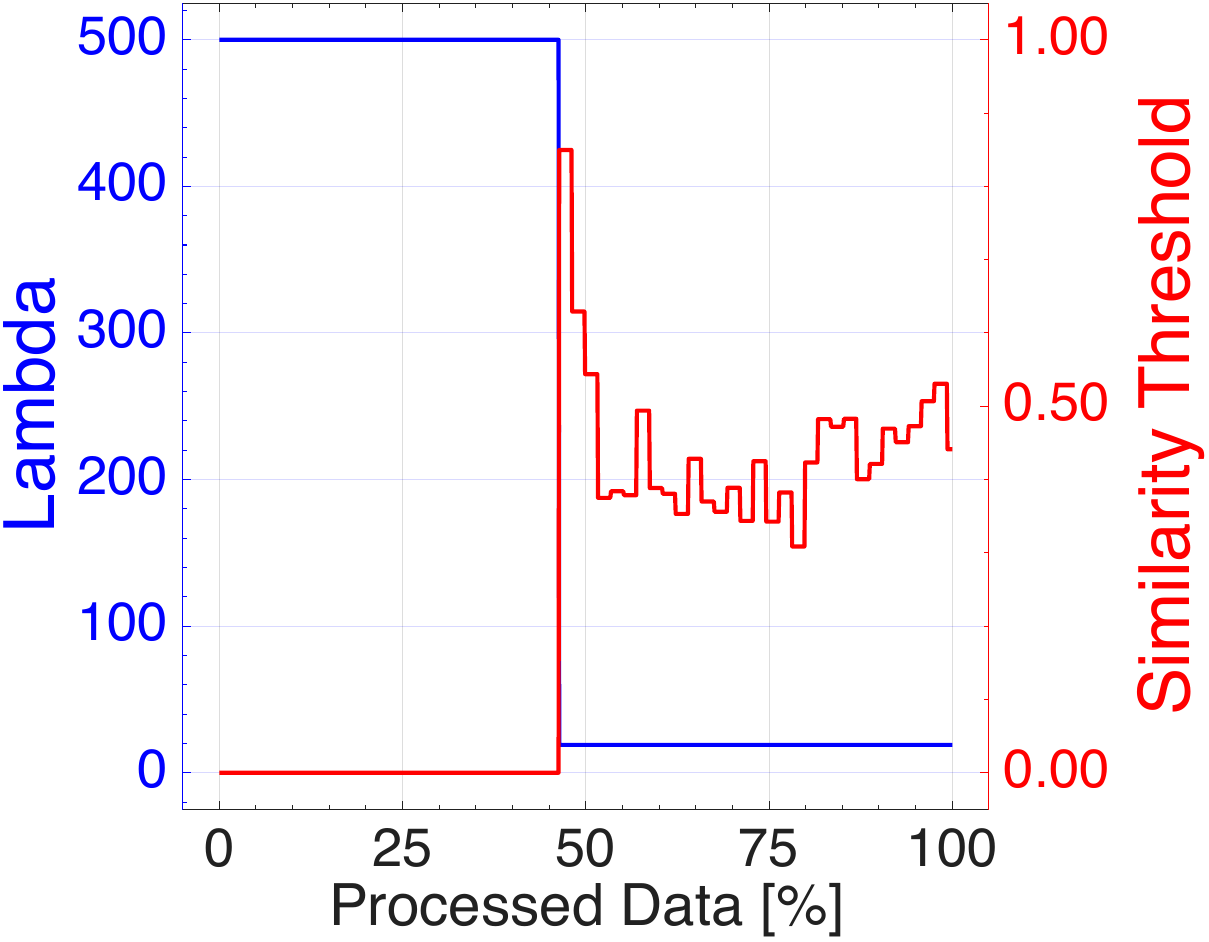}
  }\hfill
  \subfloat[Binalpha]{%
    \includegraphics[width=0.22\linewidth]{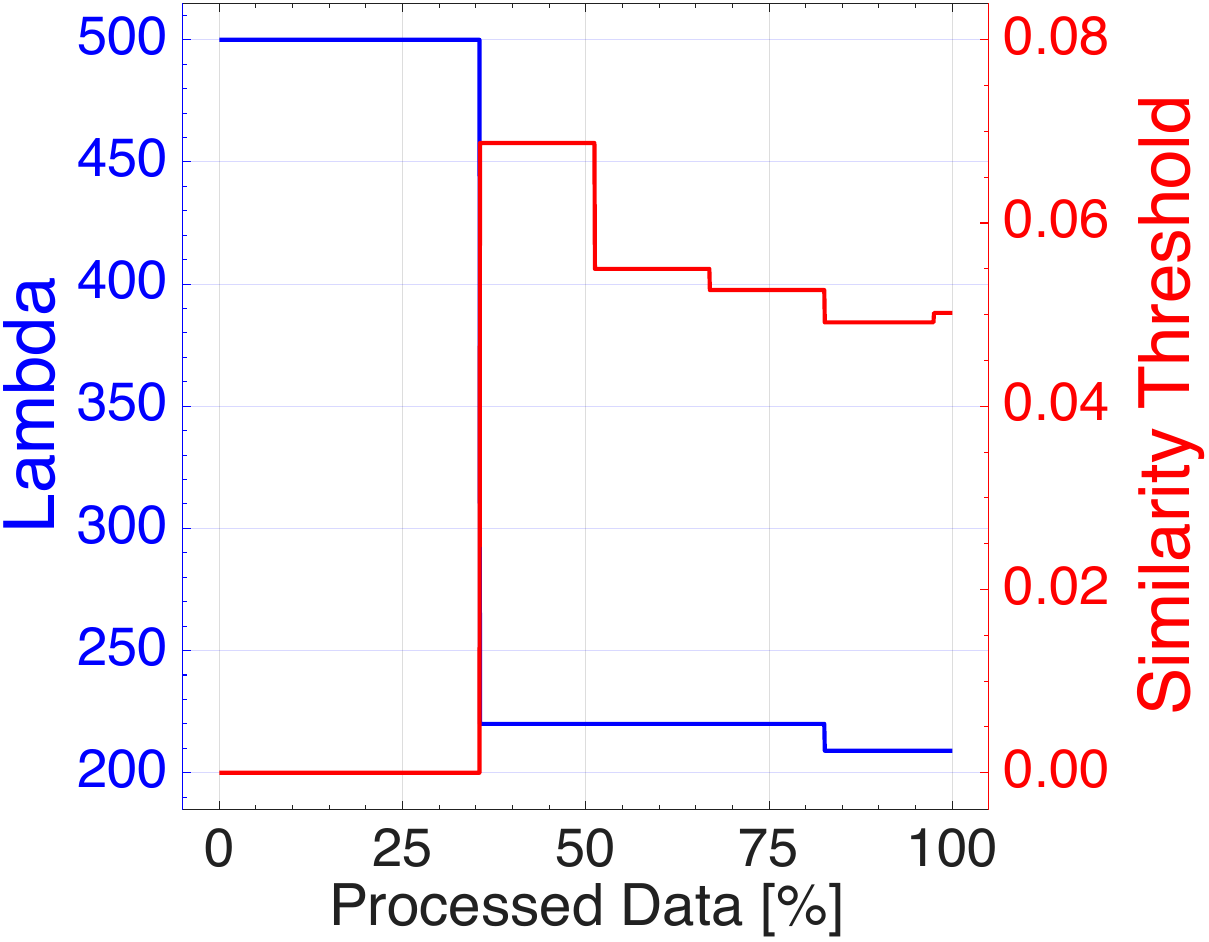}
  }\hfill
  \subfloat[Yeast]{%
    \includegraphics[width=0.22\linewidth]{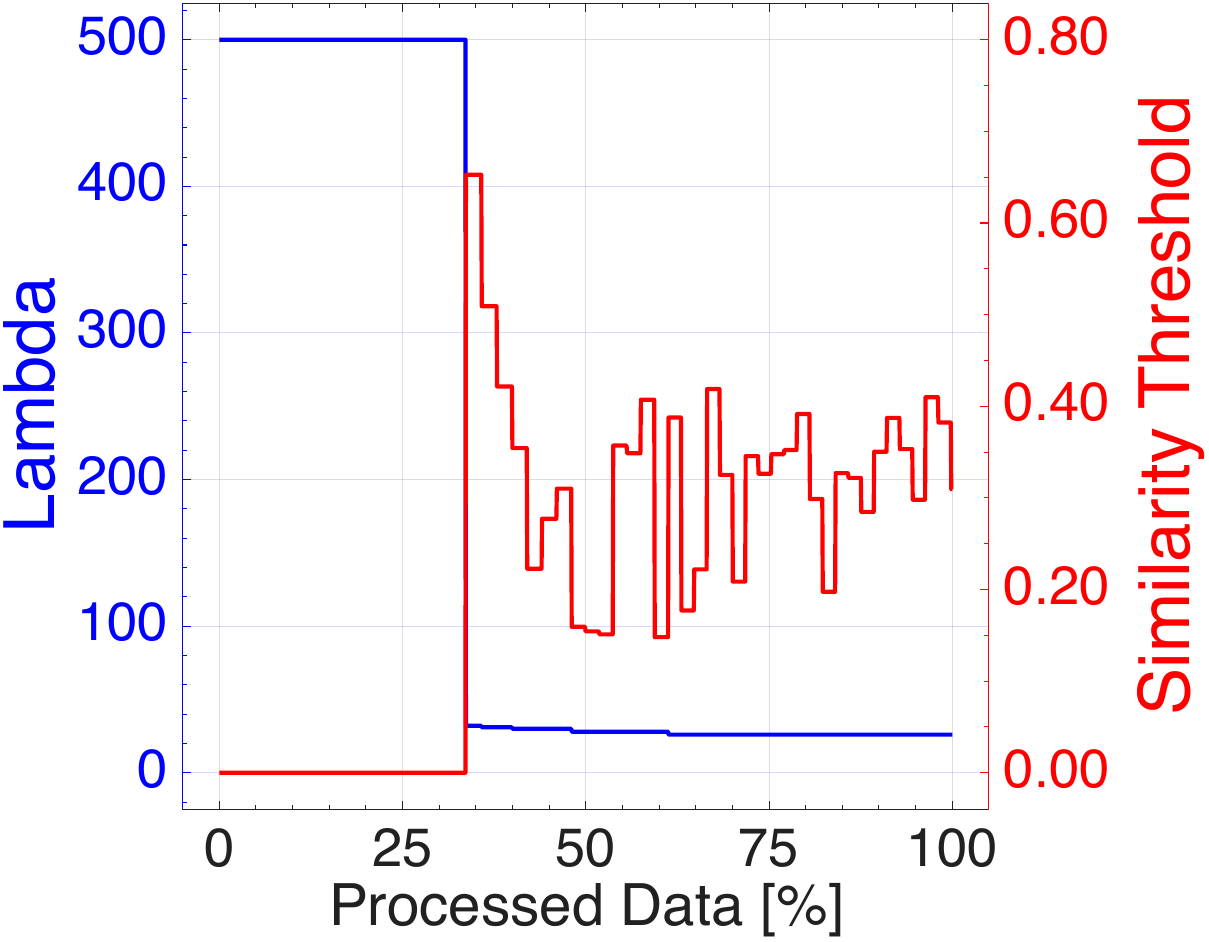}
  }\hfill
  \subfloat[Semeion]{%
    \includegraphics[width=0.22\linewidth]{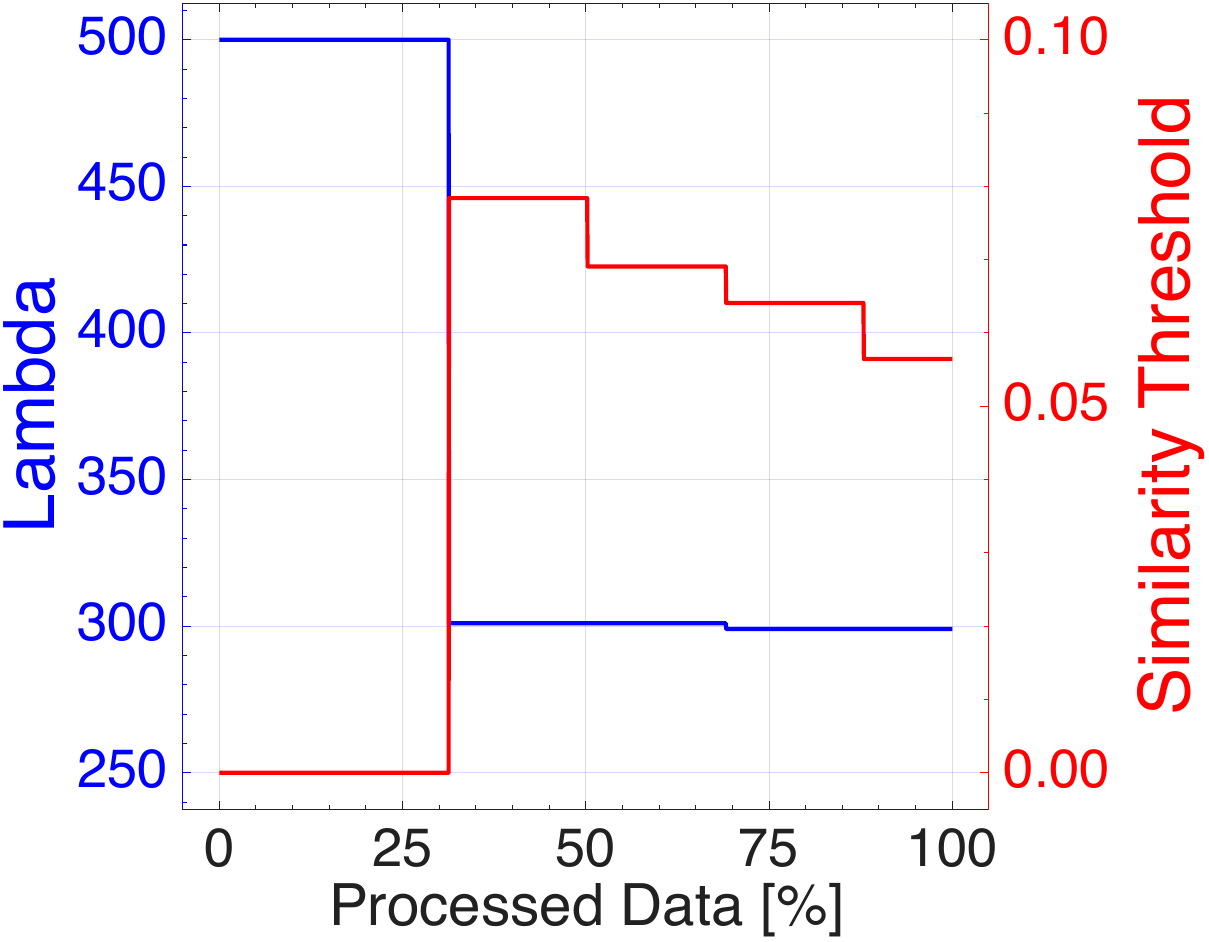}
  }\\
  \subfloat[MSRA25]{%
    \includegraphics[width=0.22\linewidth]{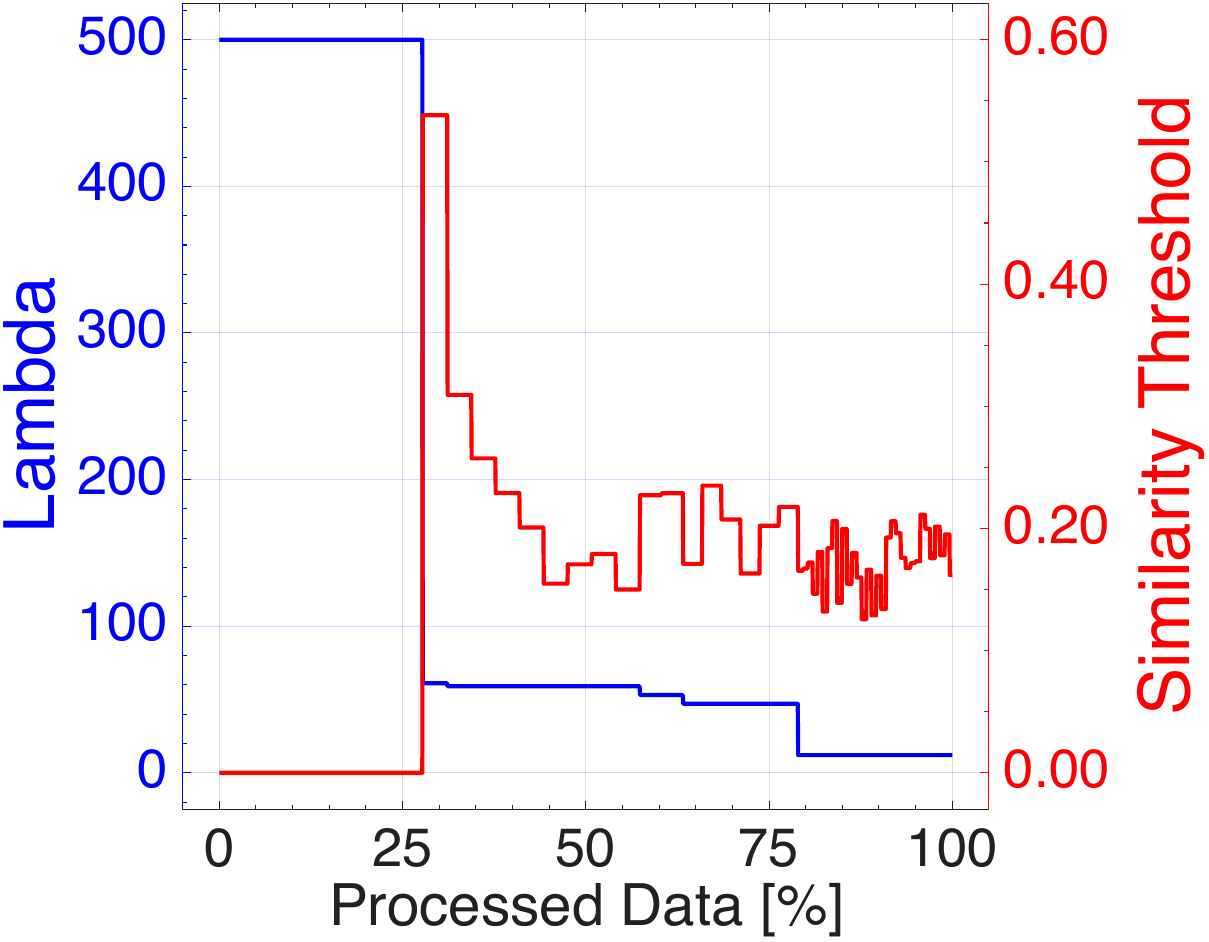}
  }\hfill
  \subfloat[Image Segmentation]{%
    \includegraphics[width=0.22\linewidth]{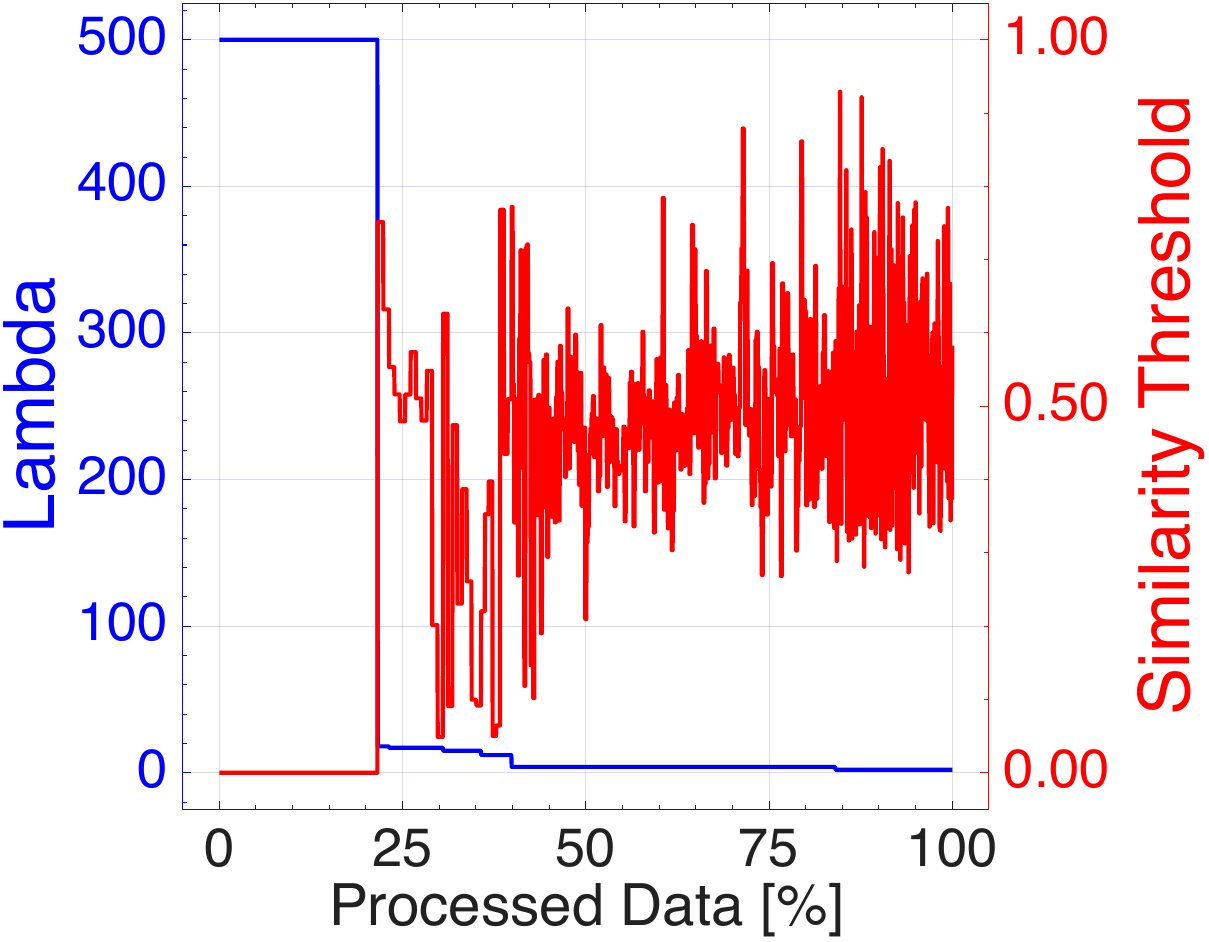}
  }\hfill
  \subfloat[Rice]{%
    \includegraphics[width=0.22\linewidth]{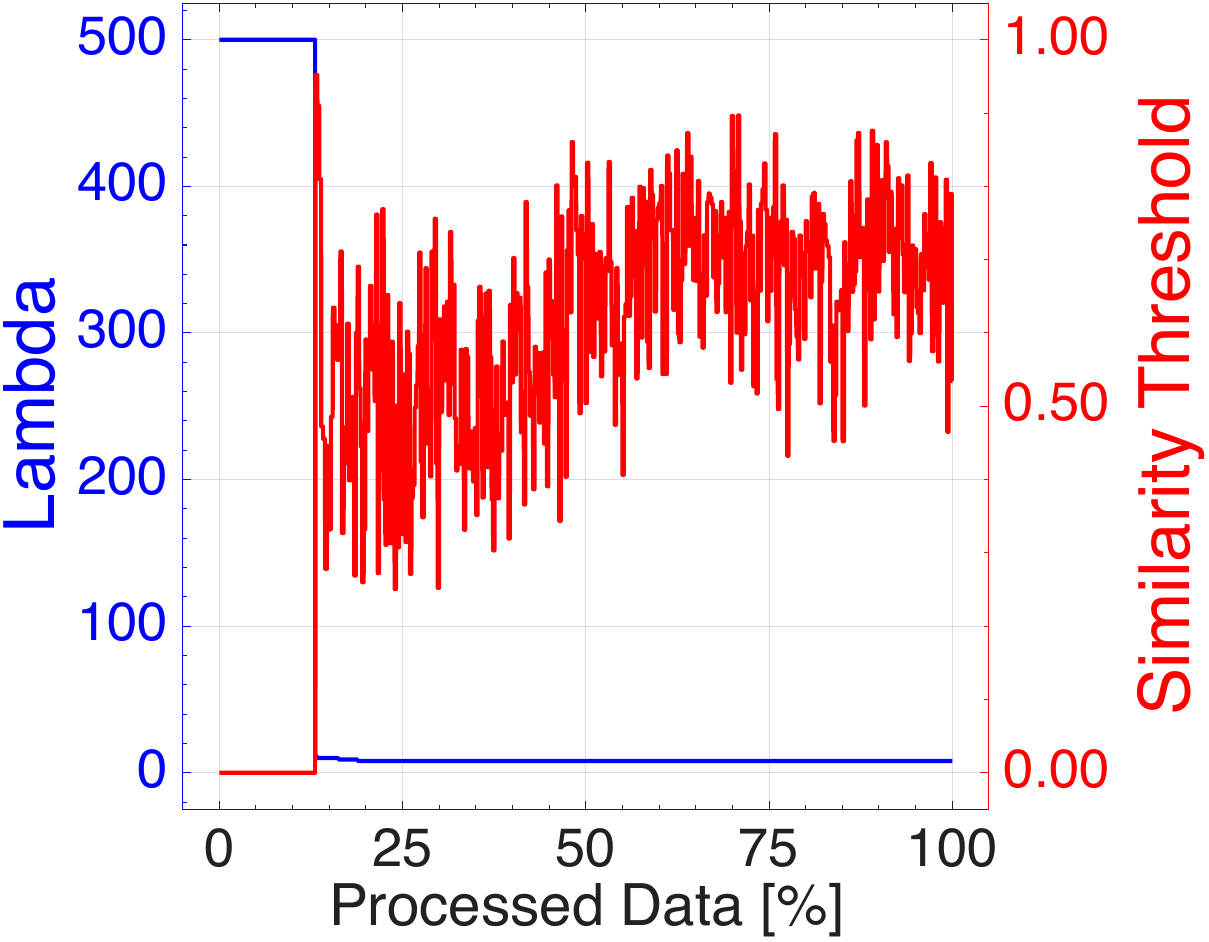}
  }\hfill
  \subfloat[TUANDROMD]{%
    \includegraphics[width=0.22\linewidth]{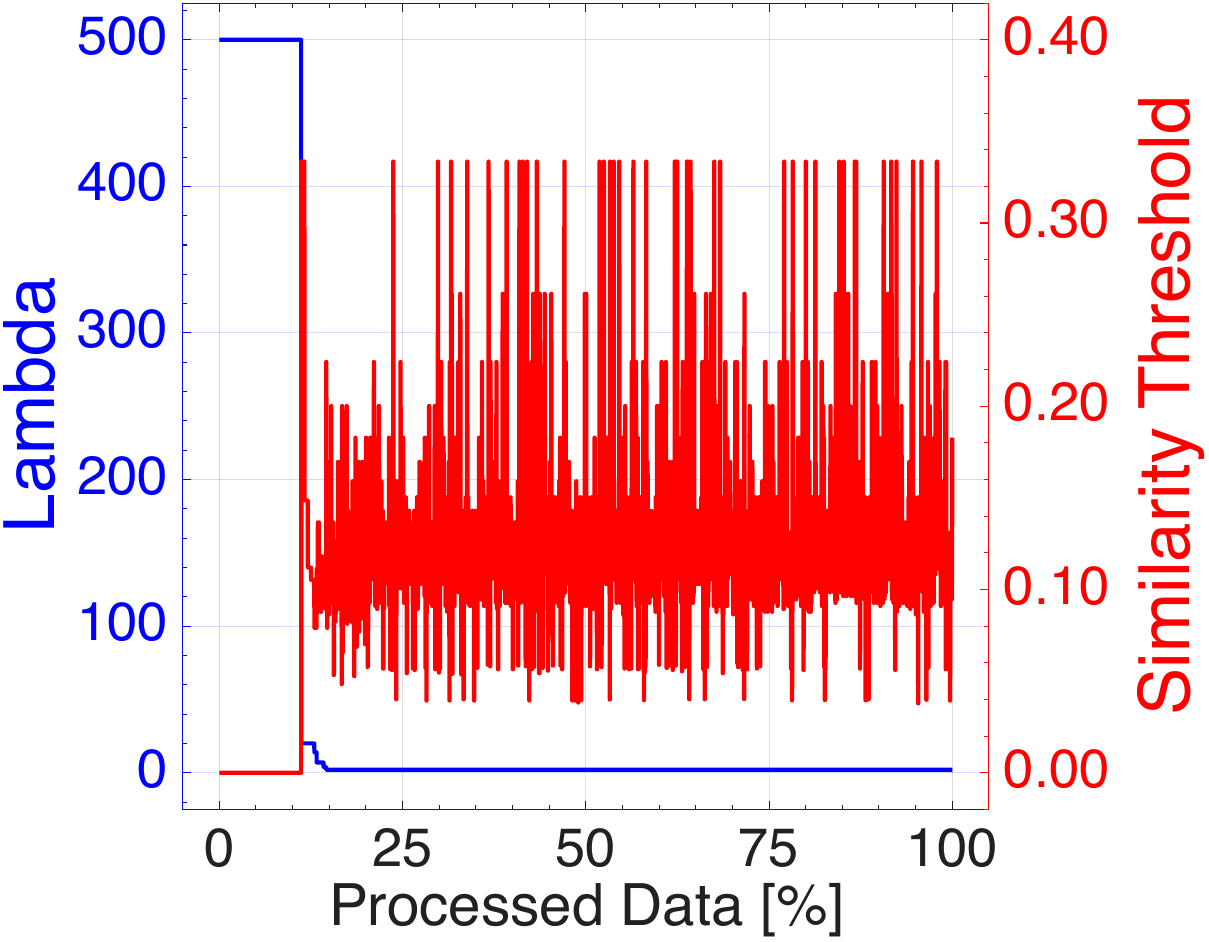}
  }\\
  \subfloat[Phoneme]{%
    \includegraphics[width=0.22\linewidth]{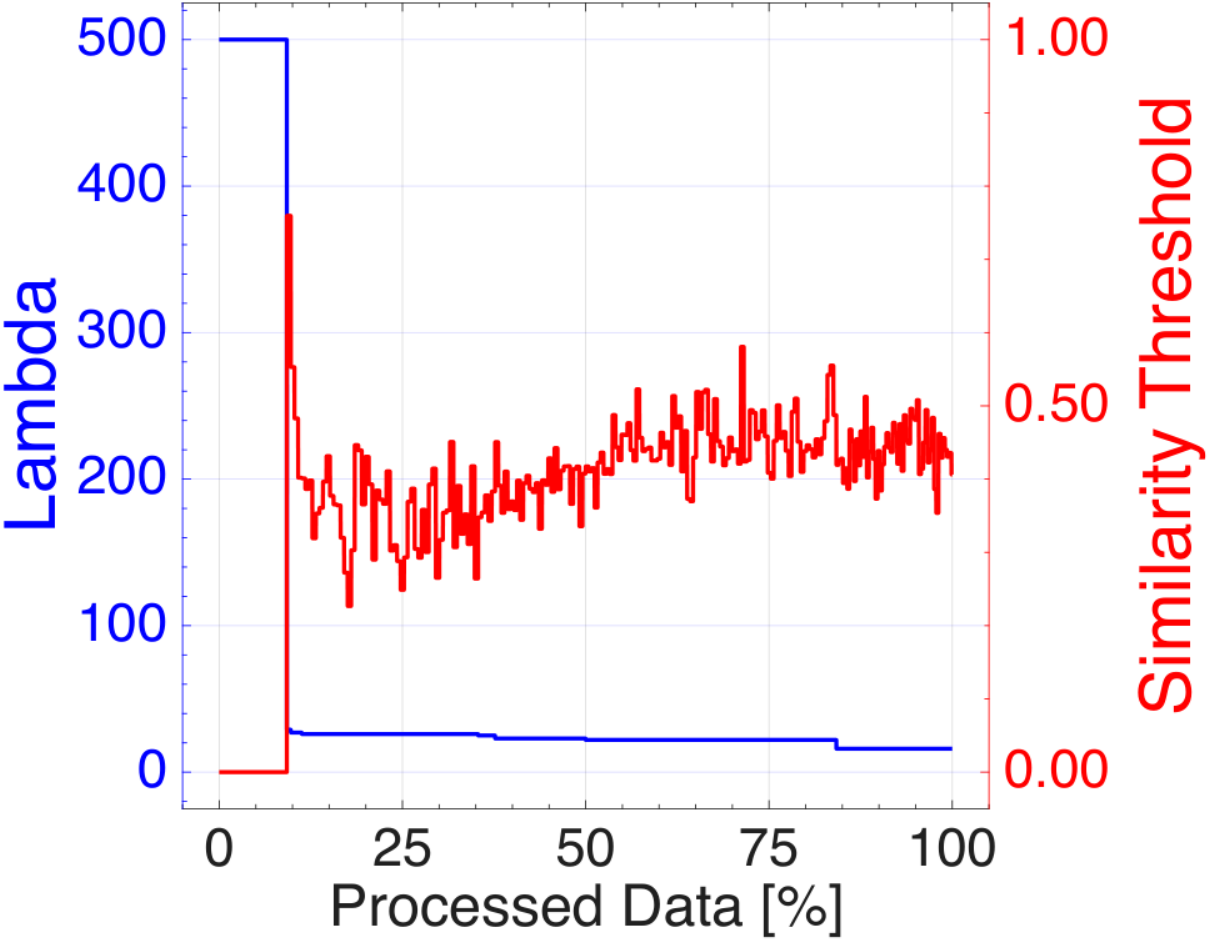}
  }\hfill
  \subfloat[Texture]{%
    \includegraphics[width=0.22\linewidth]{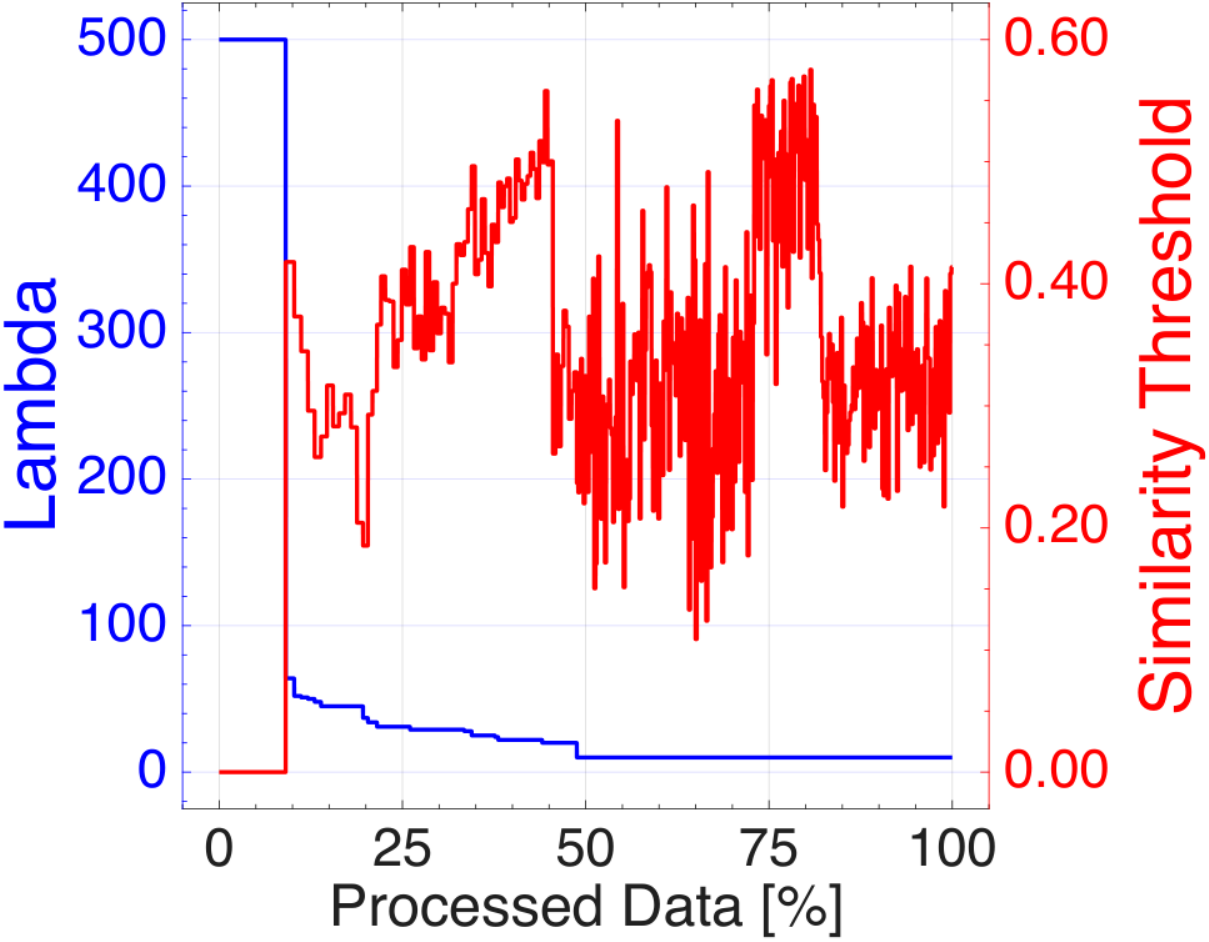}
  }\hfill
  \subfloat[OptDigits]{%
    \includegraphics[width=0.22\linewidth]{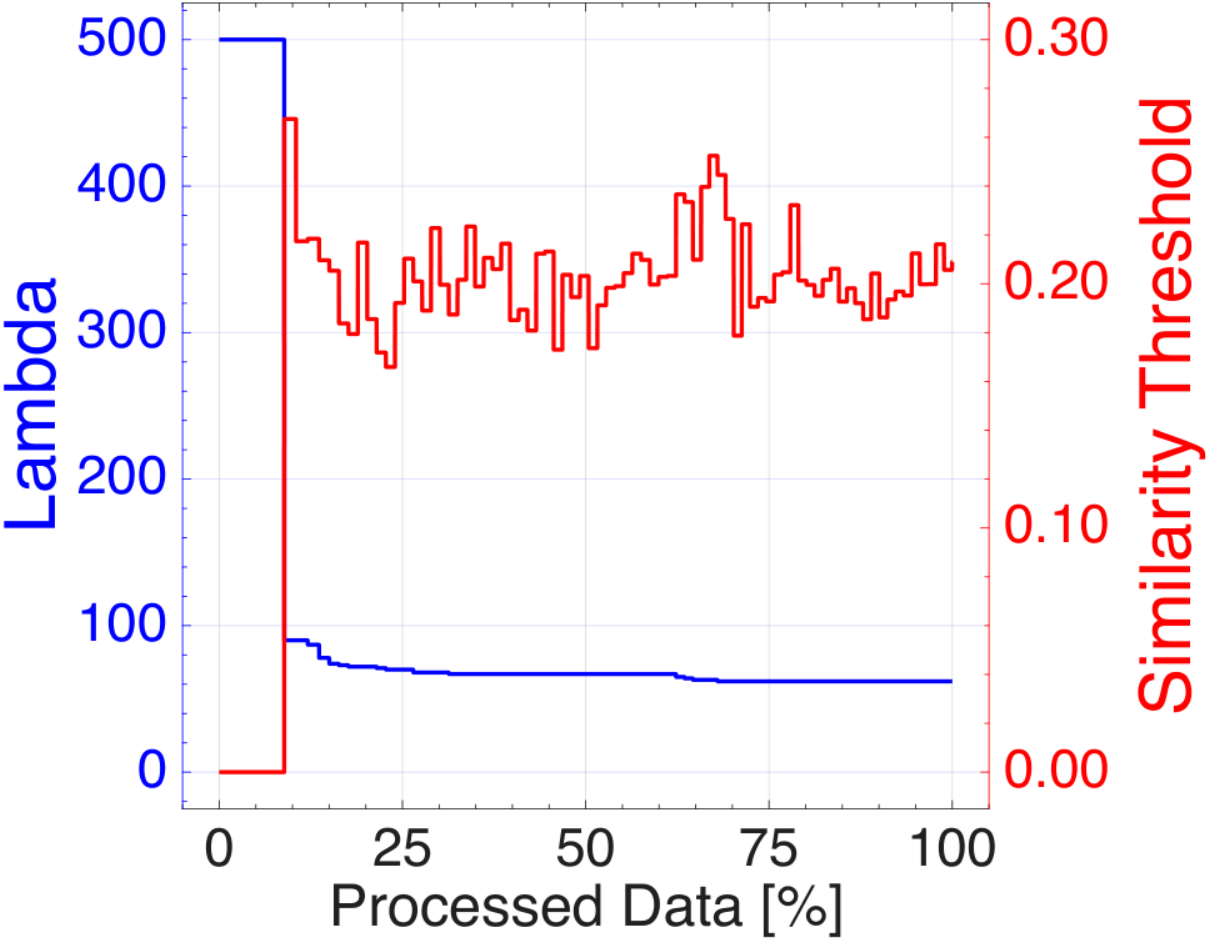}
  }\hfill
  \subfloat[Statlog]{%
    \includegraphics[width=0.22\linewidth]{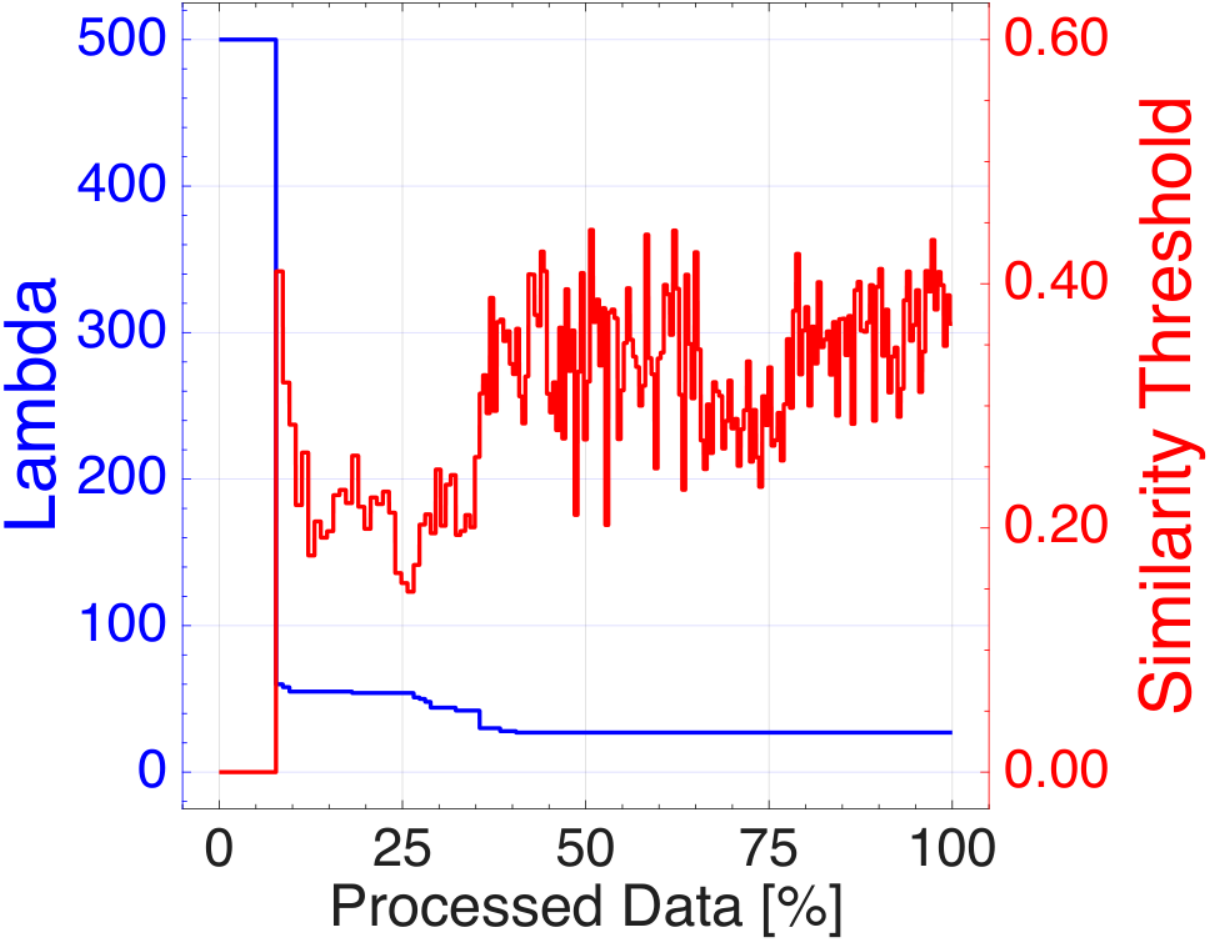}
  }\\
  \subfloat[Anuran Calls]{%
    \includegraphics[width=0.22\linewidth]{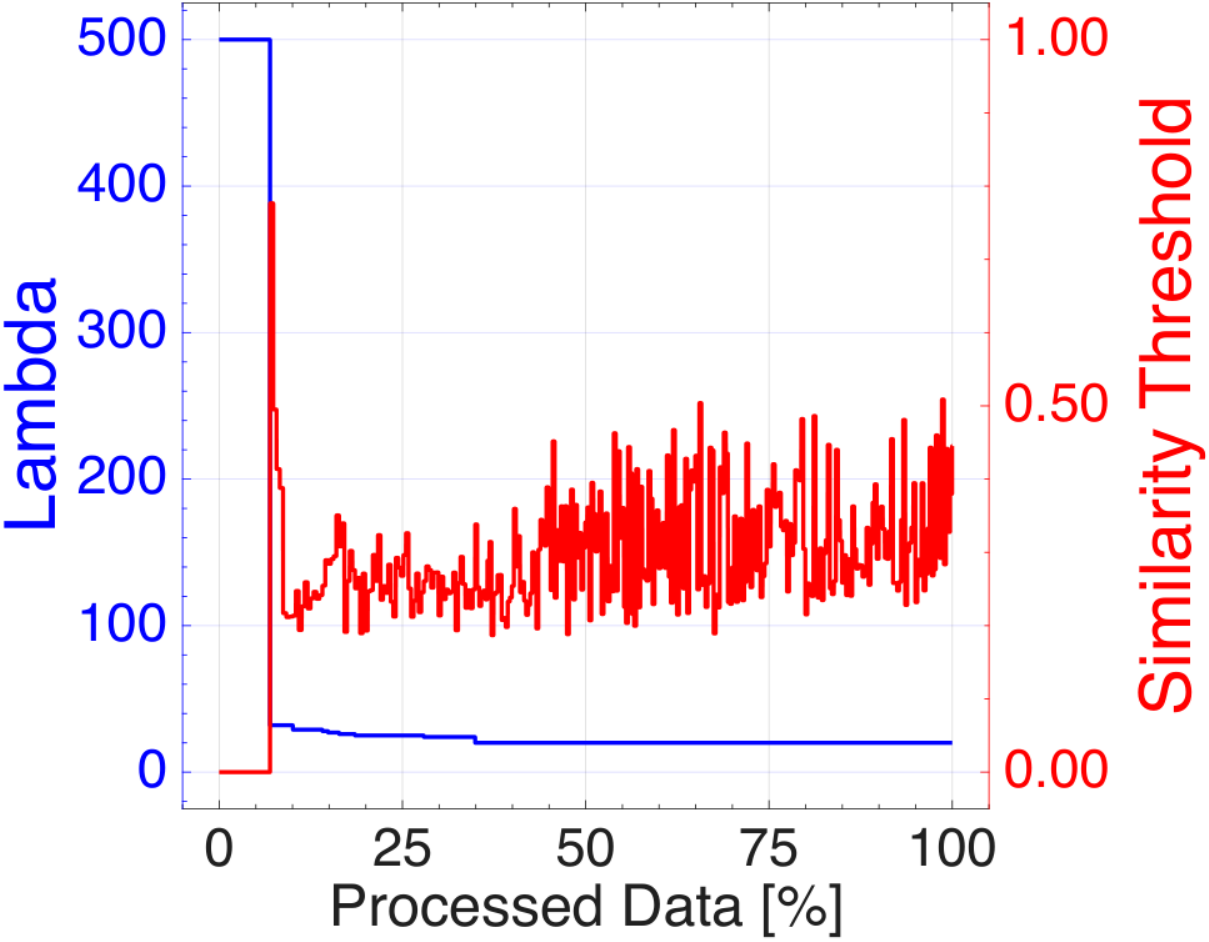}
  }\hfill
  \subfloat[Isolet]{%
    \includegraphics[width=0.22\linewidth]{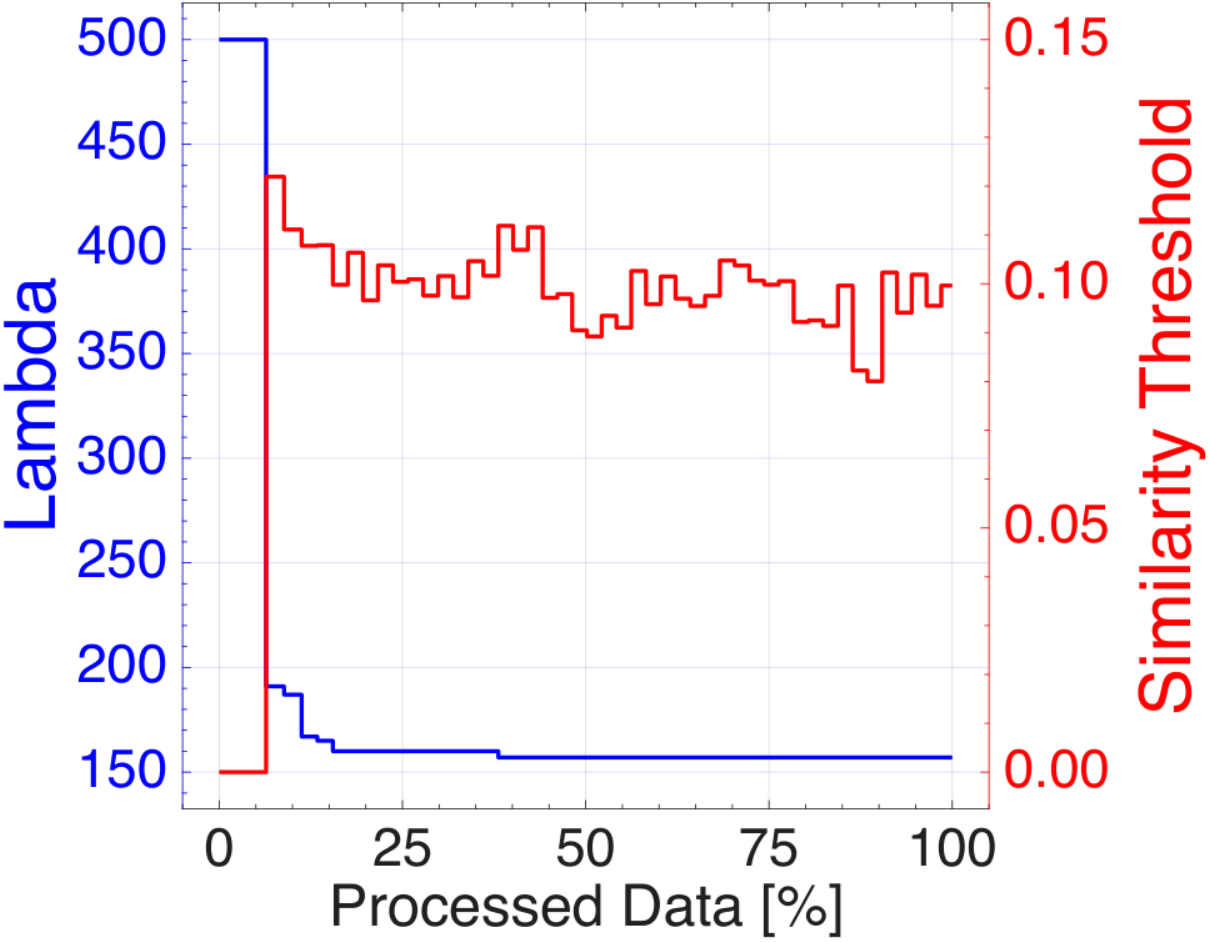}
  }\hfill
  \subfloat[MNIST10K]{%
    \includegraphics[width=0.22\linewidth]{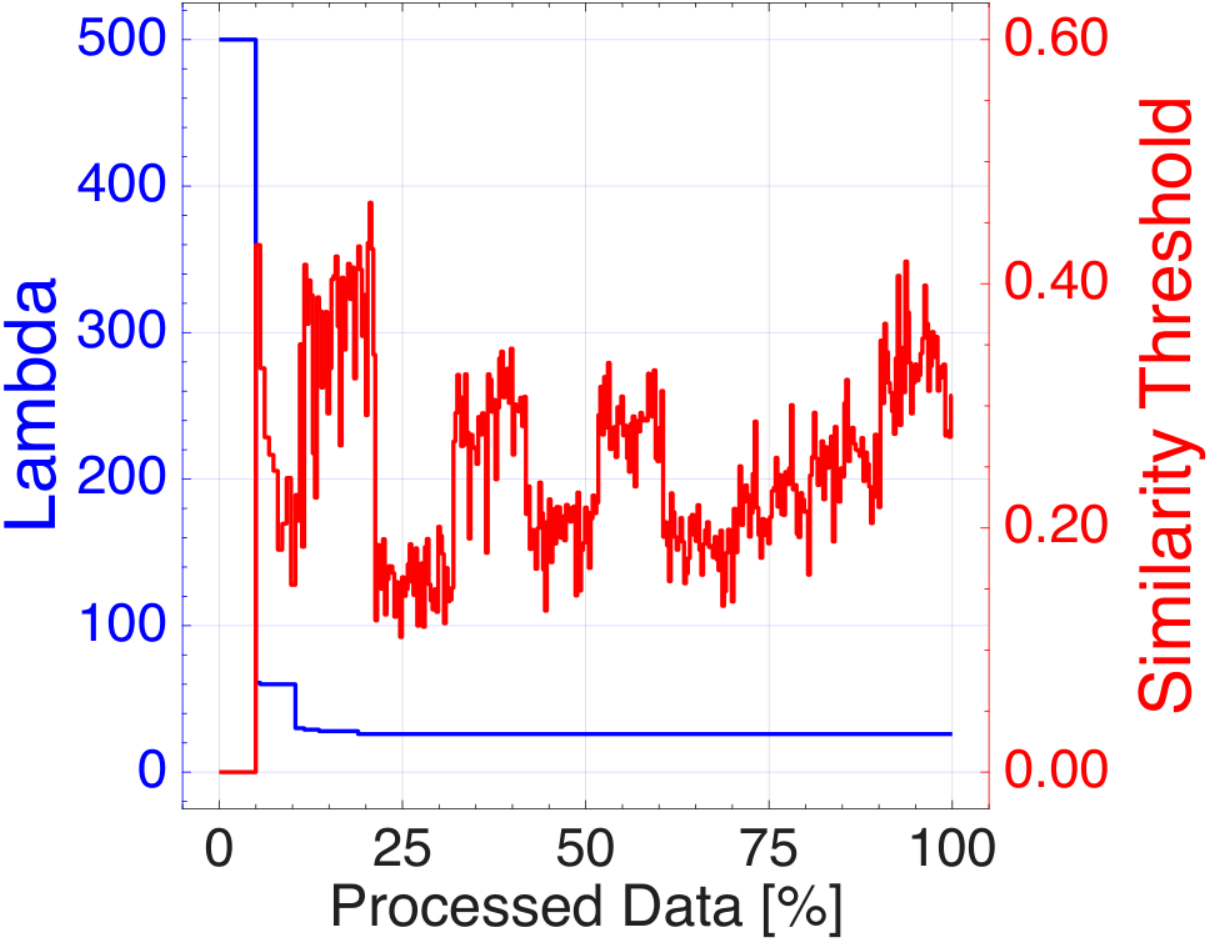}
  }\hfill
  \subfloat[PenBased]{%
    \includegraphics[width=0.22\linewidth]{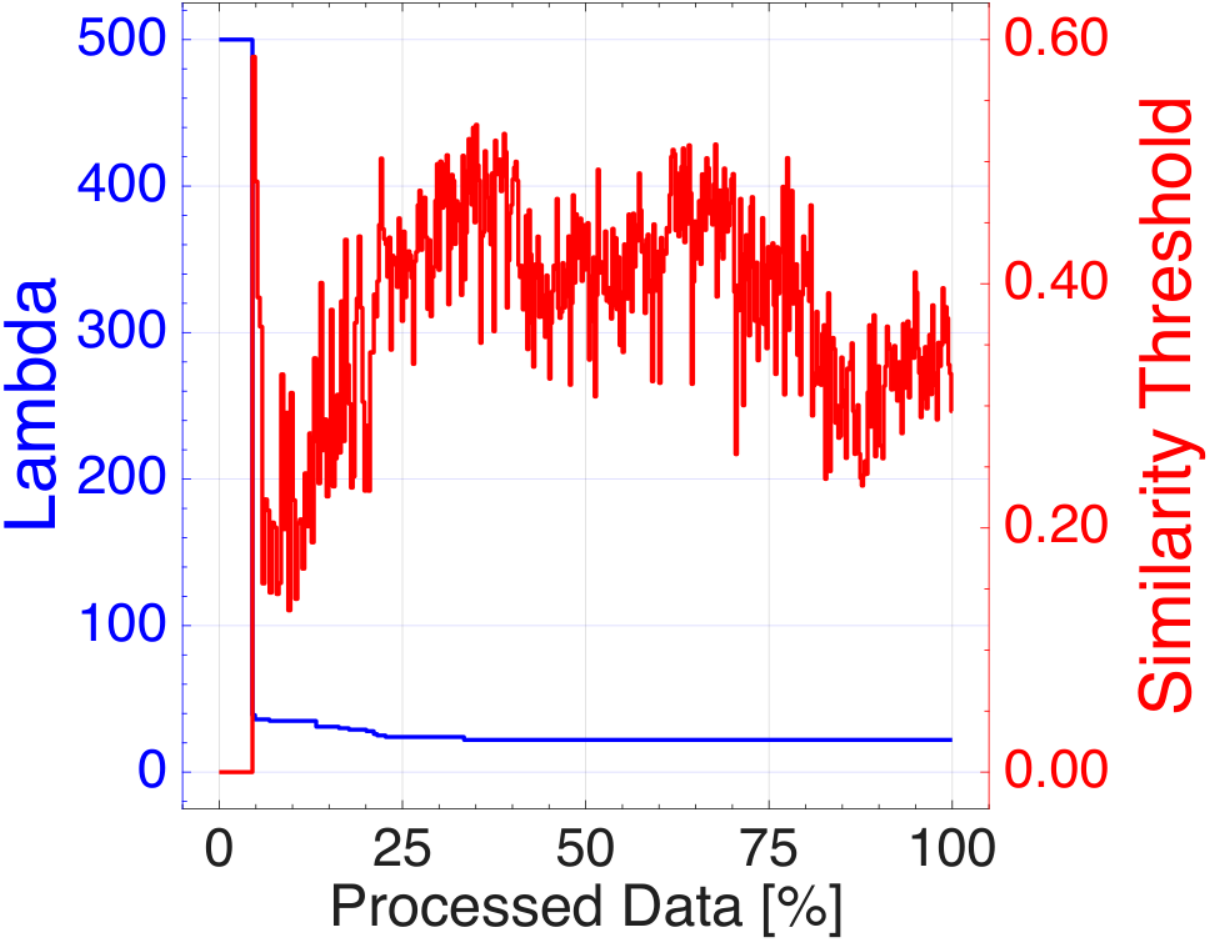}
  }\\
  \subfloat[STL10]{%
    \includegraphics[width=0.22\linewidth]{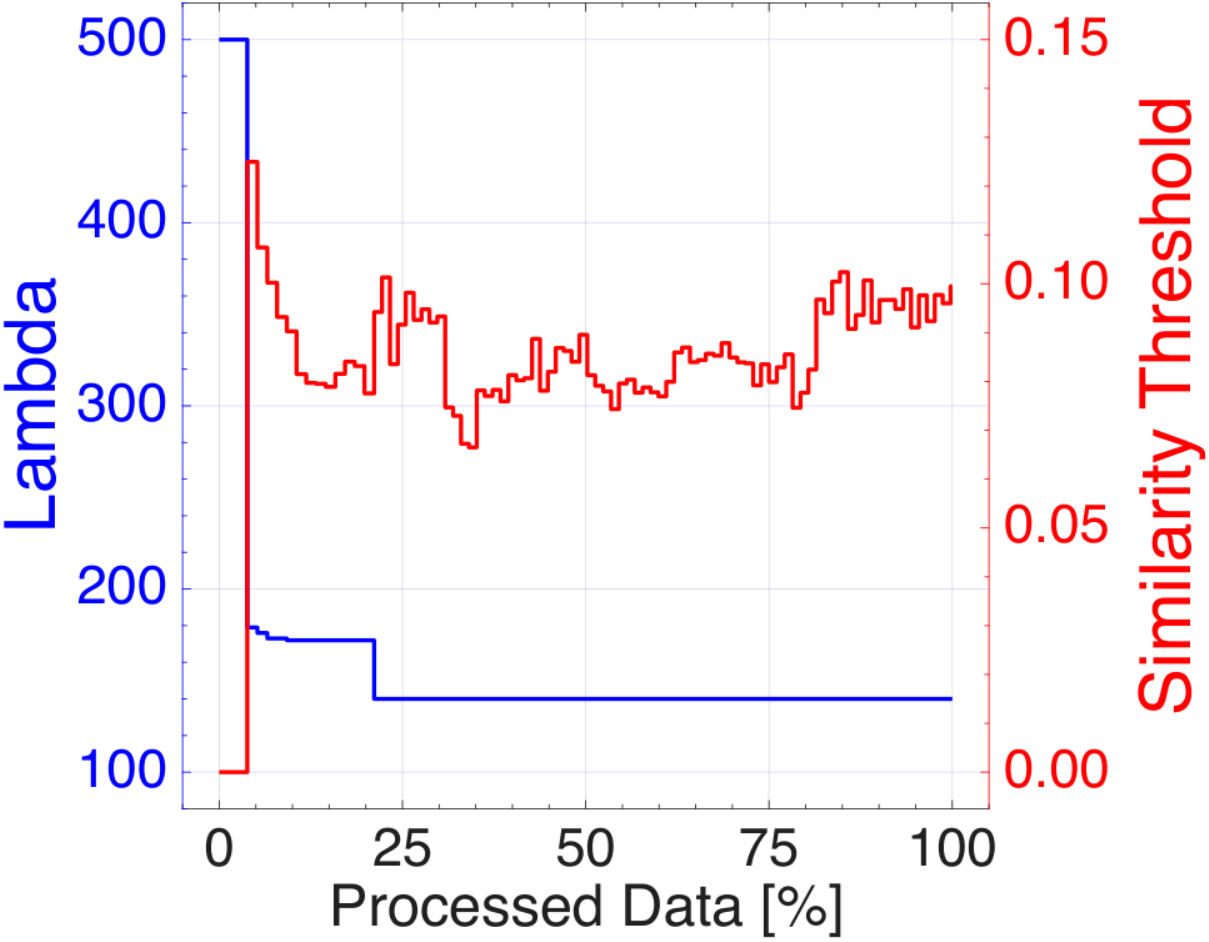}
  }\hfill
  \subfloat[Letter]{%
    \includegraphics[width=0.22\linewidth]{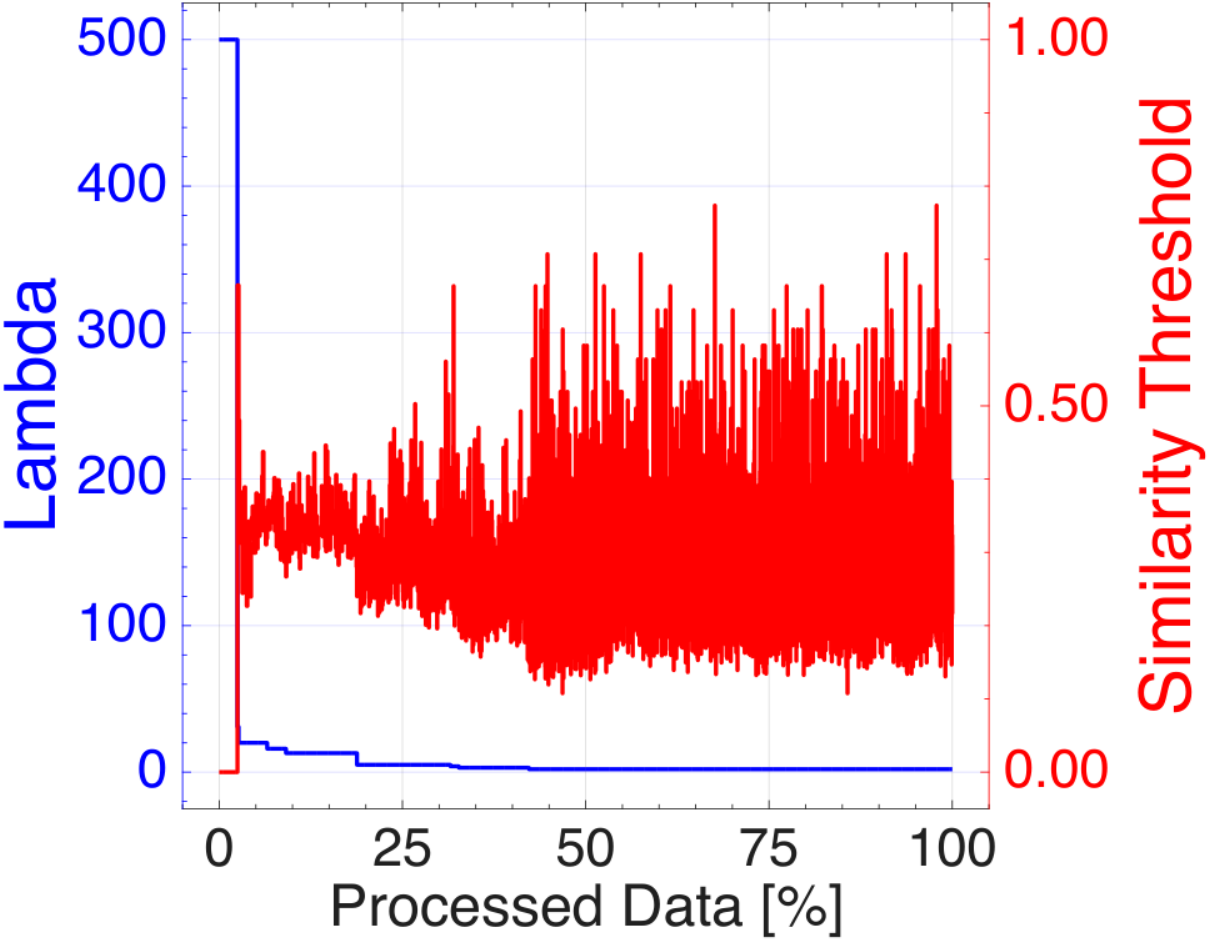}
  }\hfill
  \subfloat[Shuttle]{%
    \includegraphics[width=0.22\linewidth]{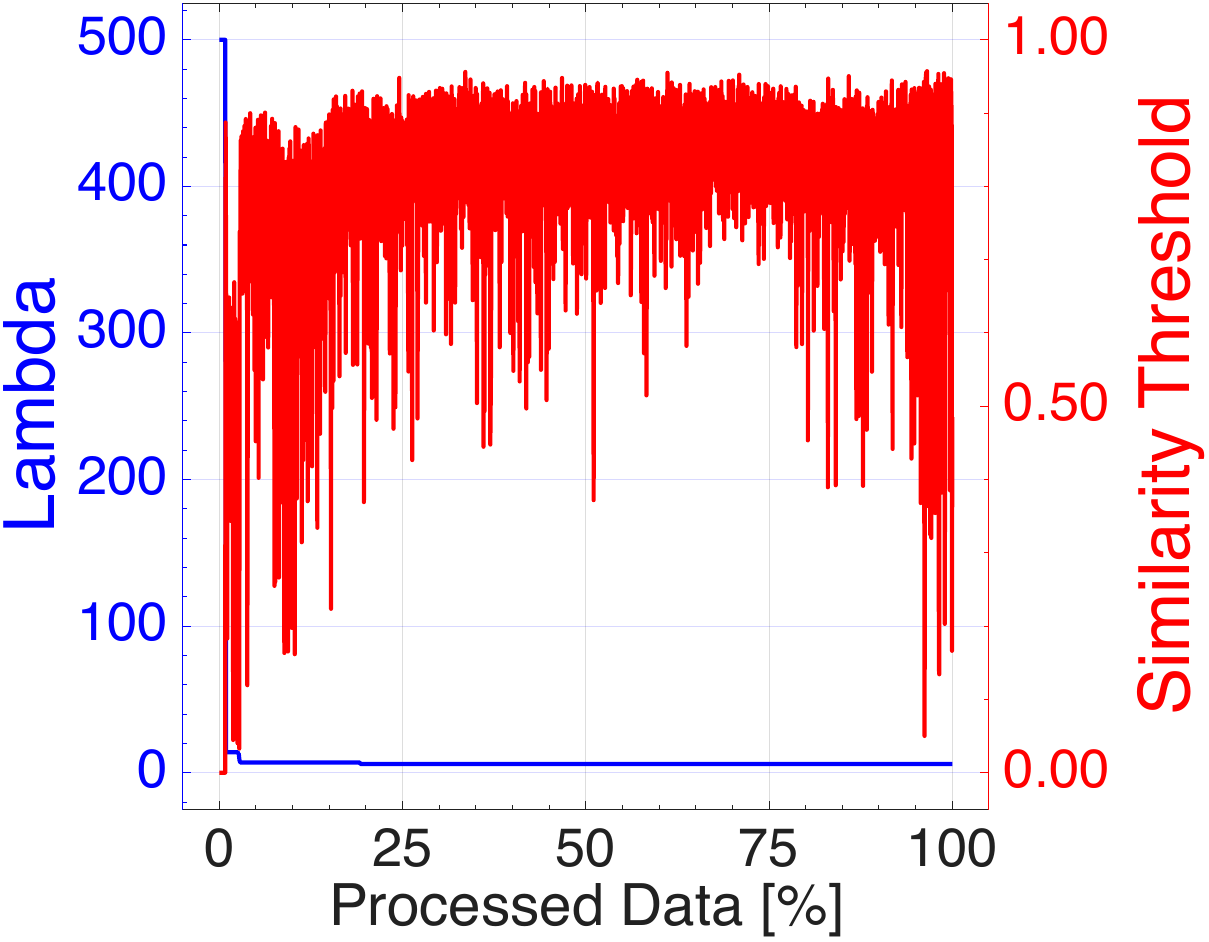}
  }\hfill
  \subfloat[Skin]{%
    \includegraphics[width=0.22\linewidth]{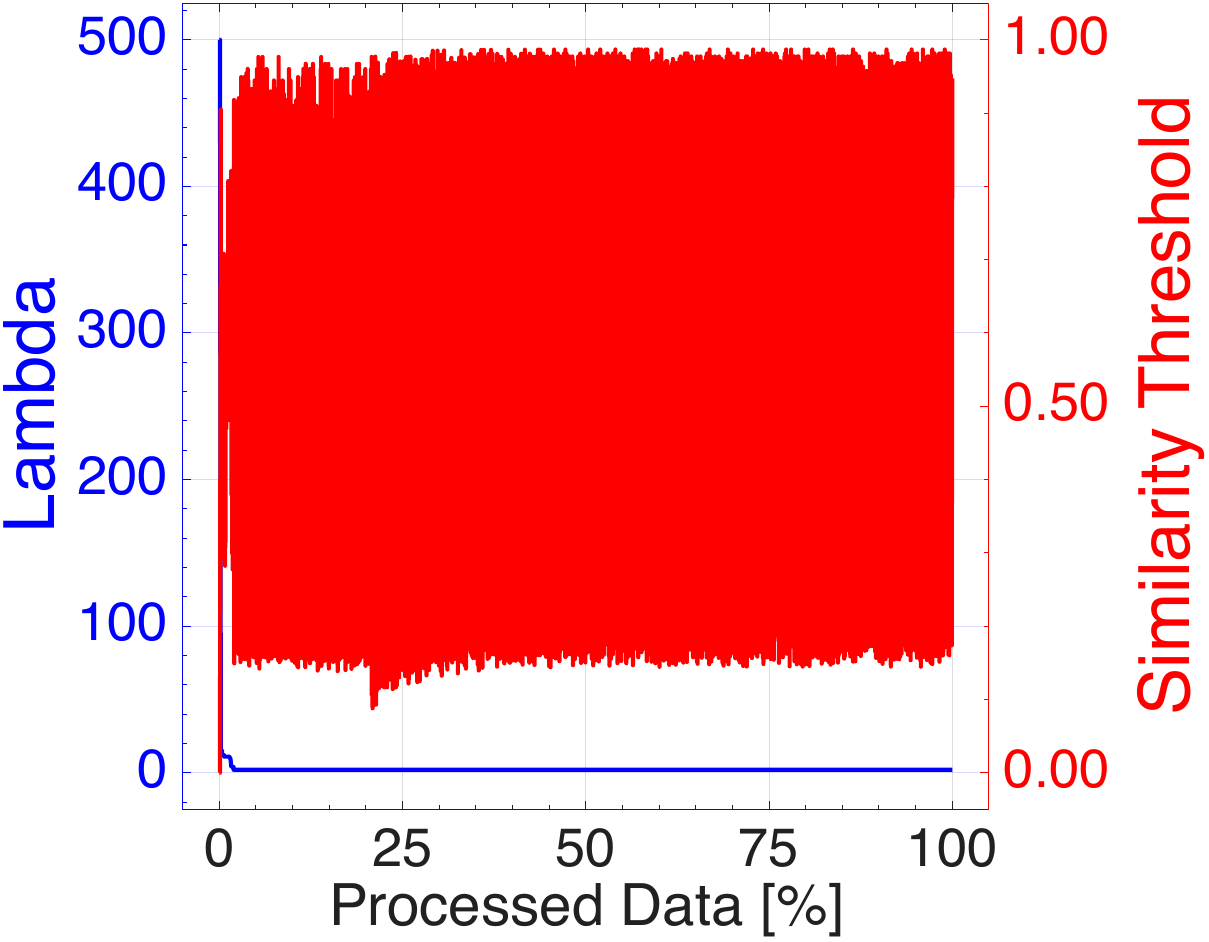}
  }
  \caption{Histories of $\Lambda$ and $V_{\text{threshold}}$ for the w/o Inc. variant in the nonstationary setting ($\Lambda_{\text{init}} = 500$).}
  \label{fig:ablation_lambda_history_noincrease_500_nonstationary}
\end{figure*}

The histories of the recalculation interval $\Lambda$ and the vigilance threshold $V_{\text{threshold}}$ in each ablation variant are also presented. Figs.~\ref{fig:ablation_lambda_history_idat_nonstationary}-\ref{fig:ablation_lambda_history_noincrease_500_nonstationary} show the histories of $\Lambda$ and $V_{\text{threshold}}$ for IDAT, w/o Decremental, and w/o Incremental, respectively. More specifically, Figs.~\ref{fig:ablation_lambda_history_idat_nonstationary}, \ref{fig:ablation_lambda_history_nodecrease_nonstationary}, and \ref{fig:ablation_lambda_history_noincrease_nonstationary} set $\Lambda_{\text{init}} = 2$, while Figs.~\ref{fig:ablation_lambda_history_idat_500_nonstationary}, \ref{fig:ablation_lambda_history_nodecrease_500_nonstationary}, and \ref{fig:ablation_lambda_history_noincrease_500_nonstationary} set $\Lambda_{\text{init}} = 500$.

% Fig. \ref{fig:ablation_lambda_history_idat_nonstationary}

% Fig. \ref{fig:ablation_lambda_history_nodecrease_nonstationary}

% Fig. \ref{fig:ablation_lambda_history_noincrease_nonstationary}

\clearpage

% ================================================================================================================ SR-PCM-HDP
\begin{table*}[htbp]
\centering
\renewcommand{\arraystretch}{1.2}
\caption{List of hyperparameters for SR-PCM-HDP}
\label{tab:srpcmdp-param-def}
\footnotesize
\begin{tabular}{llll}
\hline\hline
Parameter & Type & Search Range & Description \\
\hline
$\beta$  & real & [0.1, 5.0] & Fuzzifier that controls membership decay with distance \\
$\theta$ & real & [0.0, 0.9] & Coefficient that controls graph regularization strength \\
\hline\hline
\end{tabular}
\vspace{2mm}
\end{table*}

% ================================================================================================================ TBM-P
\begin{table*}[htbp]
\centering
\renewcommand{\arraystretch}{1.2}
\caption{List of hyperparameters for TBM-P}
\label{tab:tbmp-param-def}
\footnotesize
\begin{tabular}{llll}
\hline\hline
Parameter & Type & Search Range & Description \\
\hline
$\alpha_A$ & real & [0.0, 0.95] & Learning rate that controls prototype update (A) \\
$\rho_A$   & real & [0.0, 1.0]  & Vigilance parameter that controls match acceptance (A) \\
$\beta_A$  & real & [0.0, 5.0]  & Fuzzifier that controls category sharpness (A) \\
$\theta_A$ & real & [0.1, 0.3]  & Threshold that prunes small clusters (A) \\
$\alpha_B$ & real & [0.0, 0.95] & Learning rate that controls prototype update (B) \\
$\rho_B$   & real & [0.0, 1.0]  & Vigilance parameter that controls match acceptance (B) \\
$\beta_B$  & real & [0.0, 5.0]  & Fuzzifier that controls category sharpness (B) \\
$\theta_B$ & real & [0.1, 0.3]  & Threshold that prunes small clusters (B) \\
$\varepsilon$ & real & [0.0, 0.99] & Coupling coefficient that controls inter\mbox{-}module association \\
\hline\hline
\end{tabular}
\vspace{2mm}
\end{table*}

% ================================================================================================================ GNG
\begin{table*}[htbp]
\centering
\renewcommand{\arraystretch}{1.2}
\caption{List of hyperparameters for GNG}
\label{tab:gng-param-def}
\footnotesize
\begin{tabular}{llll}
\hline\hline
Parameter & Type & Search Range & Description \\
\hline
$\epsilon_b$ & real    & [0.01, 0.99] & Learning rate that controls winner node update \\
$\epsilon_n$ & real    & [0.01, 0.99] & Learning rate that controls neighbor node update \\
$a_{\max}$   & integer & [1, 50]      & Threshold that controls maximum edge age before removal \\
$\lambda$    & integer & [1, 50]      & Interval that controls frequency of node insertion \\
$\alpha$     & real    & [0.01, 0.5]  & Factor that controls error reduction during node splitting \\
$\delta$     & real    & [0.90, 0.99] & Factor that controls global error decay \\
\hline\hline
\end{tabular}
\vspace{2mm}
\end{table*}

% ================================================================================================================ DDVFA
\begin{table*}[htbp]
\centering
\renewcommand{\arraystretch}{1.2}
\caption{List of hyperparameters for DDVFA}
\label{tab:ddvfa-param-def}
\footnotesize
\begin{tabular}{llll}
\hline\hline
Parameter & Type & Search Range & Description \\
\hline
$\rho_1$        & real & [0.50, 0.99]                    & Vigilance parameter that controls global match acceptance \\
$\rho_2$        & real & [0.50, 0.99]                    & Vigilance parameter that controls local match acceptance \\
$\alpha$        & real & [$10^{-4}$, $10^{-1}$]          & Choice parameter that controls category preference in activation \\
$\beta$         & real & [0.5, 1.0]                      & Learning rate that controls prototype update speed \\
$\gamma$        & real & [0.5, 5.0]                      & Similarity exponent that controls activation sharpness \\
$\gamma_{\mathrm{ref}}$ & real & [0.5, 5.0]              & Reference exponent that controls normalization in threshold \\
\hline\hline
\end{tabular}
\vspace{2mm}
\end{table*}

% ================================================================================================================ CAEA
\begin{table*}[htbp]
\centering
\renewcommand{\arraystretch}{1.2}
\caption{List of hyperparameters for CAEA}
\label{tab:caea-param-def}
\footnotesize
\begin{tabular}{llll}
\hline\hline
Parameter & Type & Search Range & Description \\
\hline
$\lambda$        & integer & [2, 500] & Interval that controls deletion cycle of isolated nodes \\
$a_{\mathrm{max}}$ & integer & [2, 500] & Threshold that controls maximum edge age before deletion \\
\hline\hline
\end{tabular}
\vspace{2mm}
\end{table*}

% ================================================================================================================ SR-PCM-HDP
\begin{table*}[htbp]
\centering
\renewcommand{\arraystretch}{1.2}
\caption{Optimized hyperparameters of SR-PCM-HDP under the stationary setting}
\label{tab:srpcmdp-stationary}
\footnotesize
\begin{tabular}{lcc}
\hline\hline
\multirow{2}{*}{Dataset} & \multicolumn{2}{c}{Hyperparameter} \\
\cline{2-3}
 & $\beta$ & $\theta$ \\
\hline
Iris               &    1.375 &    0.900 \\
Seeds              &    1.079 &    0.416 \\
Dermatology        &    2.078 &    0.889 \\
Pima               &    4.608 &    0.765 \\
Mice Protein       &    4.985 &    0.899 \\
Binalpha           &    4.048 &    0.323 \\
Yeast              &    4.997 &    0.001 \\
Semeion            &    4.048 &    0.323 \\
MSRA25             &    4.789 &    0.727 \\
Image Segmentation         &    2.904 &    0.557 \\
Rice               &    0.904 &    0.090 \\
TUANDROMD          &    4.048 &    0.323 \\
Phoneme            &    2.071 &    0.882 \\
Texture            &    4.997 &    0.896 \\
OptDigits          &    4.048 &    0.323 \\
Statlog            &    4.032 &    0.899 \\
Anuran Calls       &    2.194 &    0.725 \\
Isolet             &    4.048 &    0.323 \\
MNIST10K           &    4.999 &    0.368 \\
PenBased           &    4.997 &    0.896 \\
STL10              &    4.048 &    0.323 \\
Letter             &    3.155 &    0.204 \\
Shuttle            &    1.342 &    0.769 \\
Skin               &    2.204 &    0.836 \\
\hline\hline
\end{tabular}

\begin{minipage}{\linewidth}
  \vspace{1mm}
  \setlength{\leftskip}{64mm}
    Anuran Calls dataset uses only the {\it Family} label. \\
\end{minipage}
\end{table*}

% ================================================================================================================ TBM-P
\begin{table*}[htbp]
\centering
\renewcommand{\arraystretch}{1.2}
\caption{Optimized hyperparameters of TBM-P under the stationary setting}
\label{tab:tbmp-stationary}
\footnotesize
\begin{tabular}{lccccccccc}
\hline\hline
\multirow{2}{*}{Dataset} & \multicolumn{9}{c}{Hyperparameter} \\
\cline{2-10}
 & $\alpha_A$ & $\rho_A$ & $\beta_A$ & $\theta_A$ & $\alpha_B$ & $\rho_B$ & $\beta_B$ & $\theta_B$ & $\varepsilon$ \\
\hline
Iris               &    0.070 &    0.040 &    4.367 &    0.148 &    0.212 &    0.059 &    1.026 &    0.225 &    0.631 \\
Seeds              &    0.729 &    0.863 &    4.615 &    0.101 &    0.949 &    0.350 &    3.476 &    0.145 &    0.380 \\
Dermatology        &    0.743 &    0.034 &    4.366 &    0.141 &    0.635 &    0.868 &    3.770 &    0.296 &    0.920 \\
Pima               &    0.937 &    0.885 &    4.198 &    0.124 &    0.078 &    0.025 &    3.983 &    0.105 &    0.799 \\
Mice Protein       &    0.932 &    0.681 &    3.869 &    0.210 &    0.888 &    0.137 &    3.146 &    0.173 &    0.862 \\
Binalpha           &    0.688 &    0.436 &    4.779 &    0.263 &    0.760 &    0.127 &    3.014 &    0.200 &    0.239 \\
Yeast              &    0.936 &    0.799 &    4.843 &    0.236 &    0.190 &    0.624 &    4.320 &    0.123 &    0.145 \\
Semeion            &    0.758 &    0.082 &    3.722 &    0.282 &    0.115 &    0.436 &    2.915 &    0.262 &    0.100 \\
MSRA25             &    0.891 &    0.372 &    4.373 &    0.115 &    0.316 &    0.176 &    0.141 &    0.117 &    0.626 \\
Image Segmentation         &    0.915 &    0.548 &    4.718 &    0.161 &    0.862 &    0.614 &    1.282 &    0.136 &    0.072 \\
Rice               &    0.950 &    0.638 &    3.547 &    0.142 &    0.864 &    0.968 &    2.375 &    0.184 &    0.062 \\
TUANDROMD          &    0.941 &    0.580 &    3.522 &    0.299 &    0.341 &    0.039 &    2.651 &    0.220 &    0.738 \\
Phoneme            &    0.932 &    0.681 &    3.869 &    0.210 &    0.888 &    0.137 &    3.146 &    0.173 &    0.862 \\
Texture            &    0.949 &    0.920 &    4.858 &    0.122 &    0.640 &    0.423 &    2.791 &    0.290 &    0.863 \\
OptDigits          &    0.846 &    0.359 &    3.742 &    0.133 &    0.175 &    0.187 &    2.827 &    0.267 &    0.129 \\
Statlog            &    0.942 &    0.065 &    4.987 &    0.241 &    0.271 &    0.263 &    0.006 &    0.263 &    0.423 \\
Anuran Calls       &    0.841 &    0.099 &    4.870 &    0.140 &    0.544 &    0.022 &    2.690 &    0.148 &    0.976 \\
Isolet             &    0.938 &    0.103 &    4.932 &    0.193 &    0.592 &    0.641 &    0.066 &    0.153 &    0.706 \\
MNIST10K           &    0.885 &    0.253 &    4.390 &    0.238 &    0.056 &    0.868 &    0.172 &    0.162 &    0.726 \\
PenBased           &    0.871 &    0.019 &    4.632 &    0.195 &    0.654 &    0.335 &    1.664 &    0.278 &    0.742 \\
STL10              &    0.948 &    0.069 &    2.988 &    0.197 &    0.436 &    0.891 &    2.126 &    0.148 &    0.979 \\
Letter             &    0.946 &    0.753 &    3.679 &    0.211 &    0.564 &    0.719 &    2.589 &    0.257 &    0.949 \\
Shuttle            &    0.945 &    0.790 &    2.152 &    0.107 &    0.069 &    0.182 &    1.327 &    0.103 &    0.974 \\
Skin               &       -- &       -- &       -- &       -- &       -- &       -- &       -- &       -- &       -- \\
\hline\hline
\end{tabular}

\footnotesize
\begin{minipage}{\linewidth}
  \vspace{1mm}
  \setlength{\leftskip}{29mm}
    Anuran Calls dataset uses only the {\it Family} label. \\
    The optimization for the Skin dataset failed due to excessive computational requirements. \\
\end{minipage}
\end{table*}

% ================================================================================================================ GNG
% ===================== GNG (stationary) =====================
\begin{table*}[htbp]
\centering
\renewcommand{\arraystretch}{1.2}
\caption{Optimized hyperparameters of GNG under the stationary setting}
\label{tab:gng-stationary}
\footnotesize
\begin{tabular}{lcccccc}
\hline\hline
\multirow{2}{*}{Dataset} & \multicolumn{6}{c}{Hyperparameter} \\
\cline{2-7}
 & $\epsilon_b$ & $\epsilon_n$ & $a_{\max}$ & $\lambda$ & $\alpha$ & $\delta$ \\
\hline
Iris               &    0.928 &    0.098 &       22 &       46 &    0.413 &    0.979 \\
Seeds              &    0.128 &    0.028 &       10 &        7 &    0.046 &    0.904 \\
Dermatology        &    0.816 &    0.482 &        2 &       13 &    0.111 &    0.900 \\
Pima               &    0.296 &    0.338 &        3 &       43 &    0.207 &    0.923 \\
Mice Protein       &    0.108 &    0.075 &        4 &       29 &    0.019 &    0.988 \\
Binalpha           &    0.091 &    0.071 &        1 &       14 &    0.328 &    0.905 \\
Yeast              &    0.037 &    0.056 &        1 &       46 &    0.137 &    0.989 \\
Semeion            &    0.128 &    0.028 &       10 &        7 &    0.046 &    0.904 \\
MSRA25             &    0.591 &    0.146 &        5 &       45 &    0.356 &    0.966 \\
Image Segmentation         &    0.037 &    0.058 &        1 &       47 &    0.141 &    0.927 \\
Rice               &    0.469 &    0.592 &        1 &       45 &    0.164 &    0.904 \\
TUANDROMD          &    0.039 &    0.104 &       21 &       37 &    0.475 &    0.935 \\
Phoneme            &    0.105 &    0.080 &        1 &       40 &    0.413 &    0.977 \\
Texture            &    0.182 &    0.037 &        2 &       22 &    0.412 &    0.902 \\
OptDigits          &    0.457 &    0.053 &        2 &        5 &    0.414 &    0.982 \\
Statlog            &    0.508 &    0.169 &        1 &       21 &    0.031 &    0.935 \\
Anuran Calls       &    0.234 &    0.032 &        2 &       49 &    0.258 &    0.912 \\
Isolet             &    0.038 &    0.058 &        1 &       44 &    0.133 &    0.946 \\
MNIST10K           &    0.921 &    0.051 &        2 &        3 &    0.386 &    0.927 \\
PenBased           &    0.032 &    0.062 &        1 &       50 &    0.132 &    0.970 \\
STL10              &    0.304 &    0.990 &        1 &        7 &    0.033 &    0.925 \\
Letter             &    0.072 &    0.113 &        1 &       13 &    0.151 &    0.986 \\
Shuttle            &    0.589 &    0.877 &        2 &       42 &    0.397 &    0.933 \\
Skin               &    0.158 &    0.109 &        2 &       50 &    0.405 &    0.919 \\
\hline\hline
\end{tabular}
\footnotesize
\begin{minipage}{\linewidth}
  \vspace{1mm}
  \setlength{\leftskip}{46.5mm}
    Anuran Calls dataset uses only the {\it Family} label. \\
\end{minipage}
\end{table*}

% ===================== GNG (nonstationary) =====================
\begin{table*}[htbp]
\centering
\renewcommand{\arraystretch}{1.2}
\caption{Optimized hyperparameters of GNG under the nonstationary setting}
\label{tab:gng-nonstationary}
\footnotesize
\begin{tabular}{lcccccc}
\hline\hline
\multirow{2}{*}{Dataset} & \multicolumn{6}{c}{Hyperparameter} \\
\cline{2-7}
 & $\epsilon_b$ & $\epsilon_n$ & $a_{\max}$ & $\lambda$ & $\alpha$ & $\delta$ \\
\hline
Iris               &    0.643 &    0.945 &        1 &       10 &    0.025 &    0.949 \\
Seeds              &    0.924 &    0.567 &        2 &       25 &    0.399 &    0.924 \\
Dermatology        &    0.192 &    0.045 &        8 &        6 &    0.385 &    0.983 \\
Pima               &    0.023 &    0.107 &        3 &       48 &    0.305 &    0.949 \\
Mice Protein       &    0.045 &    0.024 &        5 &       43 &    0.489 &    0.987 \\
Binalpha           &    0.558 &    0.076 &        2 &       17 &    0.031 &    0.920 \\
Yeast              &    0.787 &    0.259 &        4 &       38 &    0.010 &    0.963 \\
Semeion            &    0.239 &    0.166 &        3 &       40 &    0.384 &    0.905 \\
MSRA25             &    0.010 &    0.011 &        7 &       36 &    0.387 &    0.903 \\
Image Segmentation         &    0.989 &    0.027 &        6 &       50 &    0.291 &    0.977 \\
Rice               &    0.520 &    0.532 &        8 &       49 &    0.166 &    0.971 \\
TUANDROMD          &    0.223 &    0.032 &        2 &       49 &    0.258 &    0.957 \\
Phoneme            &    0.634 &    0.010 &        4 &       41 &    0.085 &    0.968 \\
Texture            &    0.871 &    0.045 &        3 &       31 &    0.489 &    0.940 \\
OptDigits          &    0.921 &    0.051 &        2 &        3 &    0.367 &    0.974 \\
Statlog            &    0.450 &    0.120 &        2 &       43 &    0.079 &    0.990 \\
Anuran Calls       &    0.109 &    0.168 &        3 &        9 &    0.333 &    0.976 \\
Isolet             &    0.026 &    0.648 &        1 &        2 &    0.355 &    0.952 \\
MNIST10K           &    0.021 &    0.207 &        4 &       49 &    0.094 &    0.927 \\
PenBased           &    0.123 &    0.156 &        1 &       28 &    0.491 &    0.905 \\
STL10              &    0.013 &    0.787 &        2 &        3 &    0.059 &    0.990 \\
Letter             &    0.029 &    0.057 &        6 &       37 &    0.378 &    0.907 \\
Shuttle            &    0.886 &    0.067 &        4 &       35 &    0.204 &    0.902 \\
Skin               &       -- &       -- &       -- &       -- &       -- &       -- \\
\hline\hline
\end{tabular}
\footnotesize
\begin{minipage}{\linewidth}
  \vspace{1mm}
  \setlength{\leftskip}{46.5mm}
    Anuran Calls dataset uses only the {\it Family} label. \\
    The optimization for the Skin dataset failed due to excessive computational requirements. \\
\end{minipage}
\end{table*}

% ================================================================================================================ DDVFA
% ========= DDVFA: optimized hyperparameters (stationary) =========
\begin{table*}[htbp]
\centering
\renewcommand{\arraystretch}{1.2}
\caption{Optimized hyperparameters of DDVFA under the stationary setting}
\label{tab:ddvfa-stationary}
\footnotesize
\begin{tabular}{lcccccc}
\hline\hline
\multirow{2}{*}{Dataset} & \multicolumn{6}{c}{Hyperparameter} \\
\cline{2-7}
 & $\rho_1$ & $\rho_2$ & $\alpha$ & $\beta$ & $\gamma$ & $\gamma_{\mathrm{ref}}$ \\
\hline
Iris               &    0.554 &    0.988 &    0.004 &    0.752 &    2.670 &    2.962 \\
Seeds              &    0.517 &    0.500 &    0.070 &    0.608 &    3.619 &    0.539 \\
Dermatology        &    0.502 &    0.501 &    0.037 &    0.594 &    4.907 &    0.601 \\
Pima               &    0.735 &    0.830 &    0.010 &    0.973 &    4.865 &    0.701 \\
Mice Protein       &    0.716 &    0.968 &    0.084 &    0.912 &    4.586 &    4.565 \\
Binalpha           &    0.524 &    0.512 &    0.049 &    0.524 &    2.261 &    0.869 \\
Yeast              &    0.622 &    0.604 &    0.015 &    0.517 &    4.072 &    0.682 \\
Semeion            &    0.582 &    0.701 &    0.010 &    0.659 &    2.836 &    1.471 \\
MSRA25             &    0.758 &    0.618 &    0.089 &    0.793 &    4.925 &    1.886 \\
Image Segmentation         &    0.546 &    0.548 &    0.000 &    0.819 &    4.981 &    0.731 \\
Rice               &    0.695 &    0.662 &    0.094 &    0.930 &    3.066 &    0.505 \\
TUANDROMD          &    0.579 &    0.580 &    0.022 &    0.679 &    4.995 &    2.731 \\
Phoneme            &    0.500 &    0.501 &    0.011 &    0.902 &    3.959 &    0.652 \\
Texture            &    0.752 &    0.767 &    0.051 &    0.971 &    4.387 &    2.644 \\
OptDigits          &    0.818 &    0.788 &    0.100 &    0.963 &    3.218 &    0.511 \\
Statlog            &    0.700 &    0.704 &    0.097 &    0.632 &    4.605 &    4.707 \\
Anuran Calls       &    0.732 &    0.728 &    0.038 &    0.840 &    4.516 &    1.024 \\
Isolet             &    0.802 &    0.988 &    0.050 &    0.704 &    4.012 &    1.119 \\
MNIST10K           &    0.732 &    0.718 &    0.091 &    0.842 &    4.999 &    0.596 \\
PenBased           &    0.702 &    0.721 &    0.062 &    0.728 &    4.880 &    2.682 \\
STL10              &    0.641 &    0.657 &    0.019 &    0.509 &    1.933 &    0.941 \\
Letter             &    0.776 &    0.772 &    0.058 &    0.555 &    4.998 &    3.135 \\
Shuttle            &    0.668 &    0.620 &    0.072 &    0.853 &    4.664 &    0.923 \\
Skin               &       -- &       -- &       -- &       -- &       -- &       -- \\
\hline\hline
\end{tabular}

\footnotesize
\begin{minipage}{\linewidth}
  \vspace{1mm}
  \setlength{\leftskip}{45mm}
  Anuran Calls dataset uses only the {\it Family} label. \\
  The optimization for the Skin dataset failed due to excessive computational requirements. \\
\end{minipage}
\end{table*}

% ========= DDVFA: optimized hyperparameters (nonstationary) =========
\begin{table*}[htbp]
\centering
\renewcommand{\arraystretch}{1.2}
\caption{Optimized hyperparameters of DDVFA under the nonstationary setting}
\label{tab:ddvfa-nonstationary}
\footnotesize
\begin{tabular}{lcccccc}
\hline\hline
\multirow{2}{*}{Dataset} & \multicolumn{6}{c}{Hyperparameter} \\
\cline{2-7}
 & $\rho_1$ & $\rho_2$ & $\alpha$ & $\beta$ & $\gamma$ & $\gamma_{\mathrm{ref}}$ \\
\hline
Iris               &    0.503 &    0.975 &    0.043 &    0.669 &    1.834 &    2.335 \\
Seeds              &    0.560 &    0.590 &    0.041 &    0.677 &    1.248 &    1.750 \\
Dermatology        &    0.577 &    0.501 &    0.004 &    0.960 &    2.574 &    0.649 \\
Pima               &    0.665 &    0.594 &    0.003 &    0.766 &    4.965 &    2.136 \\
Mice Protein       &    0.855 &    0.599 &    0.038 &    0.598 &    4.501 &    2.505 \\
Binalpha           &    0.952 &    0.710 &    0.012 &    0.581 &    1.100 &    3.087 \\
Yeast              &    0.546 &    0.556 &    0.089 &    0.589 &    4.036 &    1.189 \\
Semeion            &    0.515 &    0.926 &    0.013 &    0.928 &    2.931 &    4.420 \\
MSRA25             &    0.637 &    0.868 &    0.062 &    0.832 &    4.416 &    4.987 \\
Image Segmentation         &    0.553 &    0.502 &    0.074 &    0.630 &    3.607 &    0.505 \\
Rice               &    0.543 &    0.538 &    0.065 &    0.580 &    1.762 &    0.530 \\
TUANDROMD          &    0.936 &    0.988 &    0.010 &    0.871 &    4.026 &    4.827 \\
Phoneme            &    0.531 &    0.504 &    0.064 &    0.624 &    1.650 &    0.550 \\
Texture            &    0.546 &    0.540 &    0.069 &    0.679 &    3.181 &    0.507 \\
OptDigits          &    0.625 &    0.615 &    0.084 &    0.505 &    3.717 &    0.809 \\
Statlog            &    0.614 &    0.607 &    0.012 &    0.504 &    2.285 &    3.000 \\
Anuran Calls       &    0.619 &    0.560 &    0.099 &    0.895 &    3.006 &    1.248 \\
Isolet             &    0.508 &    0.844 &    0.000 &    0.506 &    3.666 &    3.094 \\
MNIST10K           &    0.840 &    0.837 &    0.015 &    0.937 &    4.319 &    1.215 \\
PenBased           &    0.524 &    0.513 &    0.073 &    0.651 &    3.420 &    0.519 \\
STL10              &    0.770 &    0.831 &    0.055 &    0.664 &    4.595 &    3.854 \\
Letter             &    0.529 &    0.535 &    0.037 &    0.557 &    4.891 &    0.612 \\
Shuttle            &    0.549 &    0.955 &    0.024 &    0.552 &    2.869 &    1.073 \\
Skin               &       -- &       -- &       -- &       -- &       -- &       -- \\
\hline\hline
\end{tabular}

\footnotesize
\begin{minipage}{\linewidth}
  \vspace{1mm}
  \setlength{\leftskip}{45mm}
  Anuran Calls dataset uses only the {\it Family} label. \\
  The optimization for the Skin dataset failed due to excessive computational requirements. \\
\end{minipage}
\end{table*}

% ================================================================================================================ CAEA
% ========= CAEA: optimized hyperparameters (stationary) =========
\begin{table}[htbp]
\centering
\renewcommand{\arraystretch}{1.2}
\caption{Optimized hyperparameters of CAEA under the stationary setting}
\label{tab:caea-stationary}
\footnotesize
\begin{tabular}{lcc}
\hline\hline
\multirow{2}{*}{Dataset} & \multicolumn{2}{c}{Hyperparameter} \\
\cline{2-3}
 & $\Lambda$ & $a_{\mathrm{max}}$ \\
\hline
Iris               &       60 &      387 \\
Seeds              &      182 &      235 \\
Dermatology        &        4 &      241 \\
Pima               &      500 &      498 \\
Mice Protein       &      337 &      497 \\
Binalpha           &       66 &      499 \\
Yeast              &        8 &       38 \\
Semeion            &      369 &      181 \\
MSRA25             &        6 &      447 \\
Image Segmentation         &      101 &      488 \\
Rice               &      183 &      200 \\
TUANDROMD          &      268 &      271 \\
Phoneme            &       10 &      330 \\
Texture            &       14 &      342 \\
OptDigits          &       59 &      500 \\
Statlog            &        4 &      483 \\
Anuran Calls       &       78 &      182 \\
Isolet             &      241 &       10 \\
MNIST10K           &      149 &       64 \\
PenBased           &      457 &      241 \\
STL10              &      141 &      224 \\
Letter             &      112 &      481 \\
Shuttle            &      415 &      484 \\
Skin               &      426 &        5 \\
\hline\hline
\end{tabular}

\footnotesize
\begin{minipage}{\linewidth}
  \vspace{1mm}
  \setlength{\leftskip}{17.5mm}
  Anuran Calls dataset uses only the {\it Family} label. \\
\end{minipage}
\vspace{5mm}
\end{table}

% ========= CAEA: optimized hyperparameters (nonstationary) =========
\begin{table}[htbp]
\centering
\renewcommand{\arraystretch}{1.2}
\caption{Optimized hyperparameters of CAEA under the nonstationary setting}
\label{tab:caea-nonstationary}
\footnotesize
\begin{tabular}{lcc}
\hline\hline
\multirow{2}{*}{Dataset} & \multicolumn{2}{c}{Hyperparameter} \\
\cline{2-3}
 & $\Lambda$ & $a_{\mathrm{max}}$ \\
\hline
Iris               &       37 &      152 \\
Seeds              &       85 &      119 \\
Dermatology        &       73 &      459 \\
Pima               &      140 &      229 \\
Mice Protein       &      403 &      181 \\
Binalpha           &      129 &      456 \\
Yeast              &      469 &       77 \\
Semeion            &      173 &       45 \\
MSRA25             &      403 &      181 \\
Image Segmentation         &      403 &      181 \\
Rice               &      403 &      181 \\
TUANDROMD          &      295 &      216 \\
Phoneme            &      499 &        3 \\
Texture            &      132 &      465 \\
OptDigits          &       38 &       18 \\
Statlog            &      282 &       26 \\
Anuran Calls       &      403 &      181 \\
Isolet             &      403 &      181 \\
MNIST10K           &      282 &        2 \\
PenBased           &      500 &       27 \\
STL10              &      226 &       84 \\
Letter             &      106 &      320 \\
Shuttle            &      282 &        2 \\
Skin               &       16 &      154 \\
\hline\hline
\end{tabular}

\footnotesize
\begin{minipage}{\linewidth}
  \vspace{1mm}
  \setlength{\leftskip}{17.5mm}
  Anuran Calls dataset uses only the {\it Family} label. \\
\end{minipage}
\vspace{5mm}
\end{table}

% Results of quantitative comparisons in the stationary setting
\begin{landscape}
\begin{table}[htbp]
	\centering
	\caption{Results of quantitative comparisons in the stationary setting}
	\label{tab:ResultsClusteringStationary}
	\footnotesize
	\renewcommand{\arraystretch}{1.4}
	\scalebox{0.67}{
	\begin{tabular}{ll|*{10}{r C{5mm}|} r C{5mm}} \hline\hline
Dataset & Metric
& \multicolumn{2}{c|}{SR-PCM-HDP} & \multicolumn{2}{c|}{TBM-P} & \multicolumn{2}{c|}{TC} & \multicolumn{2}{c|}{TableDC} & \multicolumn{2}{c|}{G-CEALS} & \multicolumn{2}{c|}{GNG} & \multicolumn{2}{c|}{SOINN+} & \multicolumn{2}{c|}{DDVFA} & \multicolumn{2}{c|}{CAEA} & \multicolumn{2}{c|}{CAE}
& \multicolumn{2}{c}{IDAT} \\ \hline
Iris&ARI&\textnormal{0.713 (0.009)}&\cellcolor{gray!100}{{1}}&\textnormal{0.568 (0.000)}&\cellcolor{gray!65}{{4.5}}&\textnormal{0.568 (0.000)}&\cellcolor{gray!65}{{4.5}}&\textnormal{0.648 (0.092)}&\cellcolor{gray!90}{{2}}&\textnormal{0.348 (0.246)}&\cellcolor{gray!20}{{9}}&\textnormal{0.480 (0.231)}&\cellcolor{gray!30}{{8}}&\textnormal{0.501 (0.096)}&\cellcolor{gray!40}{{7}}&\textnormal{0.567 (0.002)}&\cellcolor{gray!50}{{6}}&\textnormal{0.098 (0.011)}&\cellcolor{gray!0}{{11}}&\textnormal{0.182 (0.141)}&\cellcolor{gray!10}{{10}}&\textnormal{0.624 (0.084)}&\cellcolor{gray!80}{{3}} \\
 &AMI&\textnormal{0.757 (0.006)}&\cellcolor{gray!100}{{1}}&\textnormal{0.732 (0.000)}&\cellcolor{gray!85}{{2.5}}&\textnormal{0.732 (0.000)}&\cellcolor{gray!85}{{2.5}}&\textnormal{0.731 (0.080)}&\cellcolor{gray!70}{{4}}&\textnormal{0.450 (0.304)}&\cellcolor{gray!20}{{9}}&\textnormal{0.536 (0.242)}&\cellcolor{gray!30}{{8}}&\textnormal{0.603 (0.056)}&\cellcolor{gray!40}{{7}}&\textnormal{0.727 (0.012)}&\cellcolor{gray!60}{{5}}&\textnormal{0.357 (0.014)}&\cellcolor{gray!0}{{11}}&\textnormal{0.369 (0.134)}&\cellcolor{gray!10}{{10}}&\textnormal{0.693 (0.054)}&\cellcolor{gray!50}{{6}} \\
\hline\hline
\end{tabular}}
\\
\vspace{1mm}\footnotesize
\hspace*{3.0mm}
\begin{minipage}{\linewidth}
The values in parentheses indicate the standard deviation. \\
A number to the right of a metric value is the average rank of an algorithm over 30 evaluations. \\
The smaller the rank, the better the metric score. A darker tone in a cell corresponds to a smaller rank. \\
N/A indicates that an algorithm could not build a predictive model under the available computational resources. \\
OOM stands for out-of-memory. \\
\end{minipage}
\end{table}

\end{landscape}

% Results of quantitative comparisons in the nonstationary setting
\begin{table*}[htbp]
\centering
\caption{Results of quantitative comparisons in the nonstationary setting}
\label{tab:ResultsClusteringNonstationary}
\footnotesize
\renewcommand{\arraystretch}{1.3}
\scalebox{0.85}{
\begin{tabular}{ll|r C{3.6mm}| r C{3.6mm}| r C{3.6mm}| r C{3.6mm}| r C{3.6mm}| r C{3.6mm}} \hline\hline
Dataset & Metric
& \multicolumn{2}{c|}{GNG} & \multicolumn{2}{c|}{SOINN+} & \multicolumn{2}{c|}{DDVFA} & \multicolumn{2}{c|}{CAEA} & \multicolumn{2}{c|}{CAE} & \multicolumn{2}{c}{IDAT} \\ \hline
Iris & ARI
& \textnormal{0.379 (0.181)} & \cellcolor{gray!60}{{3}} & \textnormal{0.556 (0.104)} & \cellcolor{gray!80}{{2}} & \textnormal{0.048 (0.050)} & \cellcolor{gray!0}{{6}} & \textnormal{0.107 (0.041)} & \cellcolor{gray!20}{{5}} & \textnormal{0.218 (0.227)} & \cellcolor{gray!40}{{4}} & \textnormal{0.565 (0.135)} & \cellcolor{gray!100}{{1}} \\
 & AMI
& \textnormal{0.460 (0.197)} & \cellcolor{gray!60}{{3}} & \textnormal{0.582 (0.061)} & \cellcolor{gray!80}{{2}} & \textnormal{0.138 (0.103)} & \cellcolor{gray!0}{{6}} & \textnormal{0.177 (0.070)} & \cellcolor{gray!20}{{5}} & \textnormal{0.262 (0.233)} & \cellcolor{gray!40}{{4}} & \textnormal{0.611 (0.108)} & \cellcolor{gray!100}{{1}} \\
\hline
Seeds & ARI
& \textnormal{0.454 (0.195)} & \cellcolor{gray!80}{{2}} & \textnormal{0.307 (0.053)} & \cellcolor{gray!60}{{3}} & \textnormal{0.159 (0.039)} & \cellcolor{gray!20}{{5}} & \textnormal{0.305 (0.080)} & \cellcolor{gray!40}{{4}} & \textnormal{0.000 (0.000)} & \cellcolor{gray!0}{{6}} & \textnormal{0.484 (0.046)} & \cellcolor{gray!100}{{1}} \\
 & AMI
& \textnormal{0.499 (0.152)} & \cellcolor{gray!80}{{2}} & \textnormal{0.414 (0.028)} & \cellcolor{gray!60}{{3}} & \textnormal{0.180 (0.051)} & \cellcolor{gray!20}{{5}} & \textnormal{0.363 (0.004)} & \cellcolor{gray!40}{{4}} & \textnormal{0.000 (0.000)} & \cellcolor{gray!0}{{6}} & \textnormal{0.550 (0.020)} & \cellcolor{gray!100}{{1}} \\
\hline
Dermatology & ARI
& \textnormal{0.031 (0.017)} & \cellcolor{gray!20}{{5}} & \textnormal{0.171 (0.047)} & \cellcolor{gray!60}{{3}} & \textnormal{0.258 (0.072)} & \cellcolor{gray!80}{{2}} & \textnormal{0.273 (0.185)} & \cellcolor{gray!100}{{1}} & \textnormal{0.000 (0.001)} & \cellcolor{gray!0}{{6}} & \textnormal{0.072 (0.063)} & \cellcolor{gray!40}{{4}} \\
 & AMI
& \textnormal{0.085 (0.028)} & \cellcolor{gray!20}{{5}} & \textnormal{0.373 (0.045)} & \cellcolor{gray!60}{{3}} & \textnormal{0.376 (0.085)} & \cellcolor{gray!80}{{2}} & \textnormal{0.408 (0.178)} & \cellcolor{gray!100}{{1}} & \textnormal{0.003 (0.008)} & \cellcolor{gray!0}{{6}} & \textnormal{0.179 (0.087)} & \cellcolor{gray!40}{{4}} \\
\hline
Pima & ARI
& \textnormal{0.033 (0.021)} & \cellcolor{gray!80}{{2}} & \textnormal{0.013 (0.023)} & \cellcolor{gray!60}{{3}} & \textnormal{-0.002 (0.013)} & \cellcolor{gray!20}{{5}} & \textnormal{0.003 (0.001)} & \cellcolor{gray!40}{{4}} & \textnormal{-0.006 (0.008)} & \cellcolor{gray!0}{{6}} & \textnormal{0.042 (0.008)} & \cellcolor{gray!100}{{1}} \\
 & AMI
& \textnormal{0.053 (0.011)} & \cellcolor{gray!80}{{2}} & \textnormal{0.015 (0.006)} & \cellcolor{gray!20}{{5}} & \textnormal{0.008 (0.014)} & \cellcolor{gray!0}{{6}} & \textnormal{0.045 (0.001)} & \cellcolor{gray!60}{{3}} & \textnormal{0.020 (0.011)} & \cellcolor{gray!40}{{4}} & \textnormal{0.062 (0.009)} & \cellcolor{gray!100}{{1}} \\
\hline
MiceProtein & ARI
& \textnormal{0.196 (0.024)} & \cellcolor{gray!60}{{3}} & \textnormal{0.243 (0.071)} & \cellcolor{gray!80}{{2}} & \textnormal{0.003 (0.003)} & \cellcolor{gray!20}{{5}} & \textnormal{0.113 (0.017)} & \cellcolor{gray!40}{{4}} & \textnormal{0.002 (0.003)} & \cellcolor{gray!0}{{6}} & \textnormal{0.254 (0.028)} & \cellcolor{gray!100}{{1}} \\
 & AMI
& \textnormal{0.375 (0.028)} & \cellcolor{gray!40}{{4}} & \textnormal{0.493 (0.027)} & \cellcolor{gray!100}{{1}} & \textnormal{0.036 (0.010)} & \cellcolor{gray!20}{{5}} & \textnormal{0.420 (0.019)} & \cellcolor{gray!70}{{2.5}} & \textnormal{0.035 (0.046)} & \cellcolor{gray!0}{{6}} & \textnormal{0.420 (0.033)} & \cellcolor{gray!70}{{2.5}} \\
\hline
Binalpha & ARI
& \textnormal{0.024 (0.016)} & \cellcolor{gray!20}{{5}} & \textnormal{0.211 (0.037)} & \cellcolor{gray!80}{{2}} & \textnormal{0.002 (0.001)} & \cellcolor{gray!0}{{6}} & \textnormal{0.239 (0.014)} & \cellcolor{gray!100}{{1}} & \textnormal{0.095 (0.046)} & \cellcolor{gray!50}{{3.5}} & \textnormal{0.095 (0.069)} & \cellcolor{gray!50}{{3.5}} \\
 & AMI
& \textnormal{0.202 (0.062)} & \cellcolor{gray!30}{{4.5}} & \textnormal{0.444 (0.022)} & \cellcolor{gray!80}{{2}} & \textnormal{0.026 (0.014)} & \cellcolor{gray!0}{{6}} & \textnormal{0.472 (0.014)} & \cellcolor{gray!100}{{1}} & \textnormal{0.202 (0.069)} & \cellcolor{gray!30}{{4.5}} & \textnormal{0.334 (0.116)} & \cellcolor{gray!60}{{3}} \\
\hline
Yeast & ARI
& \textnormal{0.094 (0.048)} & \cellcolor{gray!80}{{2}} & \textnormal{0.034 (0.029)} & \cellcolor{gray!40}{{4}} & \textnormal{0.014 (0.013)} & \cellcolor{gray!20}{{5}} & \textnormal{0.013 (0.002)} & \cellcolor{gray!0}{{6}} & \textnormal{0.077 (0.041)} & \cellcolor{gray!60}{{3}} & \textnormal{0.122 (0.067)} & \cellcolor{gray!100}{{1}} \\
 & AMI
& \textnormal{0.190 (0.043)} & \cellcolor{gray!80}{{2}} & \textnormal{0.080 (0.050)} & \cellcolor{gray!20}{{5}} & \textnormal{0.044 (0.018)} & \cellcolor{gray!0}{{6}} & \textnormal{0.137 (0.011)} & \cellcolor{gray!60}{{3}} & \textnormal{0.120 (0.047)} & \cellcolor{gray!40}{{4}} & \textnormal{0.203 (0.060)} & \cellcolor{gray!100}{{1}} \\
\hline
Semeion & ARI
& \textnormal{0.131 (0.067)} & \cellcolor{gray!40}{{4}} & \textnormal{0.295 (0.054)} & \cellcolor{gray!80}{{2}} & \textnormal{0.029 (0.005)} & \cellcolor{gray!0}{{6}} & \textnormal{0.224 (0.023)} & \cellcolor{gray!60}{{3}} & \textnormal{0.085 (0.057)} & \cellcolor{gray!20}{{5}} & \textnormal{0.299 (0.064)} & \cellcolor{gray!100}{{1}} \\
 & AMI
& \textnormal{0.334 (0.067)} & \cellcolor{gray!40}{{4}} & \textnormal{0.495 (0.034)} & \cellcolor{gray!80}{{2}} & \textnormal{0.148 (0.044)} & \cellcolor{gray!0}{{6}} & \textnormal{0.505 (0.017)} & \cellcolor{gray!100}{{1}} & \textnormal{0.167 (0.069)} & \cellcolor{gray!20}{{5}} & \textnormal{0.462 (0.057)} & \cellcolor{gray!60}{{3}} \\
\hline
MSRA25 & ARI
& \textnormal{0.361 (0.042)} & \cellcolor{gray!60}{{3}} & \textnormal{0.324 (0.030)} & \cellcolor{gray!40}{{4}} & \textnormal{0.060 (0.019)} & \cellcolor{gray!0}{{6}} & \textnormal{0.450 (0.039)} & \cellcolor{gray!100}{{1}} & \textnormal{0.163 (0.171)} & \cellcolor{gray!20}{{5}} & \textnormal{0.369 (0.047)} & \cellcolor{gray!80}{{2}} \\
 & AMI
& \textnormal{0.611 (0.031)} & \cellcolor{gray!60}{{3}} & \textnormal{0.640 (0.017)} & \cellcolor{gray!80}{{2}} & \textnormal{0.167 (0.034)} & \cellcolor{gray!0}{{6}} & \textnormal{0.683 (0.028)} & \cellcolor{gray!100}{{1}} & \textnormal{0.328 (0.207)} & \cellcolor{gray!20}{{5}} & \textnormal{0.593 (0.032)} & \cellcolor{gray!40}{{4}} \\
\hline
Image Segmentation & ARI
& \textnormal{0.253 (0.032)} & \cellcolor{gray!40}{{4}} & \textnormal{0.294 (0.091)} & \cellcolor{gray!80}{{2}} & \textnormal{0.101 (0.043)} & \cellcolor{gray!0}{{6}} & \textnormal{0.249 (0.025)} & \cellcolor{gray!20}{{5}} & \textnormal{0.309 (0.108)} & \cellcolor{gray!100}{{1}} & \textnormal{0.280 (0.083)} & \cellcolor{gray!60}{{3}} \\
 & AMI
& \textnormal{0.484 (0.027)} & \cellcolor{gray!60}{{3}} & \textnormal{0.486 (0.045)} & \cellcolor{gray!80}{{2}} & \textnormal{0.263 (0.088)} & \cellcolor{gray!0}{{6}} & \textnormal{0.478 (0.049)} & \cellcolor{gray!20}{{5}} & \textnormal{0.543 (0.068)} & \cellcolor{gray!100}{{1}} & \textnormal{0.479 (0.080)} & \cellcolor{gray!40}{{4}} \\
\hline
Rice & ARI
& \textnormal{0.213 (0.000)} & \cellcolor{gray!100}{{1}} & \textnormal{0.193 (0.030)} & \cellcolor{gray!60}{{3}} & \textnormal{0.116 (0.013)} & \cellcolor{gray!40}{{4}} & \textnormal{0.070 (0.026)} & \cellcolor{gray!20}{{5}} & \textnormal{0.000 (0.000)} & \cellcolor{gray!0}{{6}} & \textnormal{0.196 (0.061)} & \cellcolor{gray!80}{{2}} \\
 & AMI
& \textnormal{0.272 (0.005)} & \cellcolor{gray!60}{{3}} & \textnormal{0.288 (0.024)} & \cellcolor{gray!100}{{1}} & \textnormal{0.076 (0.010)} & \cellcolor{gray!20}{{5}} & \textnormal{0.124 (0.001)} & \cellcolor{gray!40}{{4}} & \textnormal{0.000 (0.000)} & \cellcolor{gray!0}{{6}} & \textnormal{0.280 (0.026)} & \cellcolor{gray!80}{{2}} \\
\hline
TUANDROMD & ARI
& \textnormal{0.103 (0.026)} & \cellcolor{gray!40}{{4}} & \textnormal{0.095 (0.006)} & \cellcolor{gray!20}{{5}} & \textnormal{-0.020 (0.002)} & \cellcolor{gray!0}{{6}} & \textnormal{0.107 (0.025)} & \cellcolor{gray!60}{{3}} & \textnormal{0.179 (0.051)} & \cellcolor{gray!80}{{2}} & \textnormal{0.201 (0.109)} & \cellcolor{gray!100}{{1}} \\
 & AMI
& \textnormal{0.115 (0.065)} & \cellcolor{gray!20}{{5}} & \textnormal{0.243 (0.010)} & \cellcolor{gray!60}{{3}} & \textnormal{0.009 (0.004)} & \cellcolor{gray!0}{{6}} & \textnormal{0.167 (0.007)} & \cellcolor{gray!40}{{4}} & \textnormal{0.244 (0.043)} & \cellcolor{gray!80}{{2}} & \textnormal{0.282 (0.050)} & \cellcolor{gray!100}{{1}} \\
\hline
Phoneme & ARI
& \textnormal{0.025 (0.013)} & \cellcolor{gray!40}{{4}} & \textnormal{0.026 (0.014)} & \cellcolor{gray!60}{{3}} & \textnormal{0.012 (0.004)} & \cellcolor{gray!20}{{5}} & \textnormal{0.008 (0.009)} & \cellcolor{gray!0}{{6}} & \textnormal{0.081 (0.002)} & \cellcolor{gray!80}{{2}} & \textnormal{0.110 (0.041)} & \cellcolor{gray!100}{{1}} \\
 & AMI
& \textnormal{0.093 (0.005)} & \cellcolor{gray!60}{{3}} & \textnormal{0.044 (0.002)} & \cellcolor{gray!20}{{5}} & \textnormal{0.002 (0.001)} & \cellcolor{gray!0}{{6}} & \textnormal{0.076 (0.003)} & \cellcolor{gray!40}{{4}} & \textnormal{0.125 (0.011)} & \cellcolor{gray!80}{{2}} & \textnormal{0.126 (0.007)} & \cellcolor{gray!100}{{1}} \\
\hline
Texture & ARI
& \textnormal{0.309 (0.036)} & \cellcolor{gray!60}{{3}} & \textnormal{0.482 (0.082)} & \cellcolor{gray!80}{{2}} & \textnormal{0.086 (0.021)} & \cellcolor{gray!20}{{5}} & \textnormal{0.085 (0.020)} & \cellcolor{gray!0}{{6}} & \textnormal{0.307 (0.176)} & \cellcolor{gray!40}{{4}} & \textnormal{0.510 (0.061)} & \cellcolor{gray!100}{{1}} \\
 & AMI
& \textnormal{0.599 (0.014)} & \cellcolor{gray!60}{{3}} & \textnormal{0.630 (0.036)} & \cellcolor{gray!80}{{2}} & \textnormal{0.233 (0.042)} & \cellcolor{gray!0}{{6}} & \textnormal{0.463 (0.019)} & \cellcolor{gray!20}{{5}} & \textnormal{0.576 (0.087)} & \cellcolor{gray!40}{{4}} & \textnormal{0.701 (0.030)} & \cellcolor{gray!100}{{1}} \\
\hline
OptDigits & ARI
& \textnormal{0.329 (0.069)} & \cellcolor{gray!40}{{4}} & \textnormal{0.605 (0.062)} & \cellcolor{gray!80}{{2}} & \textnormal{0.064 (0.017)} & \cellcolor{gray!0}{{6}} & \textnormal{0.113 (0.037)} & \cellcolor{gray!20}{{5}} & \textnormal{0.405 (0.125)} & \cellcolor{gray!60}{{3}} & \textnormal{0.669 (0.050)} & \cellcolor{gray!100}{{1}} \\
 & AMI
& \textnormal{0.605 (0.035)} & \cellcolor{gray!60}{{3}} & \textnormal{0.695 (0.028)} & \cellcolor{gray!80}{{2}} & \textnormal{0.148 (0.037)} & \cellcolor{gray!0}{{6}} & \textnormal{0.470 (0.033)} & \cellcolor{gray!20}{{5}} & \textnormal{0.485 (0.068)} & \cellcolor{gray!40}{{4}} & \textnormal{0.763 (0.022)} & \cellcolor{gray!100}{{1}} \\
\hline
Statlog & ARI
& \textnormal{0.275 (0.060)} & \cellcolor{gray!40}{{4}} & \textnormal{0.346 (0.065)} & \cellcolor{gray!80}{{2}} & \textnormal{0.173 (0.089)} & \cellcolor{gray!20}{{5}} & \textnormal{0.087 (0.033)} & \cellcolor{gray!0}{{6}} & \textnormal{0.328 (0.061)} & \cellcolor{gray!60}{{3}} & \textnormal{0.434 (0.059)} & \cellcolor{gray!100}{{1}} \\
 & AMI
& \textnormal{0.484 (0.026)} & \cellcolor{gray!80}{{2}} & \textnormal{0.482 (0.026)} & \cellcolor{gray!60}{{3}} & \textnormal{0.250 (0.102)} & \cellcolor{gray!0}{{6}} & \textnormal{0.368 (0.021)} & \cellcolor{gray!20}{{5}} & \textnormal{0.429 (0.062)} & \cellcolor{gray!40}{{4}} & \textnormal{0.551 (0.032)} & \cellcolor{gray!100}{{1}} \\
\hline
Anuran Calls & ARI
& \textnormal{0.297 (0.090)} & \cellcolor{gray!80}{{2}} & \textnormal{0.350 (0.094)} & \cellcolor{gray!100}{{1}} & \textnormal{0.137 (0.127)} & \cellcolor{gray!20}{{5}} & \textnormal{0.090 (0.092)} & \cellcolor{gray!0}{{6}} & \textnormal{0.144 (0.083)} & \cellcolor{gray!40}{{4}} & \textnormal{0.285 (0.106)} & \cellcolor{gray!60}{{3}} \\
 & AMI
& \textnormal{0.370 (0.040)} & \cellcolor{gray!60}{{3}} & \textnormal{0.399 (0.029)} & \cellcolor{gray!100}{{1}} & \textnormal{0.107 (0.085)} & \cellcolor{gray!0}{{6}} & \textnormal{0.204 (0.038)} & \cellcolor{gray!20}{{5}} & \textnormal{0.292 (0.056)} & \cellcolor{gray!40}{{4}} & \textnormal{0.373 (0.037)} & \cellcolor{gray!80}{{2}} \\
\hline
Isolet & ARI
& \textnormal{0.295 (0.066)} & \cellcolor{gray!40}{{4}} & \textnormal{0.310 (0.039)} & \cellcolor{gray!100}{{1}} & \textnormal{0.035 (0.018)} & \cellcolor{gray!20}{{5}} & \textnormal{0.308 (0.042)} & \cellcolor{gray!80}{{2}} & \textnormal{0.000 (0.000)} & \cellcolor{gray!0}{{6}} & \textnormal{0.296 (0.071)} & \cellcolor{gray!60}{{3}} \\
 & AMI
& \textnormal{0.634 (0.026)} & \cellcolor{gray!80}{{2}} & \textnormal{0.606 (0.013)} & \cellcolor{gray!60}{{3}} & \textnormal{0.186 (0.057)} & \cellcolor{gray!20}{{5}} & \textnormal{0.571 (0.034)} & \cellcolor{gray!40}{{4}} & \textnormal{0.000 (0.000)} & \cellcolor{gray!0}{{6}} & \textnormal{0.681 (0.026)} & \cellcolor{gray!100}{{1}} \\
\hline
MNIST10K & ARI
& \textnormal{0.573 (0.056)} & \cellcolor{gray!60}{{3}} & \textnormal{0.602 (0.099)} & \cellcolor{gray!80}{{2}} & \textnormal{0.000 (0.000)} & \cellcolor{gray!0}{{6}} & \textnormal{0.363 (0.088)} & \cellcolor{gray!20}{{5}} & \textnormal{0.557 (0.108)} & \cellcolor{gray!40}{{4}} & \textnormal{0.644 (0.132)} & \cellcolor{gray!100}{{1}} \\
 & AMI
& \textnormal{0.759 (0.018)} & \cellcolor{gray!80}{{2}} & \textnormal{0.703 (0.050)} & \cellcolor{gray!40}{{4}} & \textnormal{0.000 (0.000)} & \cellcolor{gray!0}{{6}} & \textnormal{0.598 (0.039)} & \cellcolor{gray!20}{{5}} & \textnormal{0.732 (0.047)} & \cellcolor{gray!60}{{3}} & \textnormal{0.793 (0.083)} & \cellcolor{gray!100}{{1}} \\
\hline
PenBased & ARI
& \textnormal{0.383 (0.058)} & \cellcolor{gray!40}{{4}} & \textnormal{0.547 (0.045)} & \cellcolor{gray!80}{{2}} & \textnormal{0.102 (0.027)} & \cellcolor{gray!20}{{5}} & \textnormal{0.034 (0.023)} & \cellcolor{gray!0}{{6}} & \textnormal{0.449 (0.142)} & \cellcolor{gray!60}{{3}} & \textnormal{0.581 (0.045)} & \cellcolor{gray!100}{{1}} \\
 & AMI
& \textnormal{0.631 (0.027)} & \cellcolor{gray!60}{{3}} & \textnormal{0.650 (0.022)} & \cellcolor{gray!80}{{2}} & \textnormal{0.263 (0.044)} & \cellcolor{gray!0}{{6}} & \textnormal{0.423 (0.011)} & \cellcolor{gray!20}{{5}} & \textnormal{0.625 (0.080)} & \cellcolor{gray!40}{{4}} & \textnormal{0.718 (0.020)} & \cellcolor{gray!100}{{1}} \\
\hline
STL10 & ARI
& \textnormal{0.298 (0.125)} & \cellcolor{gray!60}{{3}} & \textnormal{0.454 (0.161)} & \cellcolor{gray!100}{{1}} & \textnormal{0.000 (0.000)} & \cellcolor{gray!10}{{5.5}} & \textnormal{0.214 (0.057)} & \cellcolor{gray!40}{{4}} & \textnormal{0.000 (0.000)} & \cellcolor{gray!10}{{5.5}} & \textnormal{0.341 (0.207)} & \cellcolor{gray!80}{{2}} \\
 & AMI
& \textnormal{0.573 (0.092)} & \cellcolor{gray!60}{{3}} & \textnormal{0.649 (0.087)} & \cellcolor{gray!100}{{1}} & \textnormal{0.000 (0.000)} & \cellcolor{gray!10}{{5.5}} & \textnormal{0.413 (0.058)} & \cellcolor{gray!40}{{4}} & \textnormal{0.000 (0.000)} & \cellcolor{gray!10}{{5.5}} & \textnormal{0.577 (0.185)} & \cellcolor{gray!80}{{2}} \\
\hline
Letter & ARI
& \textnormal{0.040 (0.018)} & \cellcolor{gray!20}{{5}} & \textnormal{0.107 (0.017)} & \cellcolor{gray!60}{{3}} & \textnormal{0.011 (0.007)} & \cellcolor{gray!0}{{6}} & \textnormal{0.049 (0.008)} & \cellcolor{gray!40}{{4}} & \textnormal{0.135 (0.030)} & \cellcolor{gray!100}{{1}} & \textnormal{0.108 (0.032)} & \cellcolor{gray!80}{{2}} \\
 & AMI
& \textnormal{0.376 (0.031)} & \cellcolor{gray!20}{{5}} & \textnormal{0.480 (0.012)} & \cellcolor{gray!100}{{1}} & \textnormal{0.063 (0.024)} & \cellcolor{gray!0}{{6}} & \textnormal{0.386 (0.043)} & \cellcolor{gray!40}{{4}} & \textnormal{0.478 (0.023)} & \cellcolor{gray!80}{{2}} & \textnormal{0.445 (0.016)} & \cellcolor{gray!60}{{3}} \\
\hline
Shuttle & ARI
& \textnormal{0.266 (0.197)} & \cellcolor{gray!40}{{4}} & \textnormal{0.689 (0.048)} & \cellcolor{gray!100}{{1}} & \textnormal{0.247 (0.191)} & \cellcolor{gray!20}{{5}} & \textnormal{0.020 (0.026)} & \cellcolor{gray!0}{{6}} & \textnormal{0.270 (0.194)} & \cellcolor{gray!60}{{3}} & \textnormal{0.273 (0.107)} & \cellcolor{gray!80}{{2}} \\
 & AMI
& \textnormal{0.348 (0.106)} & \cellcolor{gray!60}{{3}} & \textnormal{0.591 (0.046)} & \cellcolor{gray!100}{{1}} & \textnormal{0.248 (0.191)} & \cellcolor{gray!20}{{5}} & \textnormal{0.204 (0.014)} & \cellcolor{gray!0}{{6}} & \textnormal{0.395 (0.099)} & \cellcolor{gray!80}{{2}} & \textnormal{0.345 (0.062)} & \cellcolor{gray!40}{{4}} \\
\hline
Skin & ARI
& N/A\hspace{4.5mm} & \cellcolor{gray!10}{{5.5}} & \textnormal{0.081 (0.005)} & \cellcolor{gray!60}{{3}} & N/A\hspace{4.5mm} & \cellcolor{gray!10}{{5.5}} & \textnormal{0.016 (0.002)} & \cellcolor{gray!40}{{4}} & \textnormal{0.635 (0.036)} & \cellcolor{gray!100}{{1}} & \textnormal{0.162 (0.039)} & \cellcolor{gray!80}{{2}} \\
 & AMI
& N/A\hspace{4.5mm} & \cellcolor{gray!10}{{5.5}} & \textnormal{0.242 (0.017)} & \cellcolor{gray!60}{{3}} & N/A\hspace{4.5mm} & \cellcolor{gray!10}{{5.5}} & \textnormal{0.179 (0.007)} & \cellcolor{gray!40}{{4}} & \textnormal{0.510 (0.053)} & \cellcolor{gray!100}{{1}} & \textnormal{0.267 (0.032)} & \cellcolor{gray!80}{{2}} \\
\hline\hline
\end{tabular}}
\\
\vspace{1mm}\footnotesize
\hspace*{2.5mm}
\begin{minipage}{\linewidth}
The values in parentheses indicate the standard deviation. \\
A number to the right of a metric value is the average rank of an algorithm over 30 evaluations. \\
The smaller the rank, the better the metric score. A darker tone in a cell corresponds to a smaller rank. \\
N/A indicates that an algorithm could not build a predictive model under the available computational resources. \\
\end{minipage}
\end{table*}

% Results of AI-ARI, AI-AMI, BWT-ARI, and BWT-AMI in the nonstationary setting
\begin{table*}[htbp]
\centering
\caption{Results of AI-ARI, AI-AMI, BWT-ARI, and BWT-AMI in the nonstationary setting}
\label{tab:IncBwtNonstationary}
\footnotesize
\renewcommand{\arraystretch}{1.2}
\scalebox{0.62}{
\begin{tabular}{ll|r C{3.6mm}| r C{3.6mm}| r C{3.6mm}| r C{3.6mm}| r C{3.6mm}| r C{3.6mm}} \hline\hline
Dataset & Metric
& \multicolumn{2}{c|}{GNG} & \multicolumn{2}{c|}{SOINN+} & \multicolumn{2}{c|}{DDVFA} & \multicolumn{2}{c|}{CAEA} & \multicolumn{2}{c|}{CAE} & \multicolumn{2}{c}{IDAT} \\ \hline
Iris & AI-ARI
& \textnormal{0.576 (0.161)} & \cellcolor{gray!60}{{3}} & \textnormal{0.696 (0.070)} & \cellcolor{gray!100}{{1}} & \textnormal{0.366 (0.007)} & \cellcolor{gray!0}{{6}} & \textnormal{0.400 (0.008)} & \cellcolor{gray!20}{{5}} & \textnormal{0.440 (0.130)} & \cellcolor{gray!40}{{4}} & \textnormal{0.662 (0.074)} & \cellcolor{gray!80}{{2}} \\
 & AI-AMI
& \textnormal{0.610 (0.160)} & \cellcolor{gray!60}{{3}} & \textnormal{0.697 (0.047)} & \cellcolor{gray!100}{{1}} & \textnormal{0.421 (0.008)} & \cellcolor{gray!0}{{6}} & \textnormal{0.441 (0.022)} & \cellcolor{gray!20}{{5}} & \textnormal{0.471 (0.140)} & \cellcolor{gray!40}{{4}} & \textnormal{0.684 (0.077)} & \cellcolor{gray!80}{{2}} \\
 & BWT-ARI
& \textnormal{-0.296 (0.178)} & \cellcolor{gray!60}{{3}} & \textnormal{-0.210 (0.054)} & \cellcolor{gray!80}{{2}} & \textnormal{-0.477 (0.072)} & \cellcolor{gray!0}{{6}} & \textnormal{-0.439 (0.049)} & \cellcolor{gray!20}{{5}} & \textnormal{-0.334 (0.160)} & \cellcolor{gray!40}{{4}} & \textnormal{-0.145 (0.206)} & \cellcolor{gray!100}{{1}} \\
 & BWT-AMI
& \textnormal{-0.224 (0.180)} & \cellcolor{gray!60}{{3}} & \textnormal{-0.172 (0.031)} & \cellcolor{gray!80}{{2}} & \textnormal{-0.424 (0.150)} & \cellcolor{gray!0}{{6}} & \textnormal{-0.396 (0.075)} & \cellcolor{gray!20}{{5}} & \textnormal{-0.313 (0.150)} & \cellcolor{gray!40}{{4}} & \textnormal{-0.110 (0.163)} & \cellcolor{gray!100}{{1}} \\
\hline\hline
\end{tabular}}
\\
\vspace{1mm}\footnotesize
\hspace*{24mm}
\begin{minipage}{\linewidth}
The values in parentheses indicate the standard deviation. \\
A number to the right of a metric value is the average rank of an algorithm over 30 evaluations. \\
The smaller the rank, the better the metric score. A darker tone in a cell corresponds to a smaller rank. \\
N/A indicates that an algorithm could not build a predictive model under the available computational resources. \\
\end{minipage}
\end{table*}

\begin{landscape}

\begin{table}[htbp]
\centering
\caption{Results of quantitative comparisons in the nonstationary setting (ablation study)}
\label{tab:AblationNonstationary}
\footnotesize
\renewcommand{\arraystretch}{1.2}
\scalebox{0.82}{
\begin{tabular}{ll|rC{3.6mm}|rC{3.6mm}|rC{3.6mm}|rC{3.6mm}|rC{3.6mm}|rC{3.6mm}|rC{3.6mm}|rC{3.6mm}} \hline\hline
\multirow{2}{*}{Dataset} & \multirow{2}{*}{Metric} & \multicolumn{4}{c|}{w/o All} & \multicolumn{4}{c|}{w/o Decremental} & \multicolumn{4}{c|}{w/o Incremental} & \multicolumn{4}{c}{IDAT} \\
  &   & \multicolumn{2}{c|}{$\Lambda_{\text{init}} = 2$} & \multicolumn{2}{c|}{$\Lambda_{\text{init}} = 500$} & \multicolumn{2}{c|}{$\Lambda_{\text{init}} = 2$} & \multicolumn{2}{c|}{$\Lambda_{\text{init}} = 500$} & \multicolumn{2}{c|}{$\Lambda_{\text{init}} = 2$} & \multicolumn{2}{c|}{$\Lambda_{\text{init}} = 500$} & \multicolumn{2}{c|}{$\Lambda_{\text{init}} = 2$} & \multicolumn{2}{c}{$\Lambda_{\text{init}} = 500$} \\ \hline
Iris & ARI & \textnormal{0.313 (0.312)} & \cellcolor{gray!14}{{7}} & \textnormal{0.321 (0.179)} & \cellcolor{gray!50}{{4.5}} & \textnormal{0.468 (0.156)} & \cellcolor{gray!86}{{2}} & \textnormal{0.321 (0.179)} & \cellcolor{gray!50}{{4.5}} & \textnormal{0.281 (0.191)} & \cellcolor{gray!0}{{8}} & \textnormal{0.321 (0.179)} & \cellcolor{gray!50}{{4.5}} & \textnormal{0.565 (0.135)} & \cellcolor{gray!100}{{1}} & \textnormal{0.321 (0.179)} & \cellcolor{gray!50}{{4.5}} \\
  & AMI & \textnormal{0.342 (0.321)} & \cellcolor{gray!0}{{8}} & \textnormal{0.413 (0.166)} & \cellcolor{gray!50}{{4.5}} & \textnormal{0.551 (0.132)} & \cellcolor{gray!86}{{2}} & \textnormal{0.413 (0.166)} & \cellcolor{gray!50}{{4.5}} & \textnormal{0.379 (0.249)} & \cellcolor{gray!14}{{7}} & \textnormal{0.413 (0.166)} & \cellcolor{gray!50}{{4.5}} & \textnormal{0.611 (0.108)} & \cellcolor{gray!100}{{1}} & \textnormal{0.413 (0.166)} & \cellcolor{gray!50}{{4.5}} \\
\hline\hline
\end{tabular}}
\\
\vspace{1mm}\footnotesize
\hspace*{16mm}
\begin{minipage}{\linewidth}
The values in parentheses indicate the standard deviation. \\
A number to the right of a metric value is the average rank of an algorithm over 30 evaluations. \\
The smaller the rank, the better the metric score. A darker tone in a cell corresponds to a smaller rank. \\
\end{minipage}
\end{table}

\end{landscape}

% Results of AI-ARI, AI-AMI, BWT-ARI, and BWT-AMI in the nonstationary setting (ablation study)
\begin{table*}[htbp]
\centering
\caption{Results of AI-ARI, AI-AMI, BWT-ARI, and BWT-AMI in the nonstationary setting (ablation study)}
\label{tab:AblationNonstationary_IncBWT}
\footnotesize
\renewcommand{\arraystretch}{1.2}
\scalebox{0.62}{
\begin{tabular}{ll|rC{3.6mm}|rC{3.6mm}|rC{3.6mm}|rC{3.6mm}|rC{3.6mm}|rC{3.6mm}|rC{3.6mm}|rC{3.6mm}} \hline\hline
\multirow{2}{*}{Dataset} & \multirow{2}{*}{Metric} & \multicolumn{4}{c|}{w/o All} & \multicolumn{4}{c|}{w/o Decremental} & \multicolumn{4}{c|}{w/o Incremental} & \multicolumn{4}{c}{IDAT} \\
  &   & \multicolumn{2}{c|}{$\Lambda_{\text{init}} = 2$} & \multicolumn{2}{c|}{$\Lambda_{\text{init}} = 500$} & \multicolumn{2}{c|}{$\Lambda_{\text{init}} = 2$} & \multicolumn{2}{c|}{$\Lambda_{\text{init}} = 500$} & \multicolumn{2}{c|}{$\Lambda_{\text{init}} = 2$} & \multicolumn{2}{c|}{$\Lambda_{\text{init}} = 500$} & \multicolumn{2}{c|}{$\Lambda_{\text{init}} = 2$} & \multicolumn{2}{c}{$\Lambda_{\text{init}} = 500$} \\ \hline
Iris & AI-ARI & \textnormal{0.509 (0.153)} & \cellcolor{gray!14}{{7}} & \textnormal{0.551 (0.055)} & \cellcolor{gray!50}{{4.5}} & \textnormal{0.615 (0.097)} & \cellcolor{gray!86}{{2}} & \textnormal{0.551 (0.055)} & \cellcolor{gray!50}{{4.5}} & \textnormal{0.472 (0.081)} & \cellcolor{gray!0}{{8}} & \textnormal{0.551 (0.055)} & \cellcolor{gray!50}{{4.5}} & \textnormal{0.662 (0.074)} & \cellcolor{gray!100}{{1}} & \textnormal{0.551 (0.055)} & \cellcolor{gray!50}{{4.5}} \\
  & AI-AMI & \textnormal{0.508 (0.147)} & \cellcolor{gray!0}{{8}} & \textnormal{0.579 (0.042)} & \cellcolor{gray!50}{{4.5}} & \textnormal{0.653 (0.094)} & \cellcolor{gray!86}{{2}} & \textnormal{0.579 (0.042)} & \cellcolor{gray!50}{{4.5}} & \textnormal{0.530 (0.095)} & \cellcolor{gray!14}{{7}} & \textnormal{0.579 (0.042)} & \cellcolor{gray!50}{{4.5}} & \textnormal{0.684 (0.077)} & \cellcolor{gray!100}{{1}} & \textnormal{0.579 (0.042)} & \cellcolor{gray!50}{{4.5}} \\
\hline\hline
\end{tabular}}
\\
\vspace{1mm}\footnotesize
\hspace*{6mm}
\begin{minipage}{\linewidth}
The values in parentheses indicate the standard deviation. \\
A number to the right of a metric value is the average rank of an algorithm over 30 evaluations. \\
The smaller the rank, the better the metric score. A darker tone in a cell corresponds to a smaller rank. \\
\end{minipage}
\end{table*}

% Average number of nodes and clusters in the nonstationary setting (ablation study)
\begin{table*}[htbp]
\centering
\caption{Average number of nodes and clusters in the nonstationary setting (ablation study)}
\label{tab:AblationNonstationary_NodesClusters}
\footnotesize
\renewcommand{\arraystretch}{1.2}
\scalebox{0.8}{
\begin{tabular}{ll|>{\centering\arraybackslash}p{17mm}|>{\centering\arraybackslash}p{17mm}|>{\centering\arraybackslash}p{17mm}|>{\centering\arraybackslash}p{17mm}|>{\centering\arraybackslash}p{17mm}|>{\centering\arraybackslash}p{17mm}|>{\centering\arraybackslash}p{17mm}|>{\centering\arraybackslash}p{17mm}} \hline\hline
\multirow{2}{*}{Dataset} & \multirow{2}{*}{Metric} & \multicolumn{2}{c|}{w/o All} & \multicolumn{2}{c|}{w/o Decremental} & \multicolumn{2}{c|}{w/o Incremental} & \multicolumn{2}{c}{IDAT} \\
  &   & $\Lambda_{\text{init}} = 2$ & $\Lambda_{\text{init}} = 500$ & $\Lambda_{\text{init}} = 2$ & $\Lambda_{\text{init}} = 500$ & $\Lambda_{\text{init}} = 2$ & $\Lambda_{\text{init}} = 500$ & $\Lambda_{\text{init}} = 2$ & $\Lambda_{\text{init}} = 500$ \\ \hline
Iris & \#Nodes & \textnormal{2.3 (0.5)} & \textnormal{2.4 (0.5)} & \textnormal{8.3 (2.3)} & \textnormal{2.4 (0.5)} & \textnormal{4.6 (2.2)} & \textnormal{2.4 (0.5)} & \textnormal{7.7 (1.2)} & \textnormal{2.4 (0.5)} \\
(\#Classes: 3) & \#Clusters & \textnormal{2.0 (0.9)} & \textnormal{2.0 (0.0)} & \textnormal{4.4 (1.4)} & \textnormal{2.0 (0.0)} & \textnormal{2.5 (1.1)} & \textnormal{2.0 (0.0)} & \textnormal{4.1 (1.1)} & \textnormal{2.0 (0.0)} \\
\hline
Seeds & \#Nodes & \textnormal{2.1 (0.3)} & \textnormal{2.4 (0.5)} & \textnormal{11.6 (1.9)} & \textnormal{2.4 (0.5)} & \textnormal{5.7 (2.5)} & \textnormal{2.4 (0.5)} & \textnormal{12.7 (2.1)} & \textnormal{2.4 (0.5)} \\
(\#Classes: 3) & \#Clusters & \textnormal{1.7 (0.7)} & \textnormal{2.0 (0.0)} & \textnormal{5.7 (2.1)} & \textnormal{2.0 (0.0)} & \textnormal{2.7 (1.3)} & \textnormal{2.0 (0.0)} & \textnormal{6.0 (1.5)} & \textnormal{2.0 (0.0)} \\
\hline
Dermatology & \#Nodes & \textnormal{2.6 (0.5)} & \textnormal{2.4 (0.5)} & \textnormal{17.6 (2.9)} & \textnormal{2.4 (0.5)} & \textnormal{6.7 (2.3)} & \textnormal{2.4 (0.5)} & \textnormal{14.1 (3.7)} & \textnormal{2.4 (0.5)} \\
(\#Classes: 6) & \#Clusters & \textnormal{2.4 (0.8)} & \textnormal{2.2 (0.4)} & \textnormal{9.5 (2.2)} & \textnormal{2.2 (0.4)} & \textnormal{3.0 (1.3)} & \textnormal{2.2 (0.4)} & \textnormal{7.1 (3.1)} & \textnormal{2.2 (0.4)} \\
\hline
Pima & \#Nodes & \textnormal{2.0 (0.0)} & \textnormal{2.0 (0.0)} & \textnormal{12.6 (2.0)} & \textnormal{2.0 (0.0)} & \textnormal{7.0 (0.0)} & \textnormal{15.8 (11.5)} & \textnormal{14.0 (5.0)} & \textnormal{18.2 (13.5)} \\
(\#Classes: 2) & \#Clusters & \textnormal{2.0 (0.0)} & \textnormal{2.0 (0.0)} & \textnormal{5.8 (3.5)} & \textnormal{2.0 (0.0)} & \textnormal{2.0 (0.0)} & \textnormal{15.8 (11.5)} & \textnormal{7.4 (3.0)} & \textnormal{18.2 (13.5)} \\
\hline
MiceProtein & \#Nodes & \textnormal{2.6 (0.5)} & \textnormal{2.2 (0.4)} & \textnormal{32.7 (6.0)} & \textnormal{4.8 (14.2)} & \textnormal{15.4 (3.5)} & \textnormal{7.9 (13.0)} & \textnormal{31.0 (6.5)} & \textnormal{5.3 (2.2)} \\
(\#Classes: 8) & \#Clusters & \textnormal{2.5 (0.7)} & \textnormal{2.1 (0.3)} & \textnormal{14.8 (4.2)} & \textnormal{4.7 (14.2)} & \textnormal{5.6 (1.5)} & \textnormal{7.0 (13.1)} & \textnormal{12.8 (2.5)} & \textnormal{4.3 (2.1)} \\
\hline
Binalpha & \#Nodes & \textnormal{2.3 (0.4)} & \textnormal{2.3 (0.4)} & \textnormal{44.5 (7.8)} & \textnormal{55.9 (139.5)} & \textnormal{9.6 (2.8)} & \textnormal{19.5 (37.7)} & \textnormal{42.8 (7.3)} & \textnormal{193.3 (273.6)} \\
(\#Classes: 36) & \#Clusters & \textnormal{2.2 (0.6)} & \textnormal{2.2 (0.4)} & \textnormal{24.5 (4.6)} & \textnormal{55.8 (139.5)} & \textnormal{2.0 (0.9)} & \textnormal{17.7 (38.1)} & \textnormal{18.7 (6.5)} & \textnormal{192.4 (274.2)} \\
\hline
Yeast & \#Nodes & \textnormal{2.2 (0.4)} & \textnormal{2.3 (0.4)} & \textnormal{35.8 (5.2)} & \textnormal{2.3 (0.4)} & \textnormal{12.7 (4.3)} & \textnormal{10.8 (3.8)} & \textnormal{31.8 (6.4)} & \textnormal{14.4 (3.0)} \\
(\#Classes: 10) & \#Clusters & \textnormal{1.9 (0.7)} & \textnormal{2.0 (0.0)} & \textnormal{9.0 (3.0)} & \textnormal{2.0 (0.0)} & \textnormal{2.9 (1.3)} & \textnormal{7.9 (4.0)} & \textnormal{8.2 (3.0)} & \textnormal{11.3 (2.8)} \\
\hline
Semeion & \#Nodes & \textnormal{2.6 (0.5)} & \textnormal{2.3 (0.5)} & \textnormal{40.0 (7.6)} & \textnormal{160.0 (266.2)} & \textnormal{6.8 (1.6)} & \textnormal{15.0 (4.1)} & \textnormal{44.2 (5.6)} & \textnormal{13.9 (4.4)} \\
(\#Classes: 10) & \#Clusters & \textnormal{2.5 (0.7)} & \textnormal{2.2 (0.4)} & \textnormal{23.3 (5.4)} & \textnormal{159.9 (266.3)} & \textnormal{2.0 (0.9)} & \textnormal{10.1 (4.3)} & \textnormal{25.2 (5.9)} & \textnormal{10.7 (3.9)} \\
\hline
MSRA25 & \#Nodes & \textnormal{2.9 (0.3)} & \textnormal{2.1 (0.3)} & \textnormal{43.2 (7.7)} & \textnormal{81.3 (241.9)} & \textnormal{20.6 (6.5)} & \textnormal{12.2 (4.5)} & \textnormal{41.0 (5.3)} & \textnormal{15.3 (3.8)} \\
(\#Classes: 12) & \#Clusters & \textnormal{2.9 (0.3)} & \textnormal{2.1 (0.3)} & \textnormal{21.2 (4.1)} & \textnormal{81.3 (241.9)} & \textnormal{9.8 (2.9)} & \textnormal{6.9 (2.3)} & \textnormal{19.0 (3.4)} & \textnormal{9.3 (2.8)} \\
\hline
Image Segmentation & \#Nodes & \textnormal{2.5 (0.5)} & \textnormal{2.2 (0.4)} & \textnormal{38.2 (9.1)} & \textnormal{699.1 (354.5)} & \textnormal{21.7 (5.9)} & \textnormal{27.7 (47.3)} & \textnormal{36.7 (9.3)} & \textnormal{13.2 (4.8)} \\
(\#Classes: 7) & \#Clusters & \textnormal{2.3 (0.8)} & \textnormal{2.0 (0.2)} & \textnormal{15.8 (4.0)} & \textnormal{698.9 (354.9)} & \textnormal{7.3 (2.3)} & \textnormal{22.9 (49.0)} & \textnormal{14.5 (3.6)} & \textnormal{6.9 (2.7)} \\
\hline
Rice & \#Nodes & \textnormal{2.0 (0.0)} & \textnormal{3.0 (0.0)} & \textnormal{28.2 (1.0)} & \textnormal{857.2 (23.9)} & \textnormal{18.0 (2.5)} & \textnormal{11.6 (0.5)} & \textnormal{23.4 (0.5)} & \textnormal{15.8 (1.0)} \\
(\#Classes: 2) & \#Clusters & \textnormal{2.0 (0.0)} & \textnormal{2.0 (0.0)} & \textnormal{12.2 (1.0)} & \textnormal{857.2 (23.9)} & \textnormal{4.8 (1.0)} & \textnormal{5.6 (0.5)} & \textnormal{7.2 (3.5)} & \textnormal{6.2 (1.0)} \\
\hline
TUANDROMD & \#Nodes & \textnormal{2.0 (0.0)} & \textnormal{3.0 (0.0)} & \textnormal{17.2 (4.0)} & \textnormal{54.4 (65.3)} & \textnormal{8.4 (3.0)} & \textnormal{5.8 (3.5)} & \textnormal{9.4 (3.0)} & \textnormal{9.0 (2.5)} \\
(\#Classes: 2) & \#Clusters & \textnormal{1.6 (0.5)} & \textnormal{2.0 (0.0)} & \textnormal{9.4 (3.0)} & \textnormal{53.8 (65.8)} & \textnormal{2.8 (1.0)} & \textnormal{2.2 (1.5)} & \textnormal{5.2 (1.5)} & \textnormal{4.2 (1.0)} \\
\hline
Phoneme & \#Nodes & \textnormal{2.6 (0.5)} & \textnormal{2.0 (0.0)} & \textnormal{48.8 (4.0)} & \textnormal{986.6 (4.5)} & \textnormal{26.0 (5.0)} & \textnormal{24.2 (1.5)} & \textnormal{35.6 (3.0)} & \textnormal{34.0 (2.5)} \\
(\#Classes: 2) & \#Clusters & \textnormal{2.6 (0.5)} & \textnormal{2.0 (0.0)} & \textnormal{19.8 (1.5)} & \textnormal{986.6 (4.5)} & \textnormal{5.0 (0.0)} & \textnormal{12.2 (1.5)} & \textnormal{12.4 (0.5)} & \textnormal{13.2 (1.5)} \\
\hline
Texture & \#Nodes & \textnormal{2.2 (0.4)} & \textnormal{2.3 (0.4)} & \textnormal{135.1 (15.6)} & \textnormal{898.5 (358.4)} & \textnormal{48.2 (8.6)} & \textnormal{41.3 (15.5)} & \textnormal{128.6 (16.3)} & \textnormal{40.3 (14.2)} \\
(\#Classes: 11) & \#Clusters & \textnormal{1.9 (0.7)} & \textnormal{1.9 (0.4)} & \textnormal{39.8 (5.3)} & \textnormal{898.3 (358.7)} & \textnormal{16.8 (2.6)} & \textnormal{16.6 (4.2)} & \textnormal{34.3 (4.7)} & \textnormal{16.1 (3.9)} \\
\hline
OptDigits & \#Nodes & \textnormal{2.3 (0.5)} & \textnormal{2.5 (0.5)} & \textnormal{133.4 (14.1)} & \textnormal{615.9 (209.2)} & \textnormal{35.2 (6.7)} & \textnormal{37.5 (12.5)} & \textnormal{129.3 (15.7)} & \textnormal{43.6 (8.6)} \\
(\#Classes: 10) & \#Clusters & \textnormal{1.8 (0.9)} & \textnormal{2.0 (0.3)} & \textnormal{33.2 (6.3)} & \textnormal{615.8 (209.6)} & \textnormal{13.3 (2.2)} & \textnormal{16.0 (4.2)} & \textnormal{31.2 (5.9)} & \textnormal{17.5 (4.2)} \\
\hline
Statlog & \#Nodes & \textnormal{2.3 (0.5)} & \textnormal{2.3 (0.4)} & \textnormal{51.4 (8.6)} & \textnormal{950.1 (180.5)} & \textnormal{24.4 (5.1)} & \textnormal{31.7 (7.1)} & \textnormal{45.8 (8.5)} & \textnormal{33.7 (9.5)} \\
(\#Classes: 6) & \#Clusters & \textnormal{1.8 (0.9)} & \textnormal{1.7 (0.4)} & \textnormal{20.0 (5.1)} & \textnormal{950.1 (180.5)} & \textnormal{8.1 (1.6)} & \textnormal{13.0 (3.7)} & \textnormal{17.0 (3.8)} & \textnormal{13.9 (4.1)} \\
\hline
Anuran Calls & \#Nodes & \textnormal{2.7 (0.5)} & \textnormal{2.4 (0.5)} & \textnormal{32.1 (9.4)} & \textnormal{280.0 (370.9)} & \textnormal{26.1 (3.8)} & \textnormal{23.4 (5.7)} & \textnormal{27.5 (7.2)} & \textnormal{25.2 (8.4)} \\
(\#Classes: 4) & \#Clusters & \textnormal{2.7 (0.5)} & \textnormal{1.8 (0.4)} & \textnormal{13.1 (5.4)} & \textnormal{279.5 (371.3)} & \textnormal{8.0 (1.7)} & \textnormal{8.2 (2.0)} & \textnormal{11.1 (4.4)} & \textnormal{9.6 (2.3)} \\
\hline
Isolet & \#Nodes & \textnormal{2.7 (0.5)} & \textnormal{2.3 (0.5)} & \textnormal{82.7 (10.5)} & \textnormal{799.4 (272.7)} & \textnormal{32.4 (3.7)} & \textnormal{56.2 (11.3)} & \textnormal{76.2 (8.9)} & \textnormal{63.4 (10.4)} \\
(\#Classes: 26) & \#Clusters & \textnormal{2.4 (0.9)} & \textnormal{2.1 (0.4)} & \textnormal{30.4 (5.8)} & \textnormal{799.3 (273.0)} & \textnormal{11.6 (1.8)} & \textnormal{22.2 (4.9)} & \textnormal{24.4 (3.4)} & \textnormal{23.4 (5.5)} \\
\hline
MNIST10K & \#Nodes & \textnormal{2.0 (0.0)} & \textnormal{2.0 (0.0)} & \textnormal{137.4 (27.0)} & \textnormal{1039.5 (354.0)} & \textnormal{45.1 (12.1)} & \textnormal{73.0 (22.8)} & \textnormal{124.5 (32.8)} & \textnormal{77.0 (29.7)} \\
(\#Classes: 10) & \#Clusters & \textnormal{1.6 (0.5)} & \textnormal{1.6 (0.5)} & \textnormal{40.7 (9.3)} & \textnormal{1039.3 (354.6)} & \textnormal{15.6 (3.8)} & \textnormal{26.1 (5.9)} & \textnormal{32.3 (7.6)} & \textnormal{26.9 (6.2)} \\
\hline
PenBased & \#Nodes & \textnormal{2.5 (0.5)} & \textnormal{2.4 (0.5)} & \textnormal{150.3 (31.7)} & \textnormal{1168.6 (401.0)} & \textnormal{65.2 (6.6)} & \textnormal{62.2 (14.0)} & \textnormal{132.1 (29.2)} & \textnormal{69.7 (12.9)} \\
(\#Classes: 10) & \#Clusters & \textnormal{1.9 (1.0)} & \textnormal{1.8 (0.4)} & \textnormal{39.3 (5.1)} & \textnormal{1168.5 (401.3)} & \textnormal{20.1 (2.3)} & \textnormal{21.7 (3.2)} & \textnormal{32.5 (5.8)} & \textnormal{23.6 (4.0)} \\
\hline
STL10 & \#Nodes & \textnormal{2.7 (0.5)} & \textnormal{2.7 (0.5)} & \textnormal{47.1 (7.4)} & \textnormal{321.9 (534.0)} & \textnormal{14.6 (1.7)} & \textnormal{19.8 (2.9)} & \textnormal{37.7 (8.4)} & \textnormal{23.7 (5.3)} \\
(\#Classes: 10) & \#Clusters & \textnormal{2.7 (0.5)} & \textnormal{2.6 (0.6)} & \textnormal{15.0 (4.8)} & \textnormal{320.7 (534.7)} & \textnormal{5.5 (1.5)} & \textnormal{7.4 (2.3)} & \textnormal{10.1 (3.9)} & \textnormal{7.8 (2.8)} \\
\hline
Letter & \#Nodes & \textnormal{2.7 (0.4)} & \textnormal{2.2 (0.4)} & \textnormal{215.5 (21.1)} & \textnormal{244.1 (620.7)} & \textnormal{130.9 (11.4)} & \textnormal{98.8 (25.4)} & \textnormal{189.0 (20.4)} & \textnormal{145.4 (19.4)} \\
(\#Classes: 26) & \#Clusters & \textnormal{2.5 (0.9)} & \textnormal{2.0 (0.2)} & \textnormal{61.5 (9.1)} & \textnormal{243.5 (620.9)} & \textnormal{30.8 (6.3)} & \textnormal{24.4 (8.5)} & \textnormal{51.0 (9.0)} & \textnormal{41.5 (8.1)} \\
\hline
Shuttle & \#Nodes & \textnormal{2.1 (0.3)} & \textnormal{2.0 (0.0)} & \textnormal{182.4 (59.5)} & \textnormal{615.9 (406.3)} & \textnormal{118.9 (36.3)} & \textnormal{113.2 (42.7)} & \textnormal{131.6 (46.5)} & \textnormal{120.3 (44.6)} \\
(\#Classes: 7) & \#Clusters & \textnormal{1.9 (0.5)} & \textnormal{1.8 (0.4)} & \textnormal{29.9 (8.9)} & \textnormal{521.8 (440.4)} & \textnormal{15.1 (6.7)} & \textnormal{18.5 (6.8)} & \textnormal{20.3 (7.5)} & \textnormal{18.9 (7.3)} \\
\hline
Skin & \#Nodes & \textnormal{2.4 (0.5)} & \textnormal{2.6 (0.5)} & \textnormal{38.6 (12.0)} & \textnormal{877.4 (145.0)} & \textnormal{154.8 (30.9)} & \textnormal{119.8 (1.5)} & \textnormal{42.2 (11.5)} & \textnormal{41.2 (11.5)} \\
(\#Classes: 2) & \#Clusters & \textnormal{1.8 (1.0)} & \textnormal{2.0 (0.0)} & \textnormal{17.4 (2.0)} & \textnormal{542.6 (10.5)} & \textnormal{25.6 (4.5)} & \textnormal{24.4 (0.5)} & \textnormal{17.0 (2.5)} & \textnormal{16.4 (2.0)} \\

\hline\hline
\multicolumn{2}{c|}{Average Cluster Error} & \textnormal{0.6 (0.3)} & \textnormal{0.6 (0.3)} & \textnormal{2.3 (2.2)} & \textnormal{86.2 (129.4)} & \textnormal{1.0 (2.3)} & \textnormal{1.6 (2.6)} & \textnormal{1.7 (1.7)} & \textnormal{1.7 (2.2)} \\
\hline\hline

\end{tabular}}
\\
\vspace{1mm}\footnotesize
\hspace*{4.5mm}
\begin{minipage}{\linewidth}
The values in parentheses indicate the standard deviation. \\
\end{minipage}
\end{table*}

% History of $\Lambda$ and $V_{\text{threshold}}$ for IDAT in the nonstationary setting.
\begin{figure*}[htbp]
  \centering
  \subfloat[Iris]{%
    \includegraphics[width=0.22\linewidth]{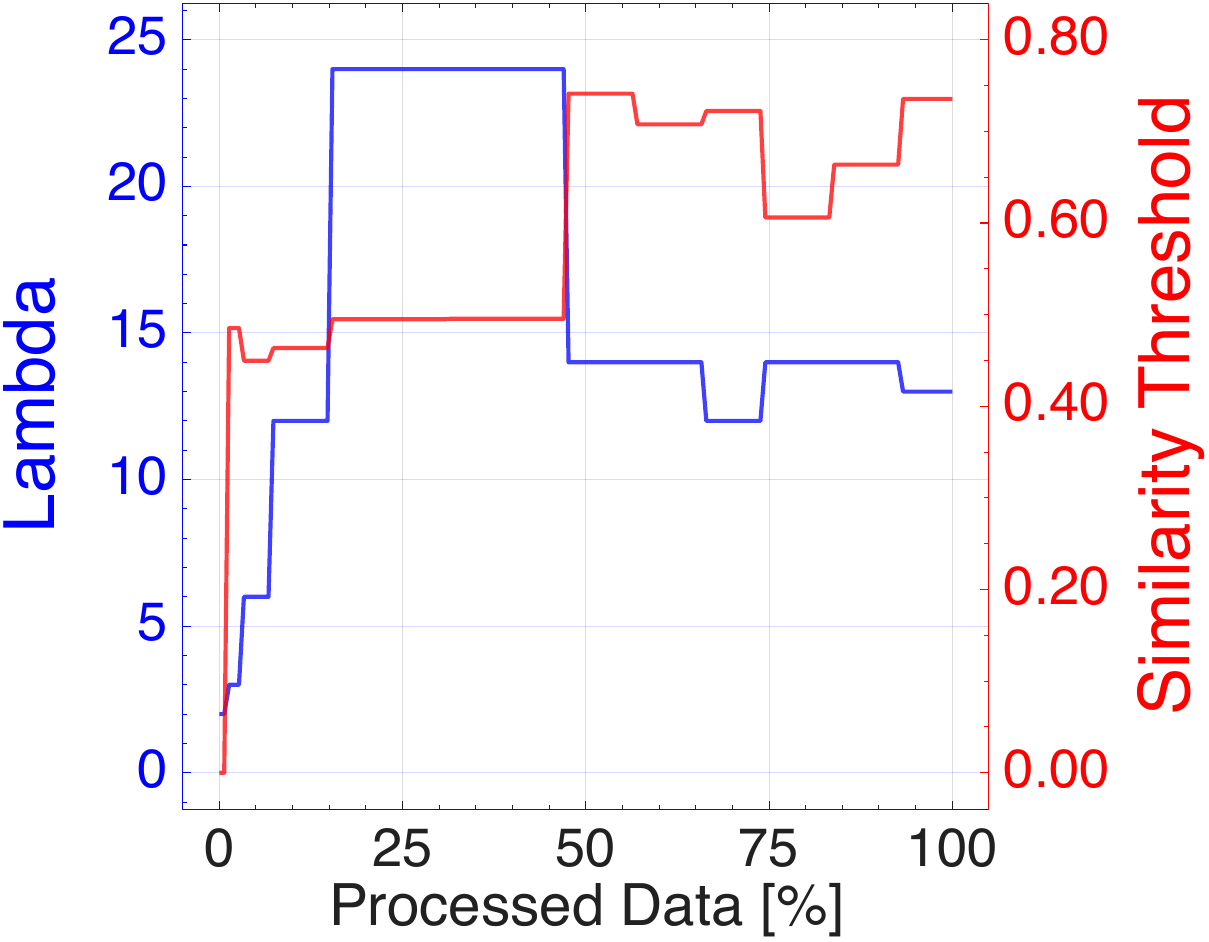}
  }\hfill
  \subfloat[Seeds]{%
    \includegraphics[width=0.22\linewidth]{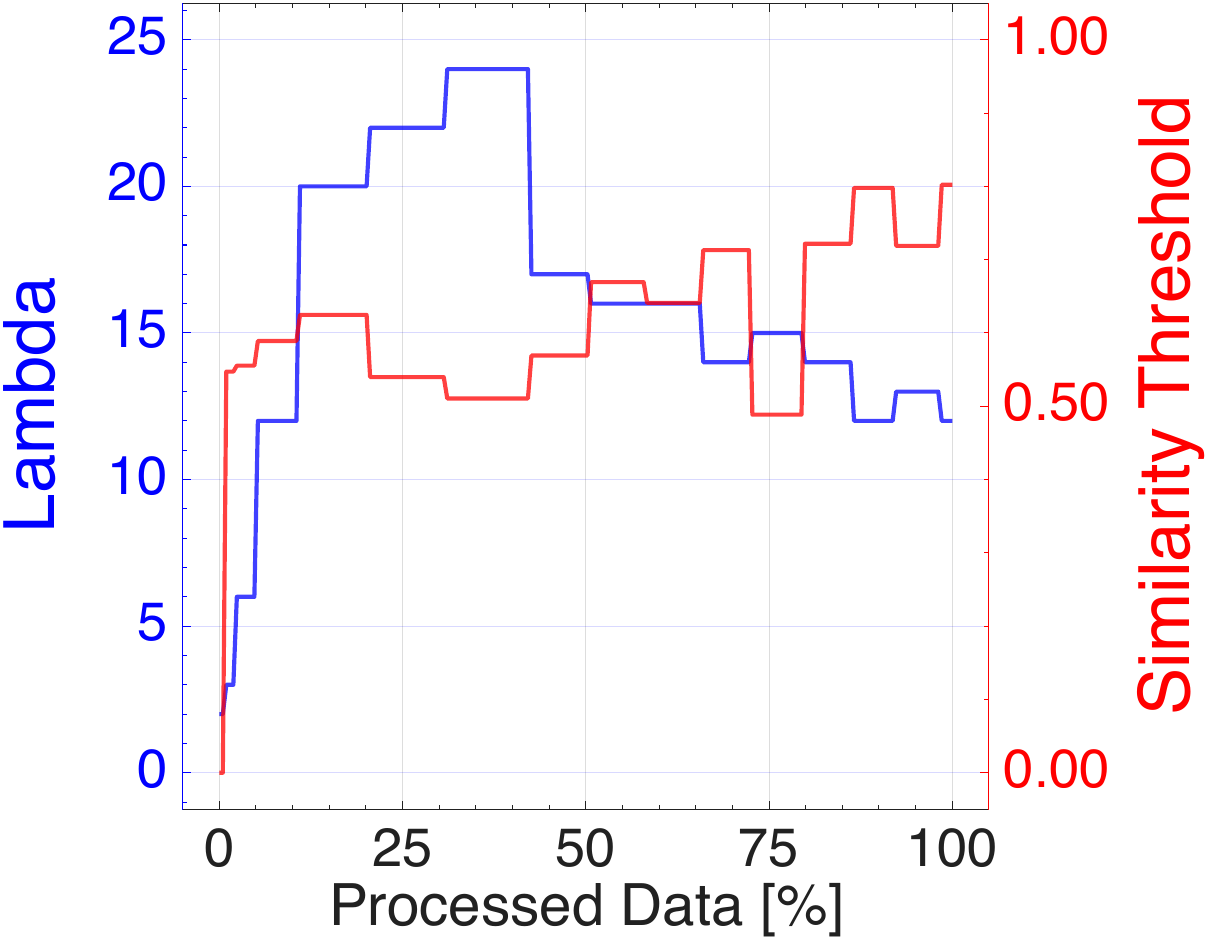}
  }\hfill
  \subfloat[Dermatology]{%
    \includegraphics[width=0.22\linewidth]{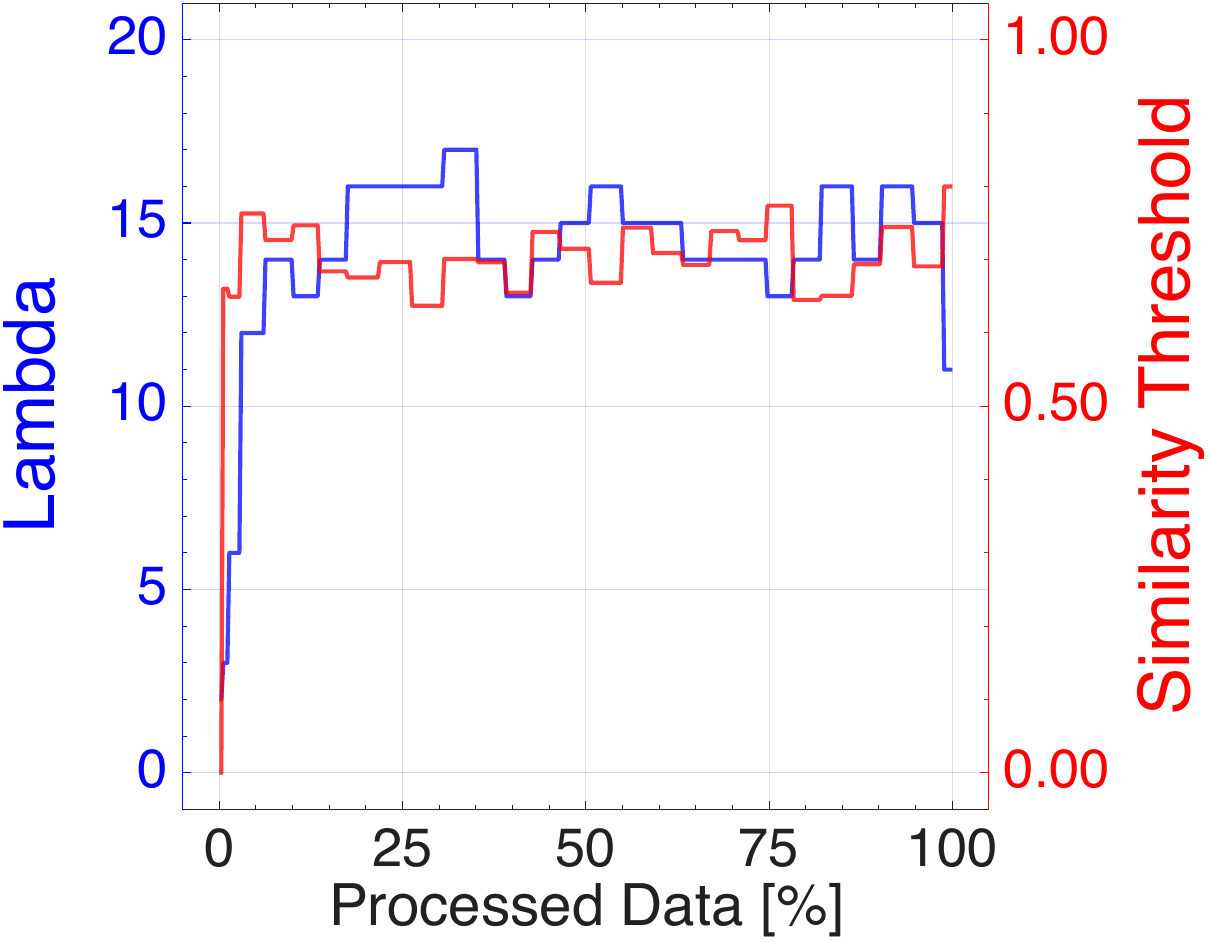}
  }\hfill
  \subfloat[Pima]{%
    \includegraphics[width=0.22\linewidth]{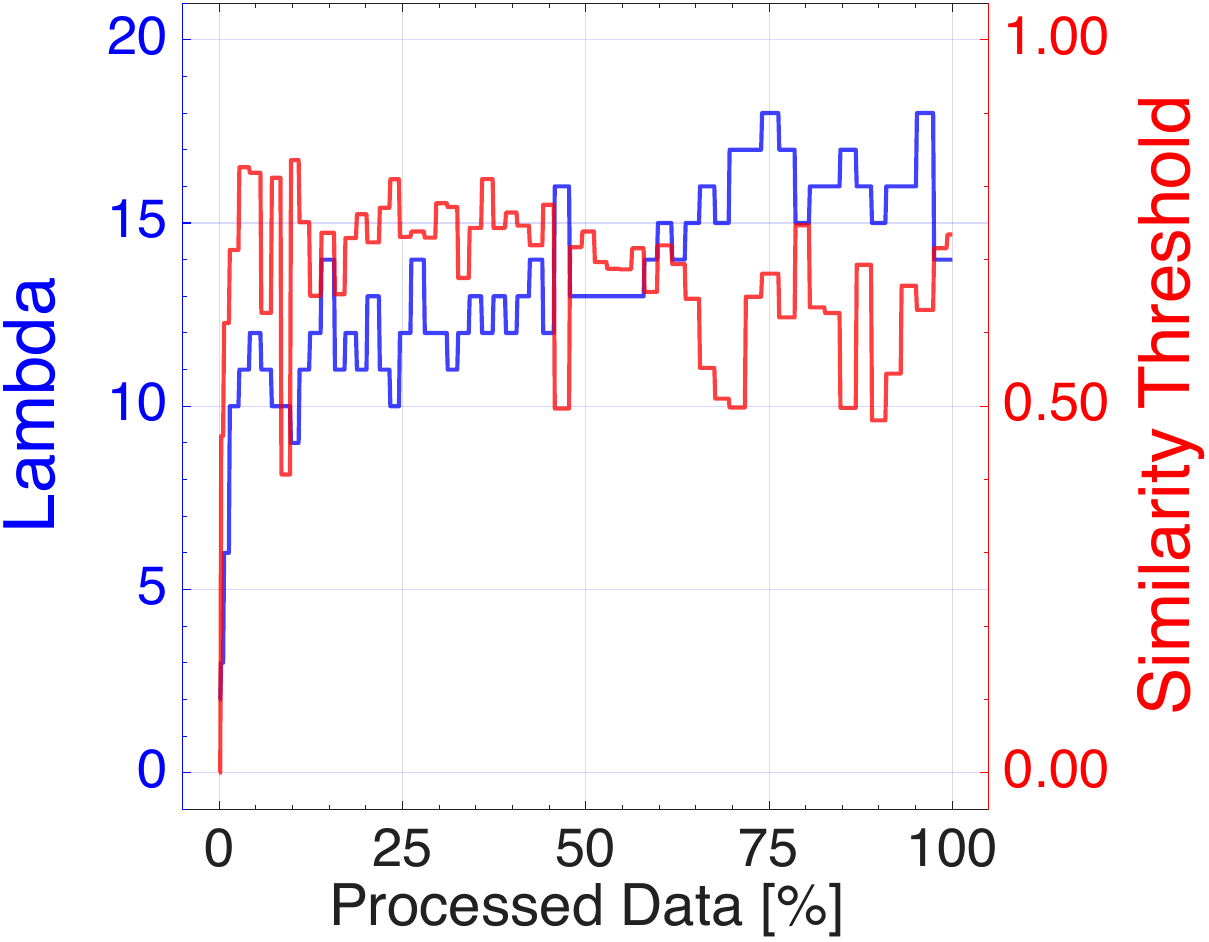}
  }\\
  \subfloat[Mice Protein]{%
    \includegraphics[width=0.22\linewidth]{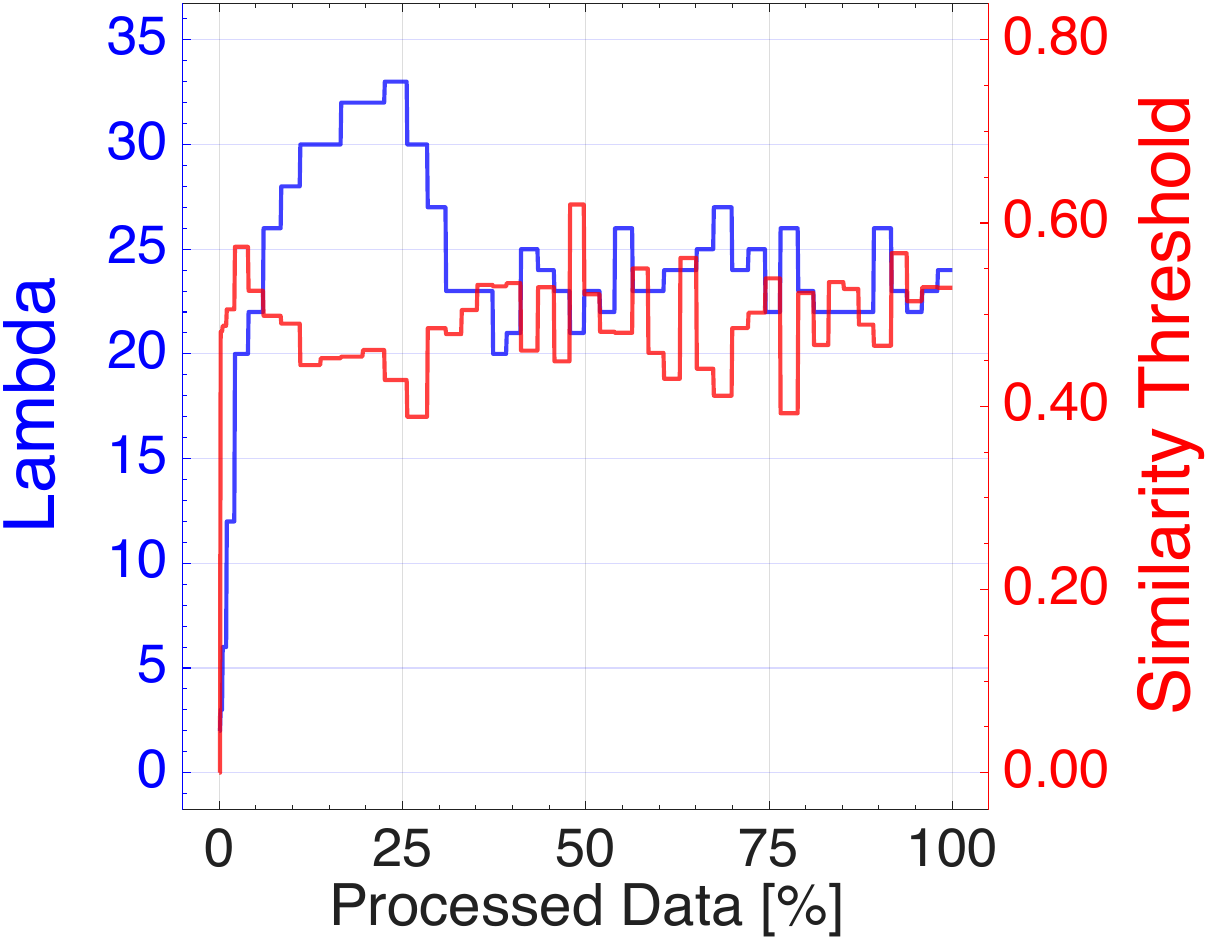}
  }\hfill
  \subfloat[Binalpha]{%
    \includegraphics[width=0.22\linewidth]{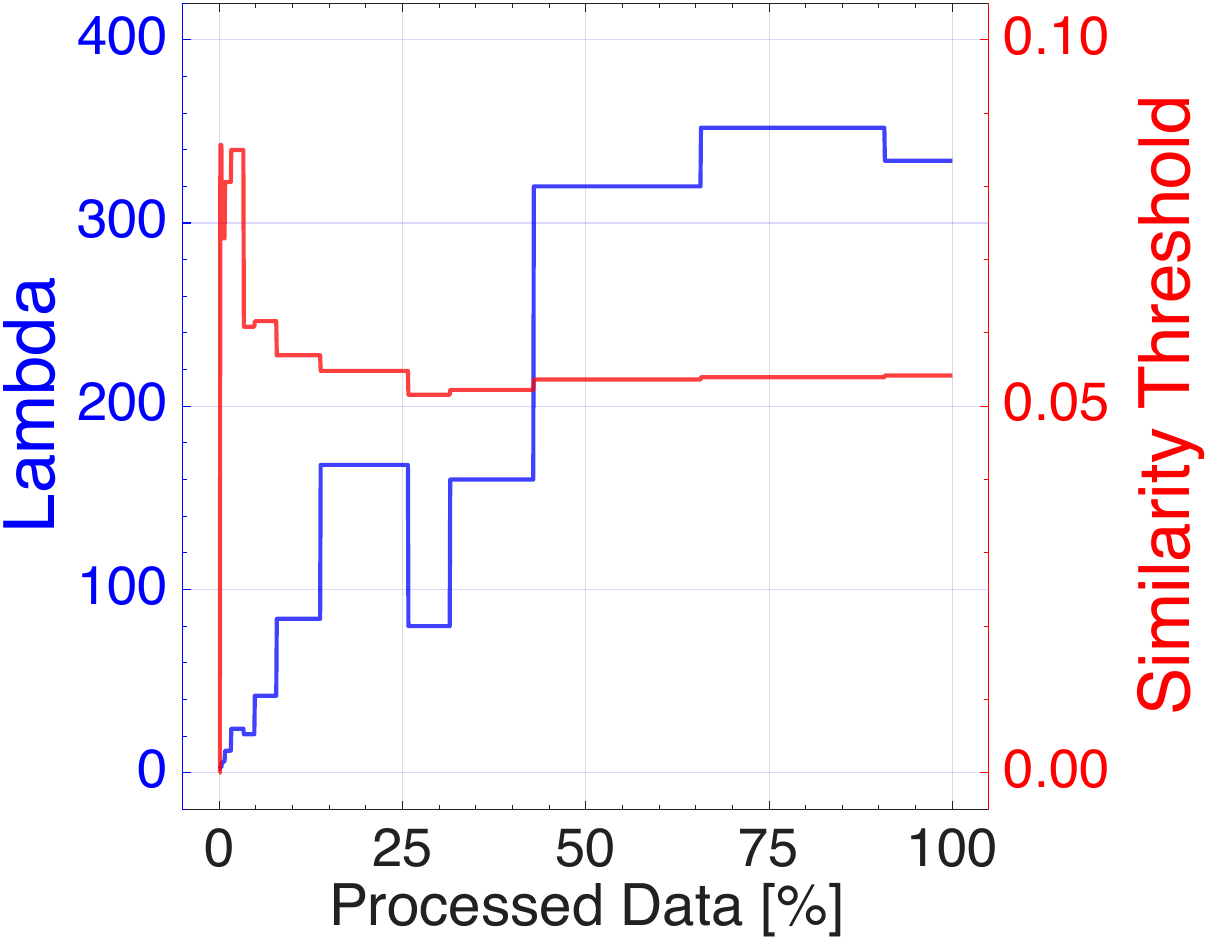}
  }\hfill
  \subfloat[Yeast]{%
    \includegraphics[width=0.22\linewidth]{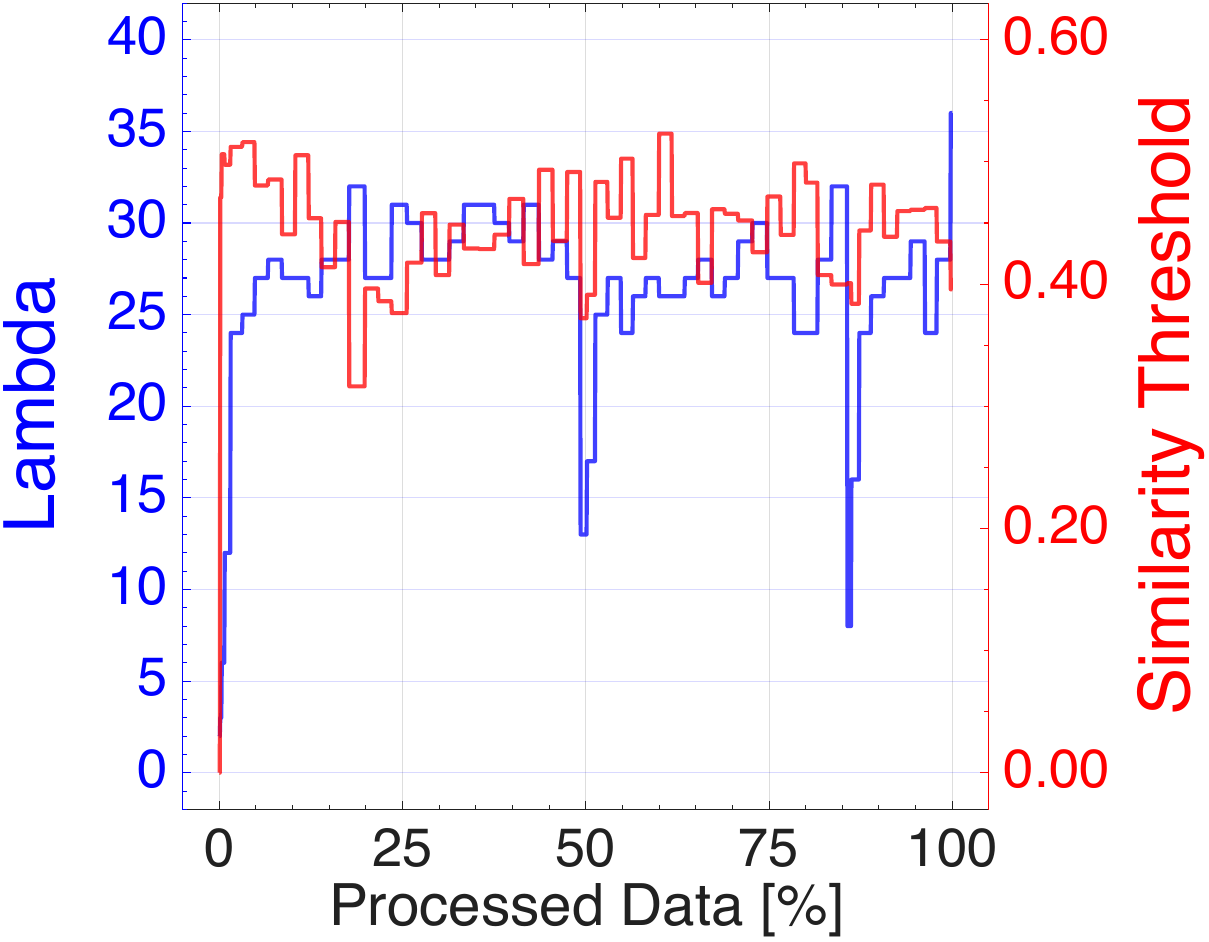}
  }\hfill
  \subfloat[Semeion]{%
    \includegraphics[width=0.22\linewidth]{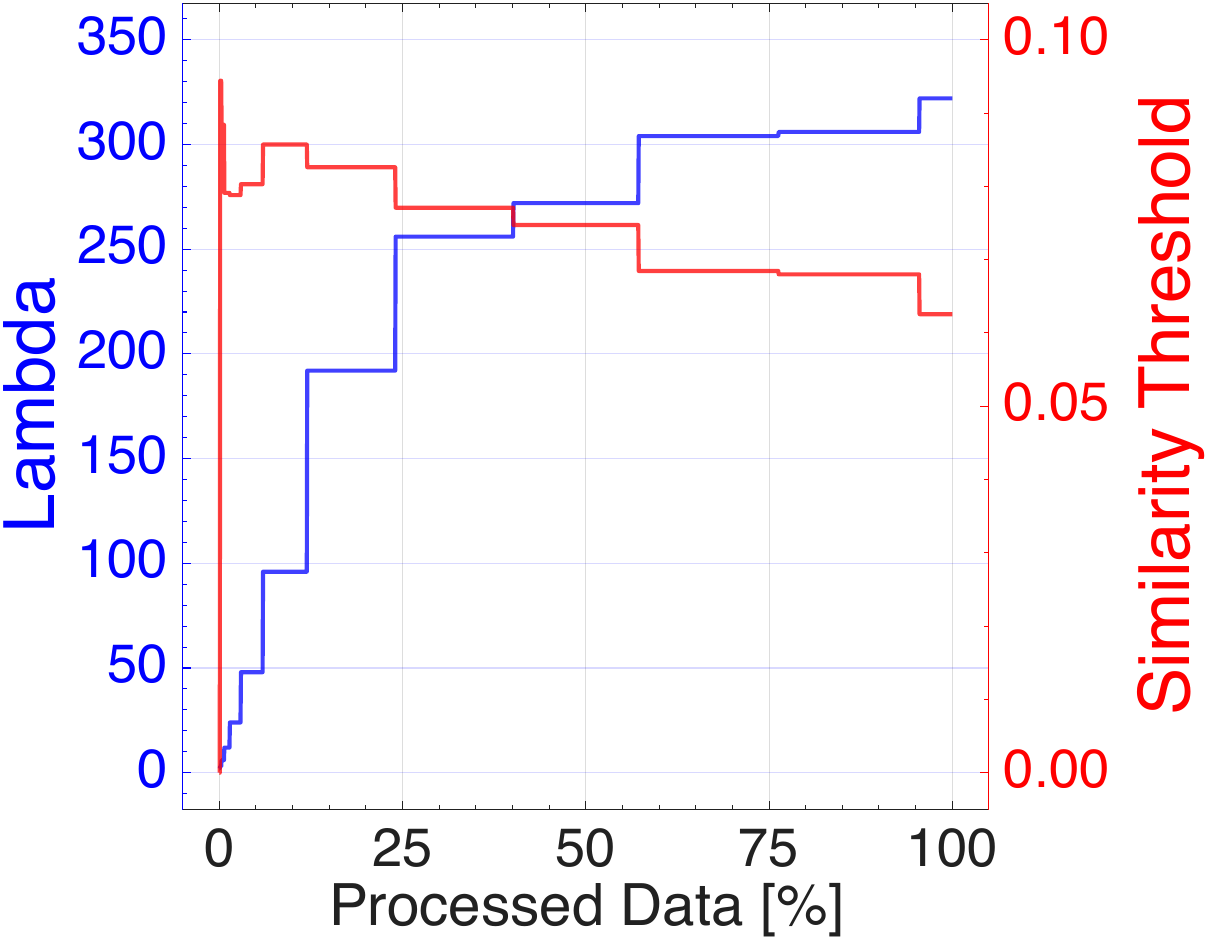}
  }\\
  \subfloat[MSRA25]{%
    \includegraphics[width=0.22\linewidth]{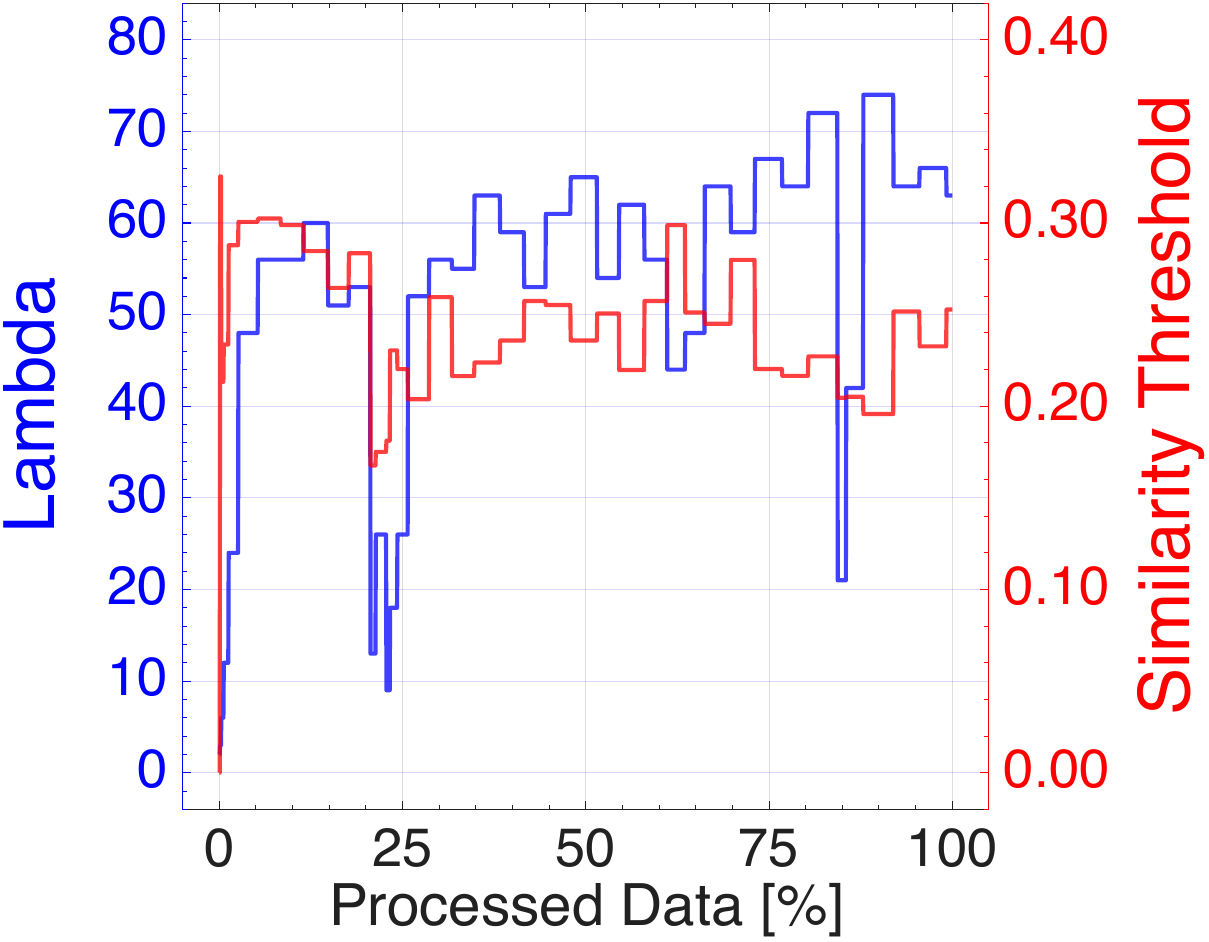}
  }\hfill
  \subfloat[Image Segmentation]{%
    \includegraphics[width=0.22\linewidth]{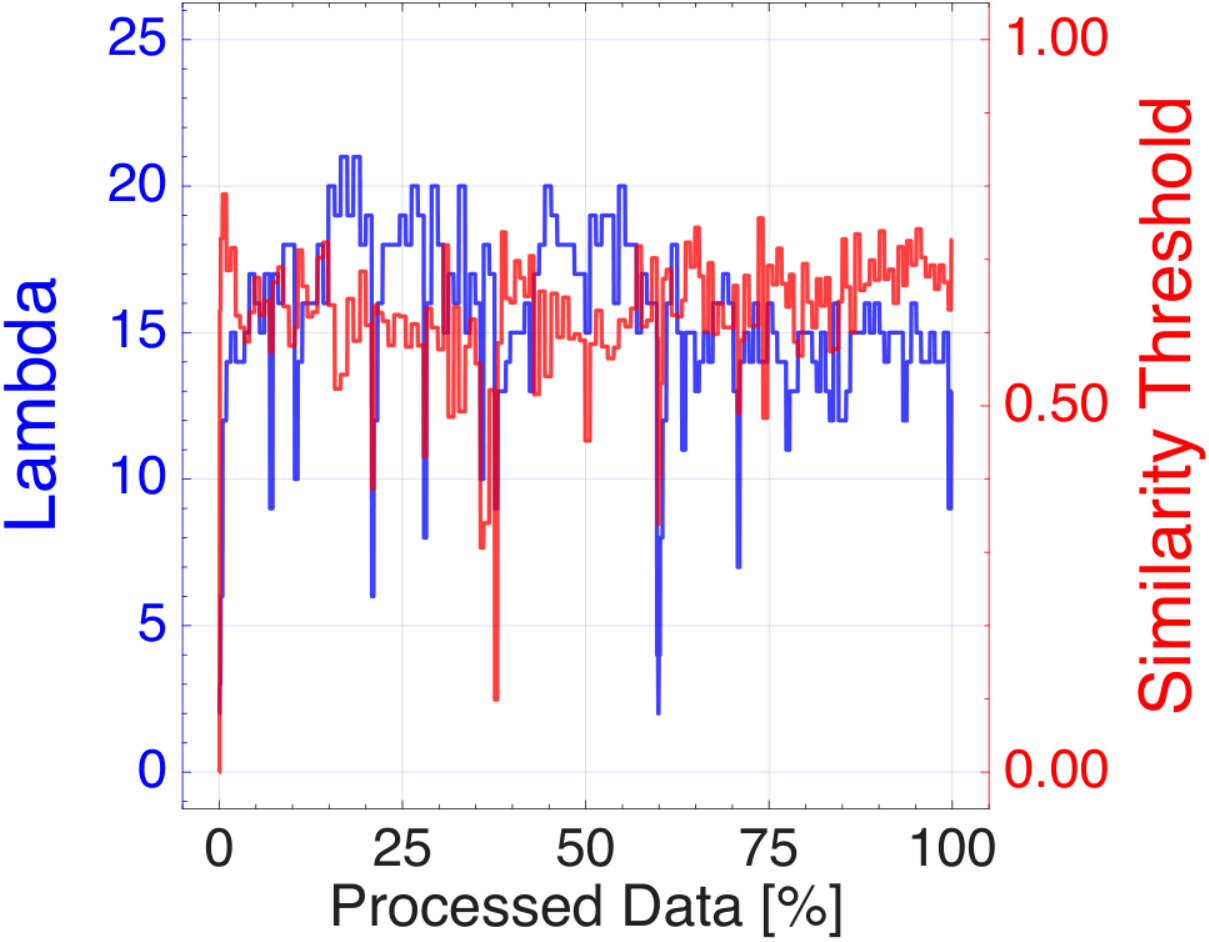}
  }\hfill
  \subfloat[Rice]{%
    \includegraphics[width=0.22\linewidth]{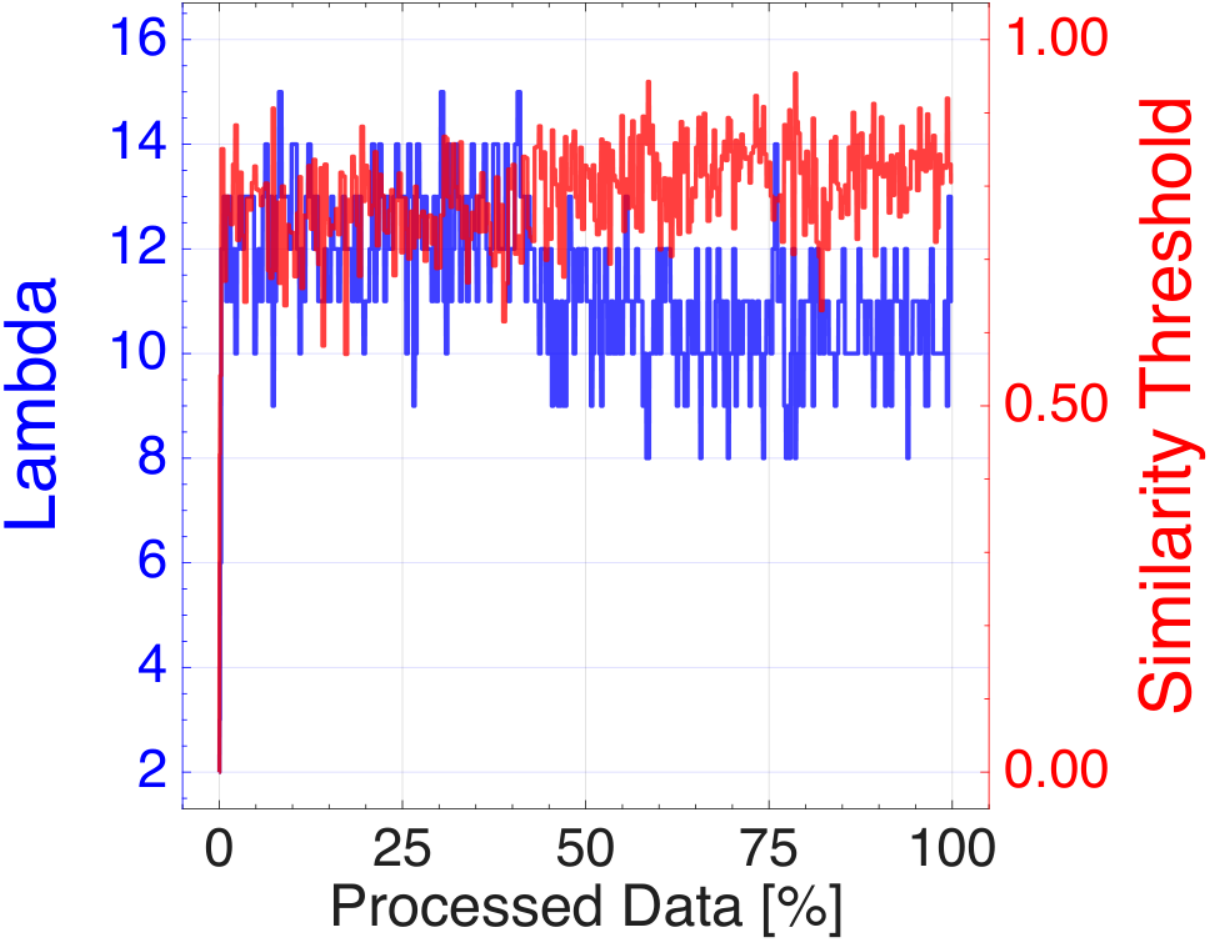}
  }\hfill
  \subfloat[TUANDROMD]{%
    \includegraphics[width=0.22\linewidth]{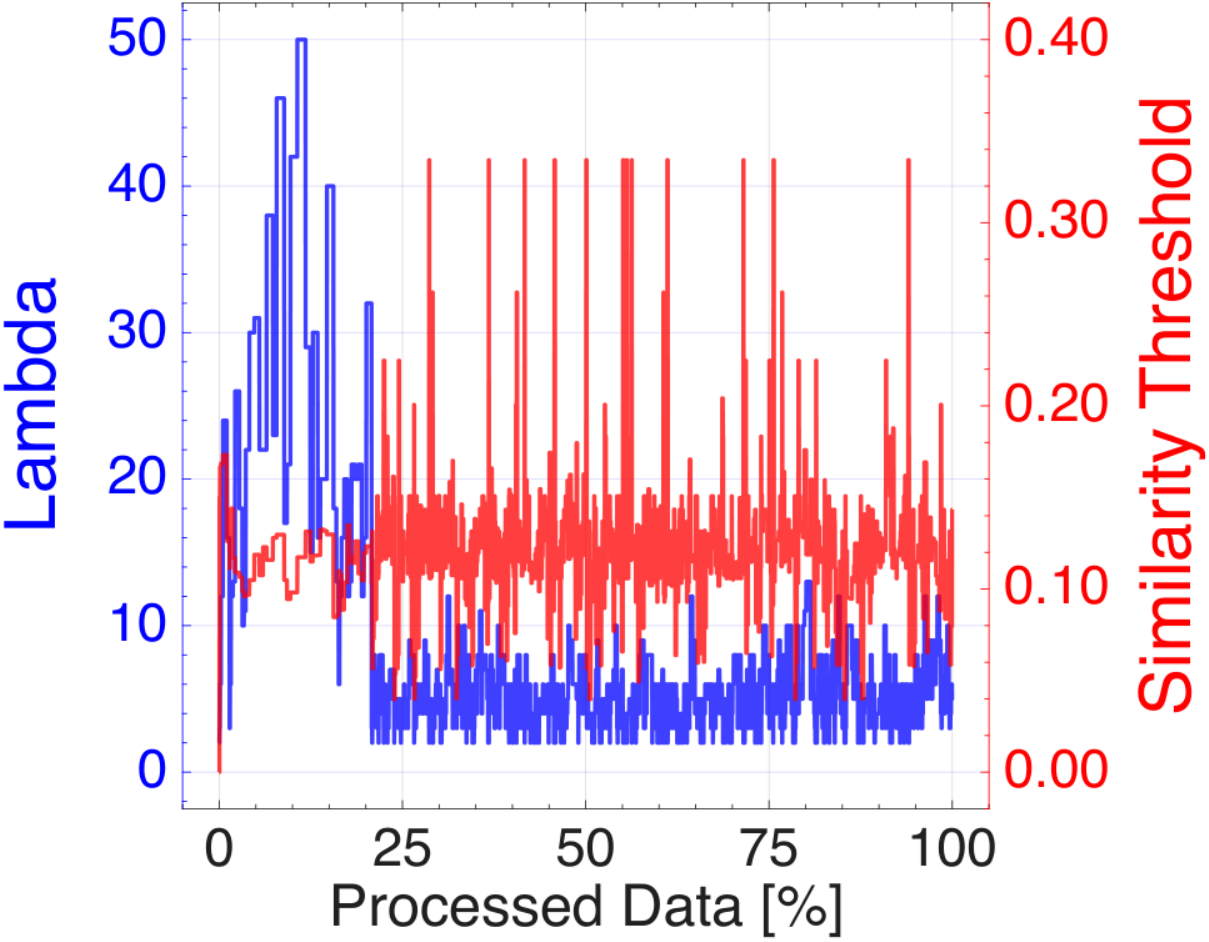}
  }\\
  \subfloat[Phoneme]{%
    \includegraphics[width=0.22\linewidth]{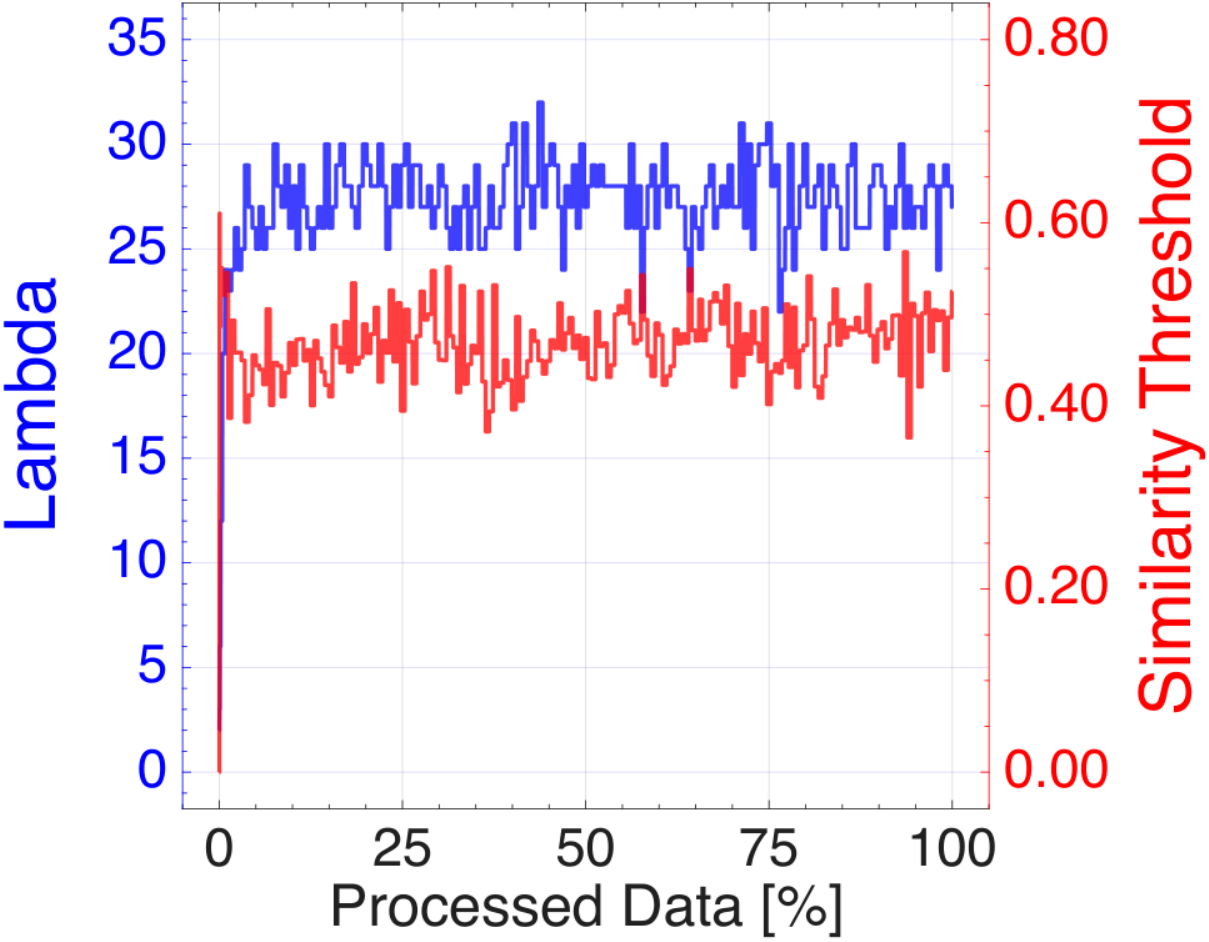}
  }\hfill
  \subfloat[Texture]{%
    \includegraphics[width=0.22\linewidth]{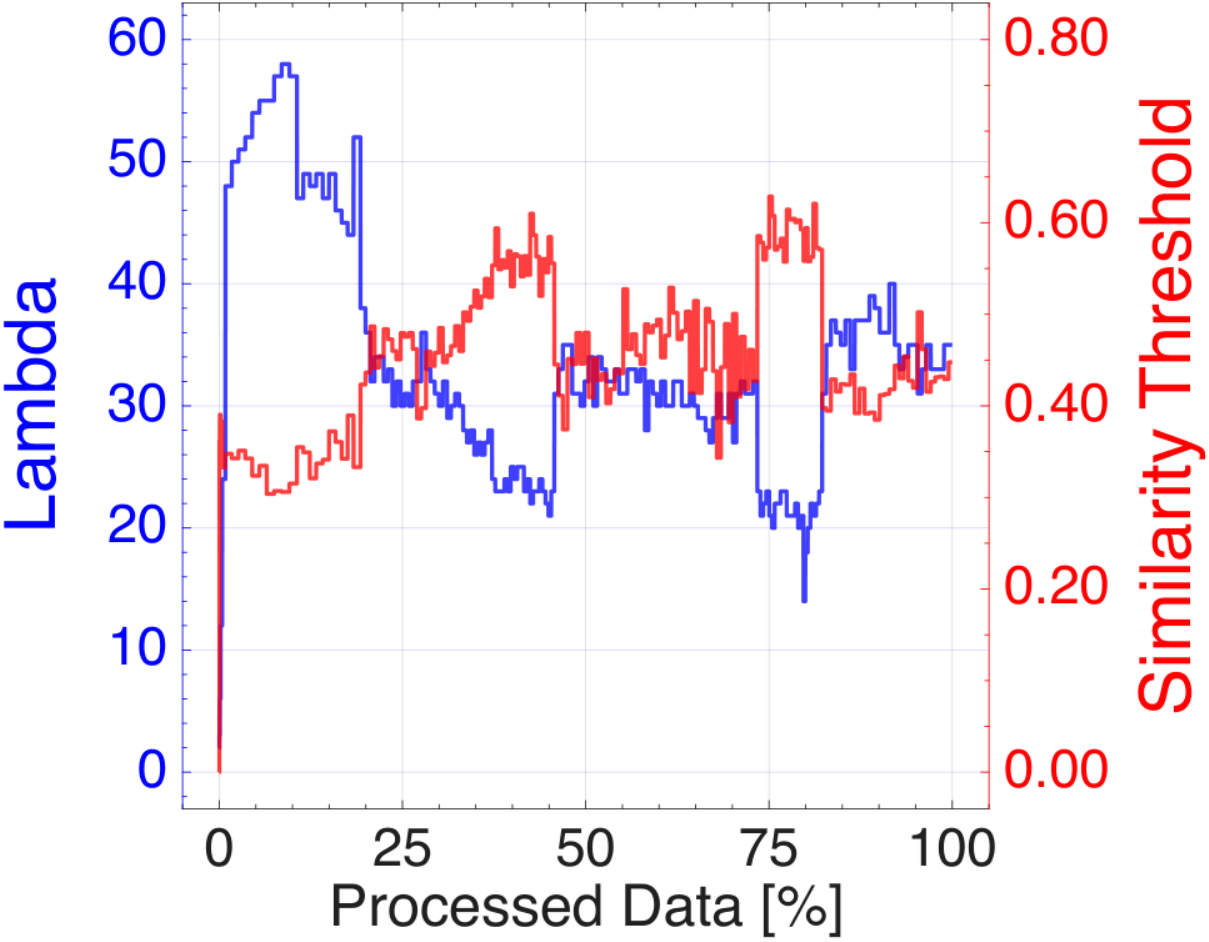}
  }\hfill
  \subfloat[OptDigits]{%
    \includegraphics[width=0.22\linewidth]{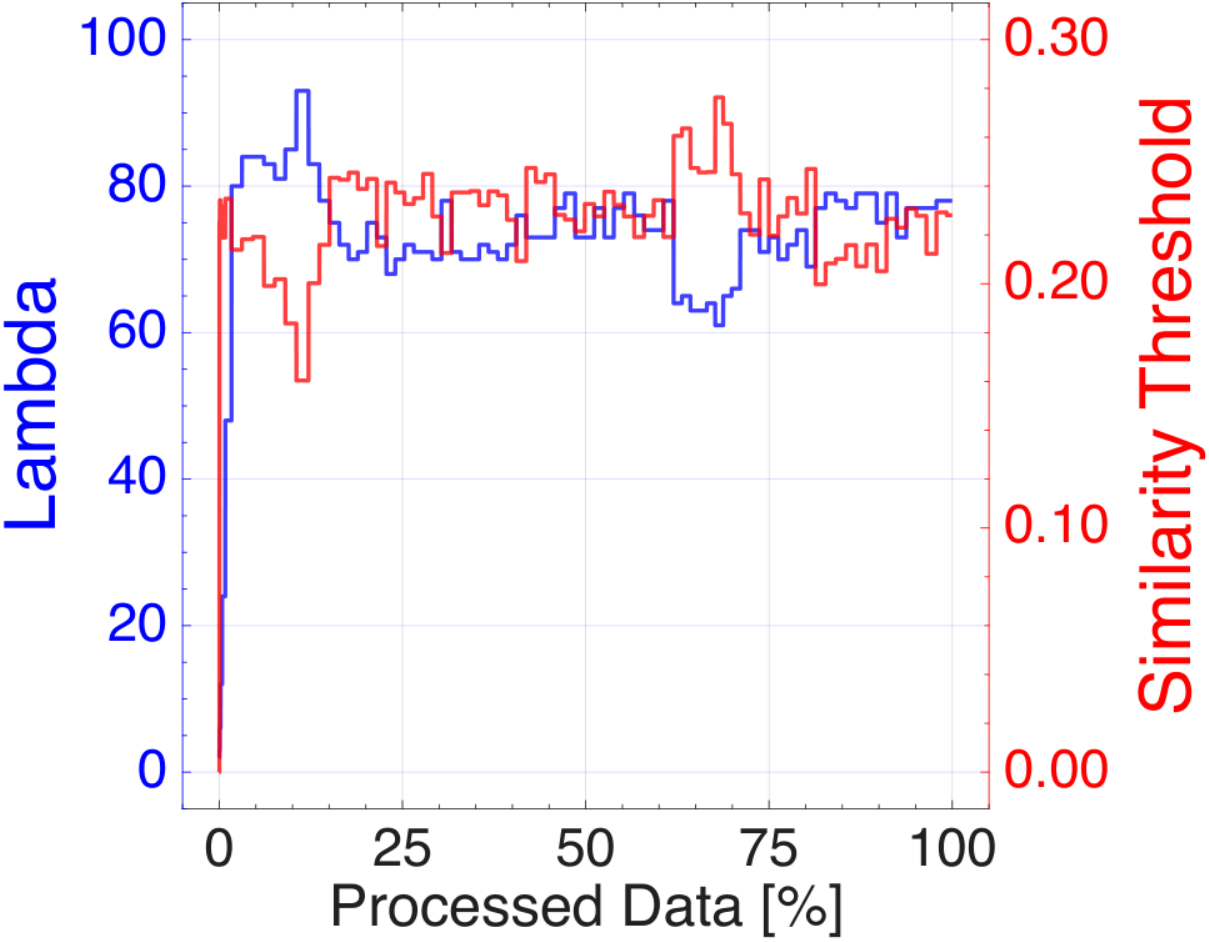}
  }\hfill
  \subfloat[Statlog]{%
    \includegraphics[width=0.22\linewidth]{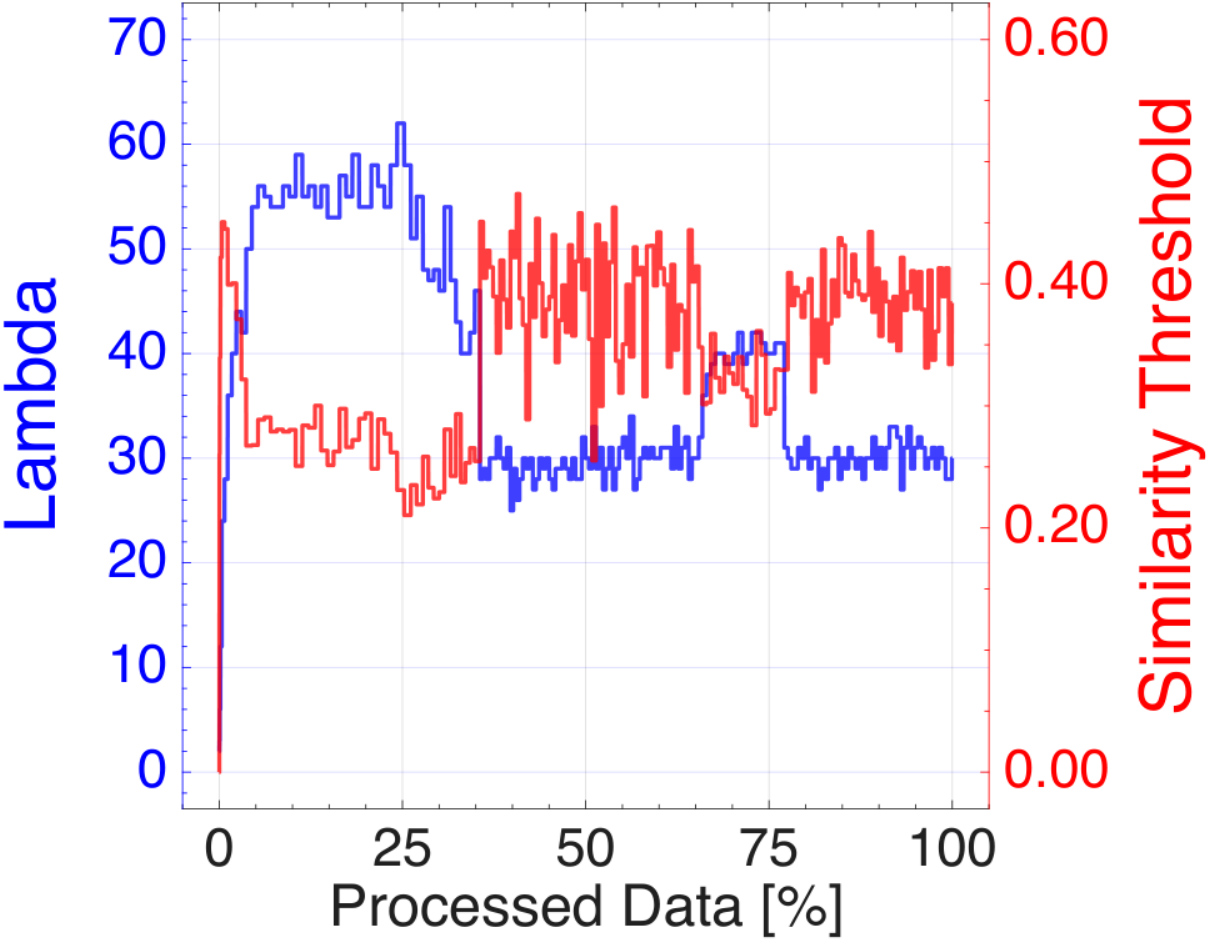}
  }\\
  \subfloat[Anuran Calls]{%
    \includegraphics[width=0.22\linewidth]{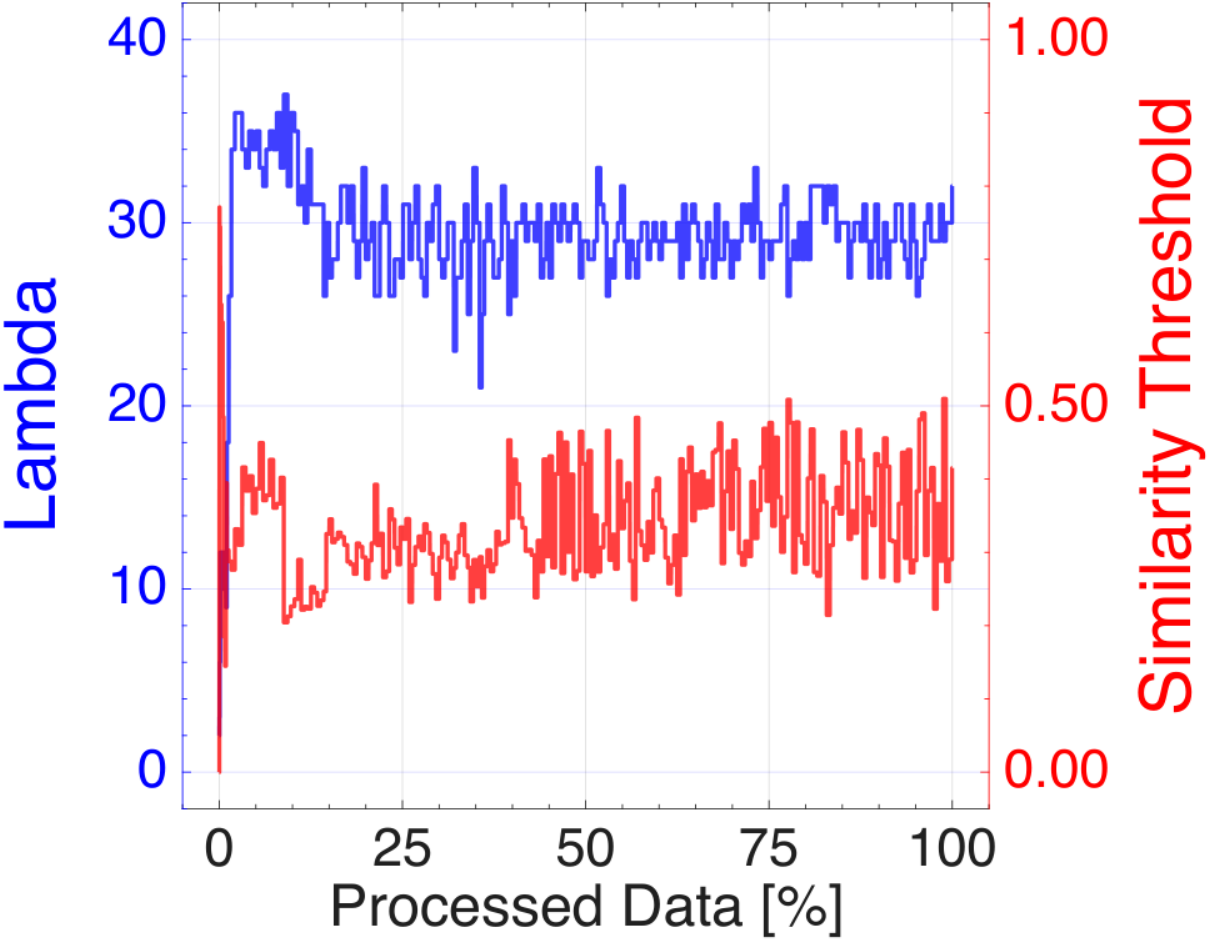}
  }\hfill
  \subfloat[Isolet]{%
    \includegraphics[width=0.22\linewidth]{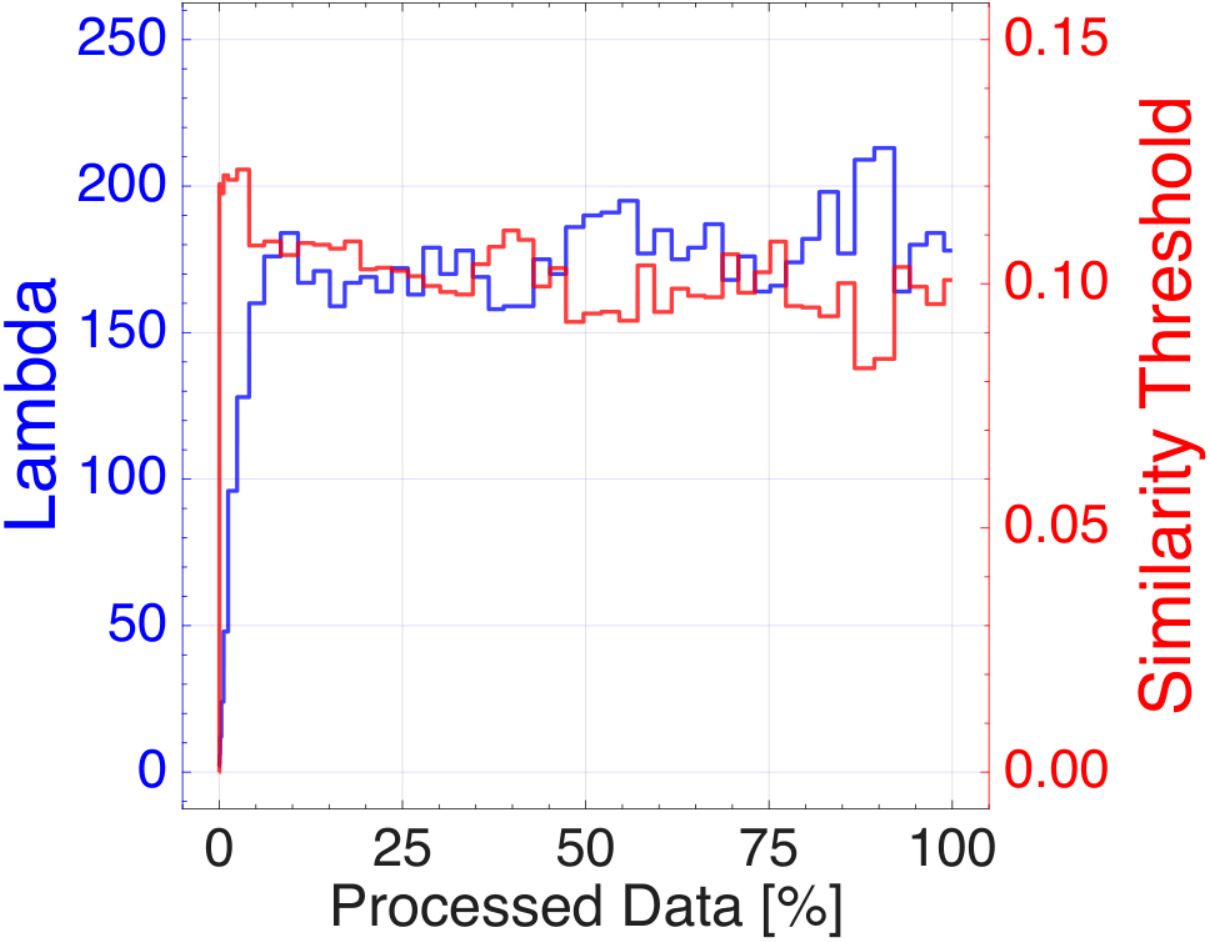}
  }\hfill
  \subfloat[MNIST10K]{%
    \includegraphics[width=0.22\linewidth]{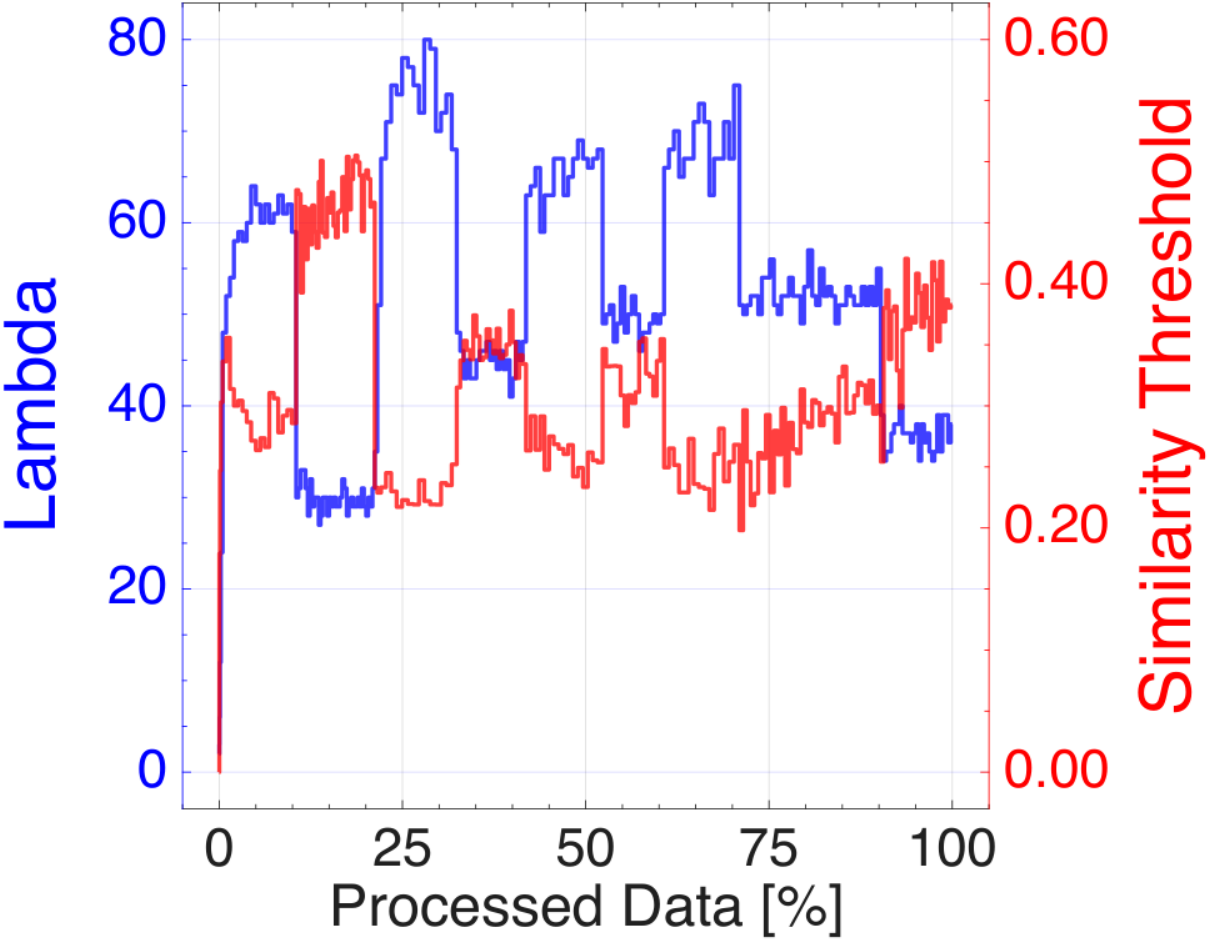}
  }\hfill
  \subfloat[PenBased]{%
    \includegraphics[width=0.22\linewidth]{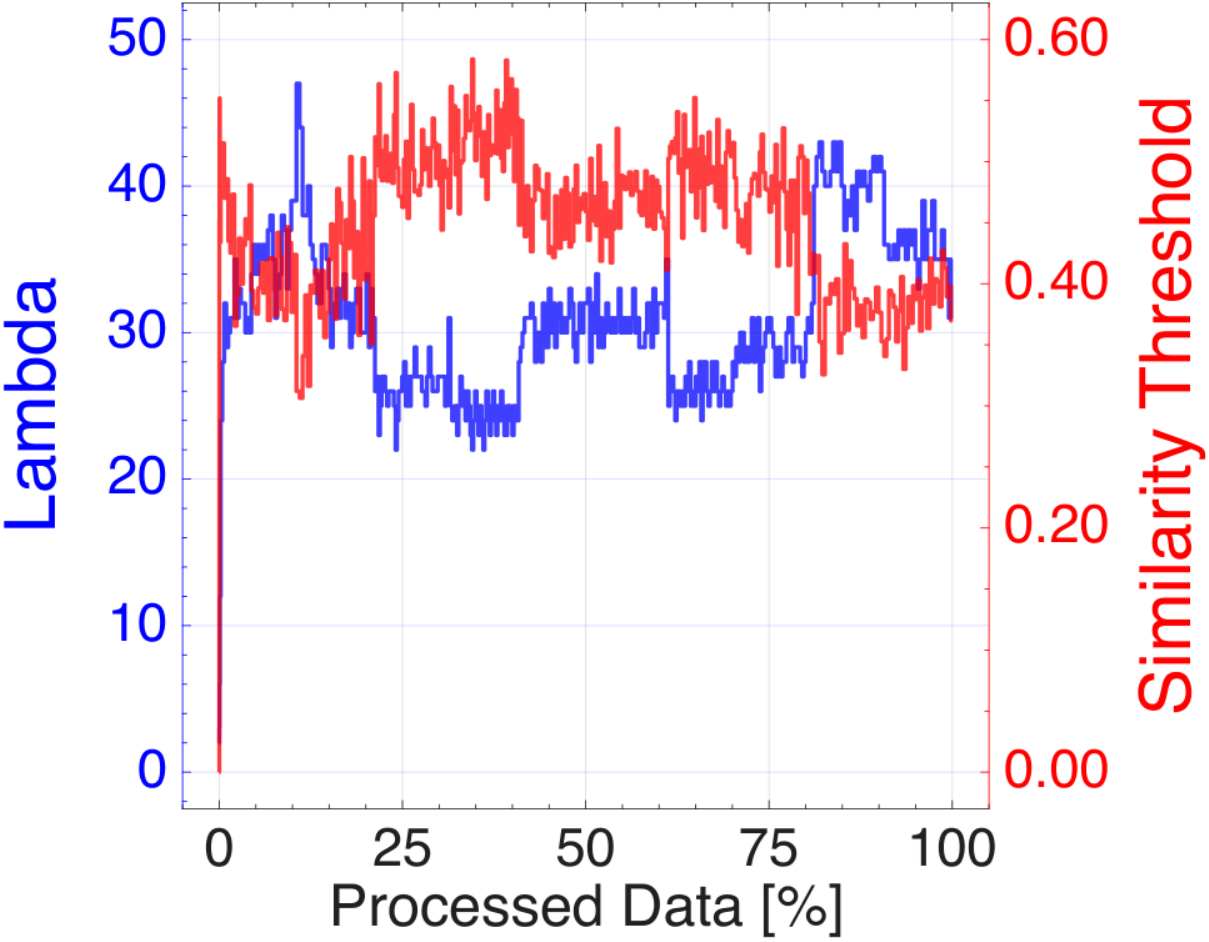}
  }\\
  \subfloat[STL10]{%
    \includegraphics[width=0.22\linewidth]{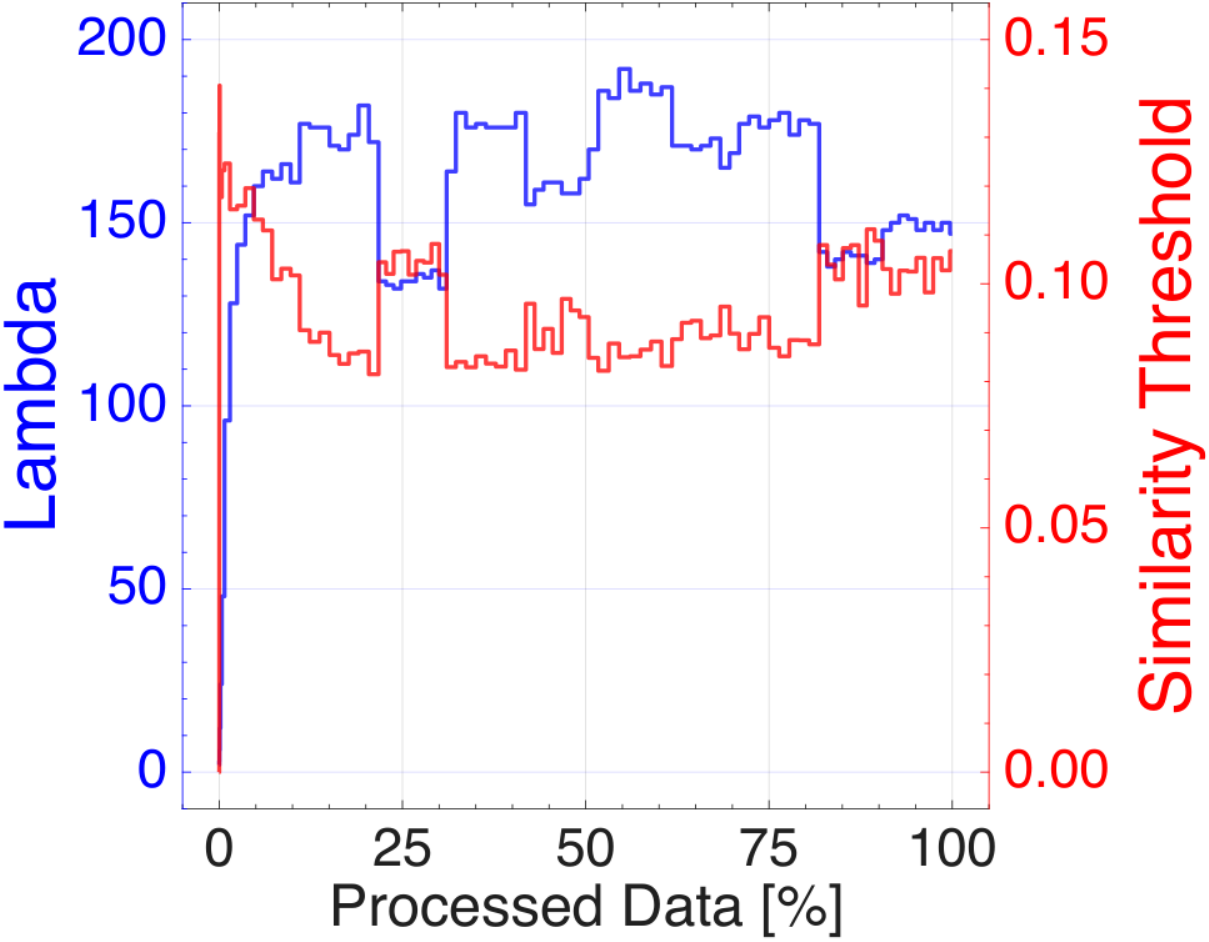}
  }\hfill
  \subfloat[Letter]{%
    \includegraphics[width=0.22\linewidth]{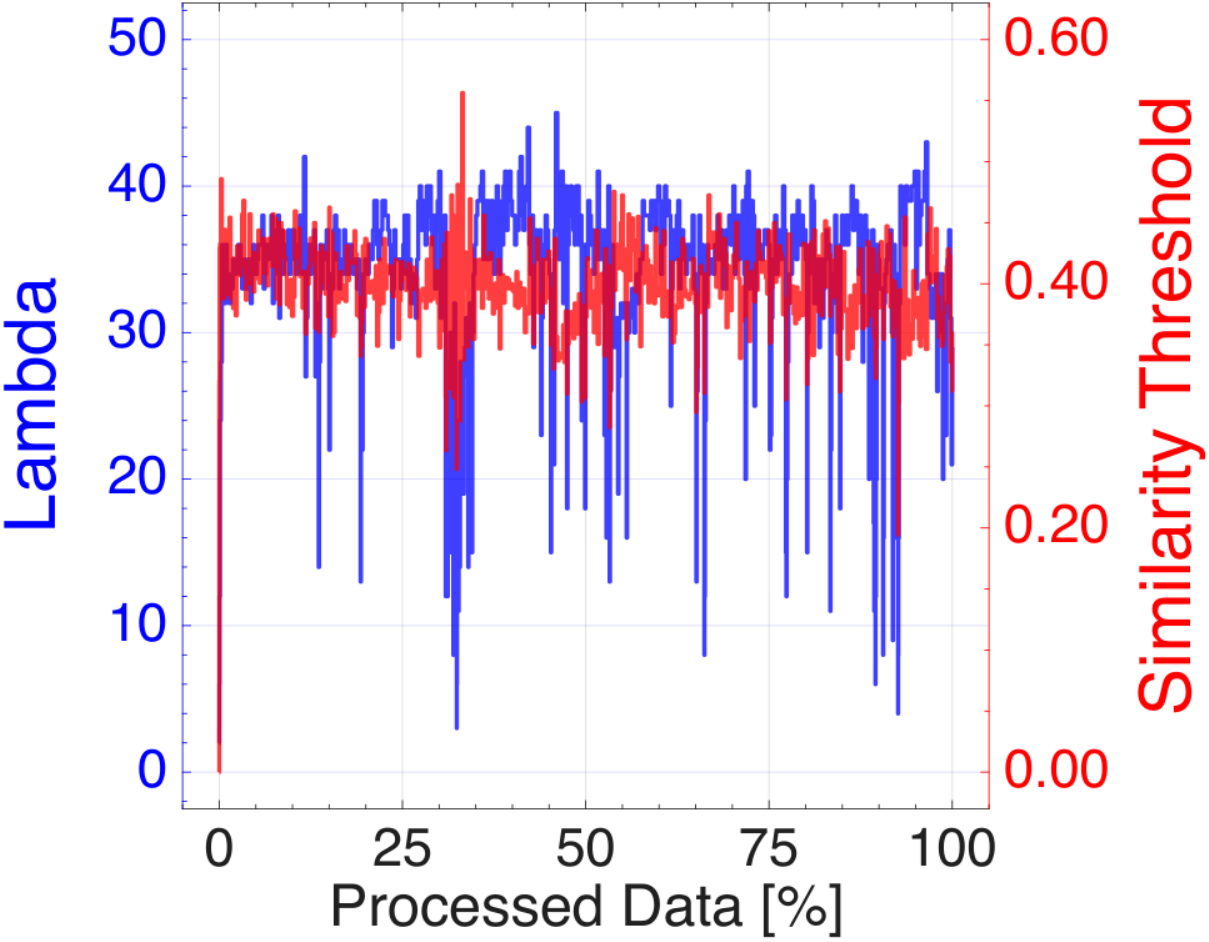}
  }\hfill
  \subfloat[Shuttle]{%
    \includegraphics[width=0.22\linewidth]{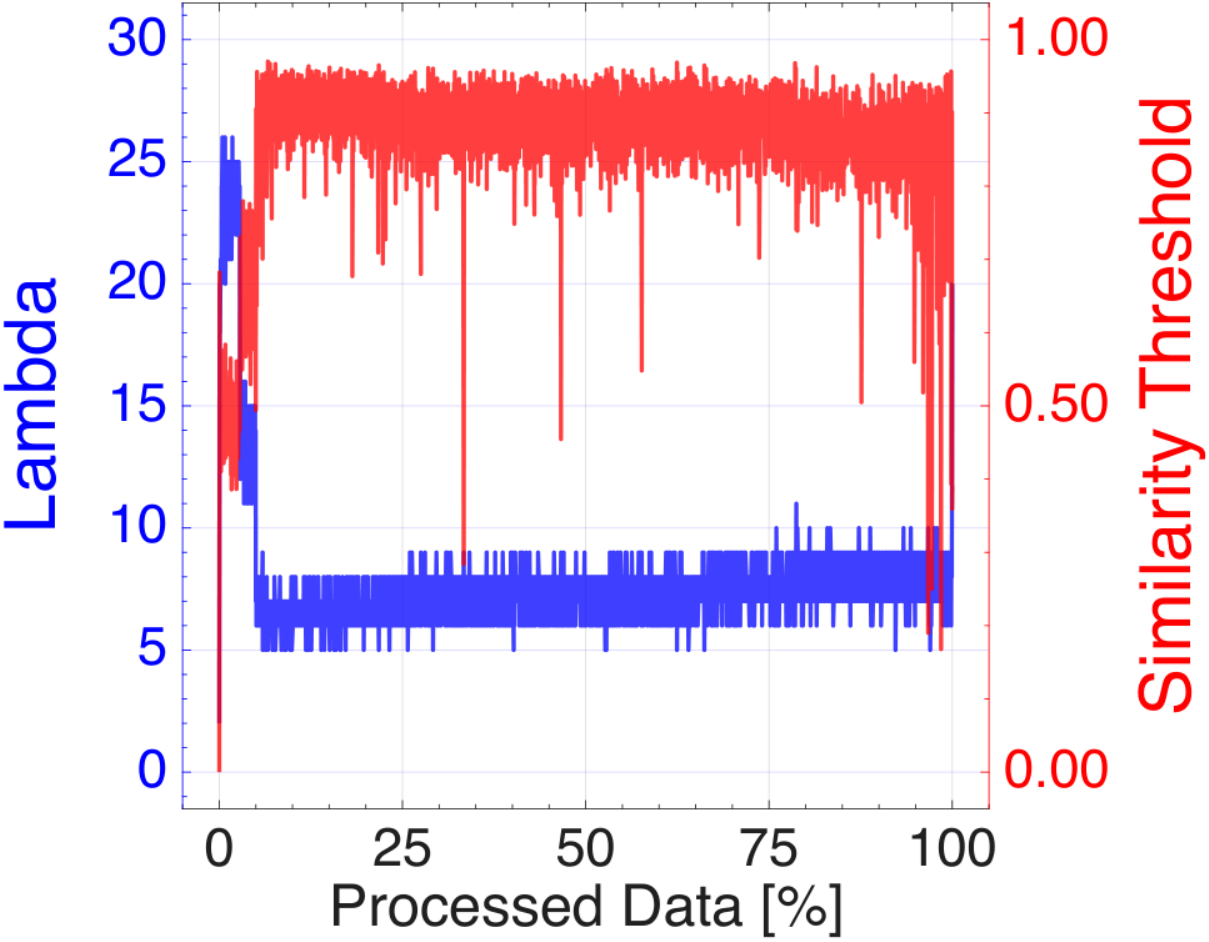}
  }\hfill
  \subfloat[Skin]{%
    \includegraphics[width=0.22\linewidth]{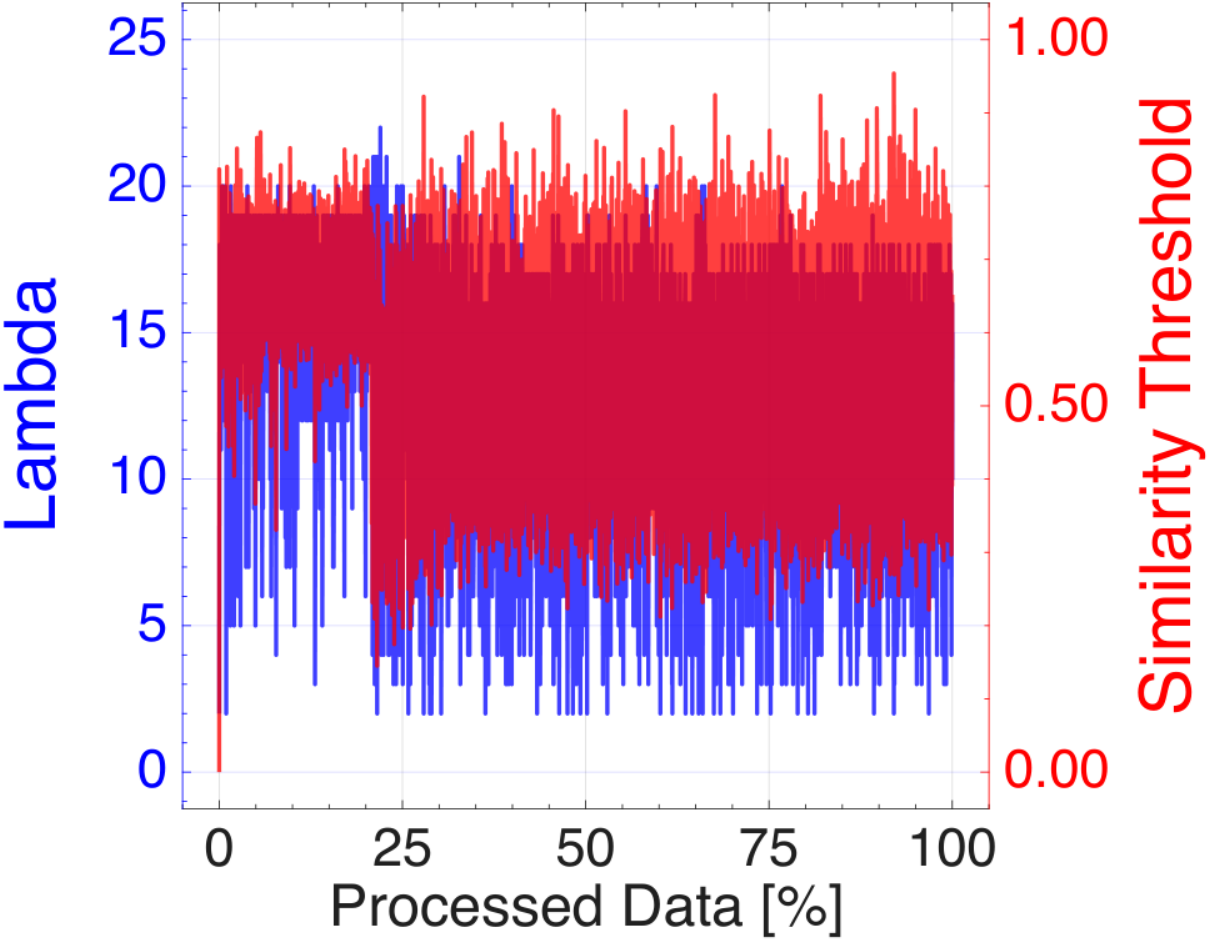}
  }
  \caption{Histories of $\Lambda$ and $V_{\text{threshold}}$ for IDAT in the nonstationary setting ($\Lambda_{\text{init}} = 2$).}
  \label{fig:ablation_lambda_history_idat_nonstationary}
\end{figure*}

% History of $\Lambda$ and $V_{\text{threshold}}$ for IDAT in the nonstationary setting.

\begin{figure*}[htbp]
  \centering
  \subfloat[Iris]{%
    \includegraphics[width=0.22\linewidth]{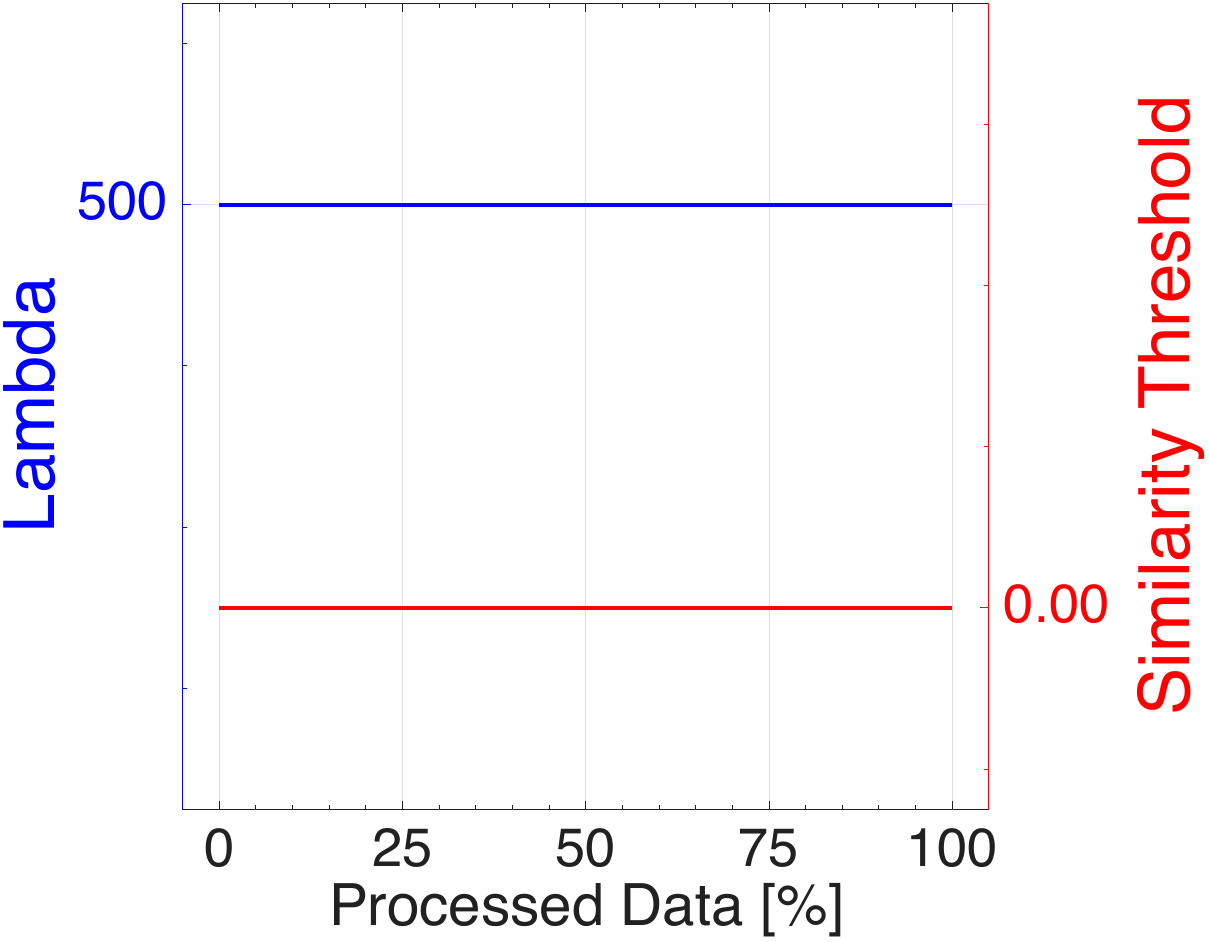}
  }\hfill
  \subfloat[Seeds]{%
    \includegraphics[width=0.22\linewidth]{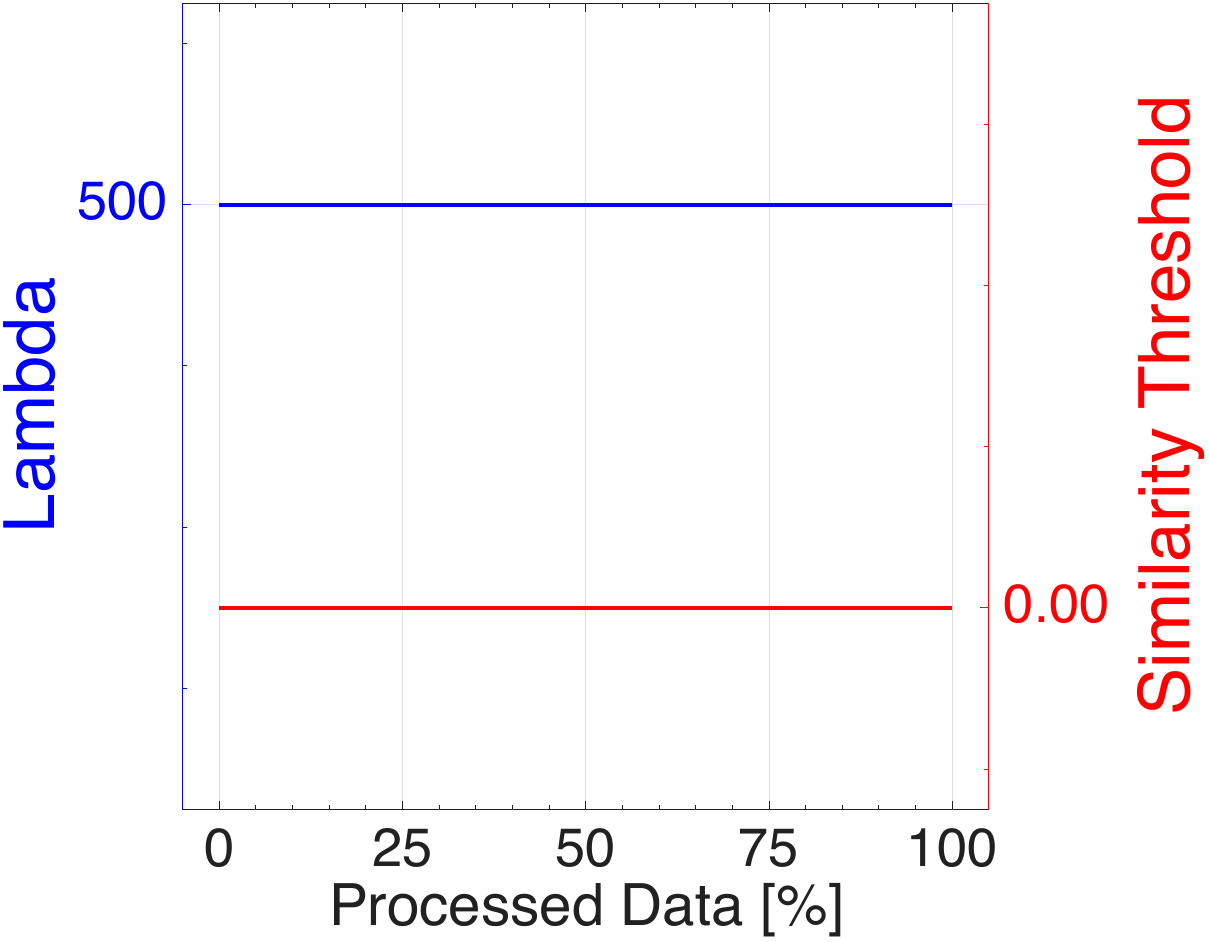}
  }\hfill
  \subfloat[Dermatology]{%
    \includegraphics[width=0.22\linewidth]{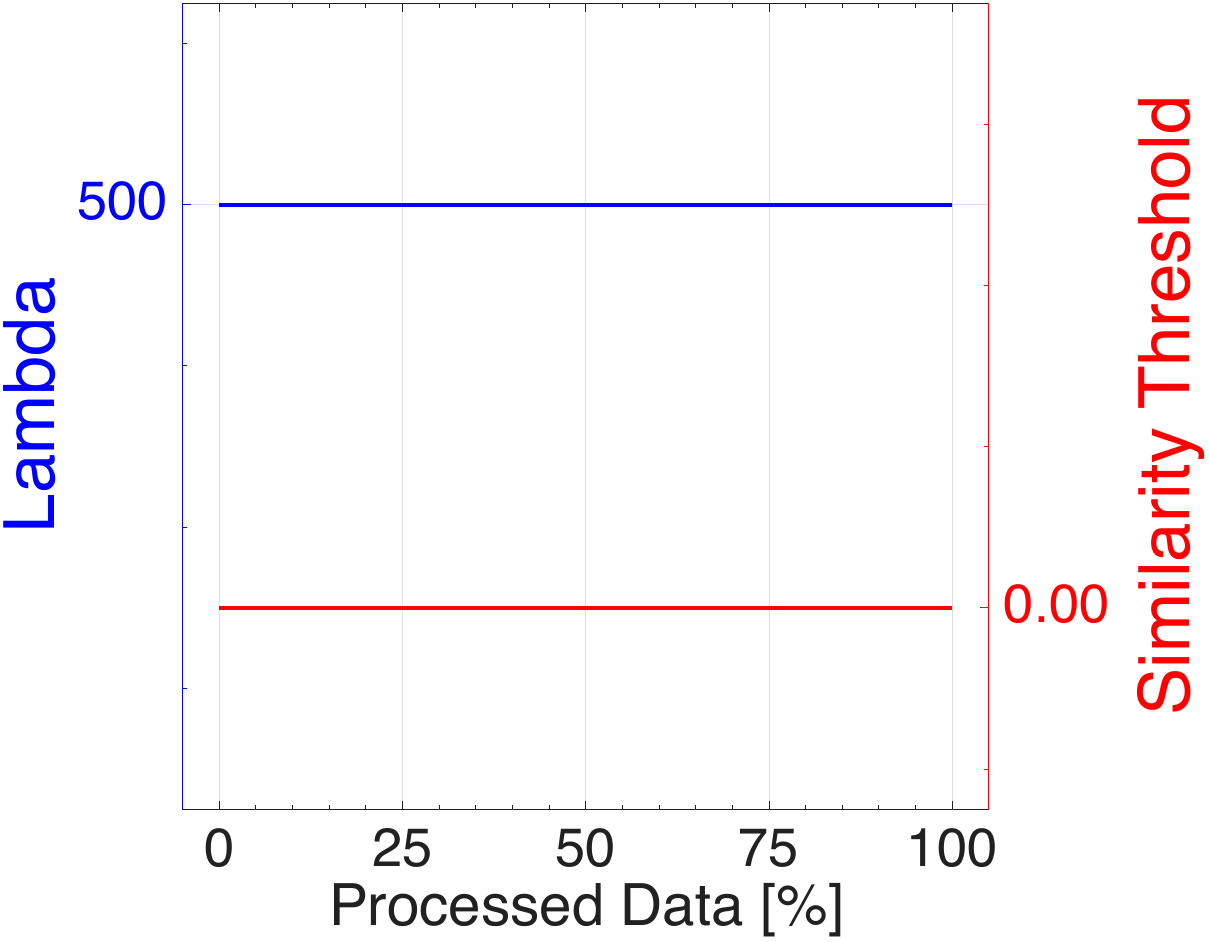}
  }\hfill
  \subfloat[Pima]{%
    \includegraphics[width=0.22\linewidth]{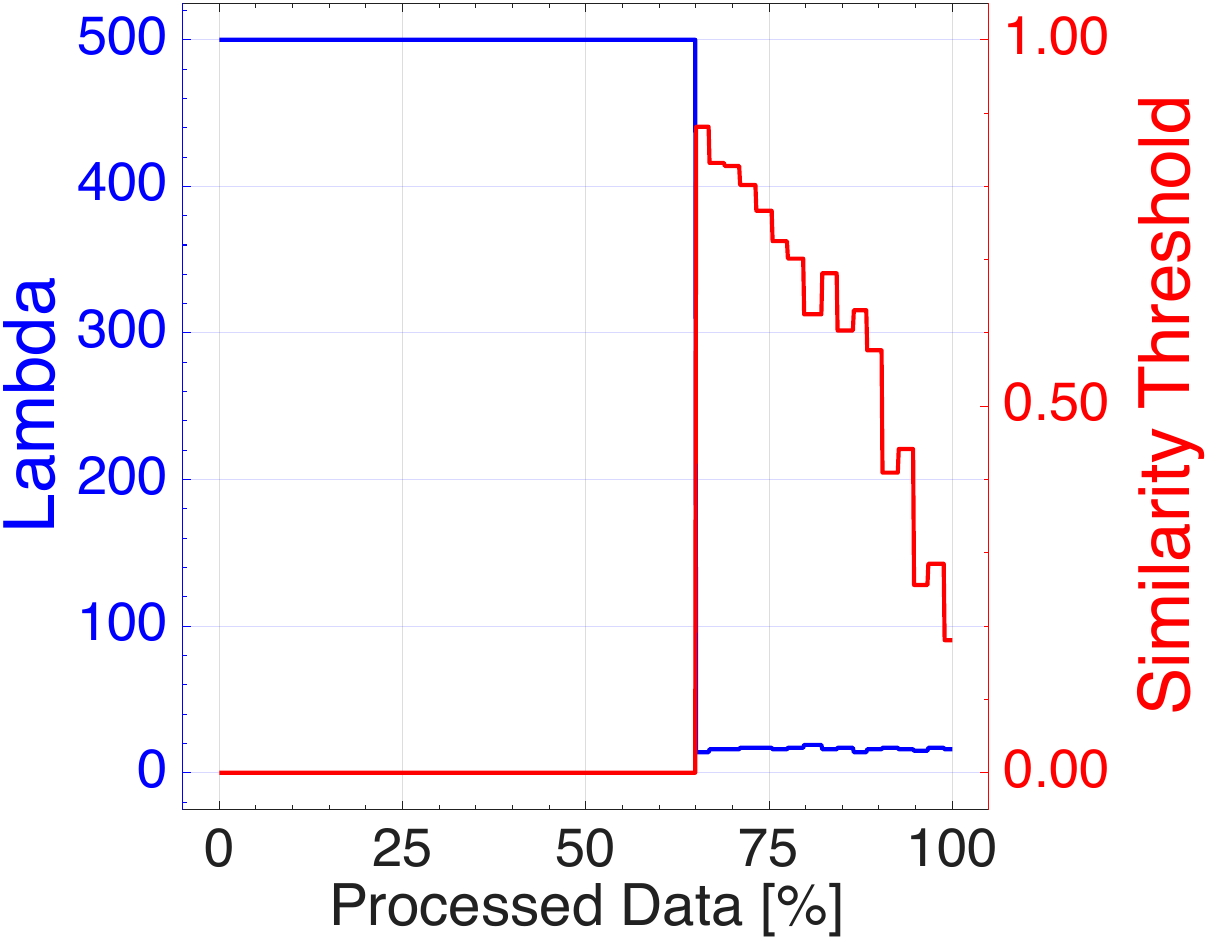}
  }\\
  \subfloat[Mice Protein]{%
    \includegraphics[width=0.22\linewidth]{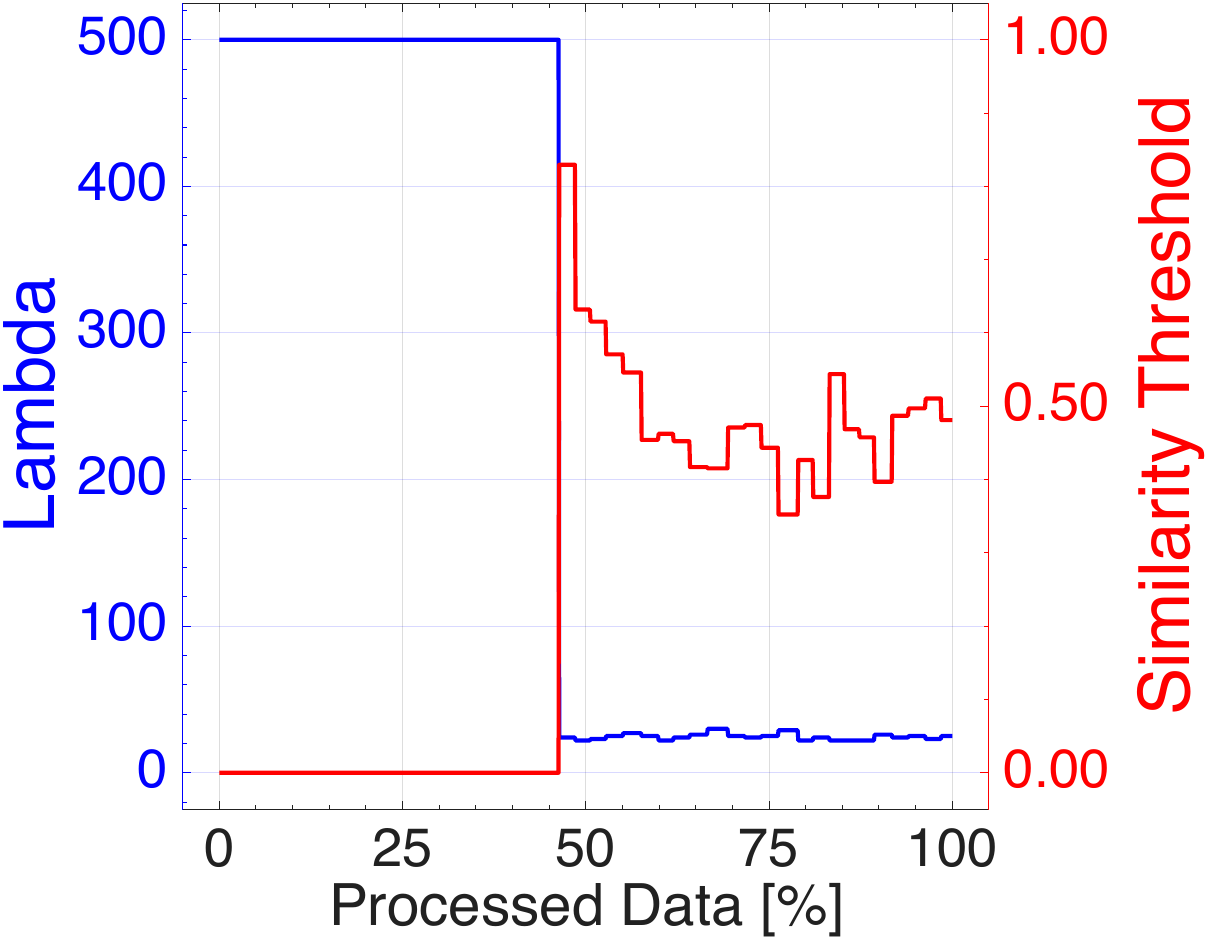}
  }\hfill
  \subfloat[Binalpha]{%
    \includegraphics[width=0.22\linewidth]{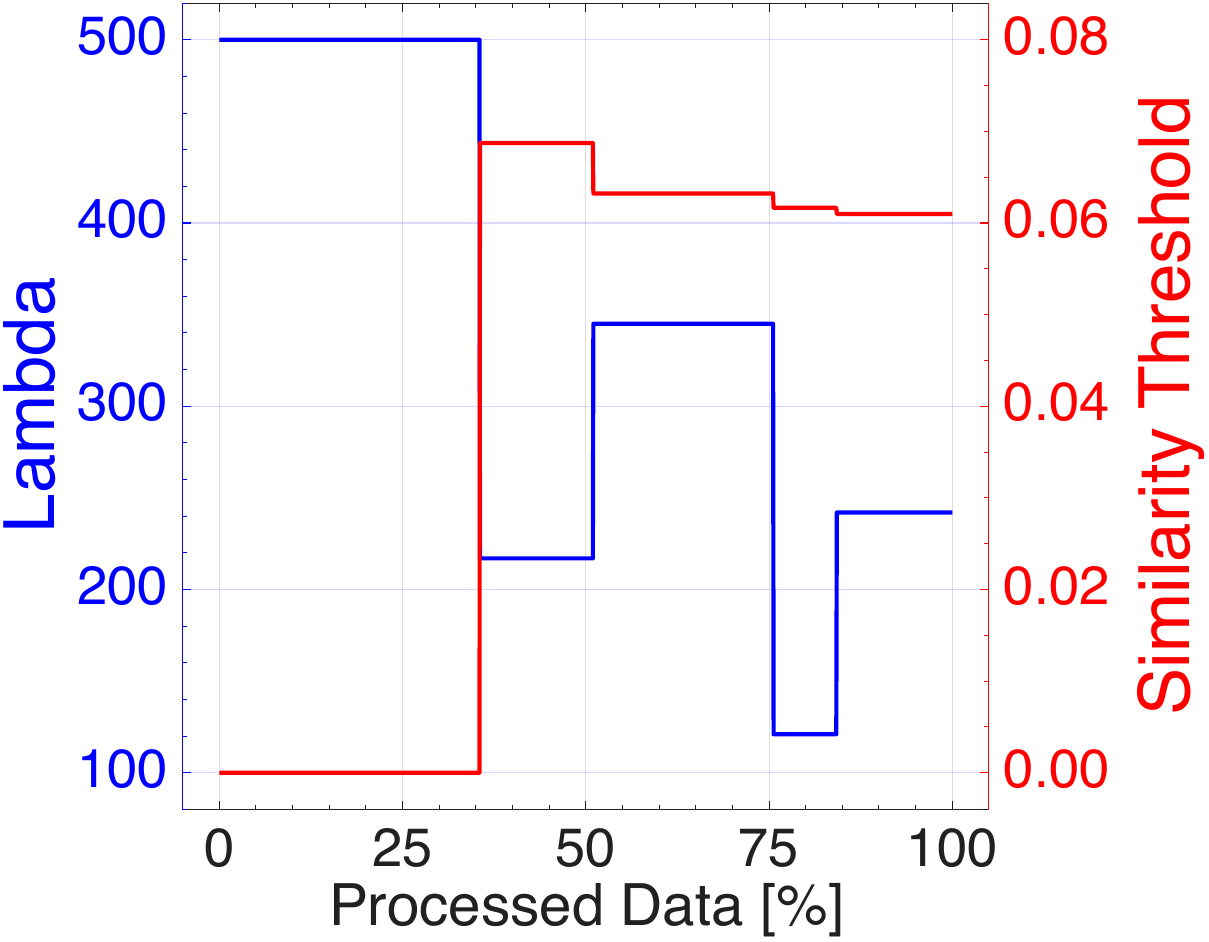}
  }\hfill
  \subfloat[Yeast]{%
    \includegraphics[width=0.22\linewidth]{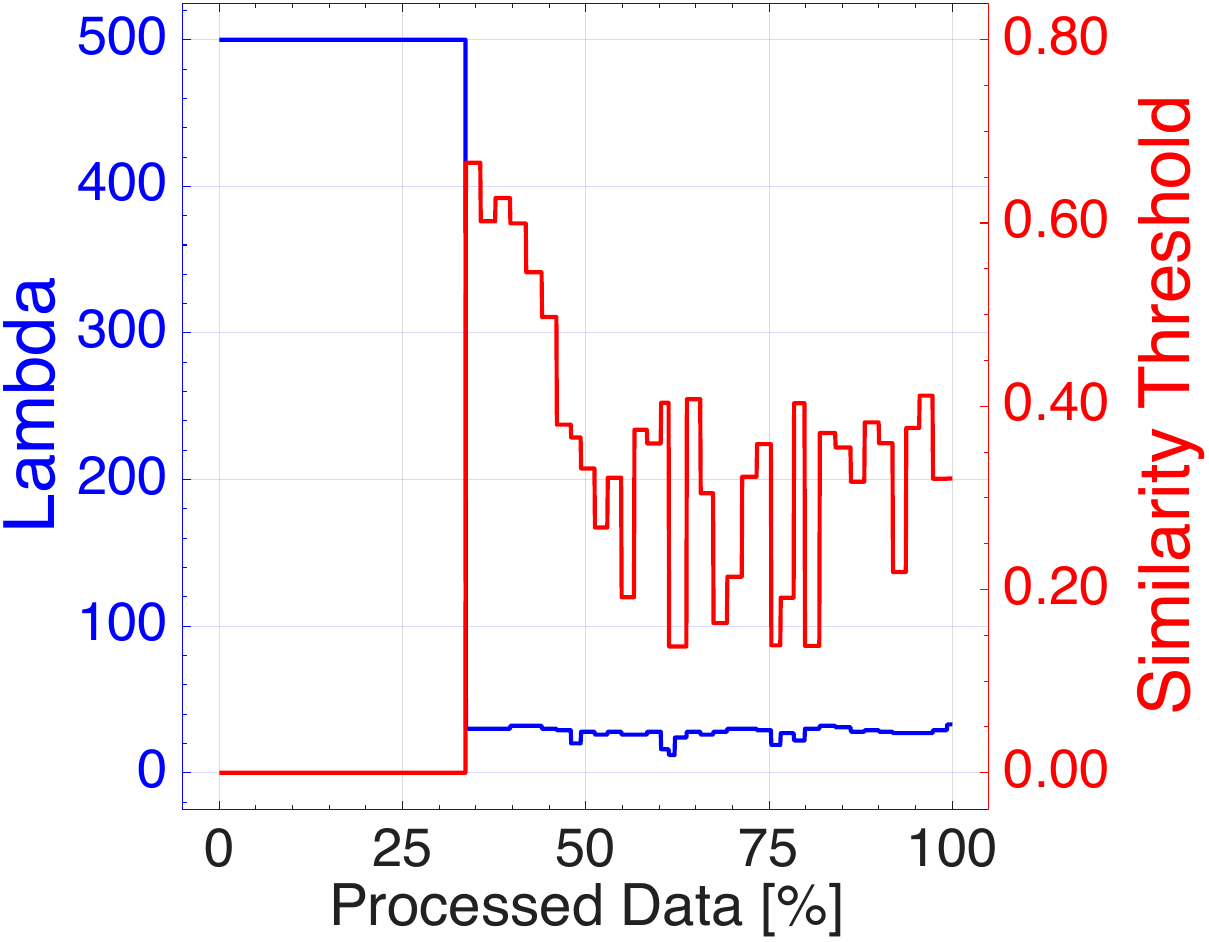}
  }\hfill
  \subfloat[Semeion]{%
    \includegraphics[width=0.22\linewidth]{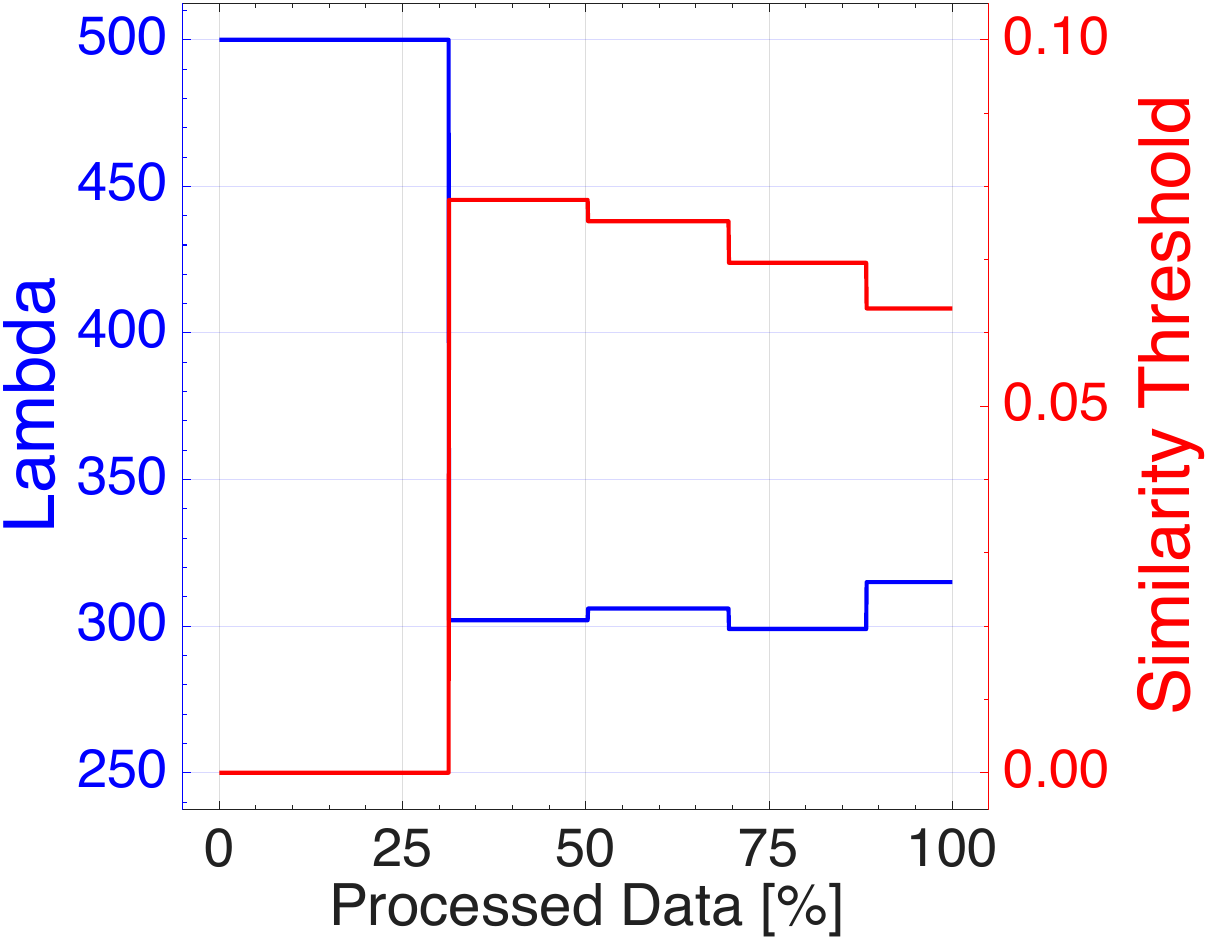}
  }\\
  \subfloat[MSRA25]{%
    \includegraphics[width=0.22\linewidth]{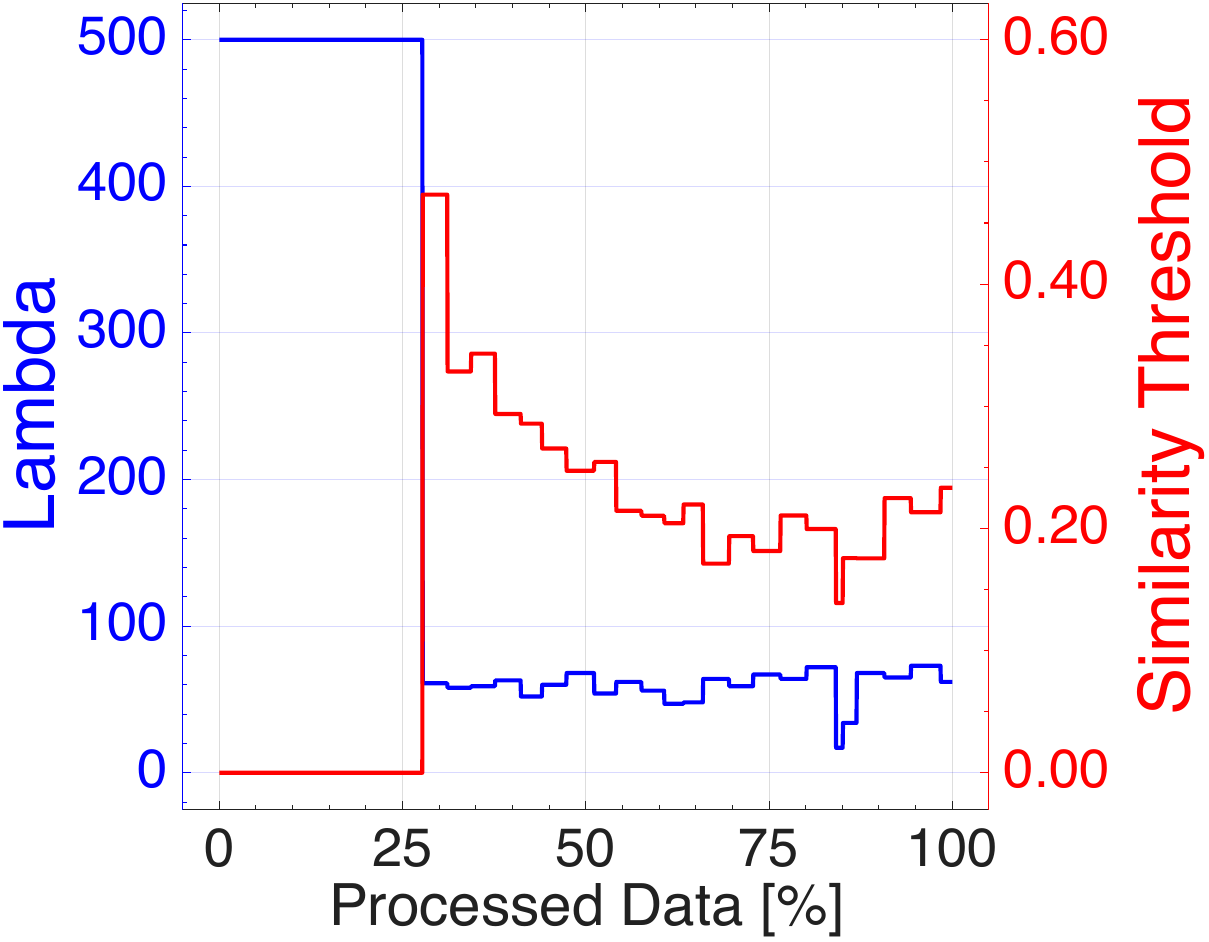}
  }\hfill
  \subfloat[Image Segmentation]{%
    \includegraphics[width=0.22\linewidth]{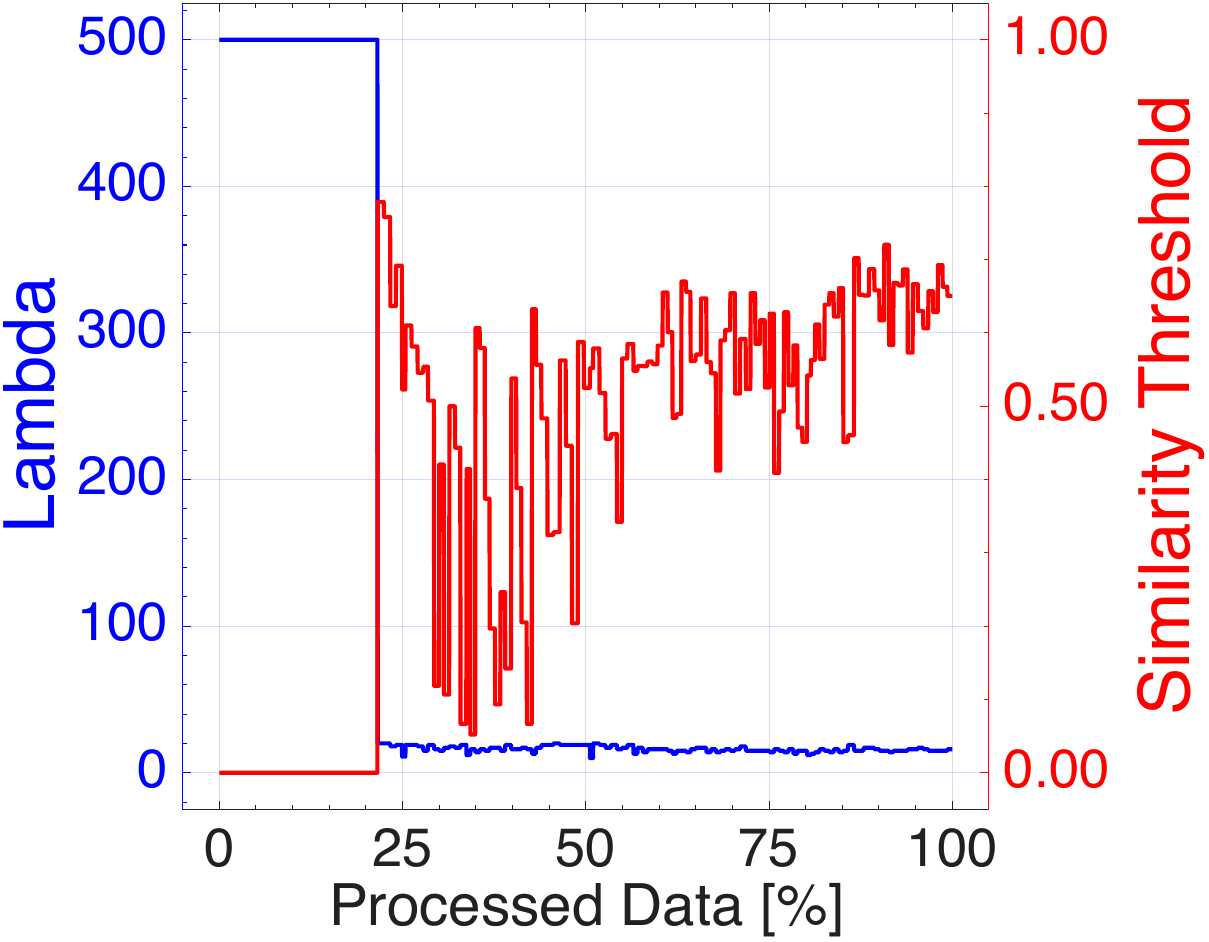}
  }\hfill
  \subfloat[Rice]{%
    \includegraphics[width=0.22\linewidth]{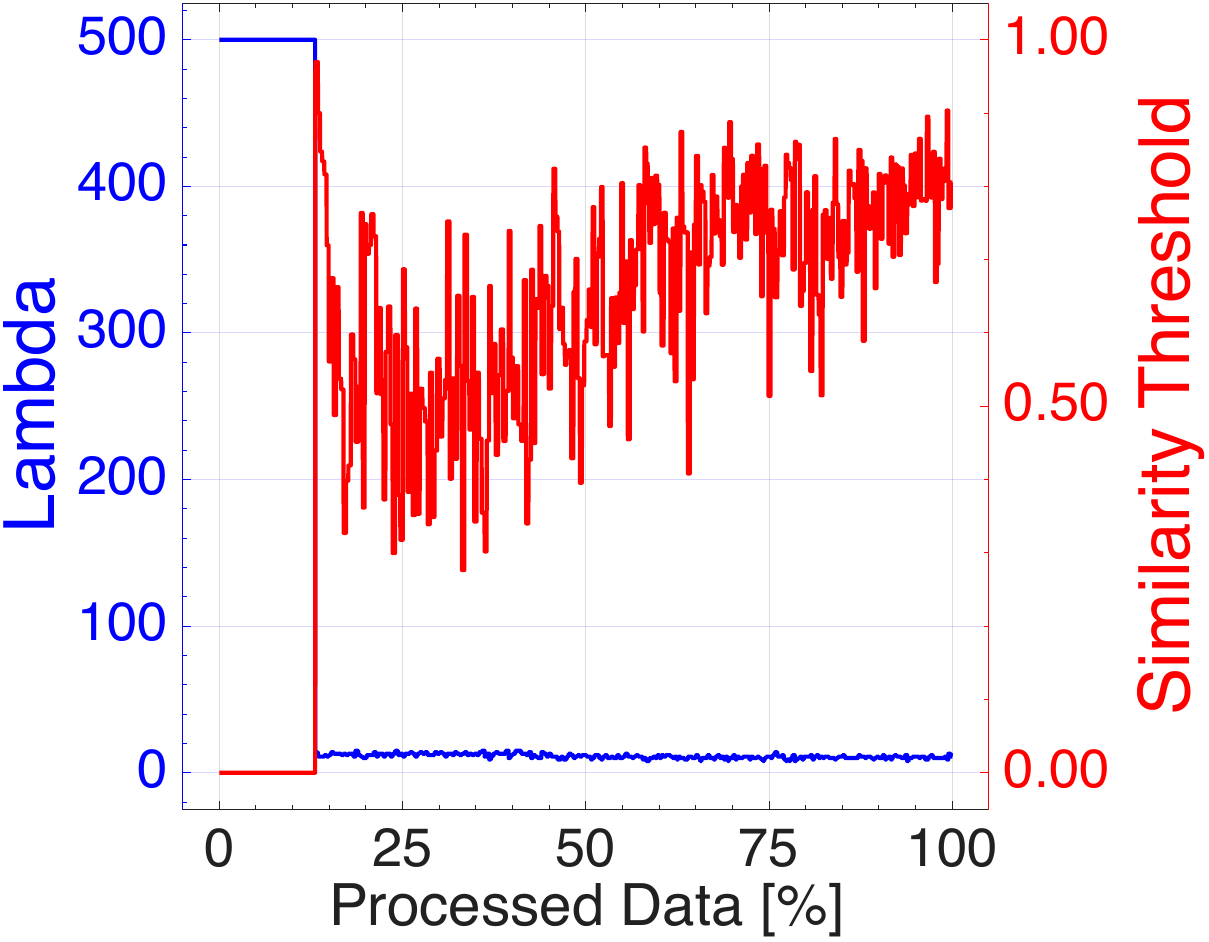}
  }\hfill
  \subfloat[TUANDROMD]{%
    \includegraphics[width=0.22\linewidth]{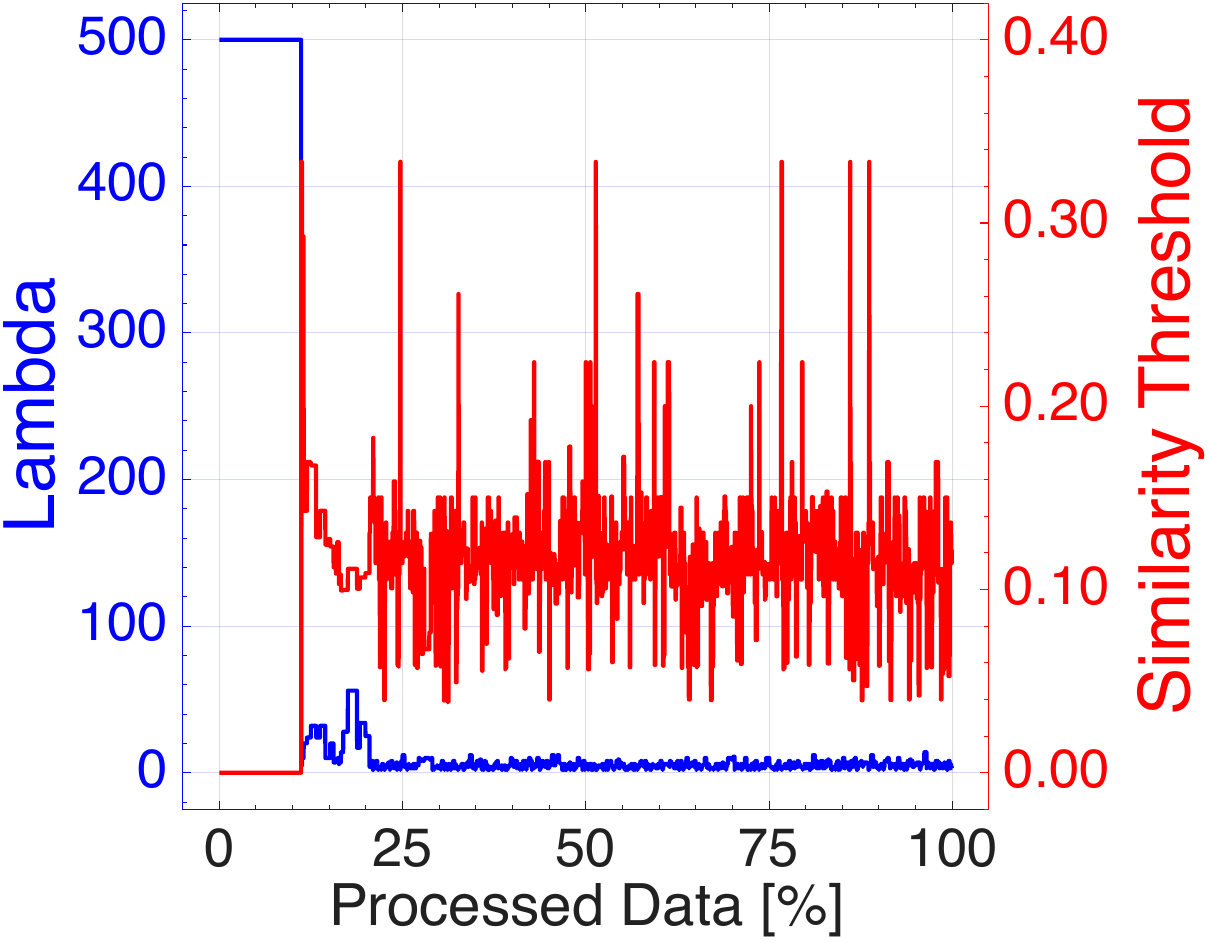}
  }\\
  \subfloat[Phoneme]{%
    \includegraphics[width=0.22\linewidth]{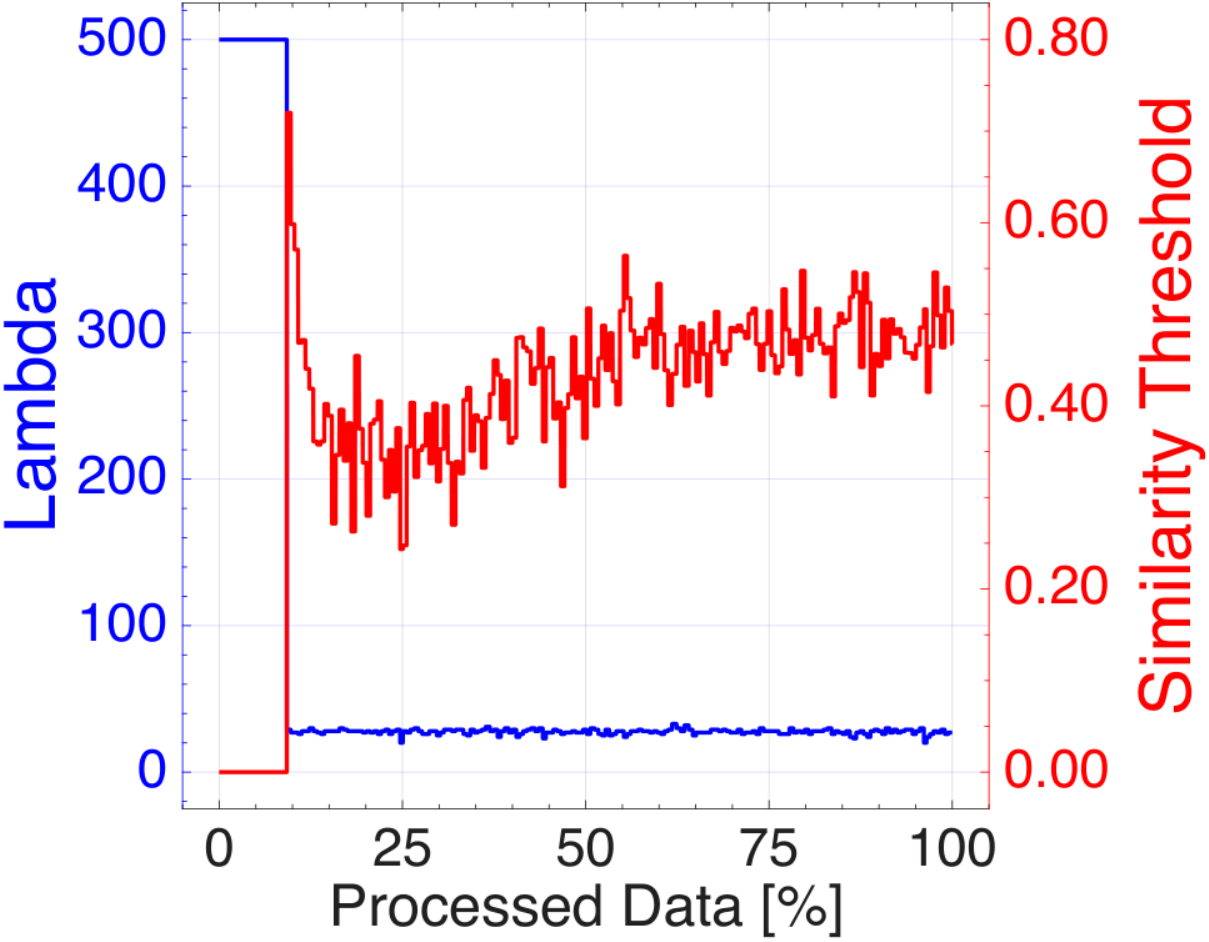}
  }\hfill
  \subfloat[Texture]{%
    \includegraphics[width=0.22\linewidth]{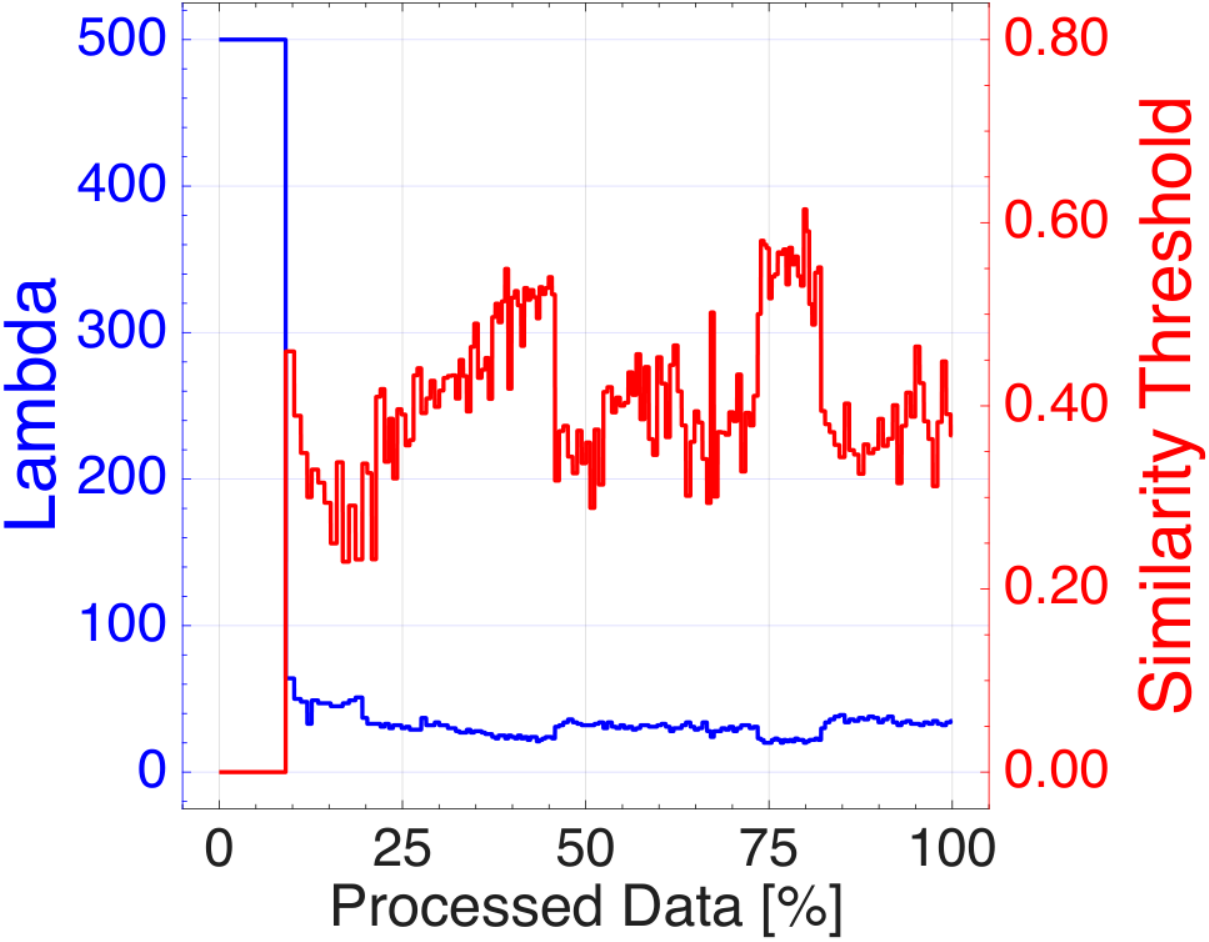}
  }\hfill
  \subfloat[OptDigits]{%
    \includegraphics[width=0.22\linewidth]{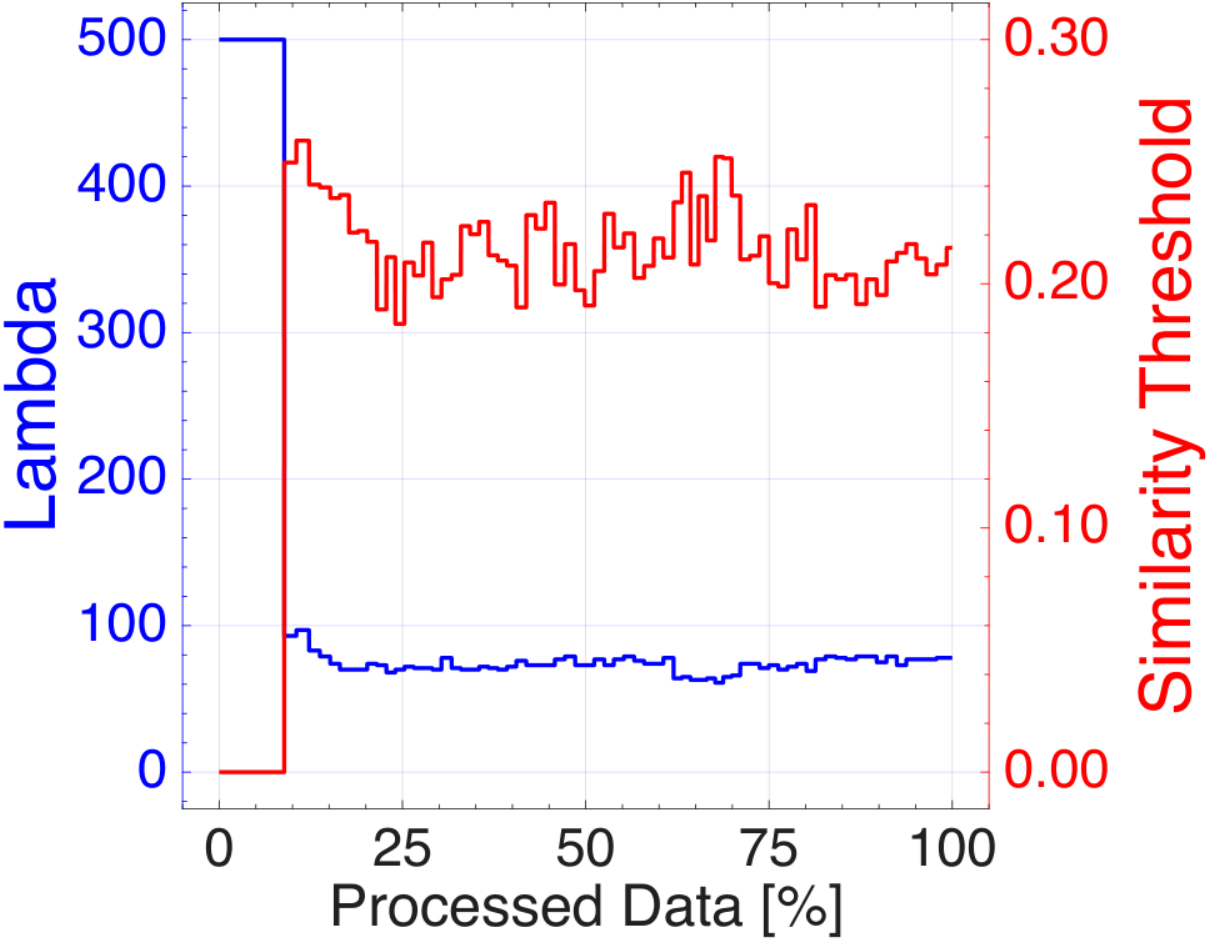}
  }\hfill
  \subfloat[Statlog]{%
    \includegraphics[width=0.22\linewidth]{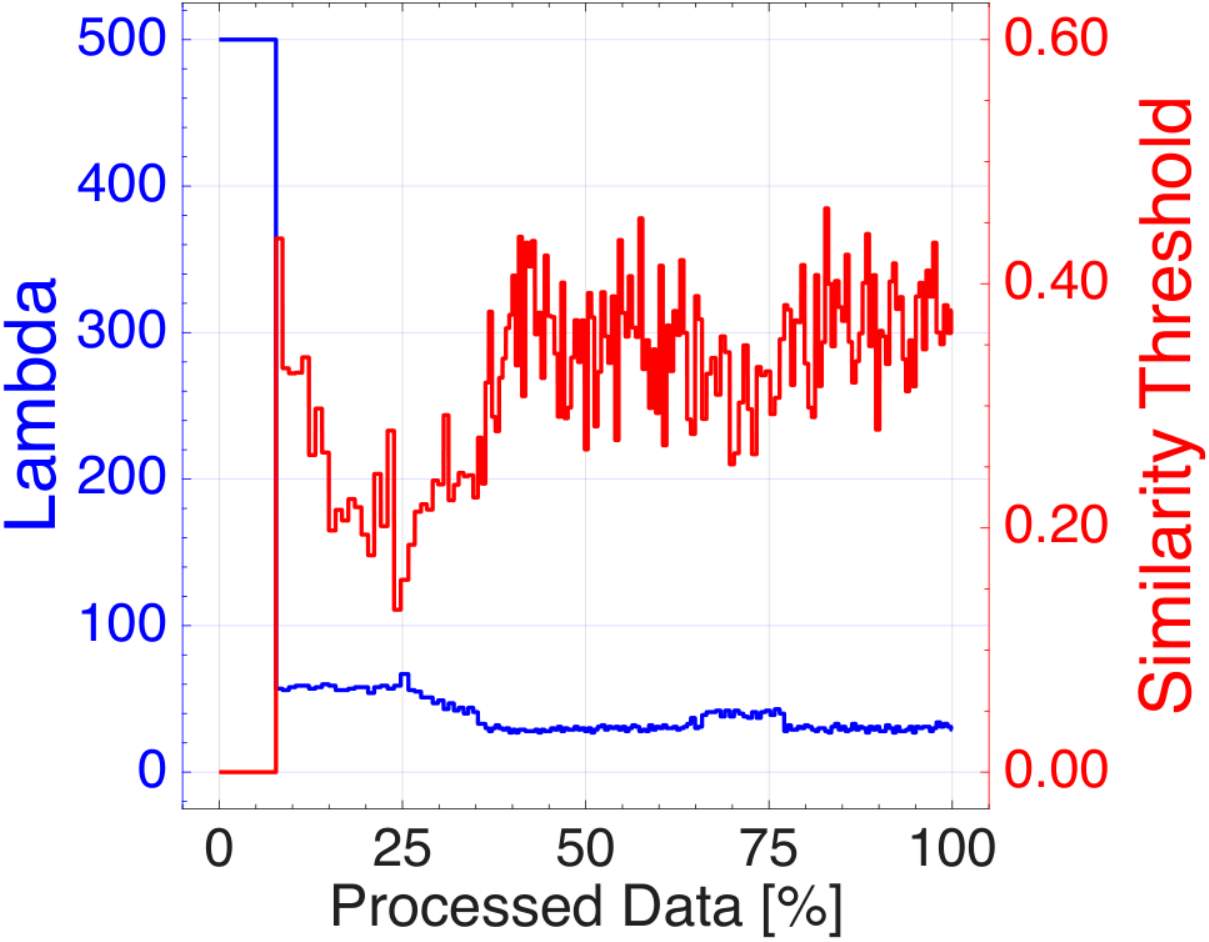}
  }\\
  \subfloat[Anuran Calls]{%
    \includegraphics[width=0.22\linewidth]{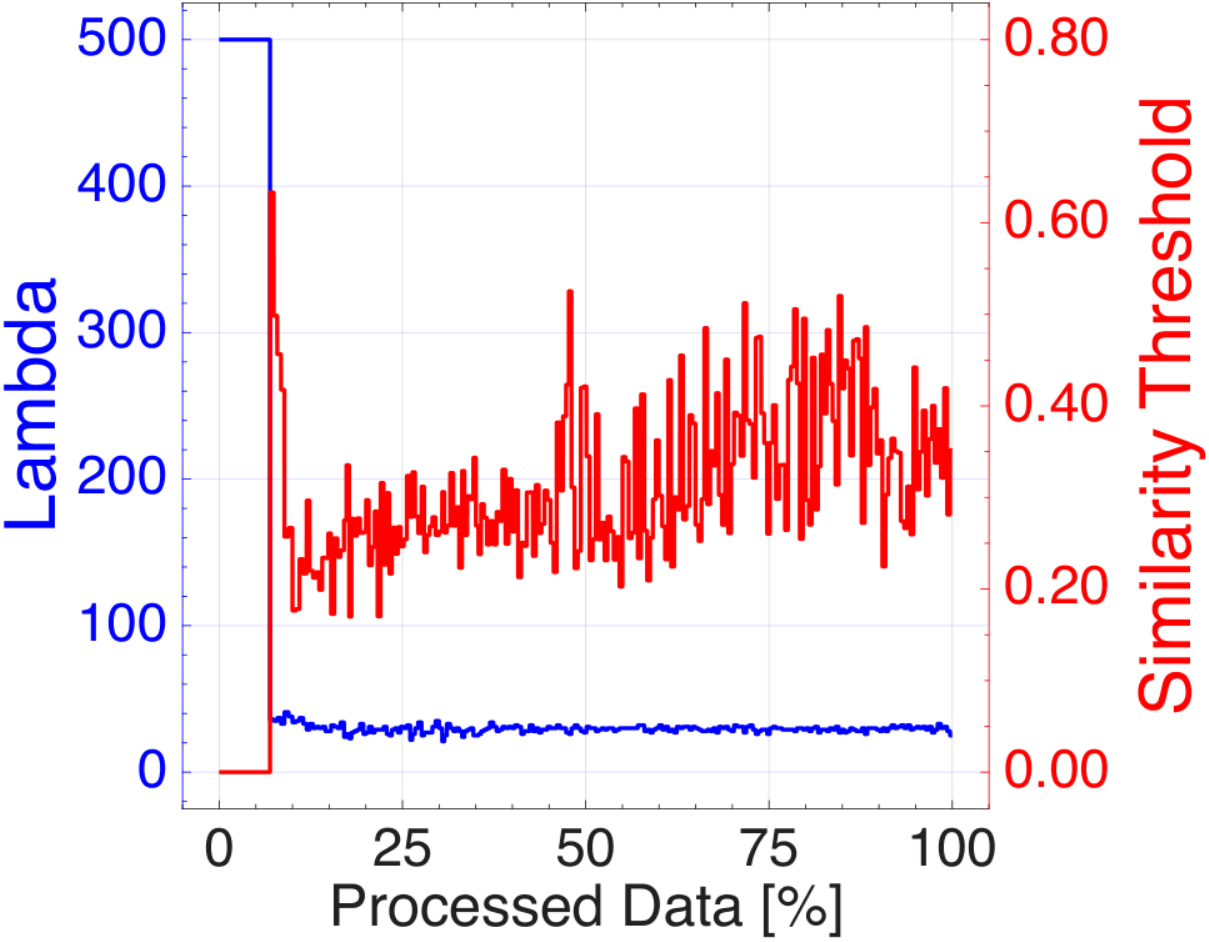}
  }\hfill
  \subfloat[Isolet]{%
    \includegraphics[width=0.22\linewidth]{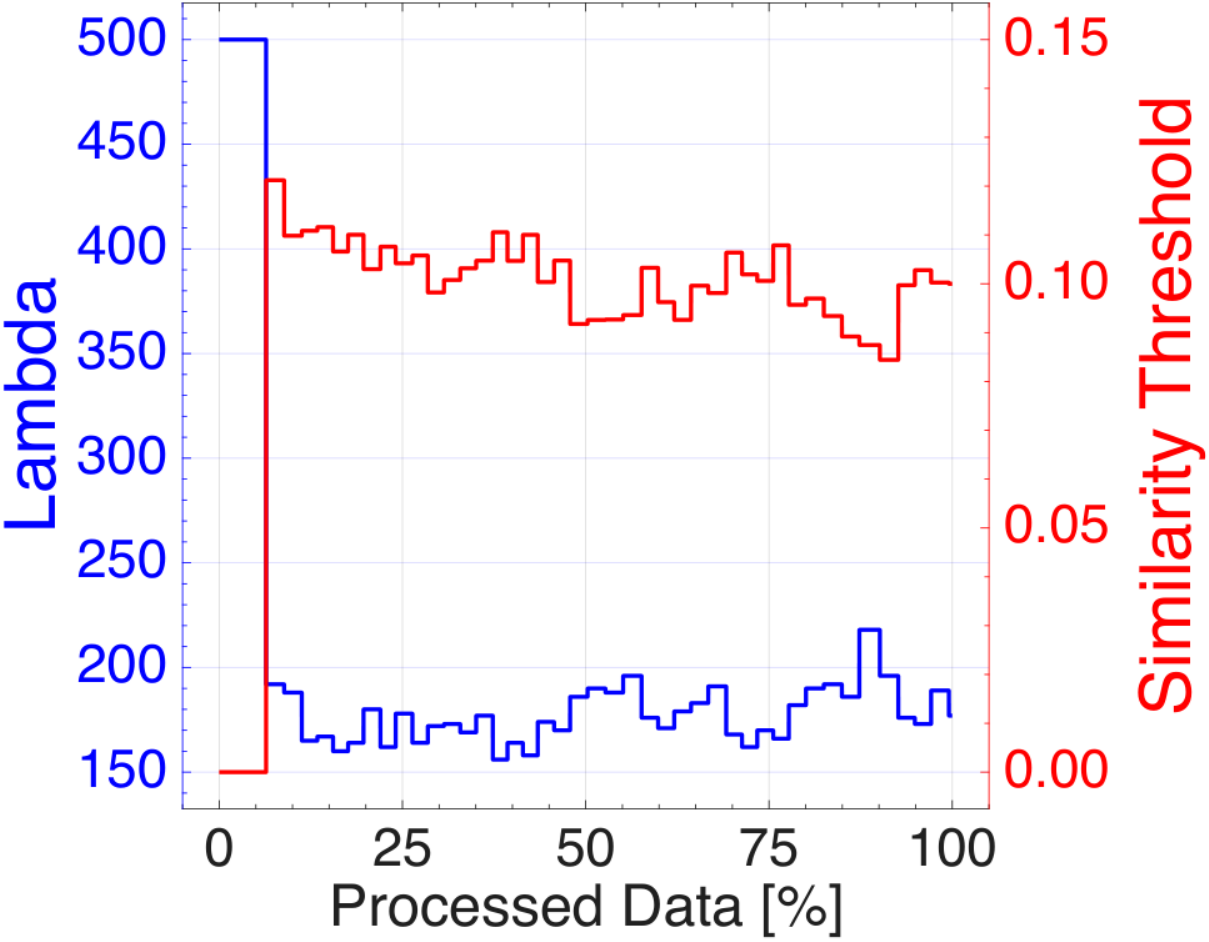}
  }\hfill
  \subfloat[MNIST10K]{%
    \includegraphics[width=0.22\linewidth]{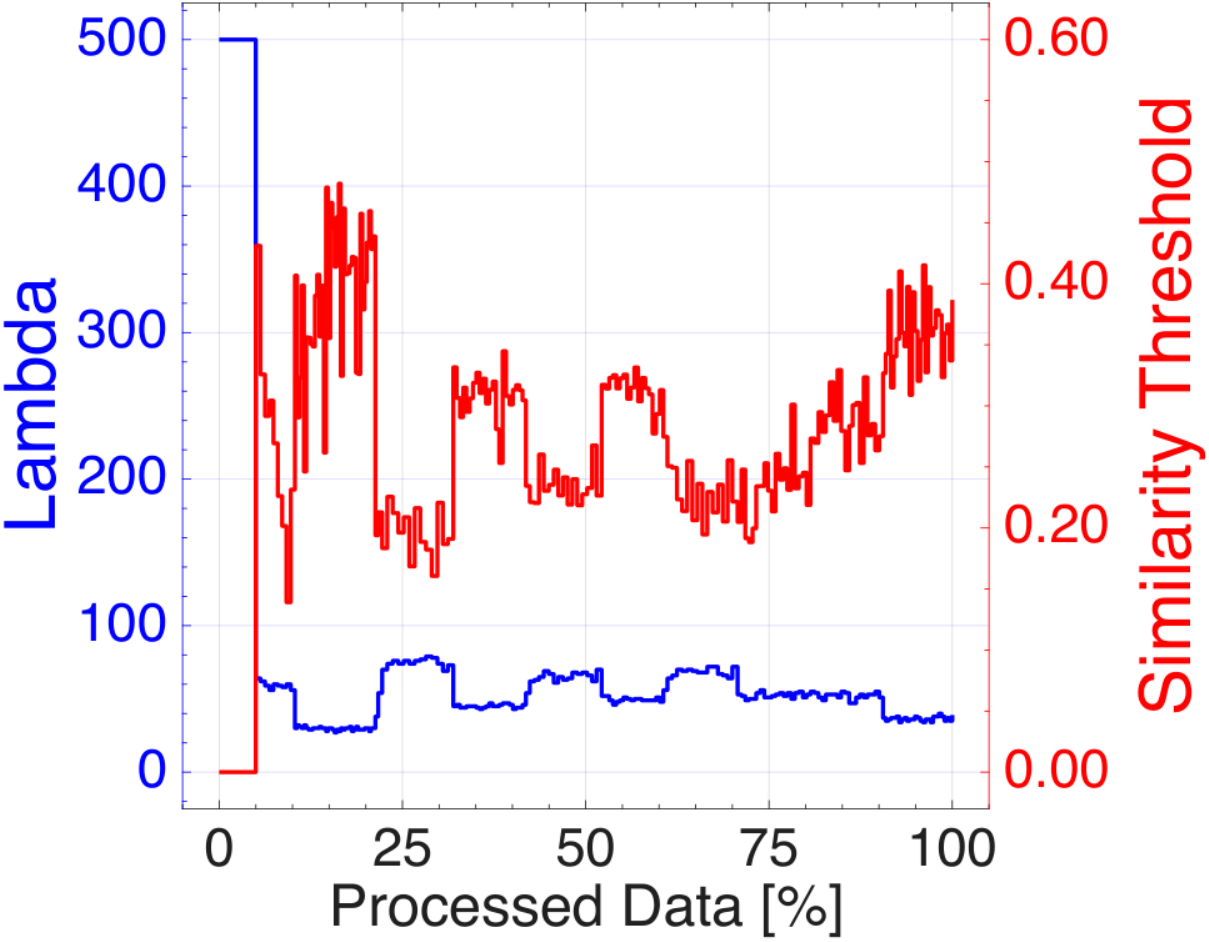}
  }\hfill
  \subfloat[PenBased]{%
    \includegraphics[width=0.22\linewidth]{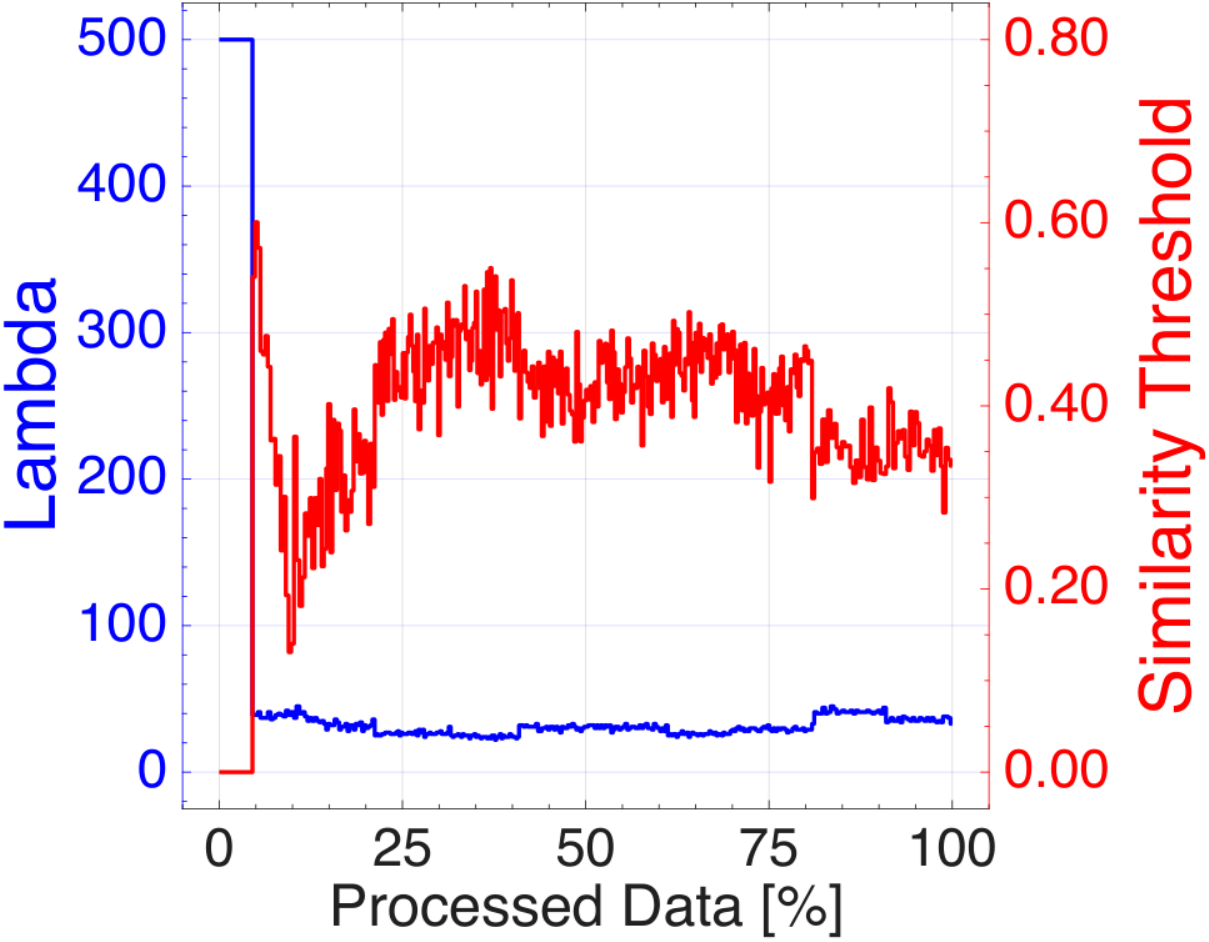}
  }\\
  \subfloat[STL10]{%
    \includegraphics[width=0.22\linewidth]{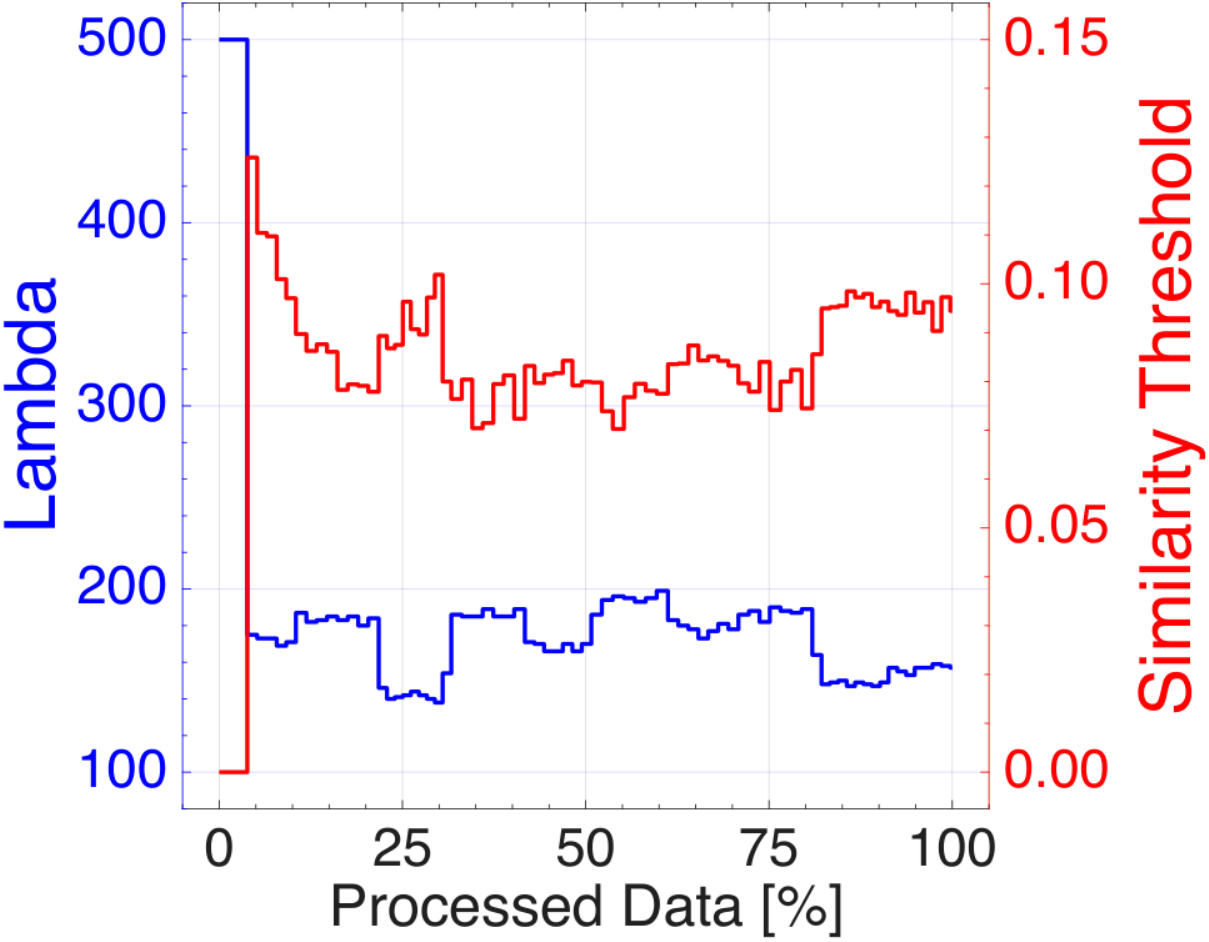}
  }\hfill
  \subfloat[Letter]{%
    \includegraphics[width=0.22\linewidth]{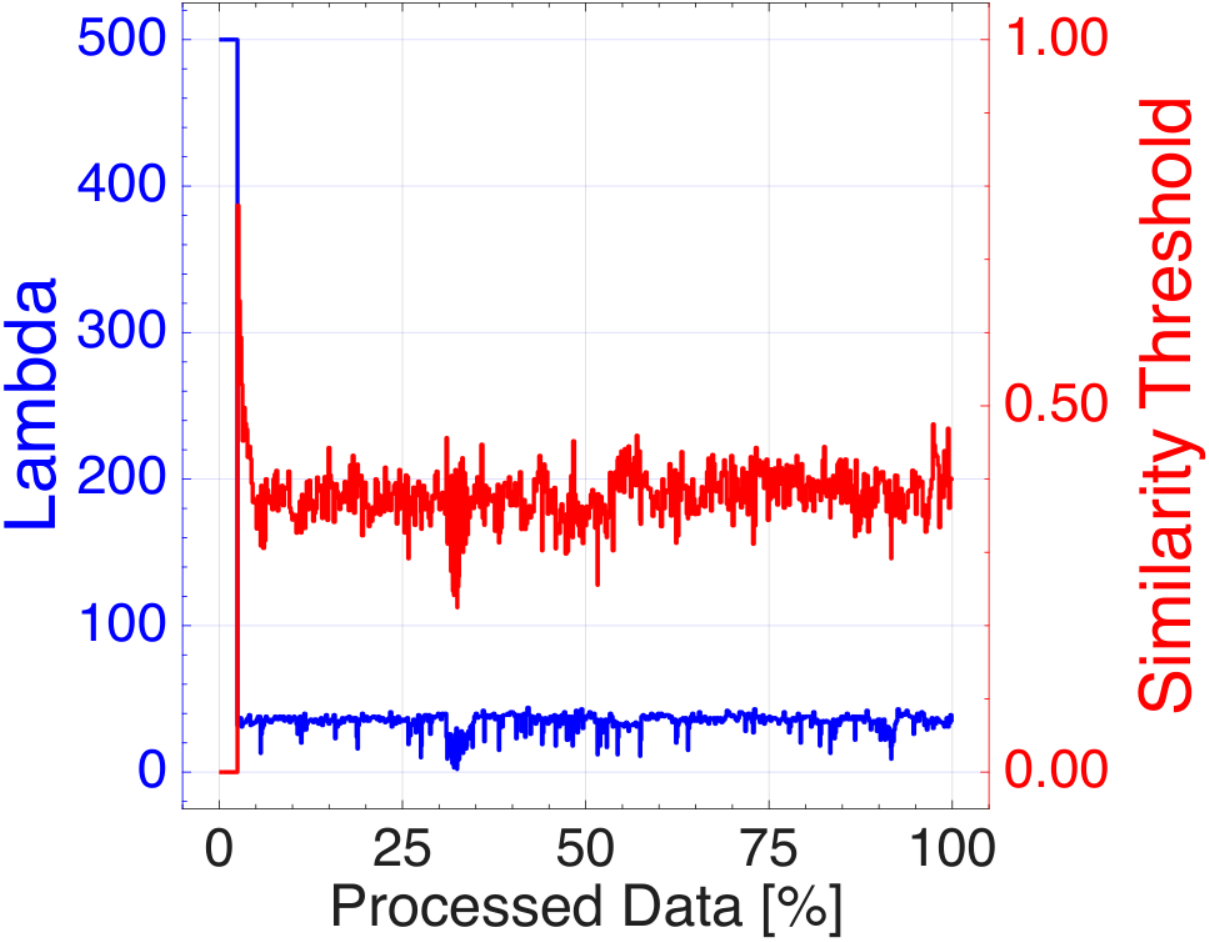}
  }\hfill
  \subfloat[Shuttle]{%
    \includegraphics[width=0.22\linewidth]{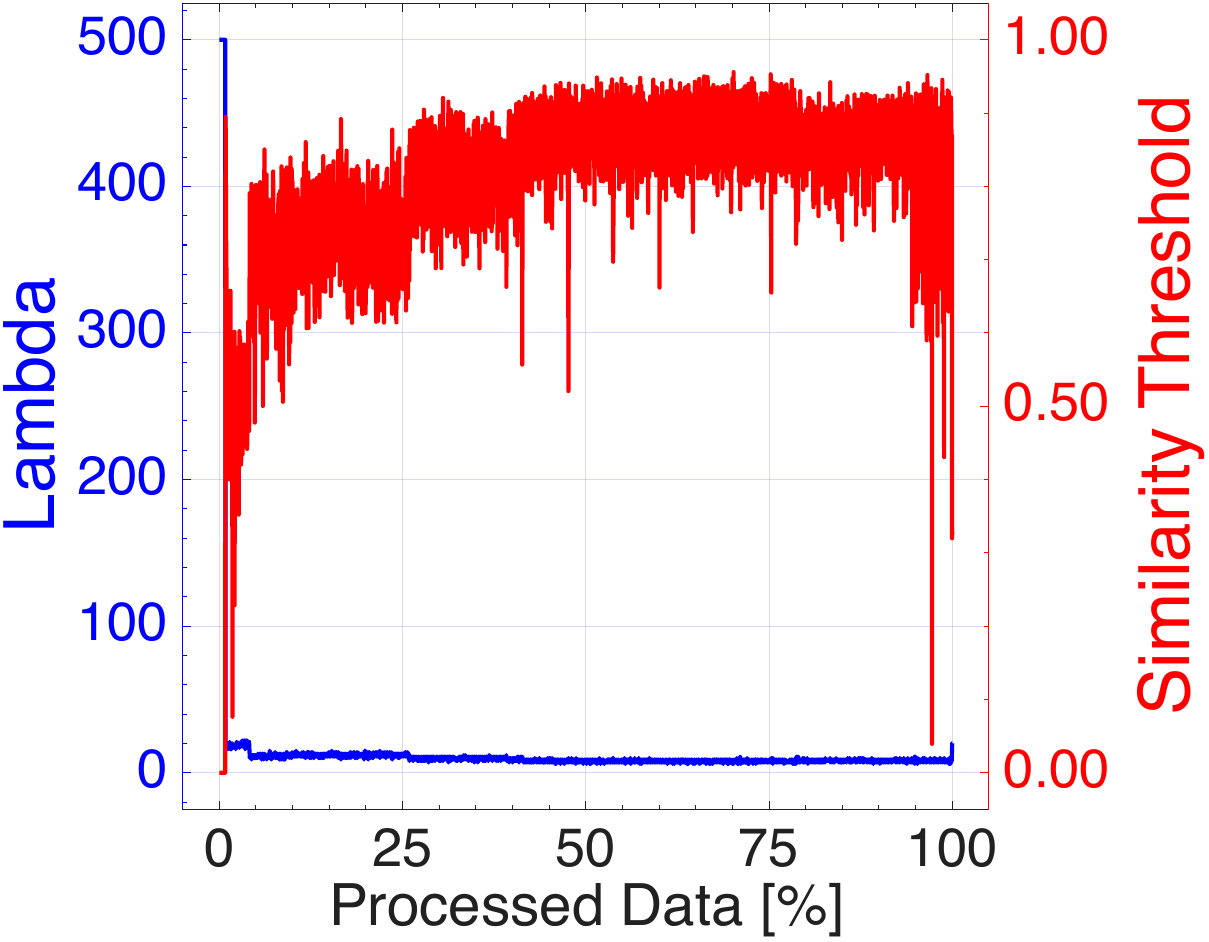}
  }\hfill
  \subfloat[Skin]{%
    \includegraphics[width=0.22\linewidth]{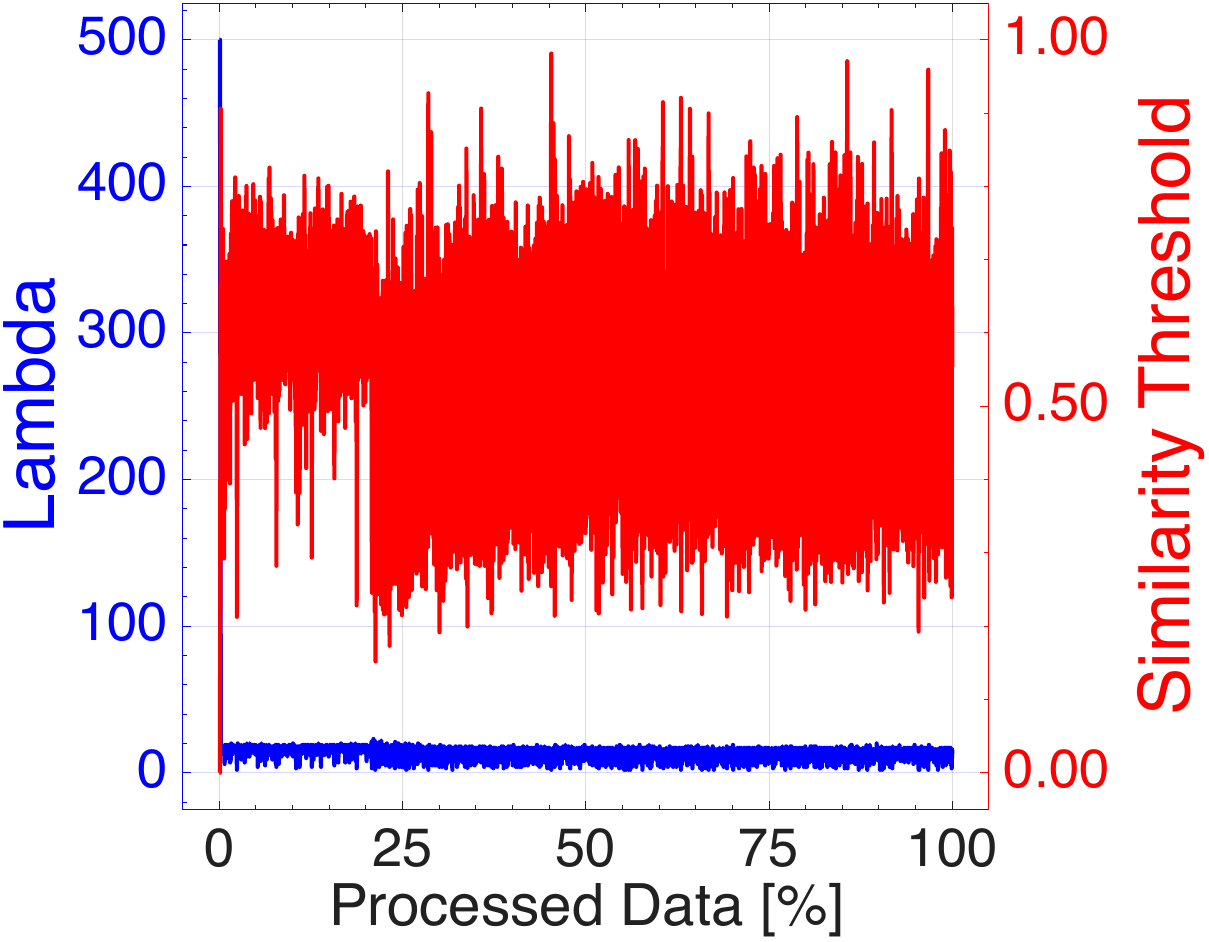}
  }
  \caption{Histories of $\Lambda$ and $V_{\text{threshold}}$ for IDAT in the nonstationary setting ($\Lambda_{\text{init}} = 500$).}
  \label{fig:ablation_lambda_history_idat_500_nonstationary}
\end{figure*}

% History of $\Lambda$ and $V_{\text{threshold}}$ for IDAT (w/o Decremental) in the nonstationary setting.

% History of $\Lambda$ and $V_{\text{threshold}}$ for IDAT (w/o Decremental) in the nonstationary setting.

% History of $\Lambda$ and $V_{\text{threshold}}$ for IDAT (w/o Incremental) in the nonstationary setting.

% History of $\Lambda$ and $V_{\text{threshold}}$ for IDAT (w/o Incremental) in the nonstationary setting.

\end{document}